\documentclass[10pt,twocolumn,letterpaper]{article}

\usepackage{cvpr}
\usepackage{times}
\usepackage{epsfig}
\usepackage{amsmath,bm}
\usepackage{amssymb}
\usepackage{graphicx}  \graphicspath{{figures/}}
\usepackage{multirow}
\usepackage{xspace}
\usepackage{paralist}
\usepackage[dvipsnames,svgnames,x11names]{xcolor}
\usepackage{algorithm}
\usepackage{algpseudocode}
\usepackage[normalem]{ulem}
\usepackage{makecell}
\usepackage{rotating}
\usepackage{booktabs}
\usepackage{gensymb}
\usepackage{subcaption}
\usepackage{placeins}

\newcommand{\rulesep}{\unskip\ \vrule\ }

\usepackage[pagebackref=true,breaklinks=true,letterpaper=true,colorlinks,bookmarks=false]{hyperref}

\usepackage{ulem}

\cvprfinalcopy 

\def\cvprPaperID{168} 

\newcommand{\setmode}[1]{\def\mode{#1}}
\setmode{final} 
\long\def\IGNORE#1{} \long\def\COMMENT#1{}
\def\authornote#1#2#3{{\textcolor{#2}{\textsl{\small#1:[*#3*]}}}}

\ifthenelse{\equal{\mode}{draft}} { 
    \newcommand{\sbnote}[1]{\authornote{SB}{Red}{#1}} 
	\newcommand{\jgnote}[1]{\authornote{JG}{Magenta}{#1}} 
    \newcommand{\khnote}[1]{\authornote{KH}{Blue}{#1}} 
    \newcommand{\advise}[1]{\authornote{NOTE}{Cyan}{#1}}
	\newcommand{\jhnote}[1]{\authornote{JH}{Green}{#1}} 
    \newcommand{\jknote}[1]{\authornote{JK}{Orange}{#1}} 
    } {}

\ifthenelse{\equal{\mode}{final}} {
    \newcommand{\advise}[1]{}
    \newcommand{\sbnote}[1]{}
    \newcommand{\jgnote}[1]{}
    \newcommand{\khnote}[1]{}
    \newcommand{\jhnote}[1]{}
    \newcommand{\jknote}[1]{}
    \typeout{************* MODE: Final}
    } {}

\newcommand{\comment}[1]{} 

\ifcvprfinal\fi
\begin{document}

\title{Geometry-Aware Learning of Maps for Camera Localization}
\author{Samarth Brahmbhatt$^{1}$ \,\,\,\, Jinwei Gu$^{2}$ \,\,\,\, Kihwan Kim$^{2}$ \,\,\,\, James Hays$^{1}$ \,\,\,\, Jan Kautz$^{2}$\\
\small{ $^1$\{\href{mailto:samarth.robo@gatech.edu}{samarth.robo},  \href{mailto:hays@gatech.edu}{hays}\}@gatech.edu\,\,\,\,\ $^2$\{\href{mailto:jinweig@nvidia.com}{jinweig}, \href{mailto:kihwank@nvidia.com}{kihwank}, \href{mailto:jkauz@nvidia.com}{jkautz}\}@nvidia.com}\\
         $^1$Georgia Institute of Technology\,\,\,\, $^2$NVIDIA $\qquad$}


\maketitle

\begin{abstract}

    Maps are a key component in image-based camera localization and visual SLAM
    systems: they are used to establish geometric constraints between images,
    correct drift in relative pose estimation, and relocalize cameras after
    lost tracking.  The exact definitions of maps, however, are often
    application-specific and hand-crafted for different scenarios (\eg 3D
    landmarks, lines, planes, bags of visual words).  We propose to represent
    maps as a deep neural net called MapNet, which enables learning a
    \textit{data-driven} map representation.  Unlike prior work on learning
    maps, MapNet exploits cheap and ubiquitous sensory inputs like visual
    odometry and GPS in addition to images and fuses them together for camera
    localization.  Geometric constraints expressed by these inputs, which have
    traditionally been used in bundle adjustment or pose-graph optimization,
    are formulated as loss terms in MapNet training and also used during
    inference.  In addition to directly improving localization accuracy, this
    allows us to update the MapNet (\ie, maps) in a self-supervised manner using
    additional unlabeled video sequences from the scene. We also propose a
    novel parameterization for camera rotation which is better suited for
    deep-learning based camera pose regression. Experimental results on both
    the indoor 7-Scenes dataset and the outdoor Oxford RobotCar dataset show significant
    performance improvement over prior work. The MapNet project webpage is
    \href{https://goo.gl/mRB3Au}{\url{https://goo.gl/mRB3Au}}.

\end{abstract}

\vspace{-1em}
\section{Introduction}
\vspace{-.5em}

Camera localization \ie recovering the 3D position and orientation of a
moving camera is one of the fundamental tasks in computer vision with a wide
variety of applications in robotics, autonomous driving, and AR/VR.
A key component in camera localization, including various visual SLAM
systems~\cite{Engel2014LSD, Mur-ArtalMT15, Zhou2015StructSLAM} and image-based
localization methods~\cite{Li2012Point, Sattler2011,Sattler2017} is the concept
of a \textit{map}. A map is an abstract summary of the input data that
establishes geometric constraints between observations and can be queried to get the camera pose
when tracking is drifting or lost. 
Maps, however, are usually defined in an application-specific manner with hand-crafted features.
Examples include 3D landmarks for
general visual SLAM methods~\cite{Klein2007PTAM, Li2012Point, Mur-ArtalMT15},
3D lines and patches in semi-dense SLAM methods and indoor
scenes~\cite{Engel2014LSD,Newcombe11,Zhou2015StructSLAM}, object-level context
in semantic SLAM methods~\cite{Civera11Semantic,Moreno13SLAMp}, bag of visual
word features on key frames for camera
relocalization~\cite{Cummins2008BOW,Sattler2011,Sattler2017}.  Being
application-specific, these map representations may ignore useful (sometimes,
the only available) features in environments they were not designed for, and
are inflexible to update as more input data come in.

\begin{figure}
\centering
    \includegraphics[width=.99\linewidth]{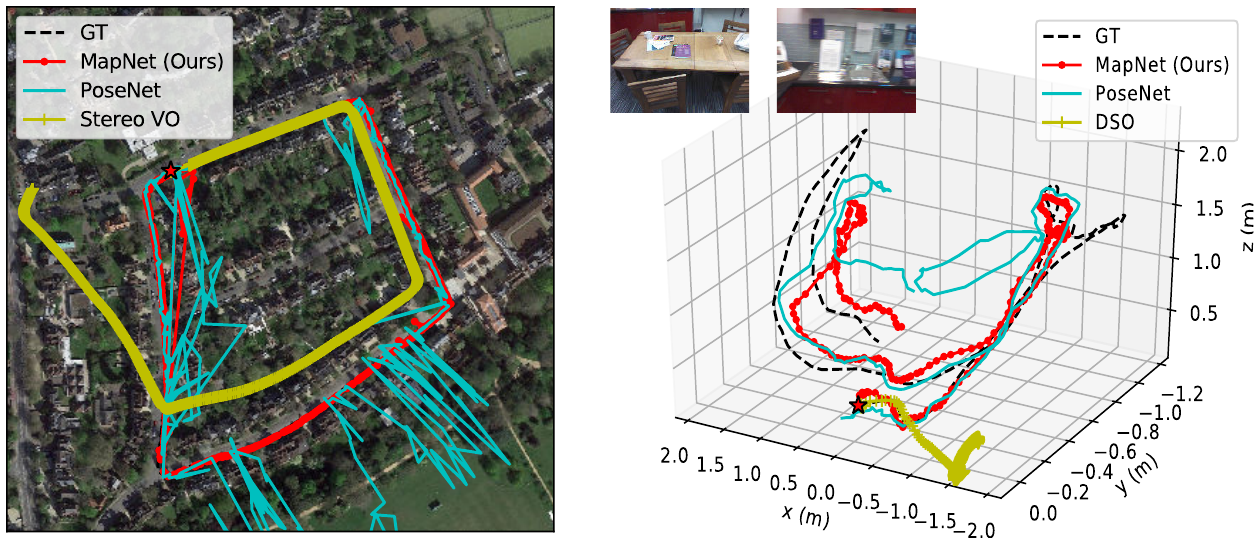}
    \vspace{-1em}
    \caption{\small Camera localization results
    for outdoor (left) and indoor (right) scenes 
    from the Oxford RobotCar~\cite{RobotCarDatasetIJRR} and 7-Scenes~\cite{Shotton13Scene7} datasets. 
    As shown, prior DNN-based methods (\eg, PoseNet~\cite{Kendall15iccv, Kendall16icra, Kendall17cvpr}) result in noisy estimations,
    while traditional visual odometry based methods (\eg, stereo VO or DSO~\cite{Engel2017DSO}) often drift over time. In contrast,
    MapNet gives accurate camera pose estimates by including various geometric constraints in DNN training and inference.}
    \label{fig:teaser}
    \vspace{-1em}
\end{figure}

Is there a general map representation for camera localization that addresses these drawbacks? In this paper,
we take a step towards answering this question. We propose to represent maps as
a DNN, called MapNet, which learns the map representation
directly from input data, with the flexibility to fuse multiple sensory inputs
and to improve over time using unlabeled data. MapNet
aims to be a part that can be easily plugged into any visual SLAM or
image-based localization systems. We are inspired by both the recent DNN-based
camera localization work (\eg, PoseNet~\cite{Kendall15iccv} and its
variants~\cite{Clark17VidLoc, Kendall17cvpr, Melekhov17Hourglass,Walch17LSTM})
in the context of structure-from-motion, as well as the traditional map
optimization methods (\eg, bundle
adjustment (BA)~\cite{Govindu01BA,Govindu04BA,Martinec07BA}, pose graph optimization
(PGO)~\cite{Carlone16PGO,Duckett02PGO, Lu97PGO}) in the context of visual SLAM.
Compared to these prior works, our approach makes the following contributions:
\begin{itemize}
    \item
        Most prior DNNs for camera localization~\cite{Kendall15iccv, Kendall16icra, Kendall17cvpr,
        Clark17VidLoc,Melekhov17Hourglass,Walch17LSTM} are trained using single images labelled
    	with absolute camera pose. In MapNet we show how the geometric constraints between
    	pairs of observations can be included as an additional loss term in training. These constraints
    	can come from a variety of sources: pose constraint from the visual odometry (VO)
    	between pairs of images, translation constraint from two GPS readings, rotation constraint from
    	two IMU readings, \etc. 
        We call this \emph{geometry-aware learning} and show that it significantly improves camera localization
        performance in Section~\ref{sec:results}.  

    \item PoseNet and its variants are \emph{offline}
        methods -- the learned DNNs are fixed after training. In contrast, we
        propose MapNet+ that can use the geometric constraints between
        pairs of observations mentioned above to continuously update the DNN weights (\ie, maps)
        without absolute camera pose supervision, as additional unlabeled data come in.
		Moreover, at runtime, we also exploit the complementary noise characteristics of MapNet predictions
		(locally noisy but drift-free) and VO (locally smooth but drifty) by fusing them
        in a moving window fashion with PGO. We call this variant MapNet+PGO. We show in Section~\ref{sec:results}
		that both MapNet+ and MapNet+PGO successively improve performance further.

    \item We propose a new parameterization for camera rotation, the logarithm of unit quaternion,
    	which is better suited for deep-learning based camera pose regression.
		This improves the performance of PoseNet and MapNet, as shown in Table~\ref{tab:result_logq}.
\end{itemize}


Figure~\ref{fig:teaser} shows two examples of camera localization.
Pure DNN-based methods (\eg, PoseNet~\cite{Kendall15iccv, Kendall16icra,
Kendall17cvpr}) result in noisy estimations, and the traditional VO-based
methods (\eg, stereo VO or DSO~\cite{Engel2017DSO}) often drift significantly
over time. By incorporting the geometric constraints into DNN-based learning
and inference, the proposed approach MapNet+PGO achieves the best result.
We evaluate the proposed methods extensively on both the indoor
7-Scenes~\cite{Shotton13Scene7} and the outdoor Oxford RobotCar
dataset~\cite{RobotCarDatasetIJRR}.

\section{Related Work}
\label{sec:related}


\begin{table*}
    \small
    \centering
    \caption{Comparison with prior DNN-based camera localization methods. Please refer to Section~\ref{sec:related} for details.}
    \vspace{-1em}
    \begin{tabular}{lllll|lll}
        \toprule
        \multirow{2}{*}{} & PoseNet & Hourglass & LSTM-Pose & VidLoc & \multicolumn{3}{c}{(Proposed)} \\
             & \cite{Kendall15iccv,Kendall16icra,Kendall17cvpr} & \cite{Melekhov17Hourglass}
             & \cite{Walch17LSTM} & \cite{Clark17VidLoc} &  MapNet & MapNet+ & MapNet+PGO
             \\
        \midrule
        Input              &  Images & Images & Images & Videos      & Images & \multicolumn{2}{c}{unlabeled videos + (VO, GPS, IMU)} \\
        Fusion ability     &  No     & No     & No     & No          & No     & Yes     & Yes \\
        Self-supervised update          &  No     & No     & No     & No          & No      & Yes      & No \\
        Temporal constraint &  No     & No     & No     & Yes         & Yes     & Yes   & Yes \\
        Geometry aware      &  Reprojection~\cite{Kendall17cvpr} & No & No & No & \multicolumn{3}{c}{Geometric constraints on camera poses}\\
        \bottomrule
    \end{tabular}
    \label{tab:comp}
    \vspace{-1em}
\end{table*}

\paragraph{Maps in Visual SLAM and Image-based Localization}

Over the years, various types of map representations and their optimization
techniques have been proposed for camera localization~\cite{Thrun02a,
Triggs99}. In visual SLAM, 3D landmarks (with feature descriptors of the
corresponding image patches) are often defined as maps in both Bayesian
filtering approaches~\cite{Davison03,Montemerlo02,Thrun05} and key-frame based
approaches~\cite{Klein2007PTAM,Mur-ArtalMT15, Strasdat12}. The features are
often modeled in various forms such as points~\cite{Davison03, Klein2007PTAM},
points and lines~\cite{gomez2017pl}, points and planes~\cite{Taguchi2013may},
or built on more semantic (object) level
features~\cite{Civera11Semantic,Moreno13SLAMp}. However, the choice of the
representation has been application-specific, and thus the performance can vary
depending on a target scene (\ie, amount of lines, planes or texture
present in the scene). To address this issue, recently,
direct~\cite{Stuhmer10, Newcombe11} and semi-direct
methods~\cite{Engel2017DSO, Engel2014LSD} utilize all the pixels with high
gradients rather than features to build maps. While they provide more stable
pose estimate and denser information of a scene, they require a higher
computational expense, and often need an accurate intrinsic calibration and
initialization because they are sensitive to photometric
consistency~\cite{bergmann17calibration}.

For map optimization,  since Lu
and Milios~\cite{Lu97PGO} first introduced a graph-based method to refine a map
with global optimization of local nodes (measurements from odometry),
various types of these local-to-global pose graph optimization methods have been
proposed~\cite{Dellaert06ijrr, grisetti2010tutorial, Thrun02a, Thrun05}.
Similarly, bundle adjustment~\cite{lour09, Triggs99} has also been
a popular choice for methods using structure from motion techniques (\ie,
keyframe-based approaches)~\cite{Klein2007PTAM,Mur-ArtalMT15, Strasdat12}.

In the context of image-based localization, visual place
recognition~\cite{Sattler2011,Sattler2017,Li2012Point} and camera
relocalization~\cite{Mur-ArtalMT15}, image descriptors (\eg, bag-of-words
(BoW) features~\cite{Cummins2008BOW}, VLAD~\cite{VLAD}, Fisher
vectors~\cite{Jegou12}, and recent DNN-based features~\cite{NetVLAD}) are used
to build maps/vocabularies for image retrieval and pose estimation.

Compared to these prior application-specific map definitions, in this paper we
primarily focus on learning a general map representation for sequential camera
localization with deep neural networks, by leveraging statistical learning from
big data and geometric constraints from pose graph optimization and
bundle adjustment.

\vspace{-1em}
\paragraph{DNN-based Camera Localization}

A few recent works use deep neural networks for image-based localization in the
context of structure-from-motion. PoseNet~\cite{Kendall15iccv} first proposed
to directly regress 6-DoF camera pose from an input image with GoogLeNet.
Kendall~\etal~\cite{Kendall16icra, Kendall17cvpr} extend PoseNet by learning
the weight between camera translation and rotation loss and incorporating the 
reprojection loss. Melekhov~\etal~\cite{Melekhov17Hourglass} improved PoseNet
with skip connections with ResNet34 architecture. Brachmann~\etal~\cite{Brachmann17RANSAC}
localize a camera in a dense 3D reconstruction by performing RANSAC on predicted 2D-3D correspondences.
Recently, RNNs (\eg, LSTM) have been introduced to spatially~\cite{Walch17LSTM} and
temporally~\cite{Clark17VidLoc} improve camera localization.

MapNet is inspired by the PoseNet line of work, but has several major
modifications. Table~\ref{tab:comp} shows a comparative summary.  
Clark~\etal~\cite{Clark17VidLoc} used an
LSTM to implicitly learn the temporal relationship between consecutive frames, but
its performance is on-par or worse than prior methods~\cite{Kendall17cvpr,Melekhov17Hourglass}.
In contrast, MapNet uses single images as input during inference but still
uses the geometric constraints between pairs as a meaningful learning signal for training.
MapNet+ and MapNet+PGO use unlabeled videos and multiple sensory input (\eg,
visual odometry, IMU, GPS) to further improve performance. Thus, they can fuse information
from multiple modalities and improve in a self-supervised manner.
In~\cite{Kendall17cvpr} Kendall~\etal make PoseNet \emph{scene}-geometry aware by
minimizing the reprojection error of 3D points in multiple images. In
contrast, MapNet is \emph{camera motion}-geometry aware by utilizing the
geometric constraints between camera poses. 

\section{Proposed Approach}
\label{sec:method}

\begin{figure*}
\centering
    \includegraphics[width=.25\linewidth]{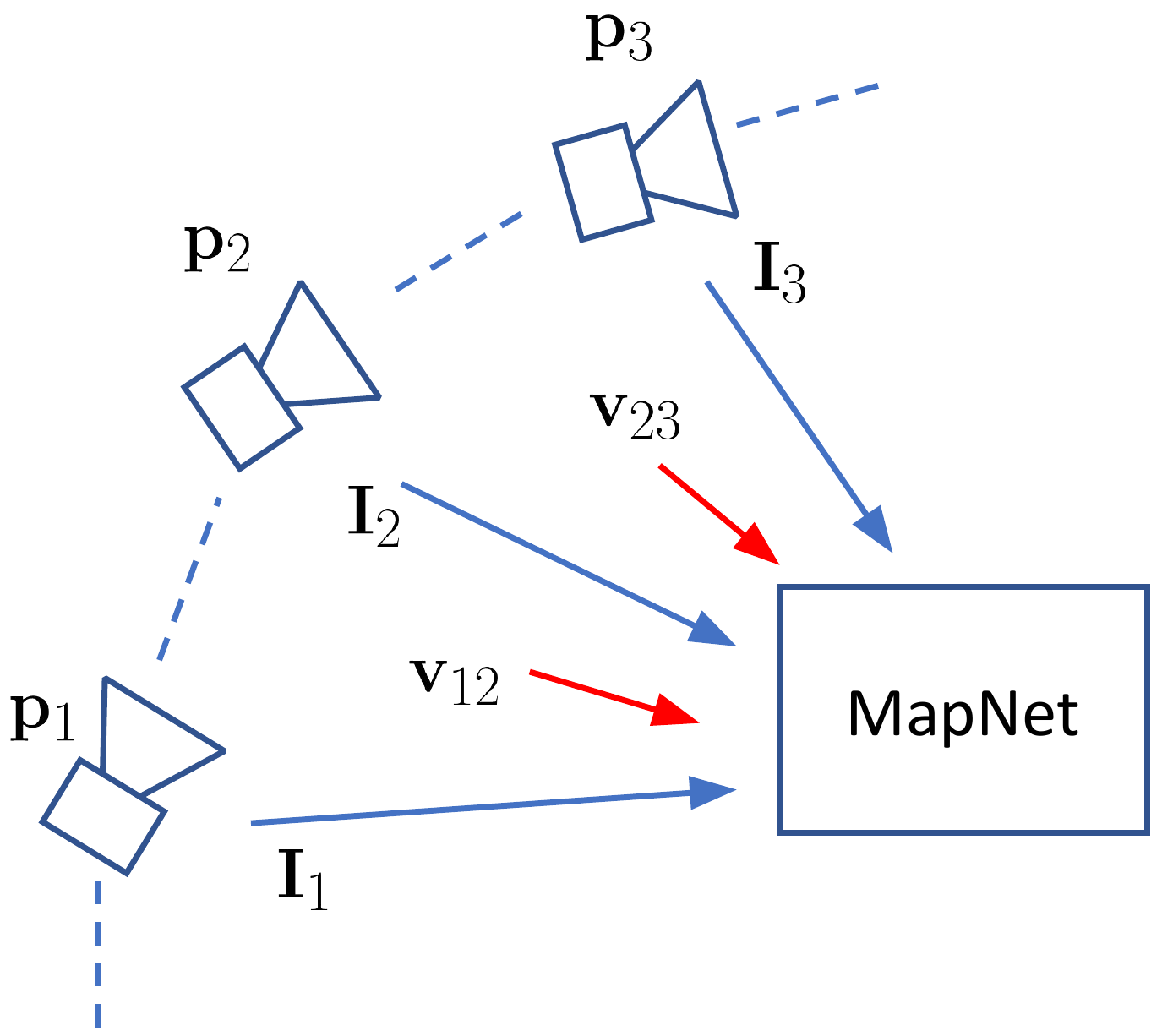}\hfill
    \includegraphics[width=.69\linewidth]{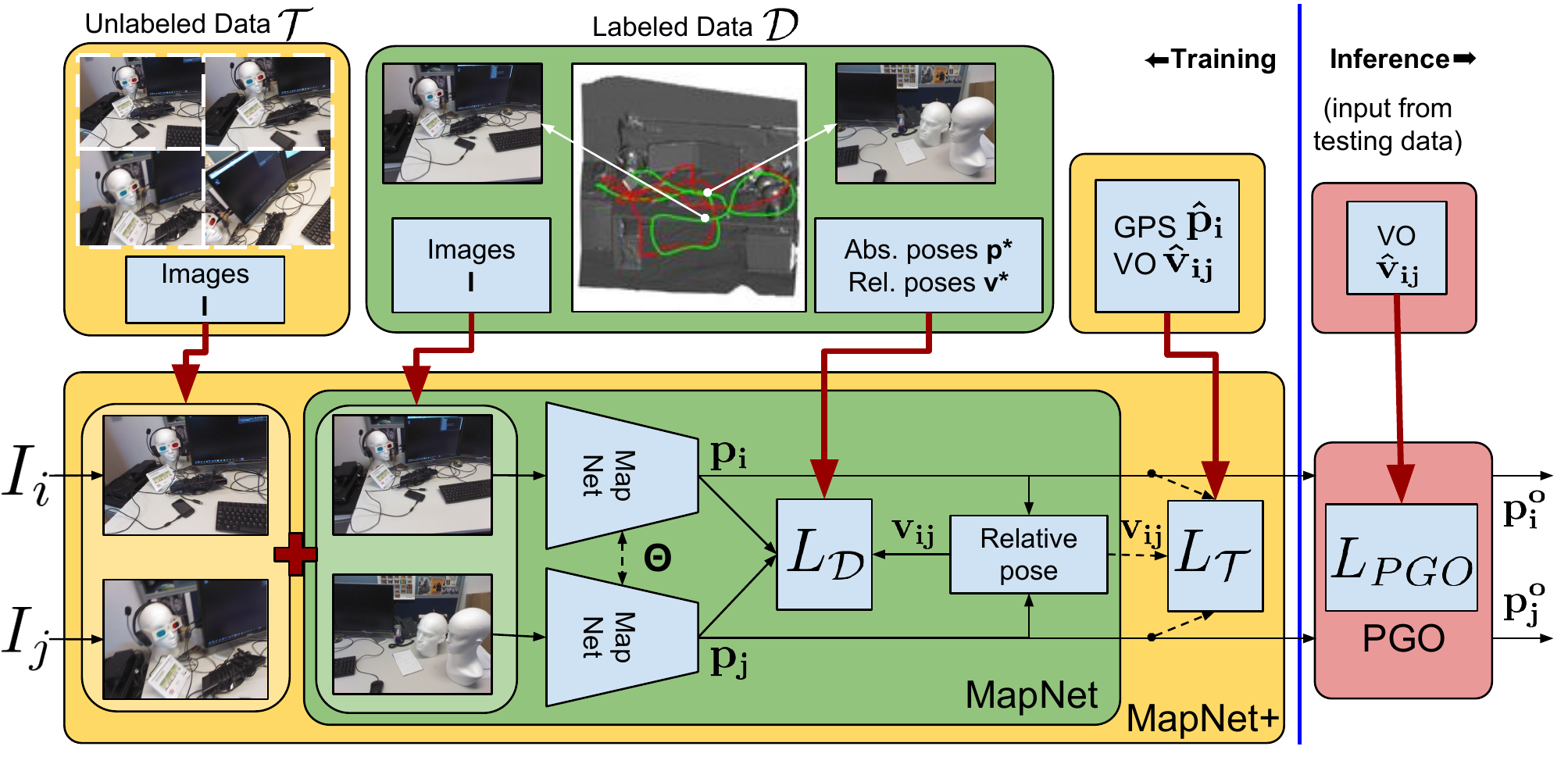}
    \caption{\small \textbf{Left}: MapNet learns a general map representation directly from input data, including images,
    visual odometry (VO), and other sensory inputs. \textbf{Right}: Data flow for our proposed algorithms. MapNet
    enforces geometric constraints between relative poses and absolute poses in network training.
    MapNet+ fuses other inputs such as visual odometry to update maps with self-supervised learning.
    MapNet+PGO performs PGO at testing time to further improve accuracy.}
    \label{fig:network}
\end{figure*}

In this paper, we learn a general map representation for sequential camera
localization with deep neural networks (DNNs).  Maps are represented as learned
weights of a DNN trained to regress camera pose. Figure~\ref{fig:network} shows all of our three proposed models.
At the heart of MapNet is a DNN that regresses absolute camera pose from an input image, which is described in detail in Section~\ref{subsec:regress}.
MapNet takes in tuples of images and additionally enforces constraints between pose predictions for pairs, as described in
Section~\ref{subsec:mapnet}. MapNet+ improves a trained MapNet by utilizing the geometric constraints expressed by visual odometry (VO) on additional
unlabeled videos from the same scene, or synchronized GPS readings (Section~\ref{subsec:mapnet+}). Finally, we employ moving-window PGO during inference to
obtain a smooth and drift free camera trajectory by fusing MapNet+ absolute pose predictions and VO (Section~\ref{subsec:mapnet+p}).

\subsection{Camera Pose Regression with DNNs}
\label{subsec:regress}

Our work is built upon prior works in DNN-based pose estimation
methods~\cite{Clark17VidLoc, Kendall16icra, Kendall17cvpr,Kendall15iccv,
Melekhov17Hourglass, Walch17LSTM}, which regress 6-DoF camera pose from an input
RGB image with a DNN. In our work, we made several modifications to
PoseNet~\cite{Kendall17cvpr,Kendall15iccv}.  First, we use 
ResNet-34~\cite{He16ResNet} and modify it by introducing a global average
pooling layer after the last conv layer, followed by a fc layer with 2048
neurons, a ReLU and dropout with $p=0.5$.  This is followed by a final fc
layer that outputs a 6-DoF camera pose.



Second, we propose to parameterize camera orientation as the logarithm of a unit quaternion~\cite{quatbook}, 
which is better suited for regression with deep learning.  PoseNet and its
variants~\cite{Clark17VidLoc,Kendall17cvpr, Kendall15iccv, Melekhov17Hourglass, Walch17LSTM} used 4-d unit quaternions to represent orientation, and regress
it with $l_1$ or $l_2$ norm.  This has two issues: (1) the quadruple 
is an over parameterization of the 3-DoF rotation, and (2)
normalization of the output quadruple is required but often 
results in worse performance~\cite{Kendall15iccv,Kendall17cvpr,Melekhov17Hourglass}.
While Euler
angles used in~\cite{Su15} are not over-parameterized, they are not suited for
regression since they wrap around $2\pi$.


The logarithm of a unit quaternion, $\log\mathbf{q}$ has 3 dimensions
and is not over-parameterized. This allows us to directly use the $l_1$ or $l_2$
distance as the loss function without normalization.
The logarithm of a unit quaternion $\mathbf{q} = (u, \mathbf{v})$, where $u$ is a scalar
and $\mathbf{v}$ is a 3-d vector, is defined as~\cite{dam1998quaternions, quaternion_math}
\begin{equation}
    \log\mathbf{q}= \begin{cases}
            \frac{\mathbf{v}}{\Vert\mathbf{v}\Vert}\cos^{-1}u,& \text{if } \Vert\mathbf{v}\Vert \neq 0\\
            \mathbf{0},& \text{otherwise}
        \end{cases}
\end{equation}
The logarithmic form $\mathbf{w}=\log\mathbf{q}$ can be converted back
to a unit quaternion by the formula $\exp\mathbf{w} = (\cos
\Vert\mathbf{w}\Vert, \frac{\mathbf{w}}{\Vert\mathbf{w}\Vert}\sin
\Vert\mathbf{w}\Vert)$.
As shown in Table~\ref{tab:result_logq} in Section~\ref{sec:results},
using this rotation parameteration achieve better results than PoseNet~\cite{Kendall17cvpr}.
We also implemented other metrics for rotation~\cite{Huynh09}, and found they did not
improve the performance.

\subsection{MapNet: Geometry-Aware Learning}
\label{subsec:mapnet}

Similar to PoseNet~\cite{Clark17VidLoc, Kendall17cvpr, Kendall15iccv, Melekhov17Hourglass, Walch17LSTM},
MapNet also learns a DNN $\Theta$ that
estimates the 6-DoF camera pose $\mathbf{p} = (\mathbf{t}, \mathbf{w})$
from an input RGB image $\mathbf{I}$ on the training set
$\mathcal{D}=\{(\mathbf{I}, \mathbf{p}^*)\}$ via supervised learning, $f(\mathbf{I};\Theta)=\mathbf{p}$.
The main difference, however, is that MapNet minimizes both the loss of the per-image absolute
pose and the loss of the relative pose between image pairs, as shown in Fig.~\ref{fig:network},
\begin{equation}\label{eq:mapnet}
    L_{\mathcal{D}}(\Theta) = \sum_{i=1}^{\vert\mathcal{D}\vert} h(\mathbf{p}_i, \mathbf{p}_i^*) + \alpha
    \sum_{i,j=1,i\neq j}^{\vert\mathcal{D}\vert} h(\mathbf{v}_{ij}, \mathbf{v}_{ij}^*),
\end{equation}
where $\mathbf{v}_{ij} = (\mathbf{t}_i\!-\!\mathbf{t}_j, \mathbf{w}_i\!-\!\mathbf{w}_j)$ is the relative camera pose between
pose predictions $\mathbf{p}_i$ and $\mathbf{p}_j$ for images $\mathbf{I}_i$ and $\mathbf{I}_j$. 
$h(\cdot)$ is a function to measure the distance between the predicted camera pose $\mathbf{p}$ and the ground truth
camera pose $\mathbf{p^*}$, defined as~\cite{Kendall17cvpr}:
\begin{equation}\label{eq:pose}
h(\mathbf{p}, \mathbf{p^*}) = \Vert \mathbf{t}-\mathbf{t}^* \Vert_{1} e^{-\beta} + \beta +
    \Vert \mathbf{w}-\mathbf{w}^* \Vert_{1} e^{-\gamma} + \gamma,
\end{equation}
where $\beta$ and $\gamma$ are the weights that balance the translation
loss and rotation loss. Both $\beta$ and $\gamma$ are learned during training with
initialization $\beta_0$ and $\gamma_0$.
$(\mathbf{I}_i,\mathbf{I}_j)$ are image pairs within each tuple of $s$ images 
sampled with a gap of $k$ frames from $\mathcal{D}$.  
Intuitively, adding the second loss of the relative camera poses between image pairs
helps to enforce global consistency, which improves the performance of camera localization (see Section.~\ref{sec:results}).

\begin{figure}
    \captionsetup[subfigure]{labelformat=empty}
    \centering
    \begin{subfigure}{0.32\linewidth}
        \centering
        \includegraphics[width=\linewidth]{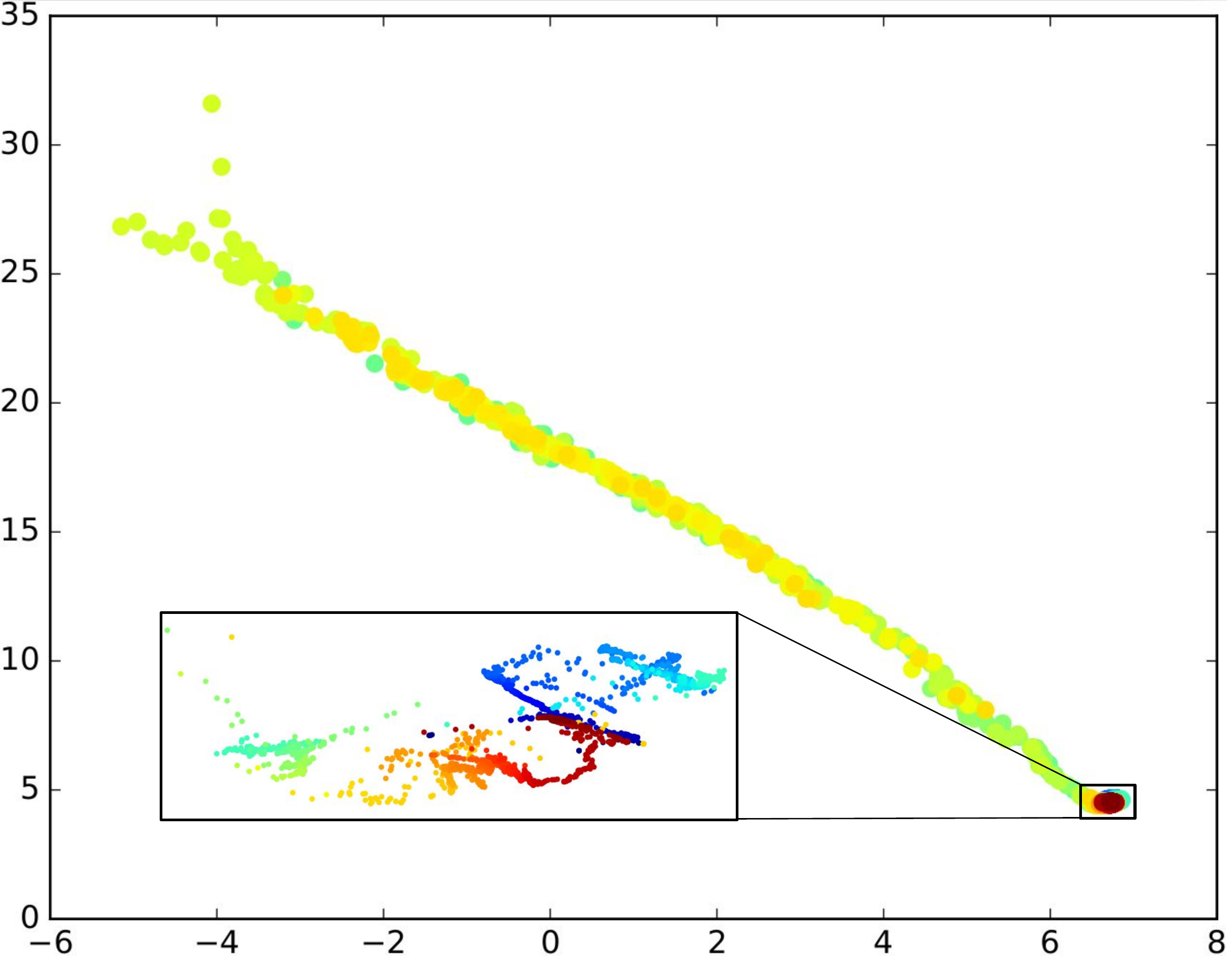}
        \vspace{-1.5em}
        \caption{PoseNet}
    \end{subfigure}
    \hfill
    \begin{subfigure}{0.32\linewidth}
        \centering
        \includegraphics[width=\linewidth]{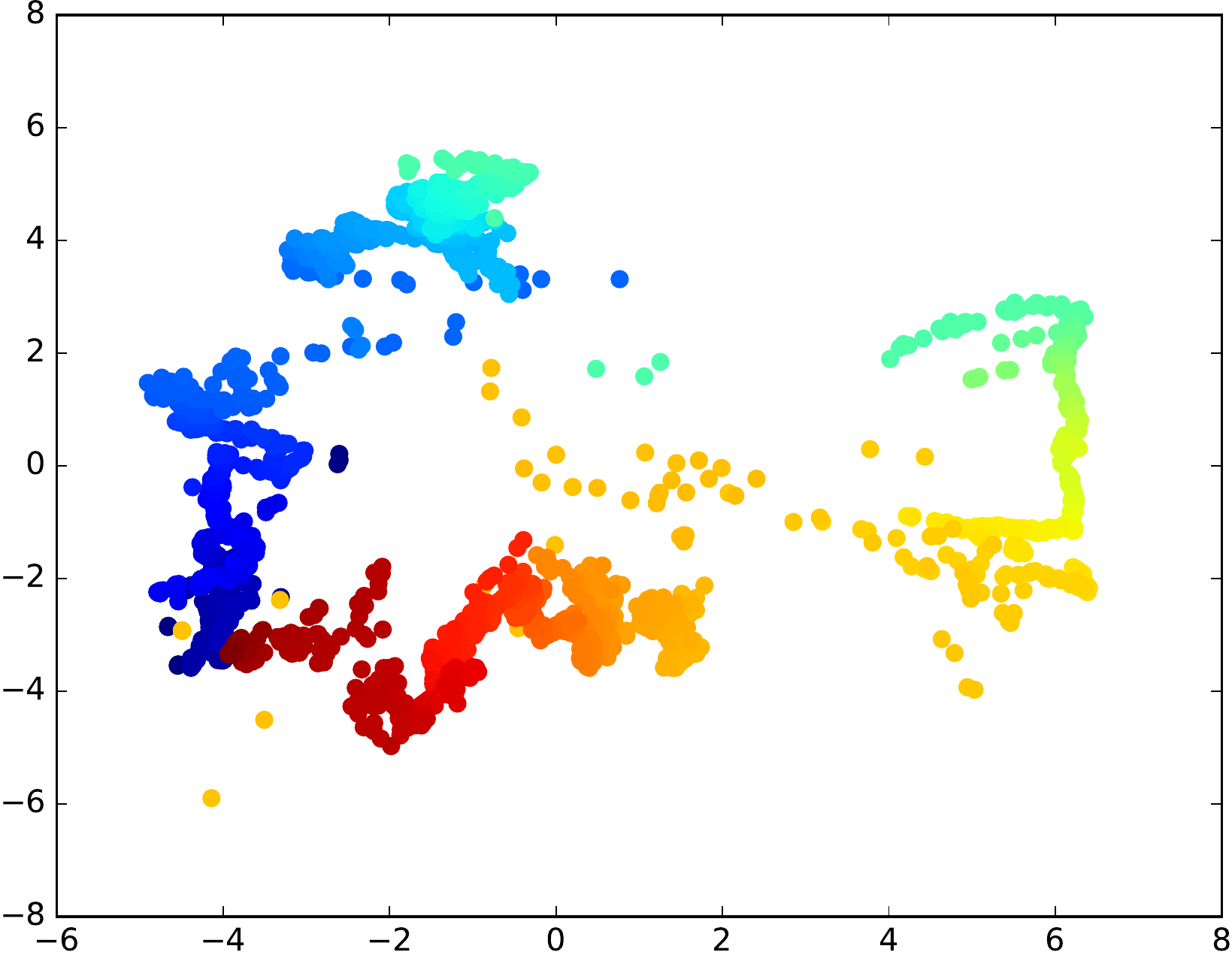}
        \vspace{-1.5em}
        \caption{MapNet}
    \end{subfigure}
    \begin{subfigure}{0.32\linewidth}
        \centering
        \includegraphics[width=\linewidth]{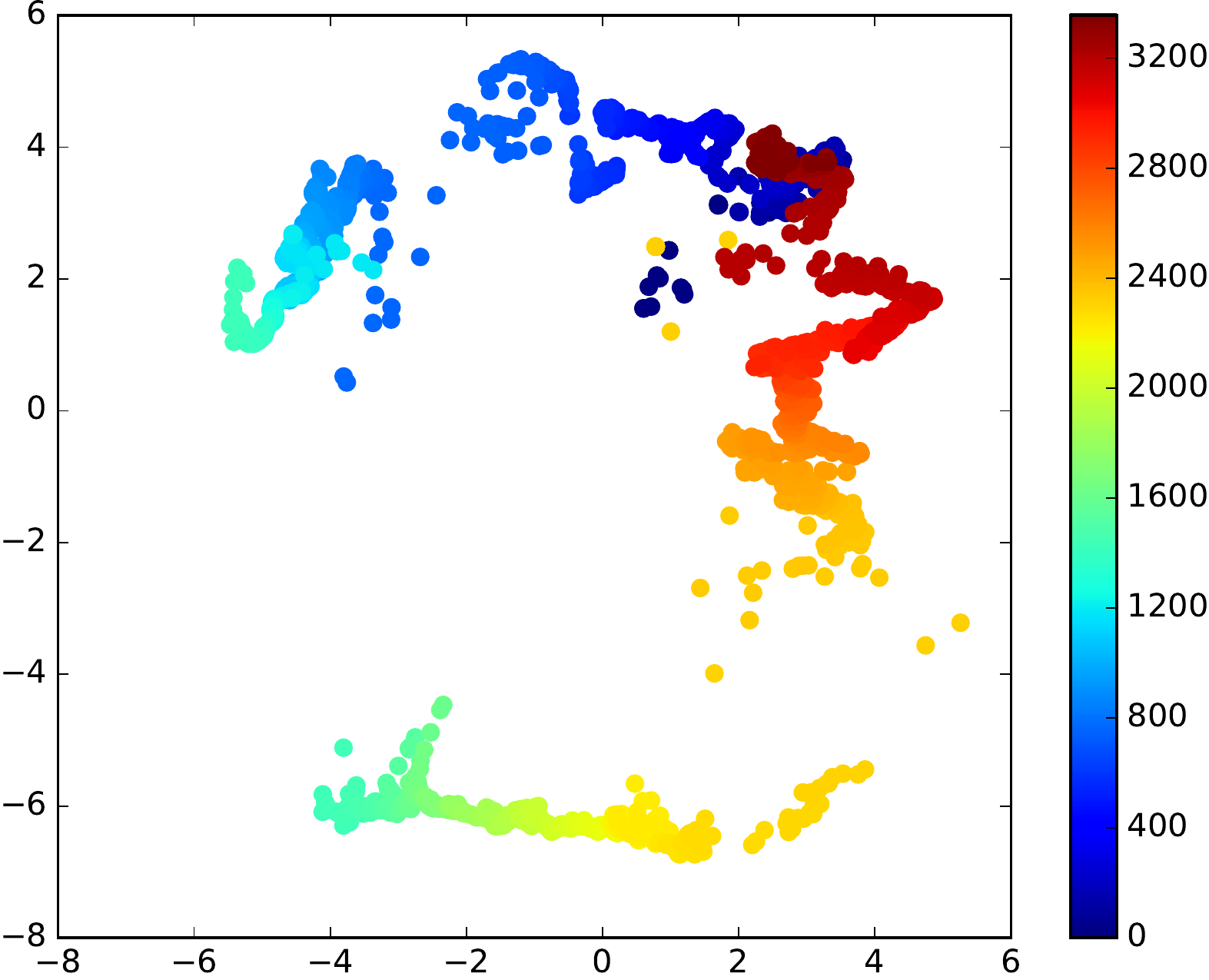}
        \vspace{-1.5em}
        \caption{MapNet+(2seq)}
    \end{subfigure}
    \vspace{-1em}
    \caption{\small 2D multi-dimensional scaling (MDS) of penultimate layer features for various
    models trained on the LOOP sequence from Oxford RobotCar~\cite{RobotCarDatasetIJRR}. Input images
    are from a held-out test sequence (see Fig.~\ref{fig:teaser} for ground-truth camera poses). The points
    are chronologically colored. Features learned by PoseNet do not correlate with the distribution of ground truth camera poses,
    while those learned by MapNet and MapNet+ show successively better correlation with the ground truth
    camera poses (see Fig.~\ref{fig:map_compare_robotcar}).}
    \label{fig:map_mds}
    \vspace{-1em}
\end{figure}


To understand more about the representation MapNet learns, we visualize
the distribution of the feature vectors of the last activation layer using 2D multi-dimensional
scaling (MDS)~\cite{Borg05mds}. We chose MDS rather than T-SNE
because it is designed to preserve global structure of the feature space.
In Fig.~\ref{fig:map_mds}, we show that PoseNet~\cite{Kendall17cvpr} feature vectors for test images
captured along a loop in the Oxford RobotCar dataset~\cite{RobotCarDatasetIJRR} do not correlate with the distribution of ground truth camera poses,
while the features learned by MapNet and MapNet+ (next subsection) 
show successively better correlation. All models use the same network architecture.

\subsection{MapNet+: Update with Unlabeled Data}
\label{subsec:mapnet+}

Both PoseNet and MapNet require labelled data (\ie, images with absolute camera
poses) to train. In many real applications, we may also have lots of unlabeled
data, \eg, videos captured at different times or camera motions in the same scene. Off-the-shelf VO 
algorithms~\cite{Engel2017DSO,Engel2013SVO} provide relative camera poses between image pairs from these videos. 
Other sensors (\eg, IMU and GPS) can also provide measurements about camera pose, especially for
challenging conditions (\eg, textureless, low-light). MapNet+ fuses these additional data $\mathcal{T}$ to update 
the weights of MapNet with self-supervised learning.

Suppose the additional data are some videos of the same
scene, $\mathcal{T}=\{\mathbf{I}_t\}$.
We can compute the relative poses $\mathbf{\hat{v}}_{ij}$ between consecutive
frames with visual odometry algorithms~\cite{Engel2017DSO,Engel2013SVO,Nister04visualodometry}.
In order to update the map with $\mathcal{T}$, we fine-tune a pre-trained MapNet $\Theta$ by minimizing
a loss function that consists of the original loss from the
labelled dataset $\mathcal{D}$ and the loss from the unlabeled data
$\mathcal{T}$, 
\begin{equation}
L(\Theta) = L_{\mathcal{D}}(\Theta) + L_{\mathcal{T}}(\Theta),
\end{equation}
where $L_{\mathcal{T}}(\Theta)$ is the distance between
the relative camera pose $\mathbf{v}_{ij}$ (from predictions $\mathbf{p}_i$, $\mathbf{p}_j$) and visual odometry $\mathbf{\hat{v}}_{ij}$,
\begin{equation}\label{eq:mapnet+}
    L_{\mathcal{T}}(\Theta) = \sum_{i,j=1,i\neq j}^{\vert\mathcal{T}\vert} h(\mathbf{v}_{ij}, \mathbf{\hat{v}}_{ij})
\end{equation}
Since VO algorithms compute $\mathbf{\hat{v}}_{ij}$ in the coordinate system of camera $i$,
the relative pose $\mathbf{v}_{ij}$ is also computed in that coordinate system:
\begin{align}\label{eq:vo}
    \mathbf{v}_{ij} = (&\exp(\mathbf{w}_j)(\mathbf{t}_i-\mathbf{t}_j)\exp(\mathbf{w}_j)^{-1},\nonumber\\
    &\log(\exp(\mathbf{w}_j)^{-1}\exp(\mathbf{w}_i)),
\end{align}
Note that we still keep the supervision loss
$L_{\mathcal{D}}(\Theta)$ from $\mathcal{D}$ --- this is important to avoid
trivial solutions if we optimize only the self-supervised loss
$L_{\mathcal{T}}(\Theta)$ from $\mathcal{T}$.  Thus each mini-batch samples half from
the labelled data $\mathcal{D}$ and half from the unlabeled data $\mathcal{T}$.
The image pairs $(\mathbf{I}_i,\mathbf{I}_j)$ are sampled similarly from tuples of $s$ images
with a gap of $k$ frames from both $\mathcal{D}$ and $\mathcal{T}$.

Intuitively, MapNet+ exploits the complimentary characteristics of VO and
DNN-based pose prediction --- VO is locally accurate but often drifts over
time, and DNN-based pose predictions are noisy but drift-free.
For other sensors such as IMU (which measures relative rotation)
and GPS (which measures 3D locations), we can define similar loss terms
$L_{\mathcal{T}}(\Theta)$ that minimize the difference between such measurements and the predictions from 
the MapNet.

\subsection{MapNet+PGO: Optimizing During Inference}
\label{subsec:mapnet+p}

During inference, MapNet+PGO fuses the absolute pose predictions from MapNet+ and the relative poses from VO using
pose graph optimization (PGO)~\cite{Carlone16PGO,Lu97PGO,Duckett02PGO} to get smooth and globally consistent pose 
predictions.
It runs in a moving-window of $T$ frames. Suppose the initial poses predicted by MapNet+
are $\{\mathbf{p}_i\}_{i=1}^T$, and the relative poses between two frames from VO are $\{\mathbf{\hat{v}}_{ij}\}$
where $i,j\in [1,T], i\neq j$.
MapNet+PGO solves for the optimal poses
$\{\mathbf{p}^o_i\}_{i=1}^T$ by minimizing the following cost: 
\begin{equation}
    L_{PGO}(\{\mathbf{p}^o_i\}_{i=1}^T) = \sum_{i=1}^{T} \bar{h}(\mathbf{p}^o_i, \mathbf{p}_i)
    +\sum_{i,j=1,i\neq j}^{T}\bar{h}(\mathbf{v}^o_{ij}, \mathbf{\hat{v}}_{ij}),
\end{equation}
where $\bar{h}(\cdot)$ is the standard pose distance function used in PGO literature~\cite{grisetti2010tutorial}.
PGO is an iterative algorithm where internally $\mathbf{v}^o_{ij}$ is derived from $\mathbf{p}^o_i$ and $\mathbf{p}^o_j$ as in Equation~(\ref{eq:vo}).
Details and derivation are included in the supplementary material.
Note here we fix the DNN weights $\Theta$ and only optimize $\{\mathbf{p}^o_i\}_{i=1}^T$.
As shown in Section~\ref{sec:results},
MapNet+PGO further improves the accuracy of pose estimation, with a minimal
extra computational cost at testing.


\subsection{Implementation Details}

We implemented our algorithms with PyTorch~\cite{pytorch}, using the Adam optimizer~\cite{kingma14adam} with a
learning rate of 1e-4 and a weight decay of 5e-4. The input images are scaled
to $341\times 256$ pixels, and normalized by pixel mean subtraction and
standard deviation division.  We set the weight
coefficient $\alpha=1$ and initializations $\beta_0=0.0$ and
$\gamma_0=-3.0$. Image pairs are sampled from tuples of size $s=3$ with spacing $k=10$ frames for MapNet and MapNet+,
and $T=7$, $k=150$ frames for PGO.\footnote{PGO for RobotCar sequences uses frame separation $T=7$, $k=10$.}
All models are trained for 300 epochs, except for MapNet+, which is finetuned with $\alpha=0$ for 5 epochs from a trained MapNet.


\section{Experimental Evaluations}
\label{sec:results}

\begin{figure*}
    \captionsetup[subfigure]{labelformat=empty}
    \centering

    \begin{subfigure}{0.19\linewidth}
        \centering
        \includegraphics[width=\linewidth]{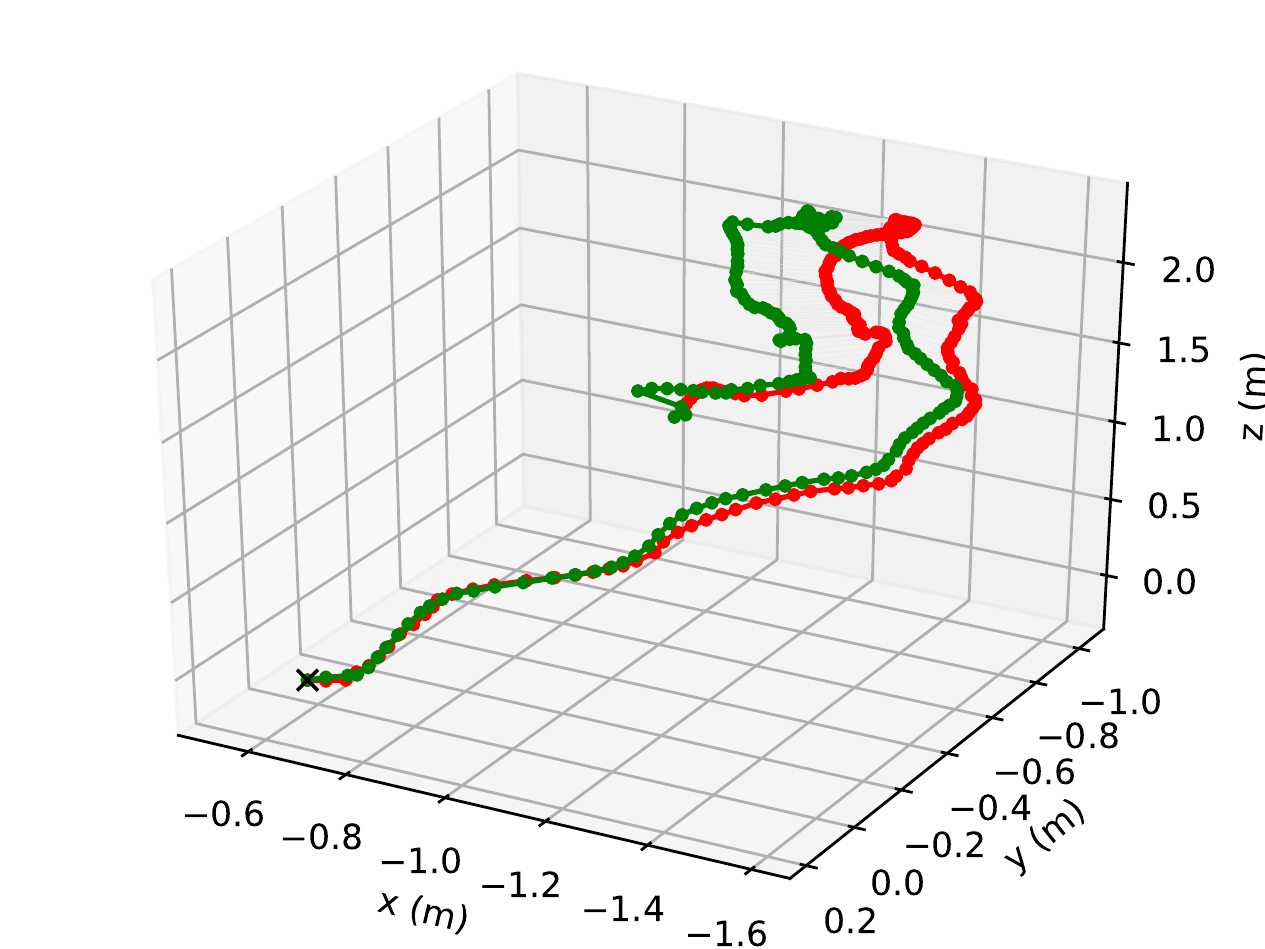}
        \includegraphics[width=\linewidth]{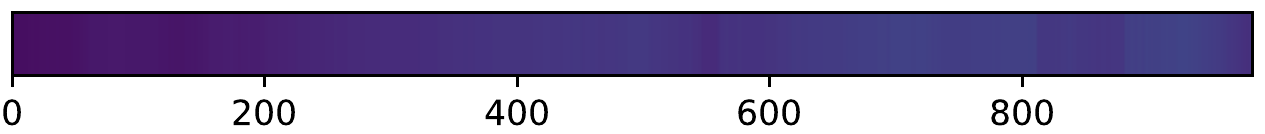}
    \end{subfigure}
    \hfill
    \begin{subfigure}{0.19\linewidth}
        \centering
        \includegraphics[width=\linewidth]{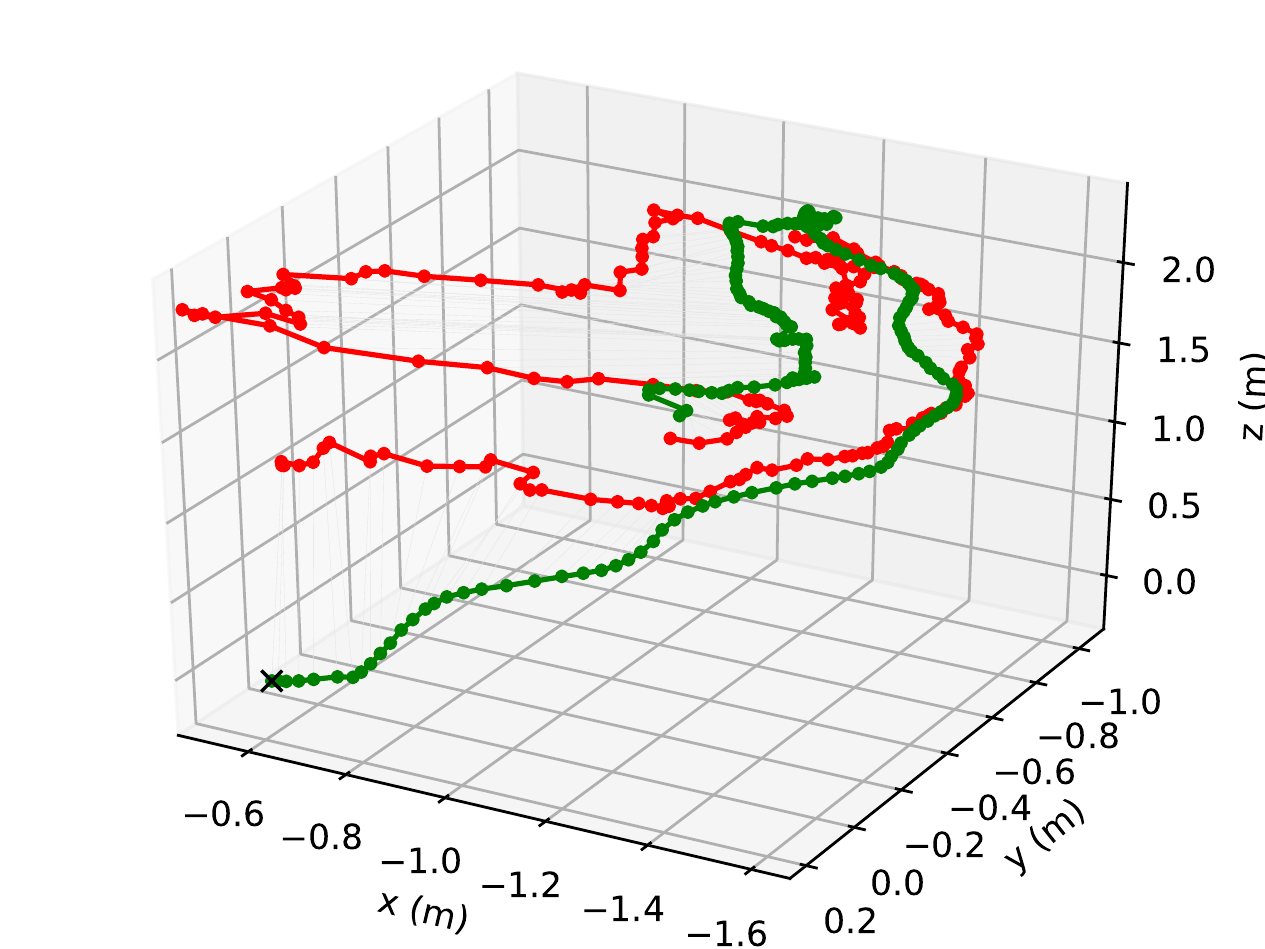}
        \includegraphics[width=\linewidth]{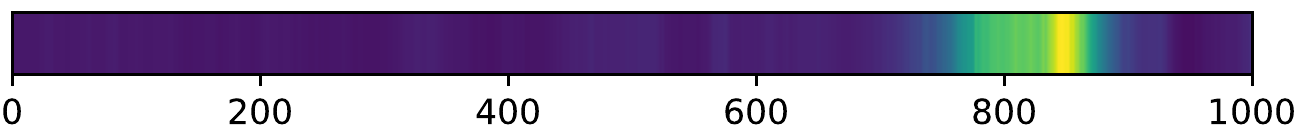}
    \end{subfigure}
    \hfill
    \begin{subfigure}{0.19\linewidth}
        \centering
        \includegraphics[width=\linewidth]{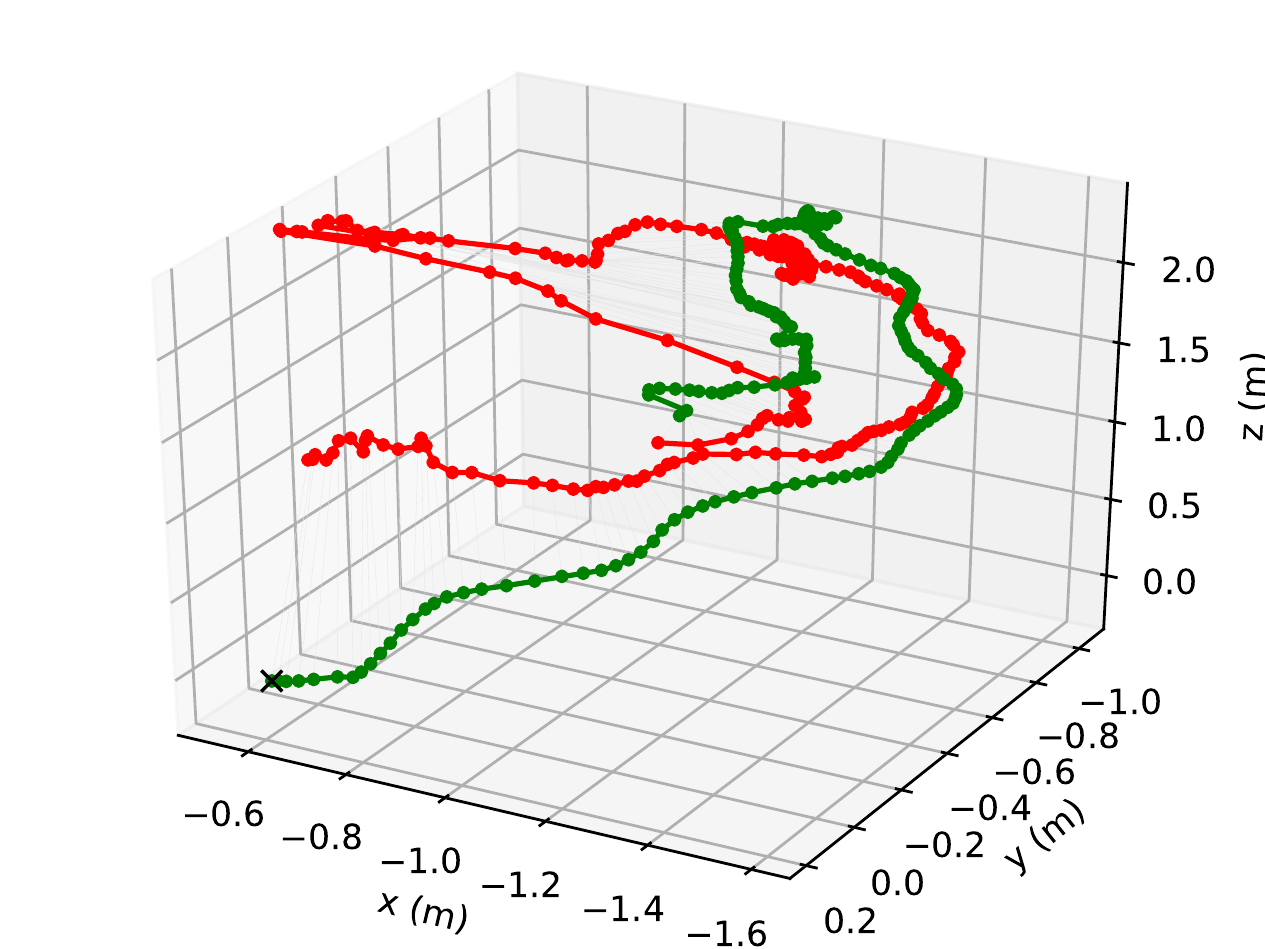}
        \includegraphics[width=\linewidth]{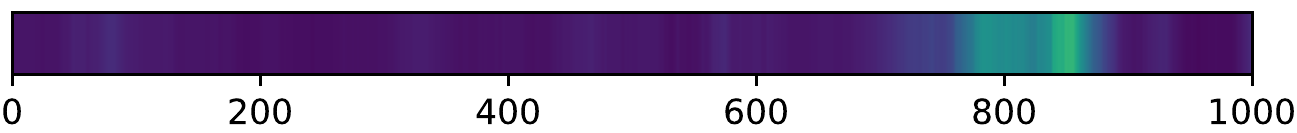}
    \end{subfigure}
    \hfill
    \begin{subfigure}{0.19\linewidth}
        \centering
        \includegraphics[width=\linewidth]{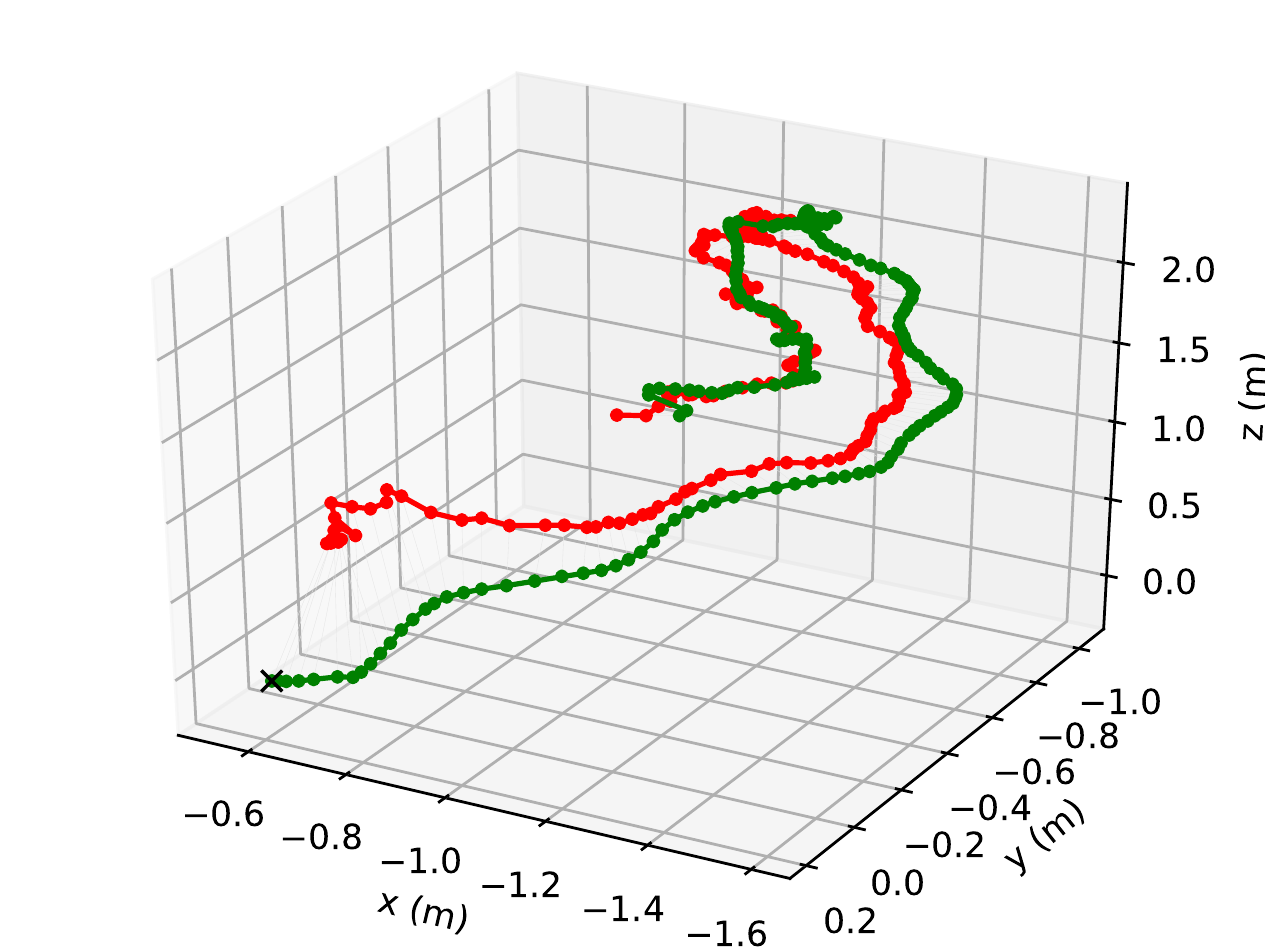}
        \includegraphics[width=\linewidth]{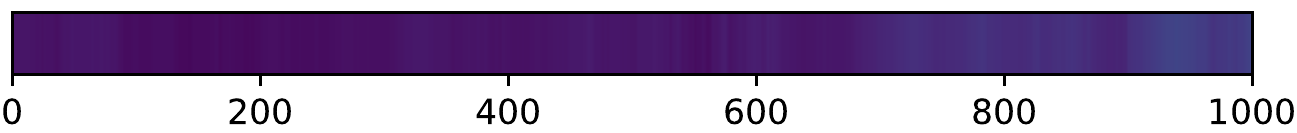}
    \end{subfigure}
    \hfill
    \begin{subfigure}{0.19\linewidth}
        \centering
        \includegraphics[width=\linewidth]{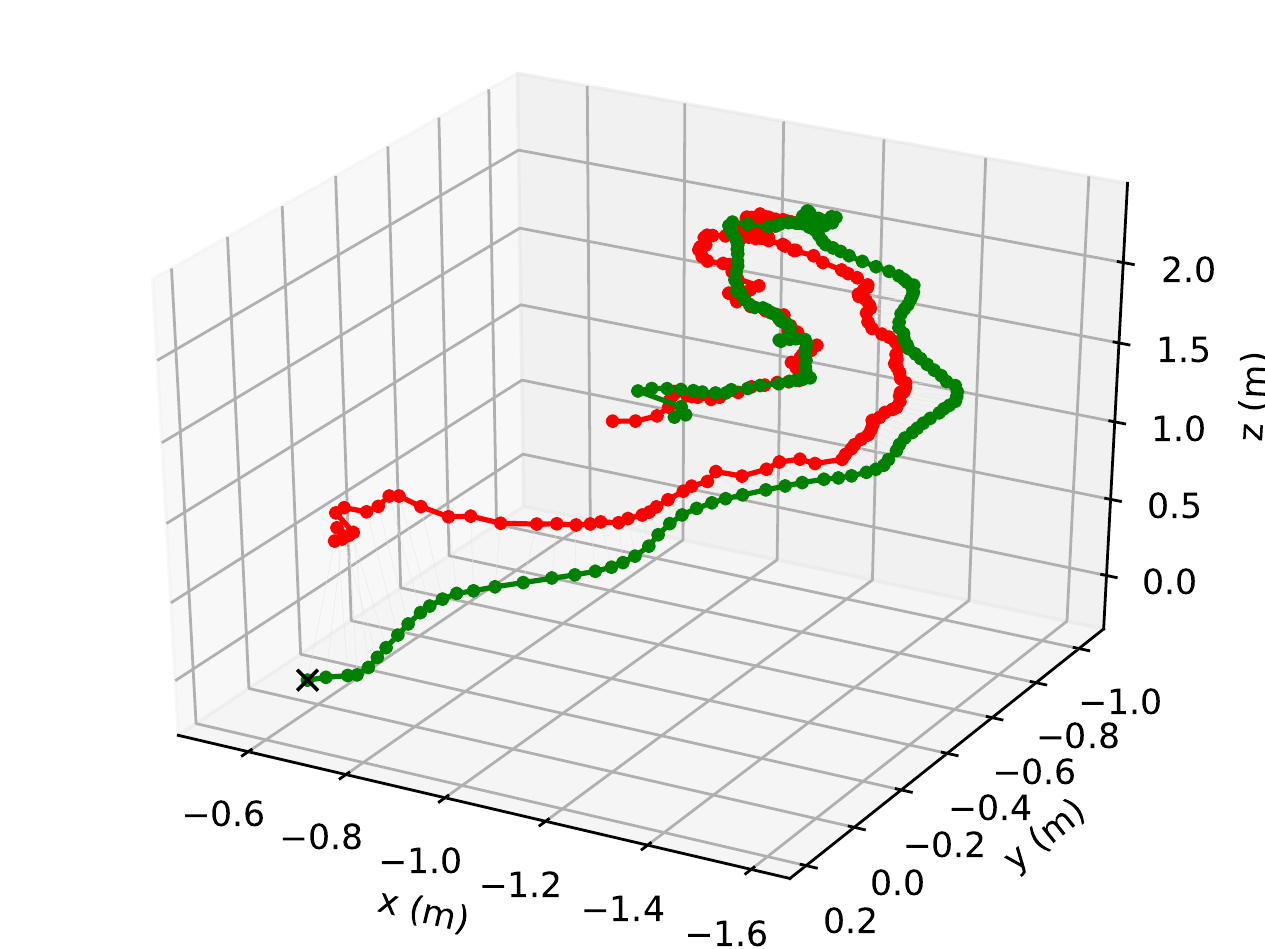}
        \includegraphics[width=\linewidth]{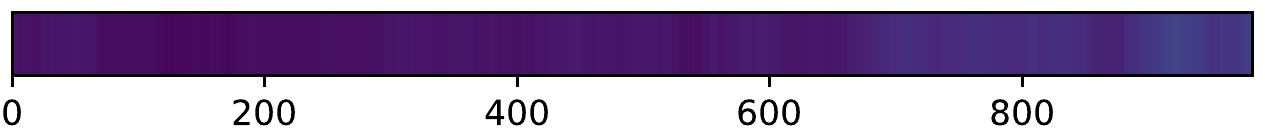}
    \end{subfigure}

    \begin{subfigure}{0.19\linewidth}
        \centering
        \includegraphics[width=\linewidth]{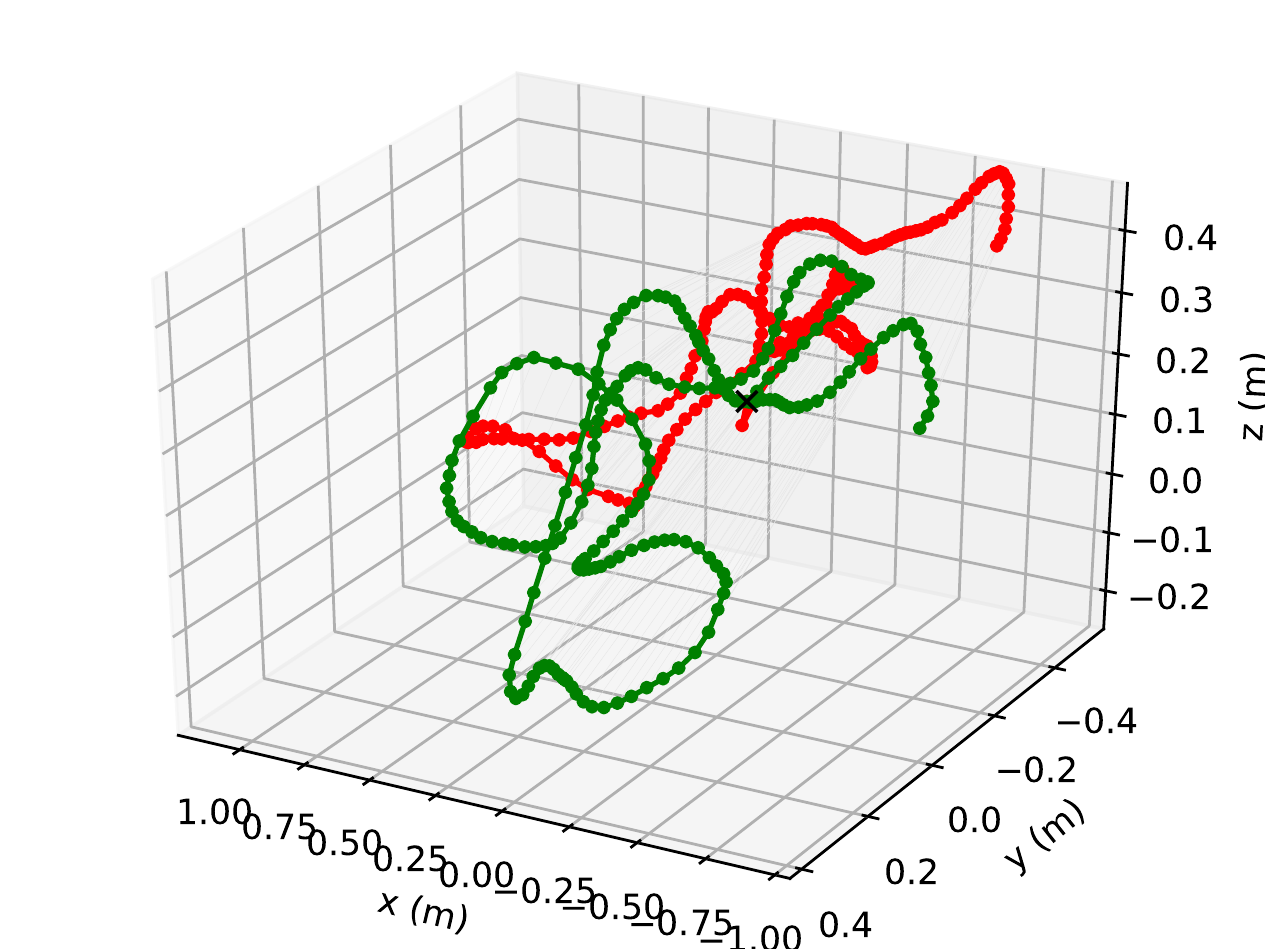}
        \includegraphics[width=\linewidth]{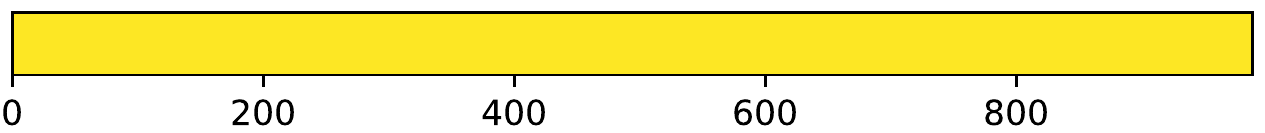}
    \end{subfigure}
    \hfill
    \begin{subfigure}{0.19\linewidth}
        \centering
        \includegraphics[width=\linewidth]{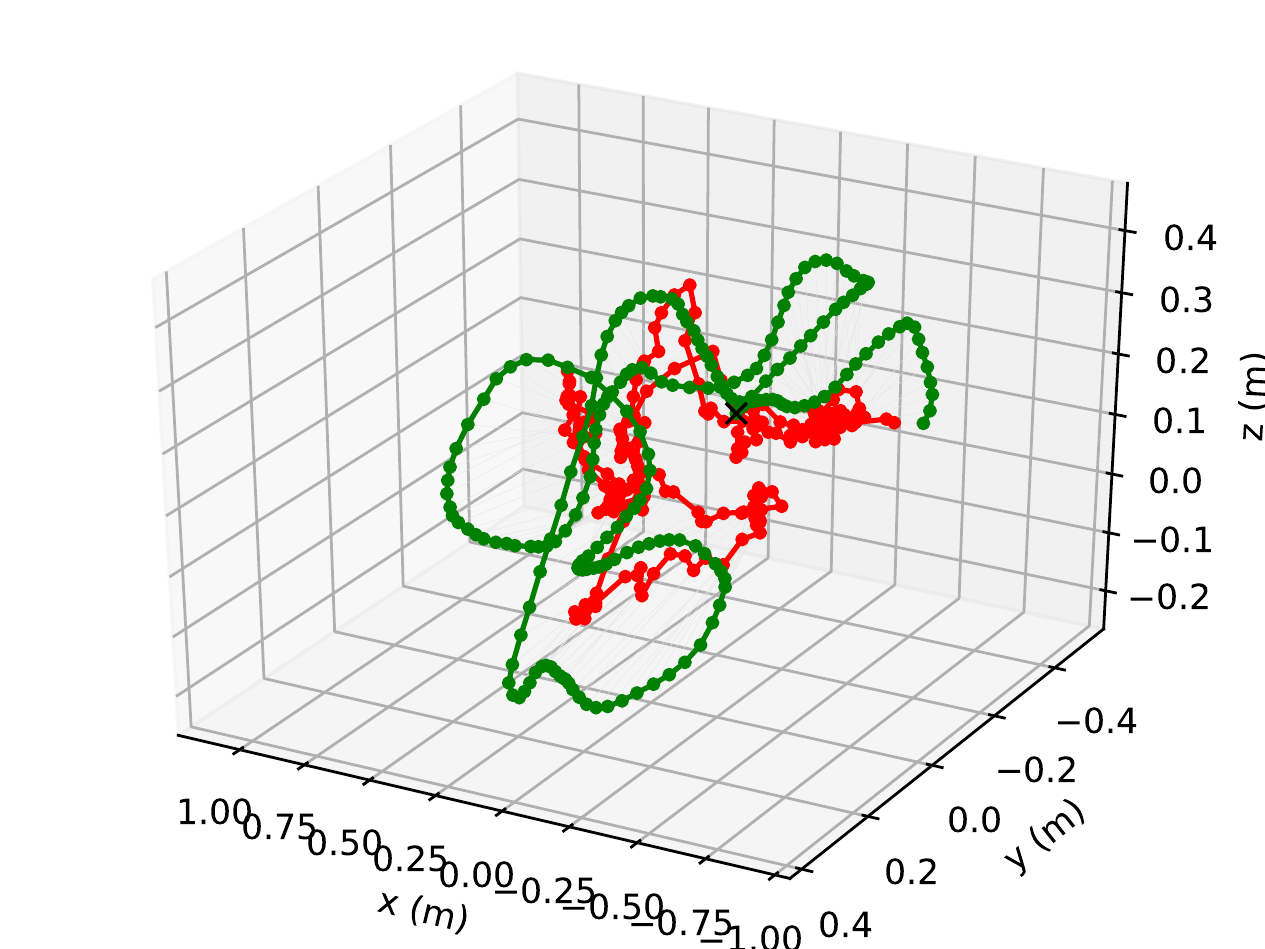}
        \includegraphics[width=\linewidth]{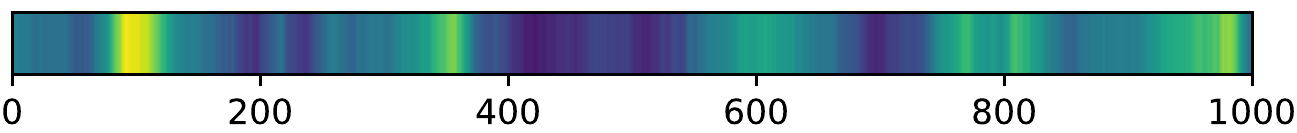}
    \end{subfigure}
    \hfill
    \begin{subfigure}{0.19\linewidth}
        \centering
        \includegraphics[width=\linewidth]{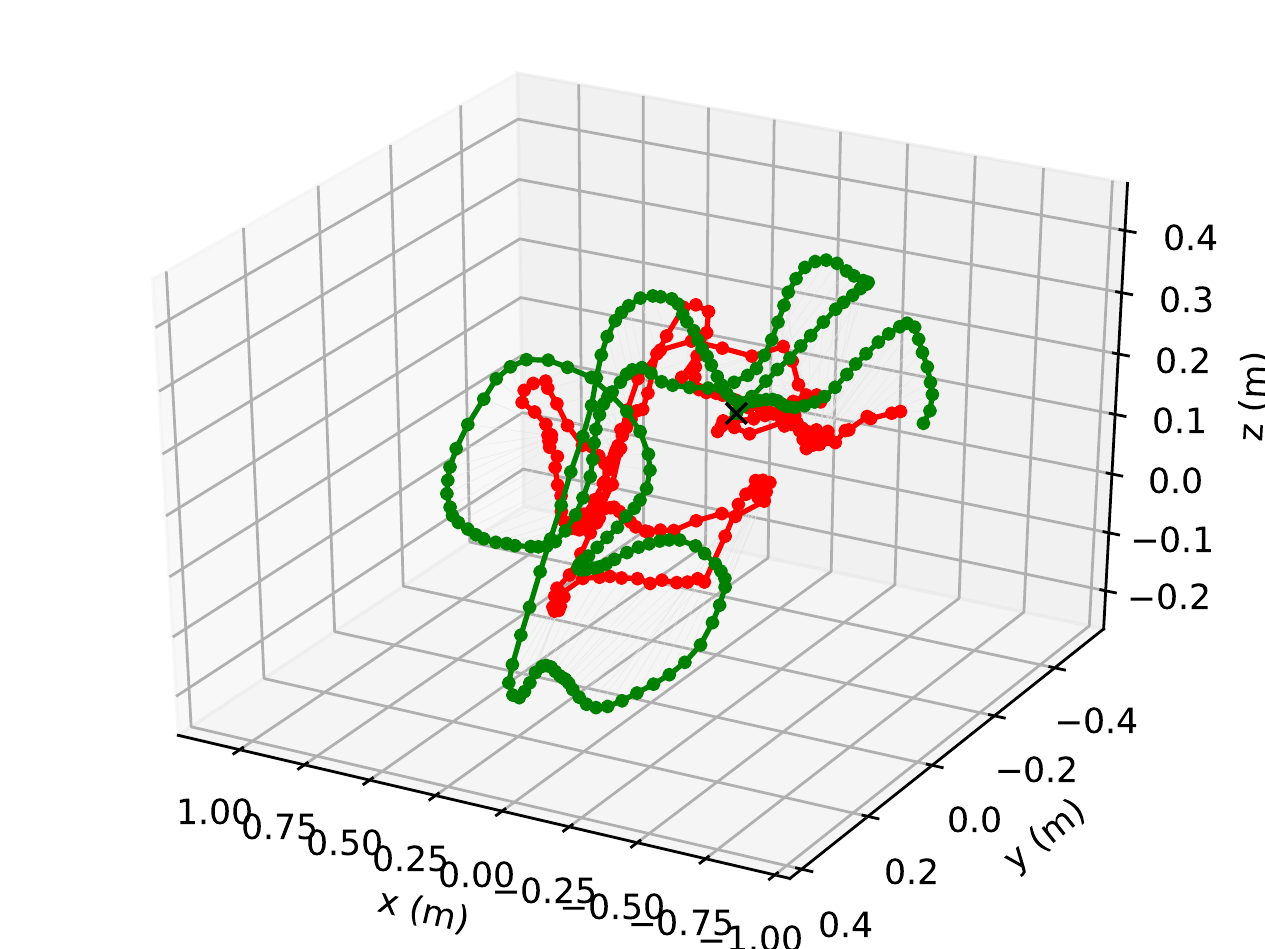}
        \includegraphics[width=\linewidth]{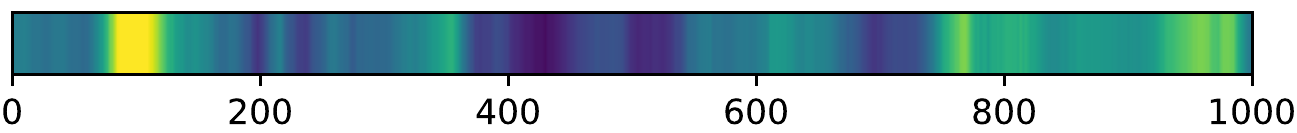}
    \end{subfigure}
    \hfill
    \begin{subfigure}{0.19\linewidth}
        \centering
        \includegraphics[width=\linewidth]{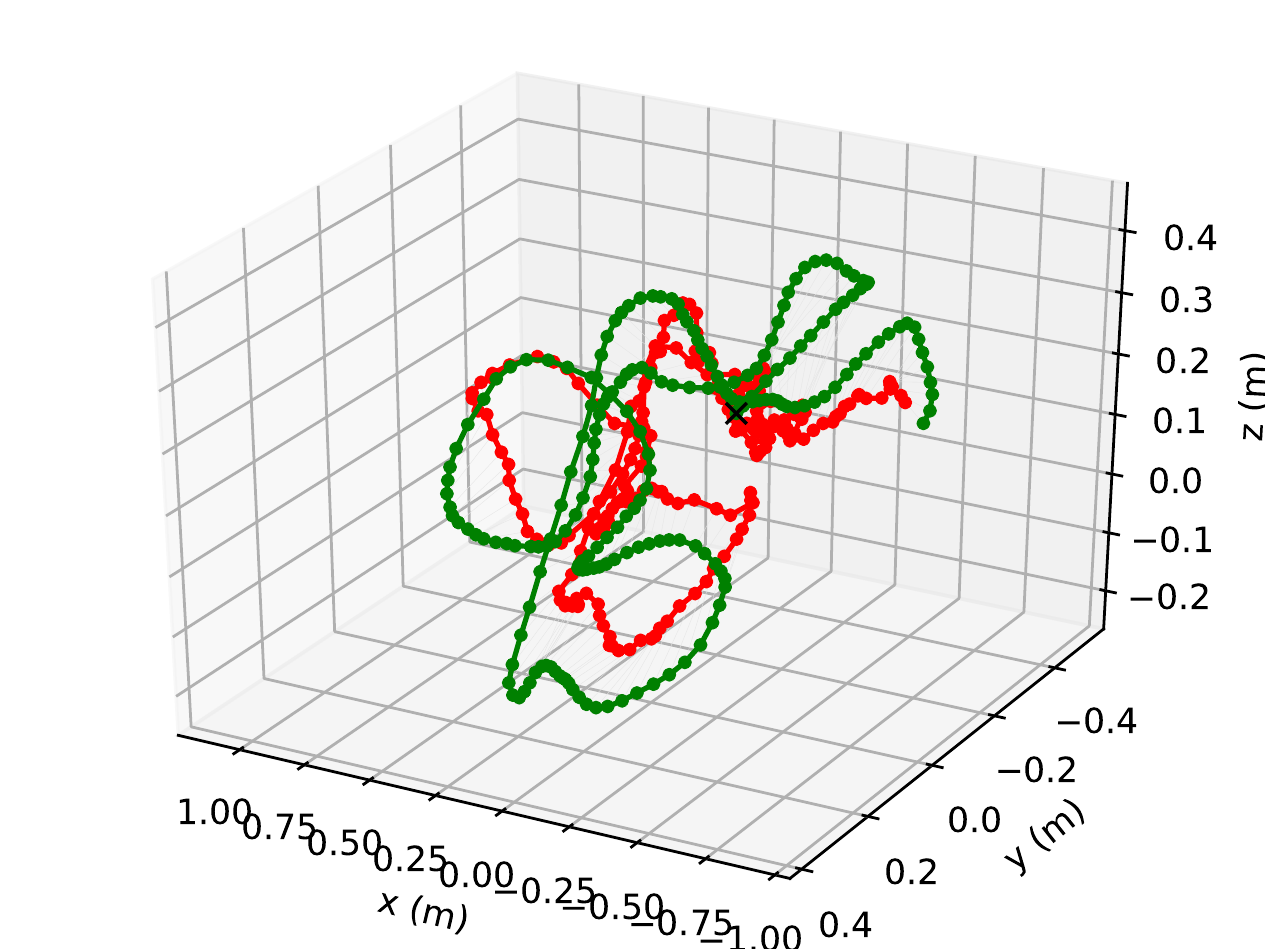}
        \includegraphics[width=\linewidth]{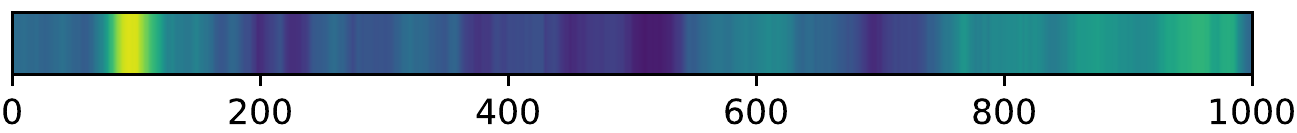}
    \end{subfigure}
    \hfill
    \begin{subfigure}{0.19\linewidth}
        \centering
        \includegraphics[width=\linewidth]{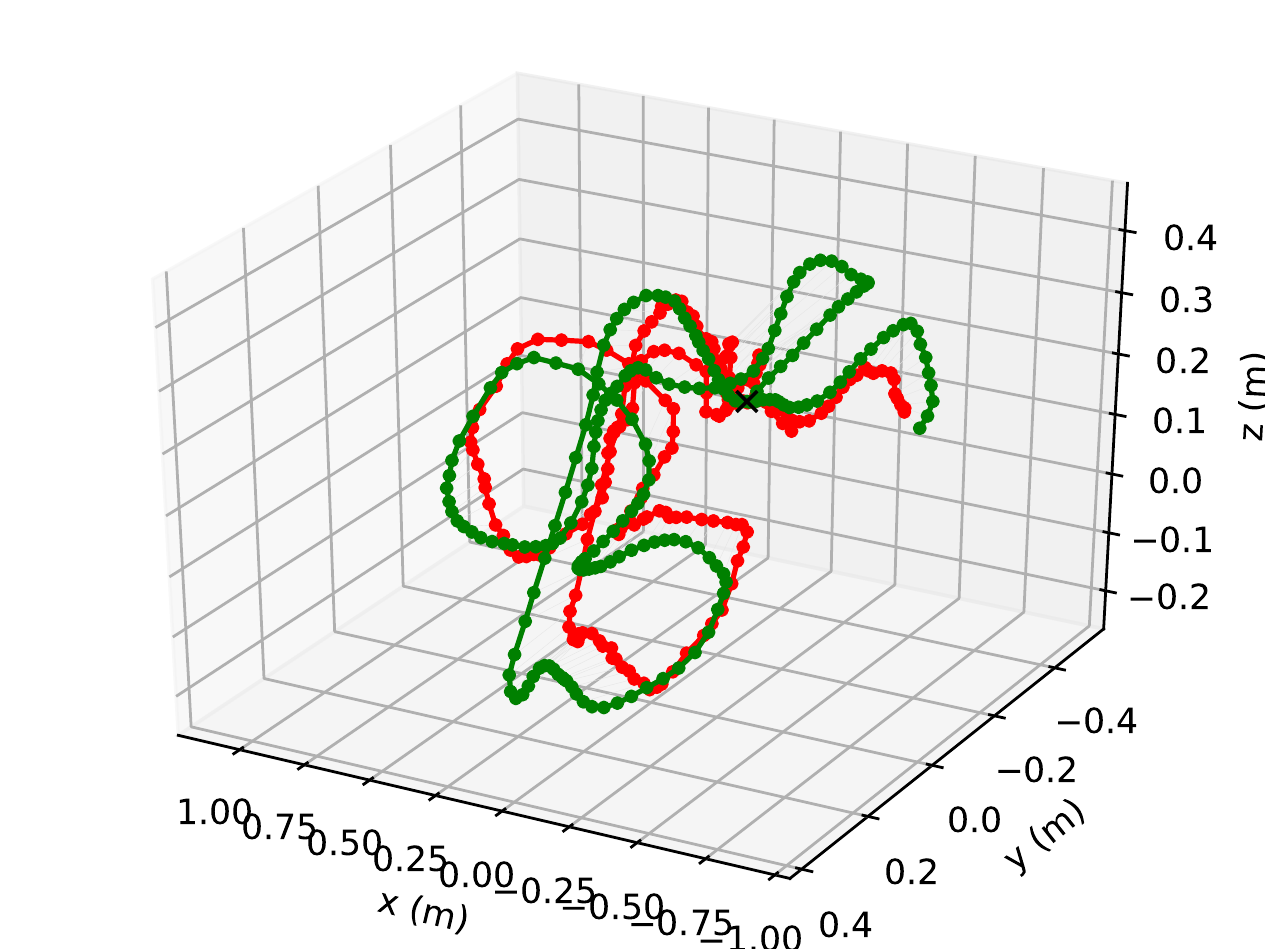}
        \includegraphics[width=\linewidth]{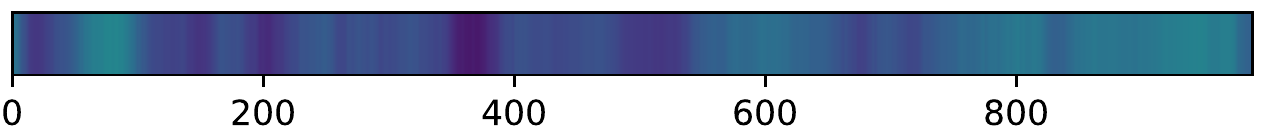}
    \end{subfigure}

    \begin{subfigure}{0.19\linewidth}
        \centering
        \includegraphics[width=\linewidth]{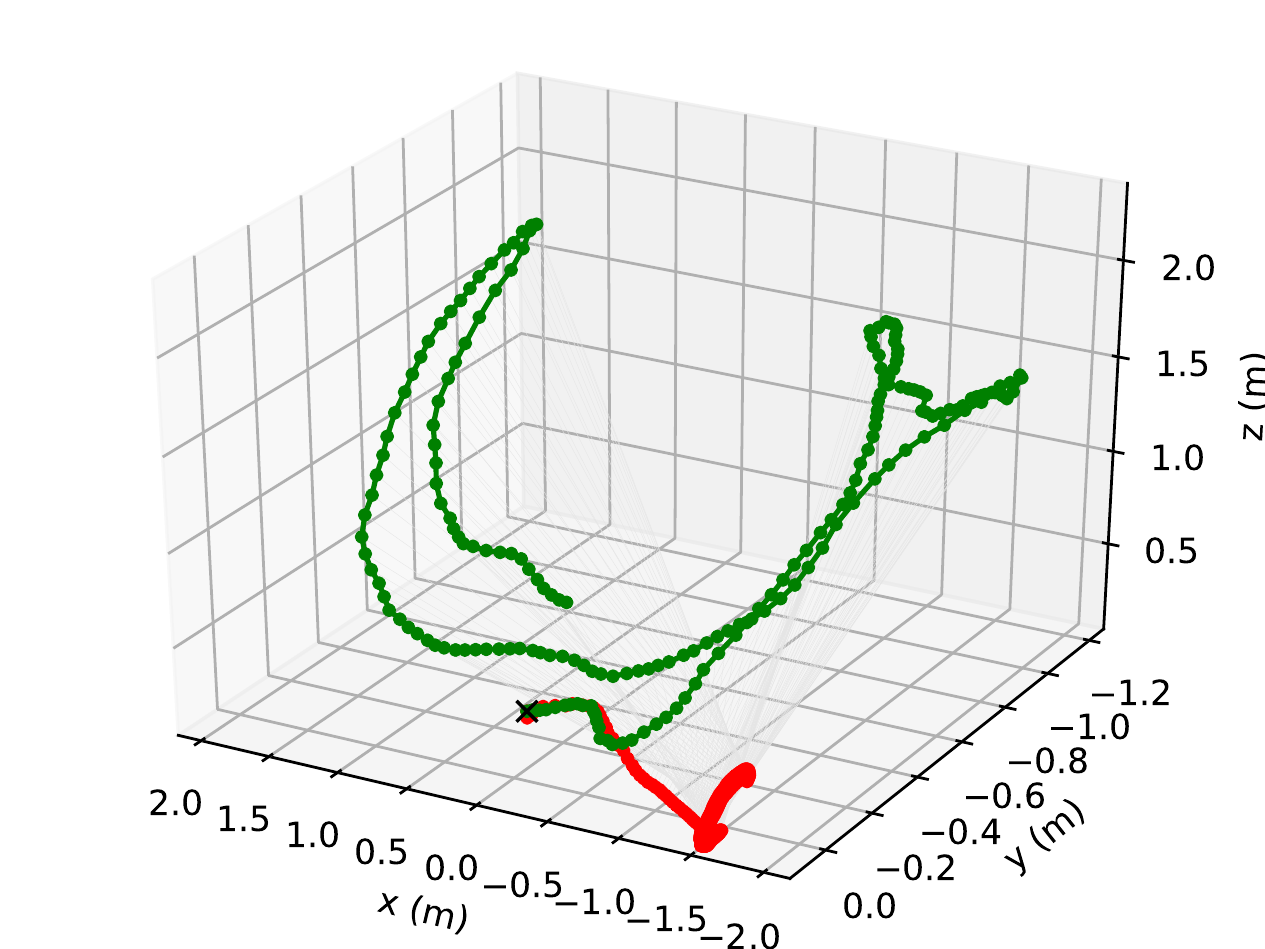}
        \includegraphics[width=\linewidth]{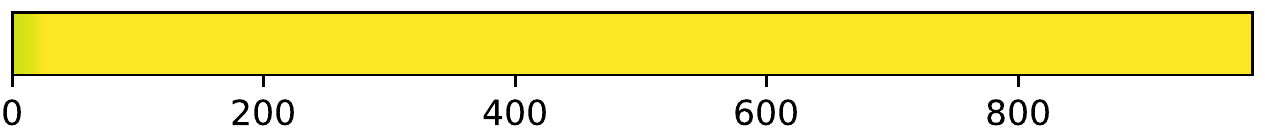}
        \caption{DSO~\cite{Engel2017DSO}}
    \end{subfigure}
    \hfill
    \begin{subfigure}{0.19\linewidth}
        \centering
        \includegraphics[width=\linewidth]{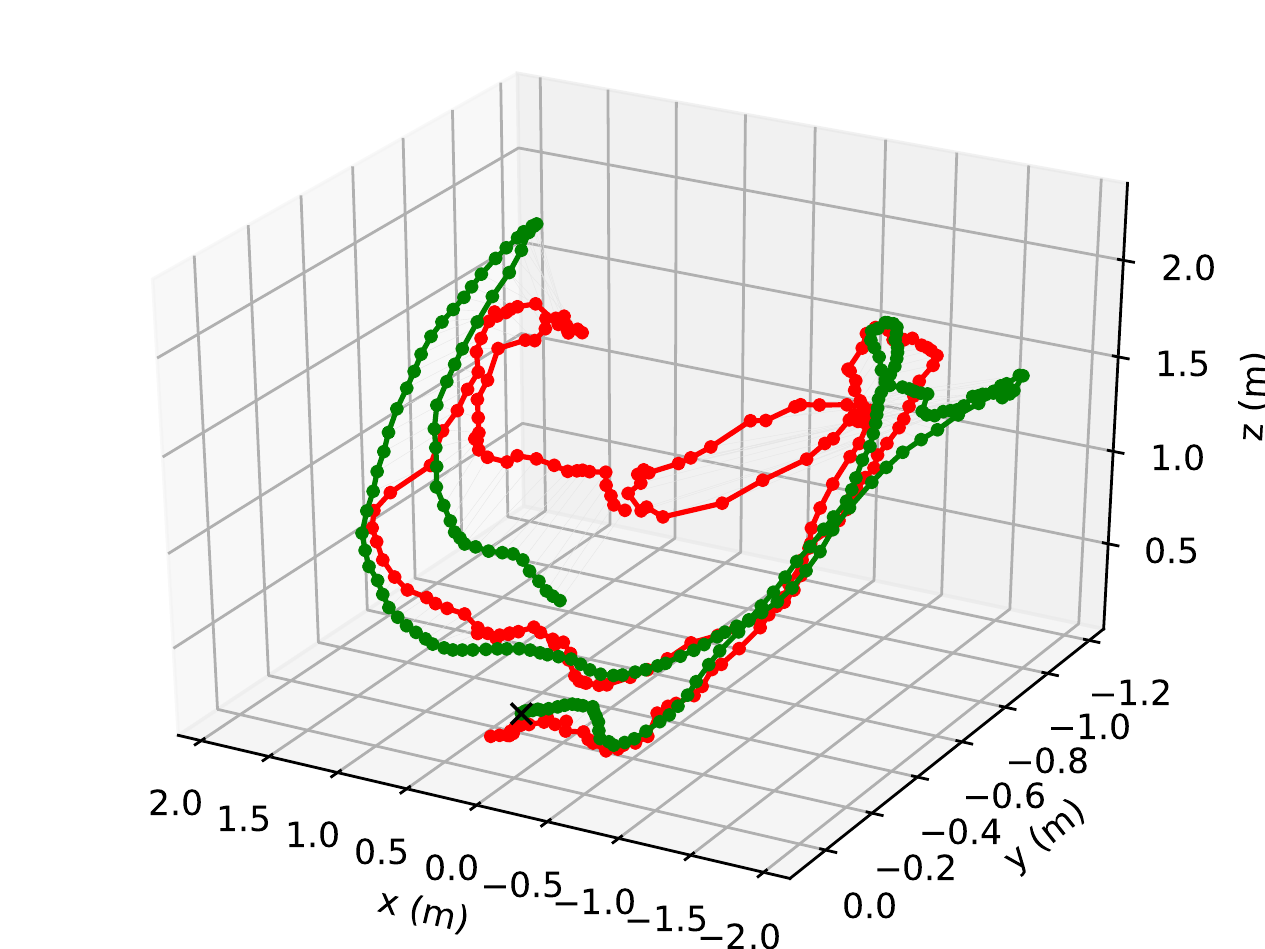}
        \includegraphics[width=\linewidth]{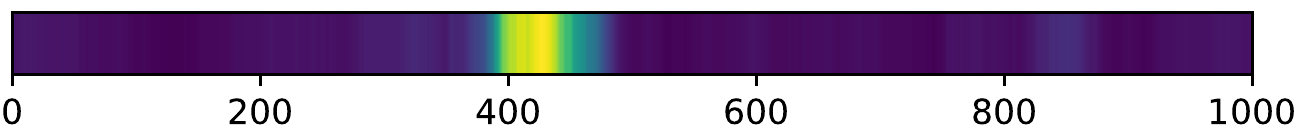}
        \caption{PoseNet~\cite{Kendall17cvpr, Kendall15iccv, Kendall16icra}}
    \end{subfigure}
    \hfill
    \begin{subfigure}{0.19\linewidth}
        \centering
        \includegraphics[width=\linewidth]{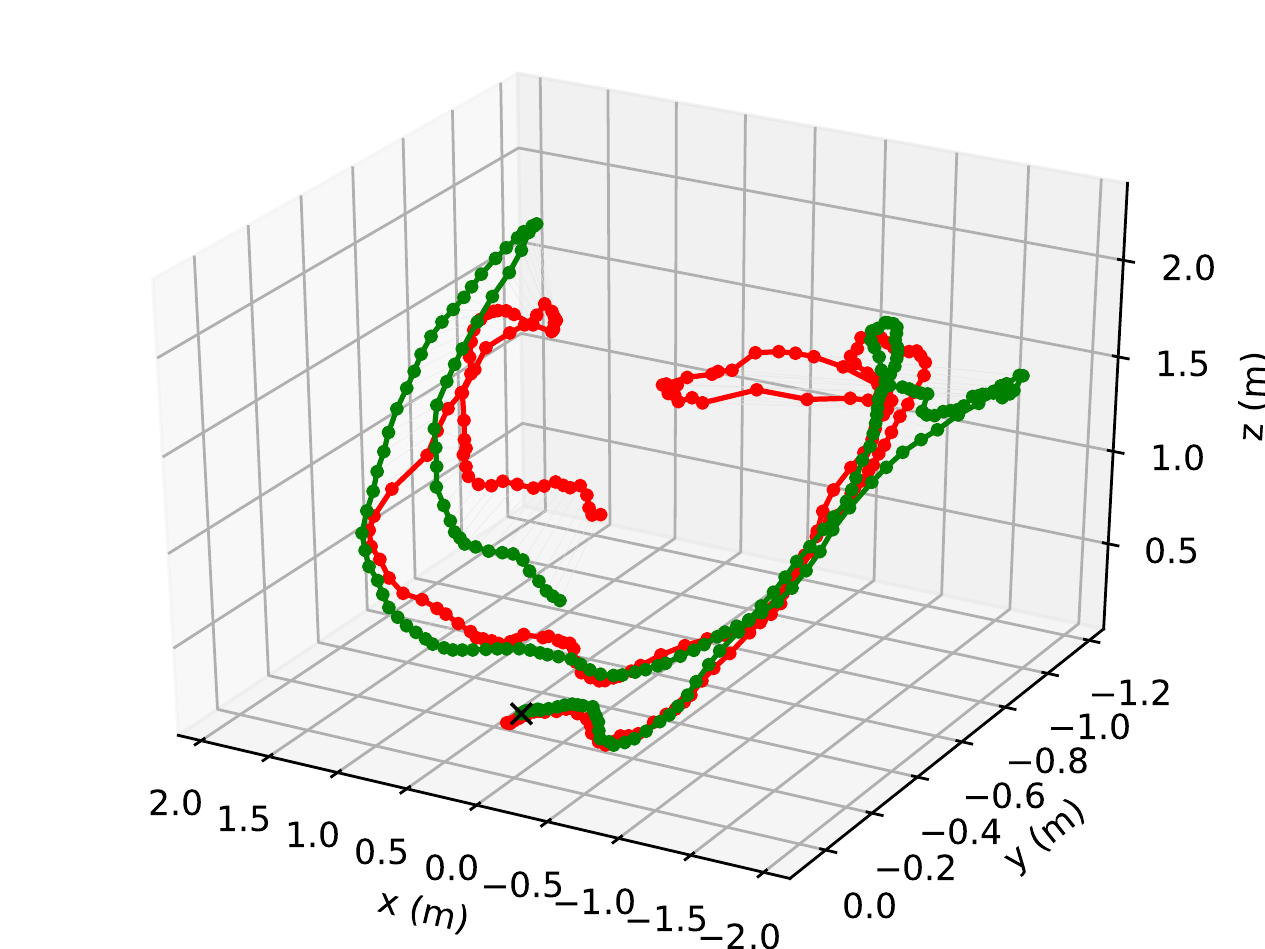}
        \includegraphics[width=\linewidth]{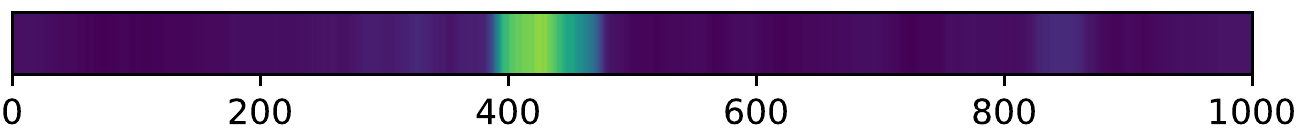}
        \caption{MapNet}
    \end{subfigure}
    \hfill
    \begin{subfigure}{0.19\linewidth}
        \centering
        \includegraphics[width=\linewidth]{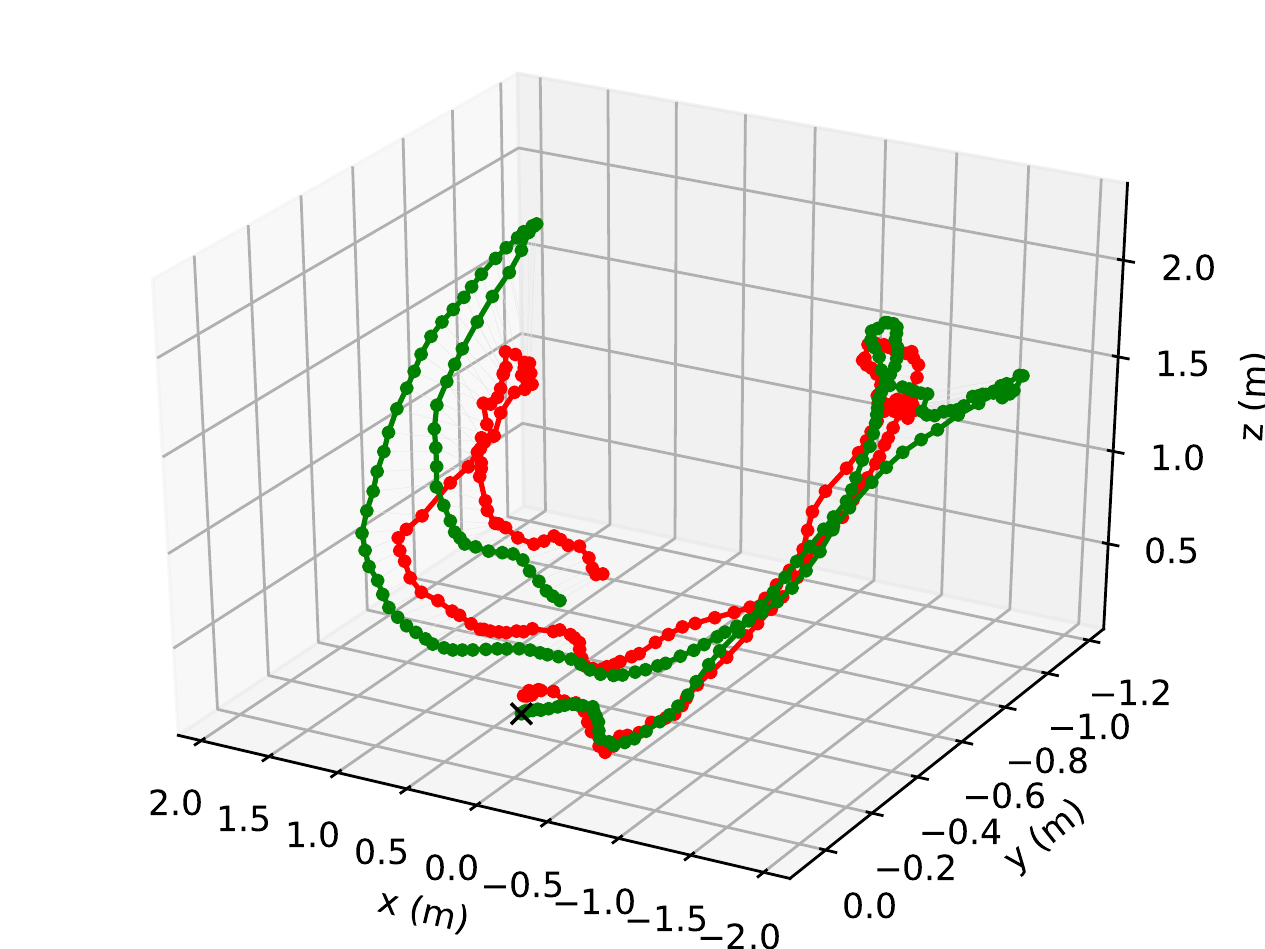}
        \includegraphics[width=\linewidth]{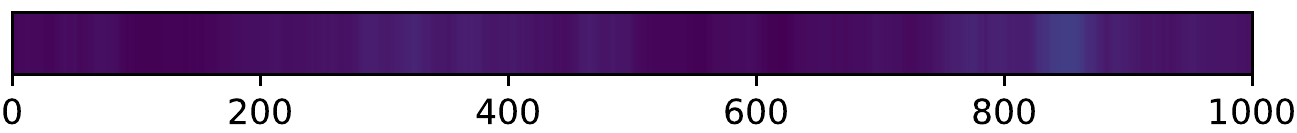}
        \caption{MapNet+}
    \end{subfigure}
    \hfill
    \begin{subfigure}{0.19\linewidth}
        \centering
        \includegraphics[width=\linewidth]{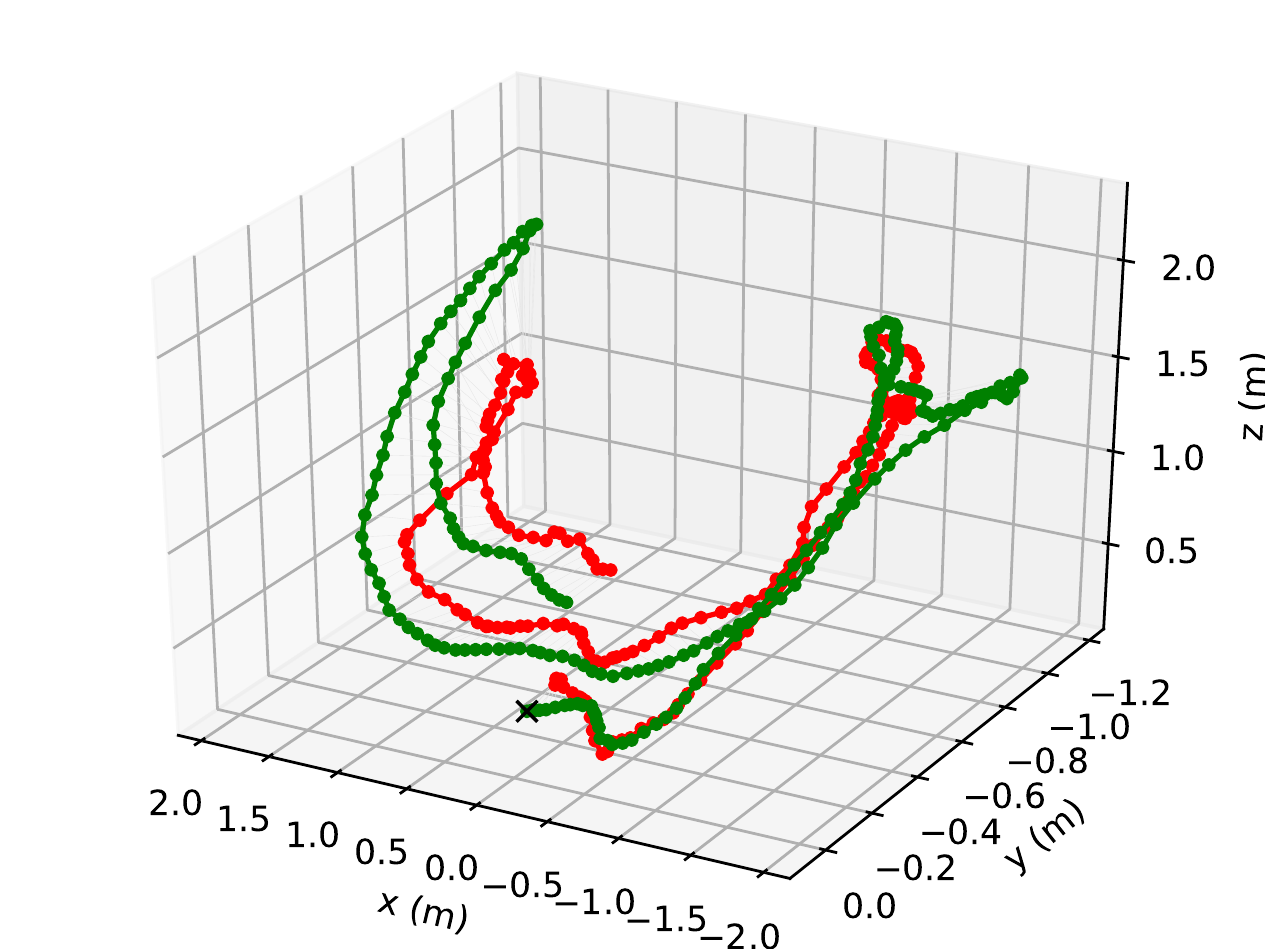}
        \includegraphics[width=\linewidth]{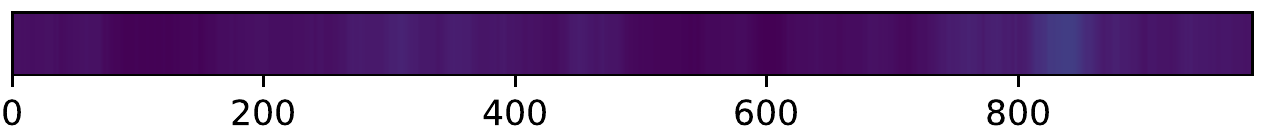}
        \caption{MapNet+PGO}
    \end{subfigure}
    \vspace{-1em}
    \caption{\small \textbf{Camera localization results on 7-Scenes dataset~\cite{Shotton13Scene7}}.
    For each subfigure, the top 3D plot shows the camera trajectory (green for the ground truth and red for the prediction), and the bottom color bar 
    shows rotation error for all the frames. From top to bottom, the three testing sequences are: Redkitchen-seq-03, Heads-seq-01, and Redkitchen-seq-12.
    See Table~\ref{tab:result_7scenes} for quantitative comparison.}
    \label{fig:map_compare}
    \vspace{-1em}
\end{figure*}

\paragraph{Datasets} We evaluate our algorithms on two well-known public datasets ---
7-Scenes~\cite{Shotton13Scene7} for small-scale, indoor, AR/VR-type
scenarios, and Oxford RobotCar~\cite{RobotCarDatasetIJRR} for
large-scale, outdoor, autonomous driving scenarios. 7-Scenes contains RGB-D image sequences of seven indoor
environments (with the spatial extent less than 4 meters) captured with a
Kinect sensor. Multiple sequences were captured for each environment, and each
sequence is 500 or 1000 frames. The ground truth camera poses are
obtained with KinectFusion. The 7-Scenes dataset has recently
been  evaluated extensively as a benchmark~\cite{Kendall15iccv, Kendall16icra,
Kendall17cvpr,
Melekhov17Hourglass,Walch17LSTM,Clark17VidLoc,Brachmann17RANSAC}, which makes
it ideal for us to compare with prior state-of-the-art methods.

Oxford RobotCar contains over 100 repetitions of a consistent route (about 10km)
through central Oxford captured twice a week over a period of over a year. Thus
the dataset captures different combinations of weather, traffic, pedestrians,
construction and roadworks. In addition to the images captured with the six
cameras mounted on the car, the dataset also contains LIDAR, GPS and INS
measurements, as well as stereo visual odometry (VO).
We extracted two subsets from this dataset: LOOP (Fig.~\ref{fig:teaser})
with a total length of 1120m, which was also used in VidLoc~\cite{Clark17VidLoc}, and
FULL (Fig.~\ref{fig:map_compare_robotcar_full}) with a total length
of 9562m. Details of the training, validation, and testing sequences are provided
in the supplementary material.

\vspace{-1em}
\paragraph{Baselines and Data Augmentation}

We compare our approach with two groups of prior methods on 7-Scenes. For the DNN-based
prior work, we compare with PoseNet15~\cite{Kendall15iccv},
PoseNet16~\cite{Kendall16icra}, PoseNet17~\cite{Kendall17cvpr},
Hourglass~\cite{Melekhov17Hourglass}, LSTM-PoseNet~\cite{Walch17LSTM}, and
VidLoc~\cite{Clark17VidLoc}.  For the traditional visual odometry based
methods, we used DSO~\cite{Engel2017DSO} to 
compute the VO and integrate to obtain camera poses. We run the DSO with images at the same spatial resolution as 
MapNet. On Oxford RobotCar, only VidLoc~\cite{Clark17VidLoc} reported
results on the LOOP scene but it did not provide training and
testing sequences. Thus, the two baselines to compare are 
the provided stereo VO, as well as our version of
PoseNet (with $\log\mathbf{q}$). In RobotCar, we randomly perturb the brightness,
saturation, hue and contrast of images during training for experiments, which
we found essential for performing cross-weather and cross-time
localization.


\subsection{Experiments on the 7-Scenes Dataset}

\begin{table}
    \footnotesize
    \centering
    \caption{\small Translation and rotation error on the 7-Scenes dataset.}
    \vspace{-1em}
    \begin{tabular}{ll|ll}
        \toprule
        \multirow{2}{*}{Scene}       & PoseNet17            & PoseNet      & PoseNet+$\log\mathbf{q}$ \\ 
                                     & \cite{Kendall17cvpr}  & (ResNet34)   & (ResNet34)              \\
        \midrule
        Chess       & 0.13m, 4.48\degree	 & {\bf 0.11m, 4.24\degree}  & {\bf 0.11m}, 4.29\degree  \\
        Fire        & {\bf 0.27m, 11.30\degree} & 0.29m, 11.68\degree & {\bf 0.27m}, 12.13\degree \\
        Heads       & {\bf 0.17m}, 13.00\degree	 & 0.20m, 13.11\degree & 0.19m, {\bf 12.15\degree} \\
        Office      & {\bf 0.19m, 5.55\degree}	 & {\bf 0.19m}, 6.40\degree  & {\bf 0.19m}, 6.35\degree  \\
        Pumpkin     & 0.26m, {\bf 4.75\degree}	 & 0.23m, 5.77\degree  & {\bf 0.22m}, 5.05\degree  \\
        Red Kitchen & {\bf 0.23m}, 5.35\degree	 & 0.27m, 5.81\degree  & 0.25m, {\bf 5.27\degree}  \\
        Stairs      & 0.35m, 12.40\degree   & 0.31m, 12.43\degree & {\bf 0.30m, 11.29\degree} \\
        \midrule
        Average     & 0.23m, 8.12\degree    & 0.23m, 8.49\degree  & {\bf 0.22m, 8.07\degree}   \\
        \bottomrule
    \end{tabular}
    \label{tab:result_logq}
    \vspace{-1em}
\end{table}

\vspace{-.5em}
\paragraph{Effects of Rotation Parameteriation} 

In Section~\ref{subsec:regress}, we introduced a new parameterization of camera
orientation for PoseNet and used ResNet34 as the base network.
Table~\ref{tab:result_logq} shows the quantitative results of these
modifications to the baseline PoseNet. Following the same convention of prior
work~\cite{Kendall15iccv,Kendall16icra,Kendall17cvpr,Melekhov17Hourglass,Walch17LSTM,Clark17VidLoc},
we compute the median error for camera translation and rotation.\footnote{Other statistics
of the camera pose estimation errors are also provided in the supplementary material, which
support the same conclusion.}
As shown, our proposed rotation parameterization does improve
performance. 

\begin{table*}
    \footnotesize
    \centering
    \caption{\small Translation error (m) and rotation error (\degree) for various methods on the 7-Scenes dataset~\cite{Shotton13Scene7}.}
    \vspace{-1em}
    \begin{tabular}{llllll|lll}
        \toprule
        \multirow{2}{*}{Scene}   & PoseNet17 & Hourglass & LSTM-Pose & VidLoc & DSO & MapNet & MapNet+ & MapNet+PGO \\
                & \cite{Kendall17cvpr} & \cite{Melekhov17Hourglass} & \cite{Walch17LSTM} & \cite{Clark17VidLoc} & \cite{Engel2017DSO} &
                 &  &  \\
        \midrule
        Chess       &  0.13m, 4.48\degree   & 0.15m, 6.17\degree  & 0.24m, 5.77\degree & 0.18m, NA & 0.17m, 8.13\degree & {\bf 0.08m}, 3.25\degree  & 0.10m, {\bf 3.17\degree}  & 0.09m, 3.24\degree \\
        Fire        &  0.27m, 11.30\degree  & 0.27m, 10.84\degree & 0.34m, 11.9\degree & 0.26m, NA & {\bf 0.19m}, 65.0\degree & 0.27m, 11.69\degree & 0.20m, {\bf 9.04\degree}  & 0.20m, 9.29\degree \\
        Heads       &  0.17m, 13.00\degree  & 0.19m, 11.63\degree & 0.21m, 13.7\degree & 0.14m, NA & 0.61m, 68.2\degree & 0.18m, 13.25\degree & 0.13m, 11.13\degree & {\bf 0.12m, 8.45\degree} \\
        Office      &  0.19m, 5.55\degree   & 0.21m, 8.48\degree  & 0.30m, 8.08\degree & 0.26m, NA & 1.51m, 16.8\degree & {\bf 0.17m, 5.15\degree}  & 0.18m, 5.38\degree  & 0.19m, 5.42\degree\\
        Pumpkin     &  0.26m, 4.75\degree   & 0.25m, 7.01\degree  & 0.33m, 7.00\degree & 0.36m, NA & 0.61m, 15.8\degree & 0.22m, 4.02\degree  & {\bf 0.19m, 3.92\degree}  & {\bf 0.19m}, 3.96\degree \\
        Kitchen &  0.23m, 5.35\degree   & 0.27m, 10.15\degree & 0.37m, 8.83\degree & 0.31m, NA & 0.23m, 10.9\degree & 0.23m, 4.93\degree  & 0.20m, 5.01\degree  & {\bf 0.20m, 4.94\degree} \\
        Stairs      &  0.35m, 12.40\degree  & 0.29m, 12.46\degree & 0.40m, 13.7\degree & {\bf 0.26m}, NA & 0.26m, 21.3\degree & 0.30m, 12.08\degree & 0.30m, 13.37\degree & 0.27m, {\bf 10.57\degree} \\
        \midrule
        Average     &  0.23m, 8.12\degree   & 0.23m, 9.53\degree  & 0.31m, 9.85\degree & 0.25m, NA & 0.26m, 29.4\degree & 0.21m, 7.77\degree & 0.19m, 7.29\degree   & {\bf 0.18m, 6.55\degree}\\
        \bottomrule
    \end{tabular}
    \label{tab:result_7scenes}
    \vspace{-.5em}
\end{table*}

\vspace{-1em}
\paragraph{Comparison with Prior Methods}

Figure~\ref{fig:map_compare} shows the camera trajectories for several testing
sequences from the 7-Scenes dataset for DSO VO, PoseNet, MapNet, MapNet+, and MapNet+PGO.
Table~\ref{tab:result_7scenes} shows quantitative comparisons.
The unlabeled data used to fine-tune MapNet+ for these experiments are the unlabeled test sequences.
This is a transductive learning scenario~\cite{Chapelle06book, Segonne08transduction}.
As shown, DSO often drifts over time and PoseNet results in noisy predictions. In contrast, by
including various geometric constraints into network training and inference our proposed approaches
MapNet, MapNet+ and MapNet+PGO successively improve the performance.
A complete table for all testing sequences from the 7-Scenes dataset is included in the
supplementary material.

\begin{figure*}[h!]
    \small
    \captionsetup[subfigure]{labelformat=empty}
    \centering
    \begin{subfigure}{0.24\linewidth}
        \centering
        \includegraphics[width=\linewidth]{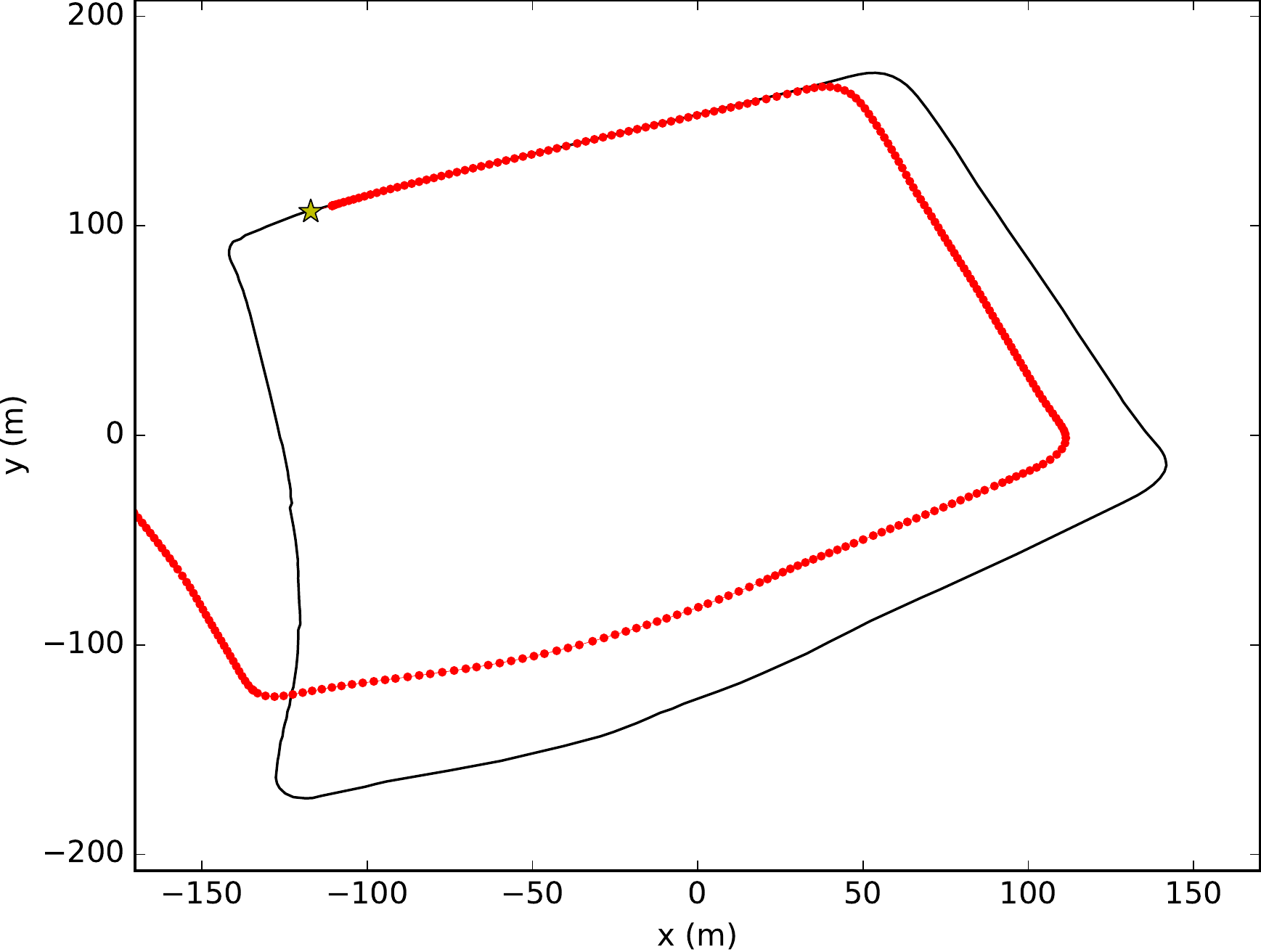}
        \vspace{-1.5em}
        \caption{\small Stereo VO (40.20m, 12.85\degree) }
    \end{subfigure}
    \hfill
    \begin{subfigure}{0.24\linewidth}
        \centering
        \includegraphics[width=\linewidth]{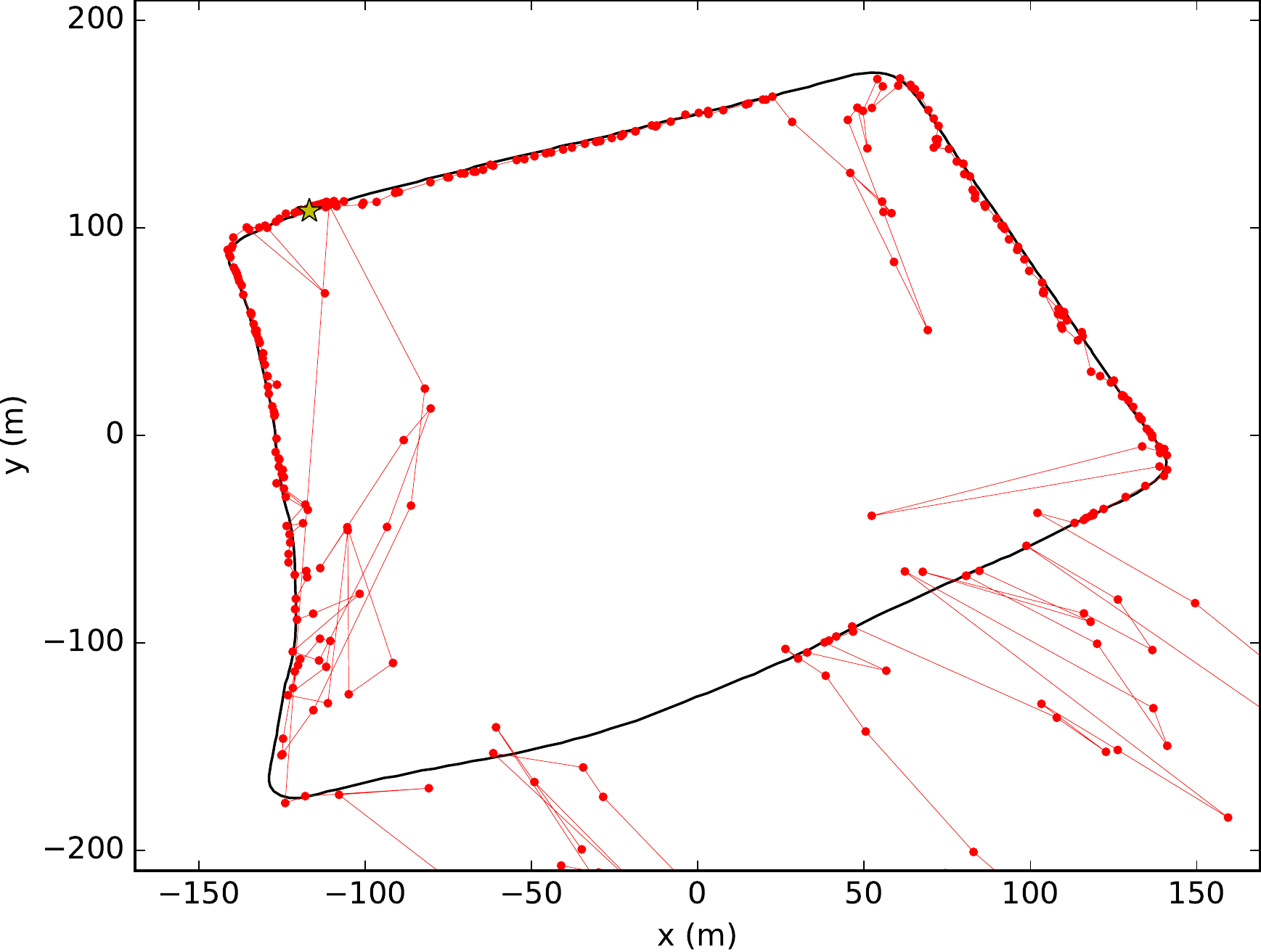}
        \vspace{-1.5em}
        \caption{\small PoseNet (25.29m, 17.45\degree)}
    \end{subfigure}
    \hfill
    \begin{subfigure}{0.24\linewidth}
        \centering
        \includegraphics[width=\linewidth]{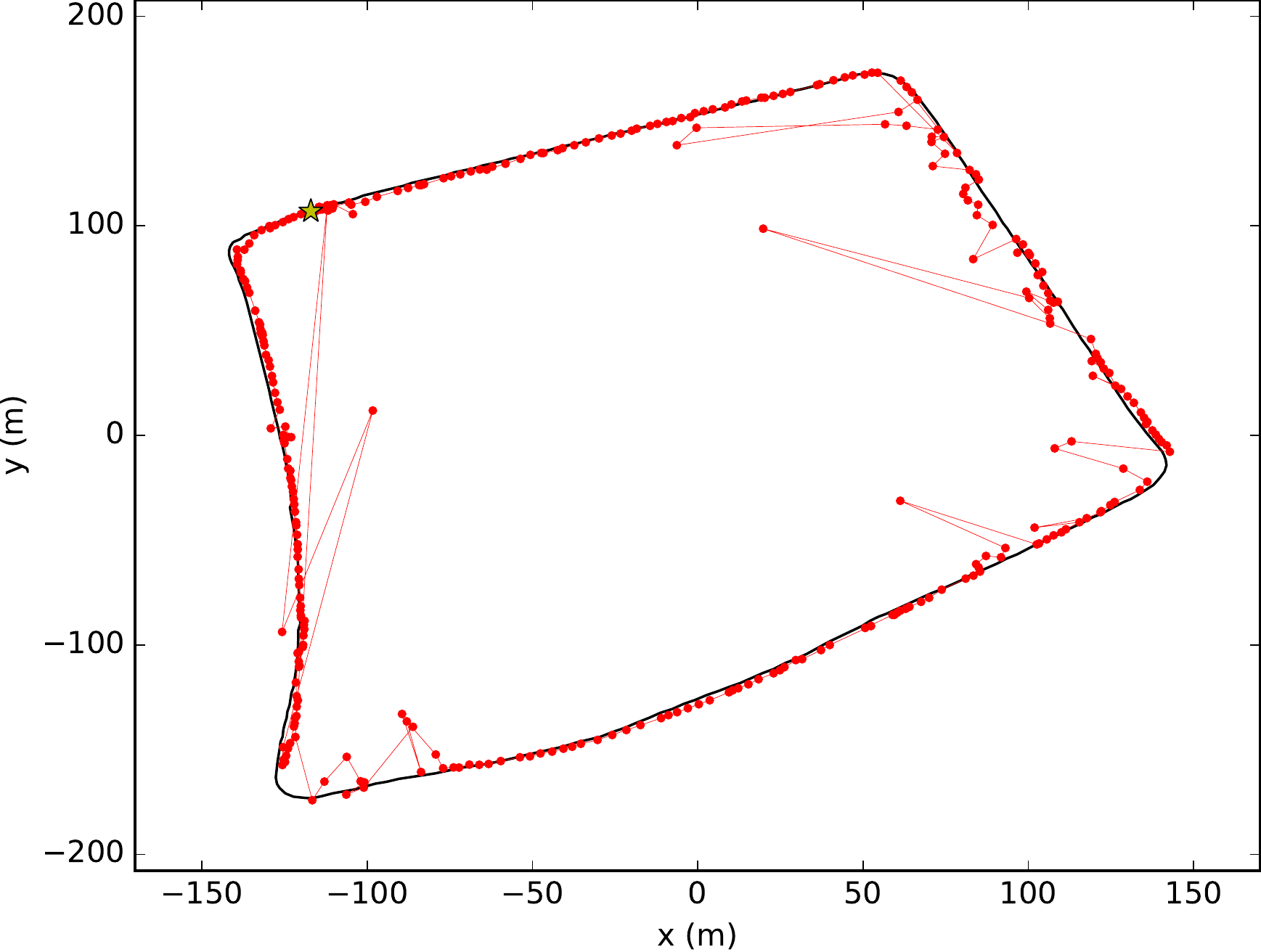}
        \vspace{-1.5em}
        \caption{\small MapNet (9.84m, 3.96\degree)}
    \end{subfigure}
    \hfill
    \rulesep
    \begin{subfigure}{0.24\linewidth}
        \centering
        \includegraphics[width=\linewidth]{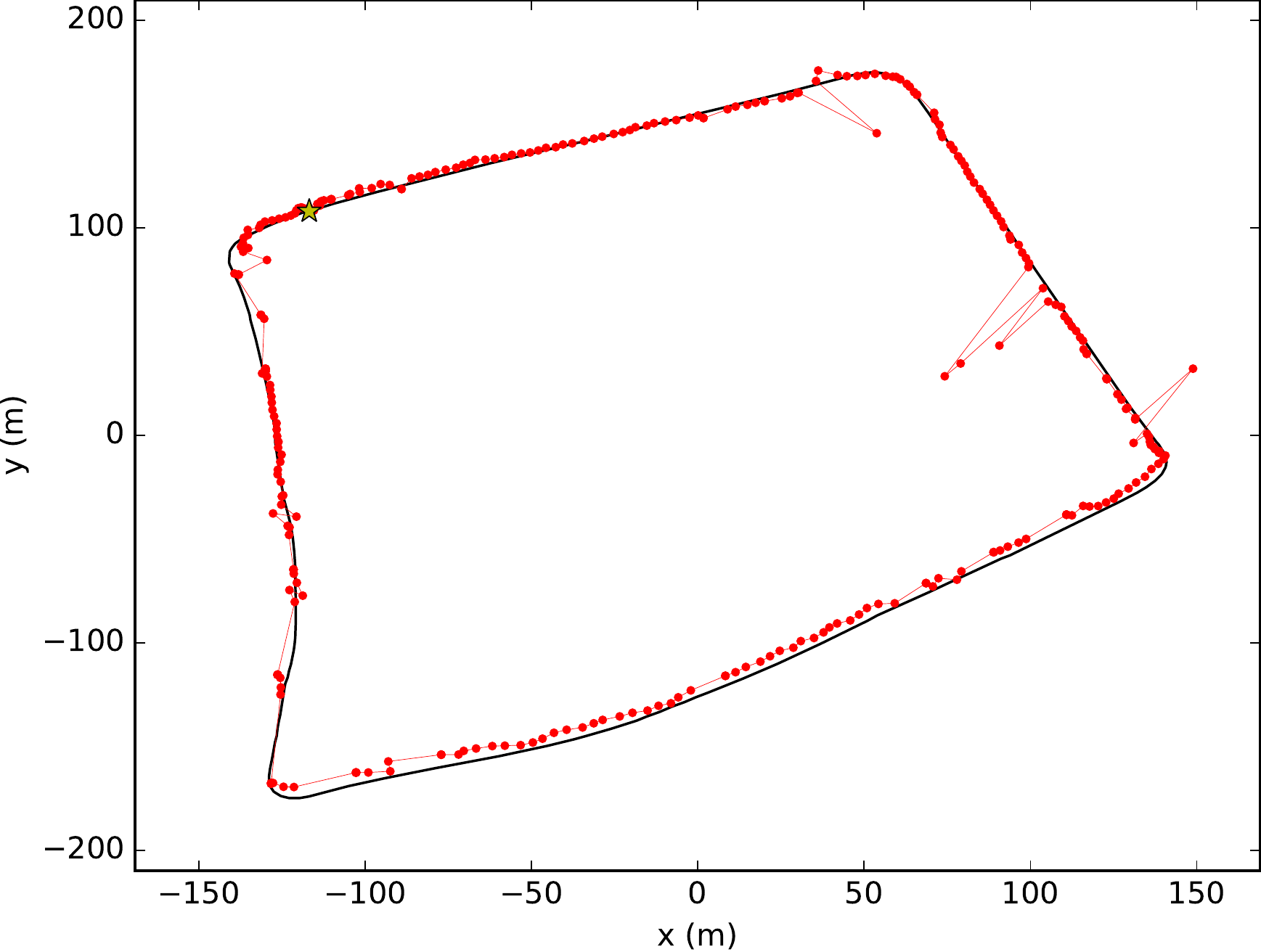}
        \vspace{-1.5em}
        \caption{\small GPS (7.03m, NA) }
    \end{subfigure}

    \begin{subfigure}{0.24\linewidth}
        \centering
        \includegraphics[width=\linewidth]{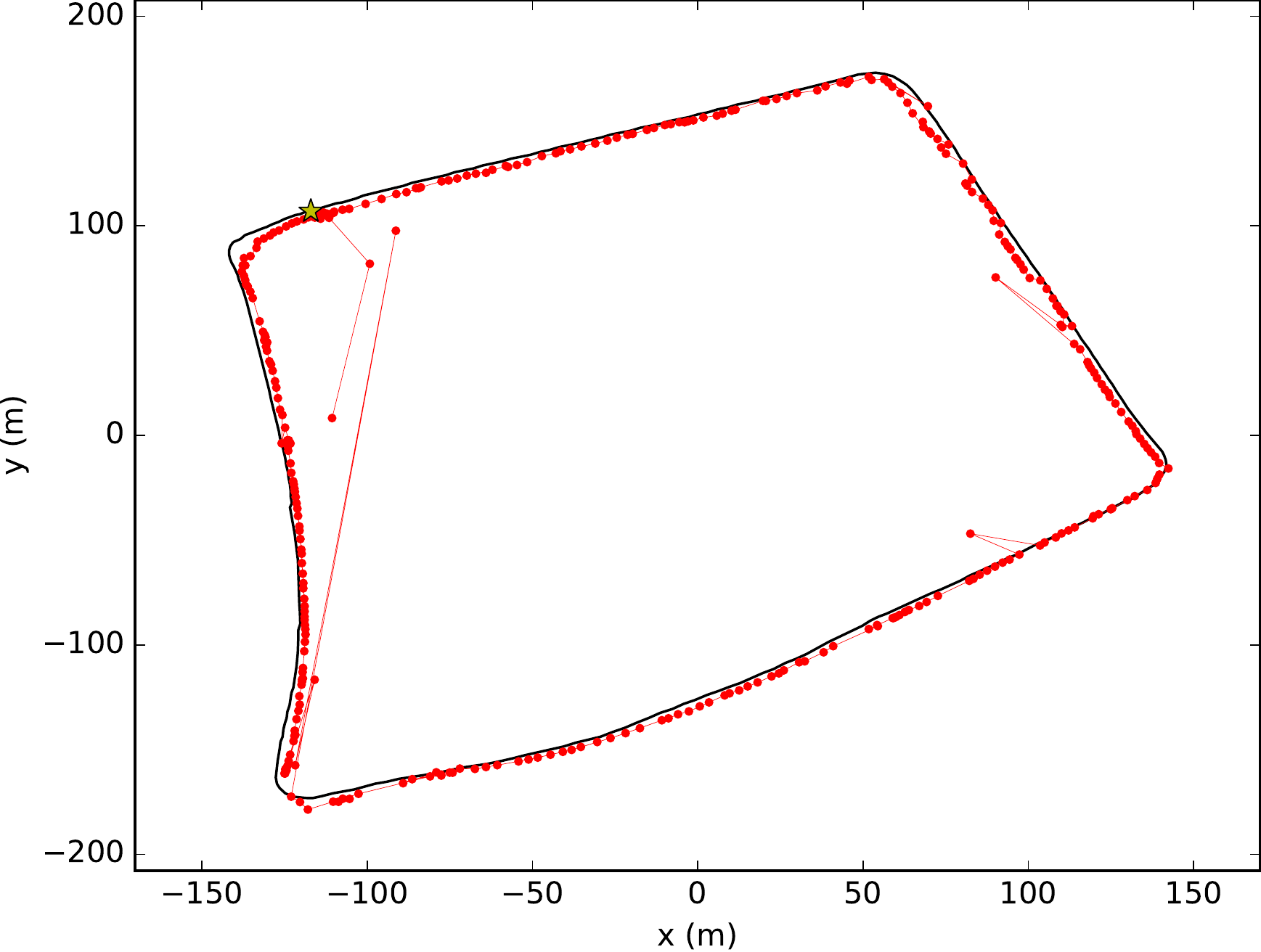}
        \vspace{-1.5em}
        \caption{\small MapNet+(1seq) (8.17m, 2.62\degree)}
    \end{subfigure}
    \hfill
    \begin{subfigure}{0.24\linewidth}
        \centering
        \includegraphics[width=\linewidth]{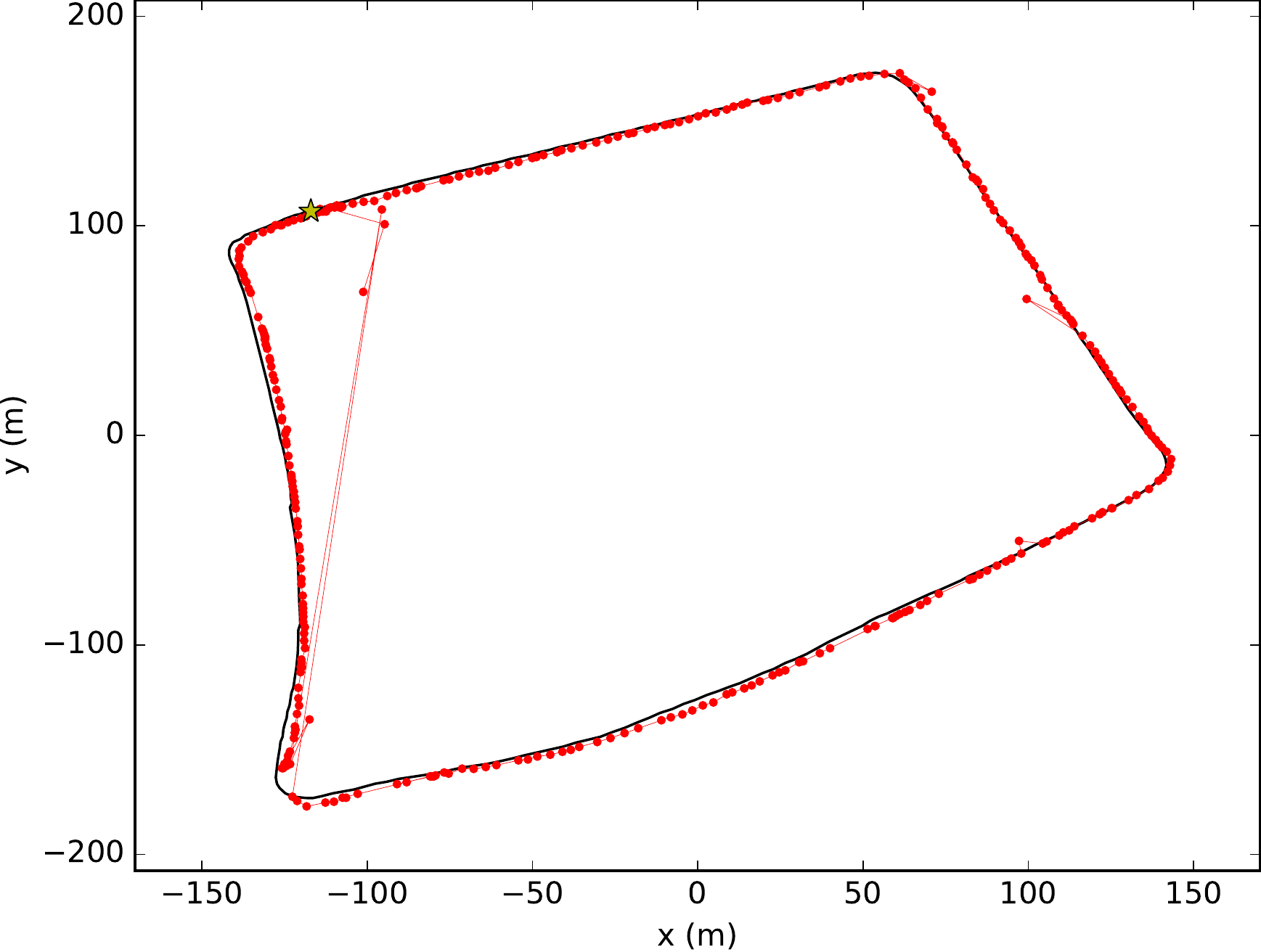}
        \vspace{-1.5em}
        \caption{\small MapNet+(2seq) (6.95m, 2.38\degree)}
    \end{subfigure}
    \hfill
    \begin{subfigure}{0.24\linewidth}
        \centering
        \includegraphics[width=\linewidth]{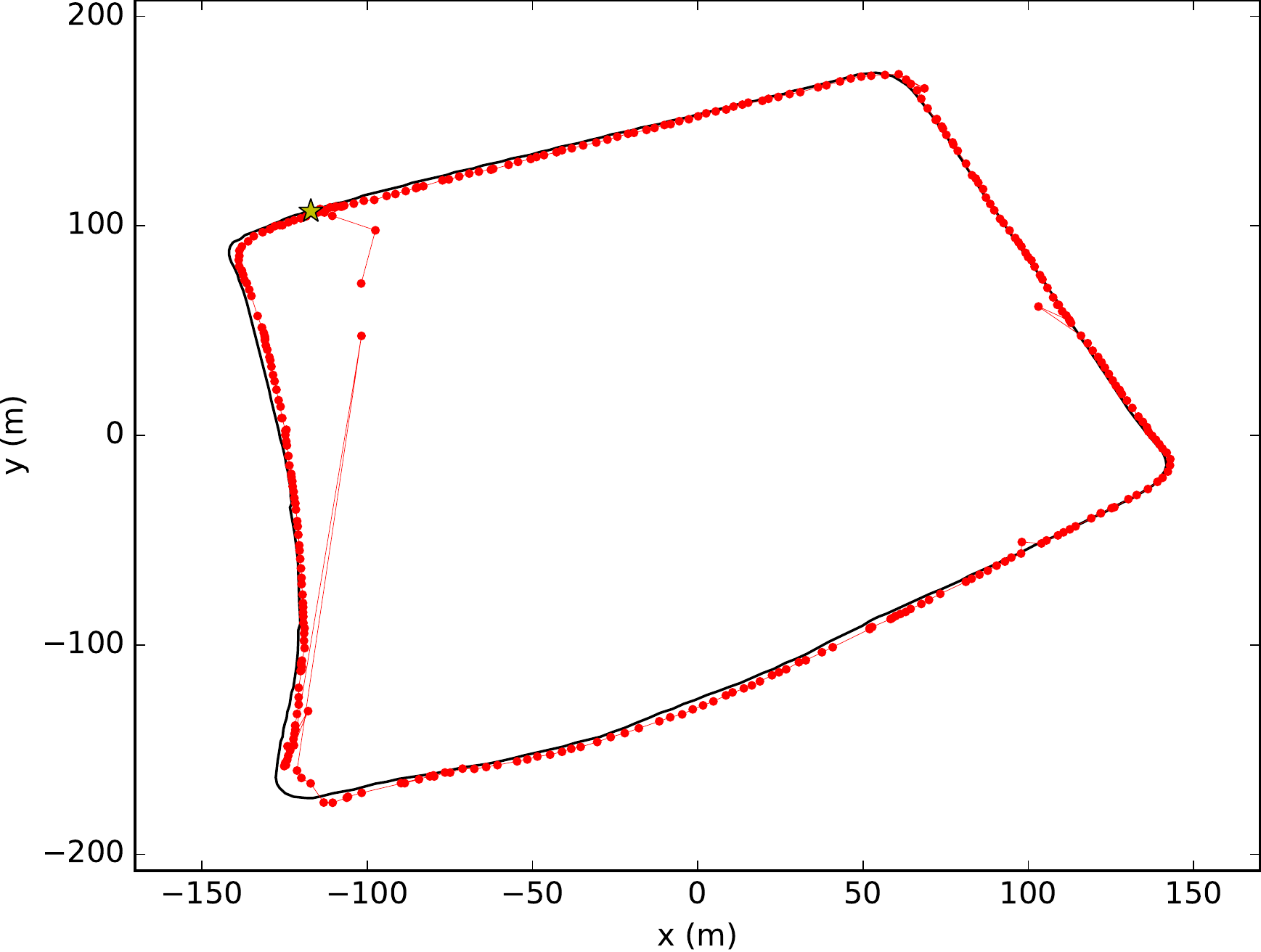}
        \vspace{-1.5em}
        \caption{\small MapNet+PGO ({\bf 6.73m, 2.23\degree})}
    \end{subfigure}
    \hfill
    \rulesep
    \begin{subfigure}{0.24\linewidth}
        \centering
        \includegraphics[width=\linewidth]{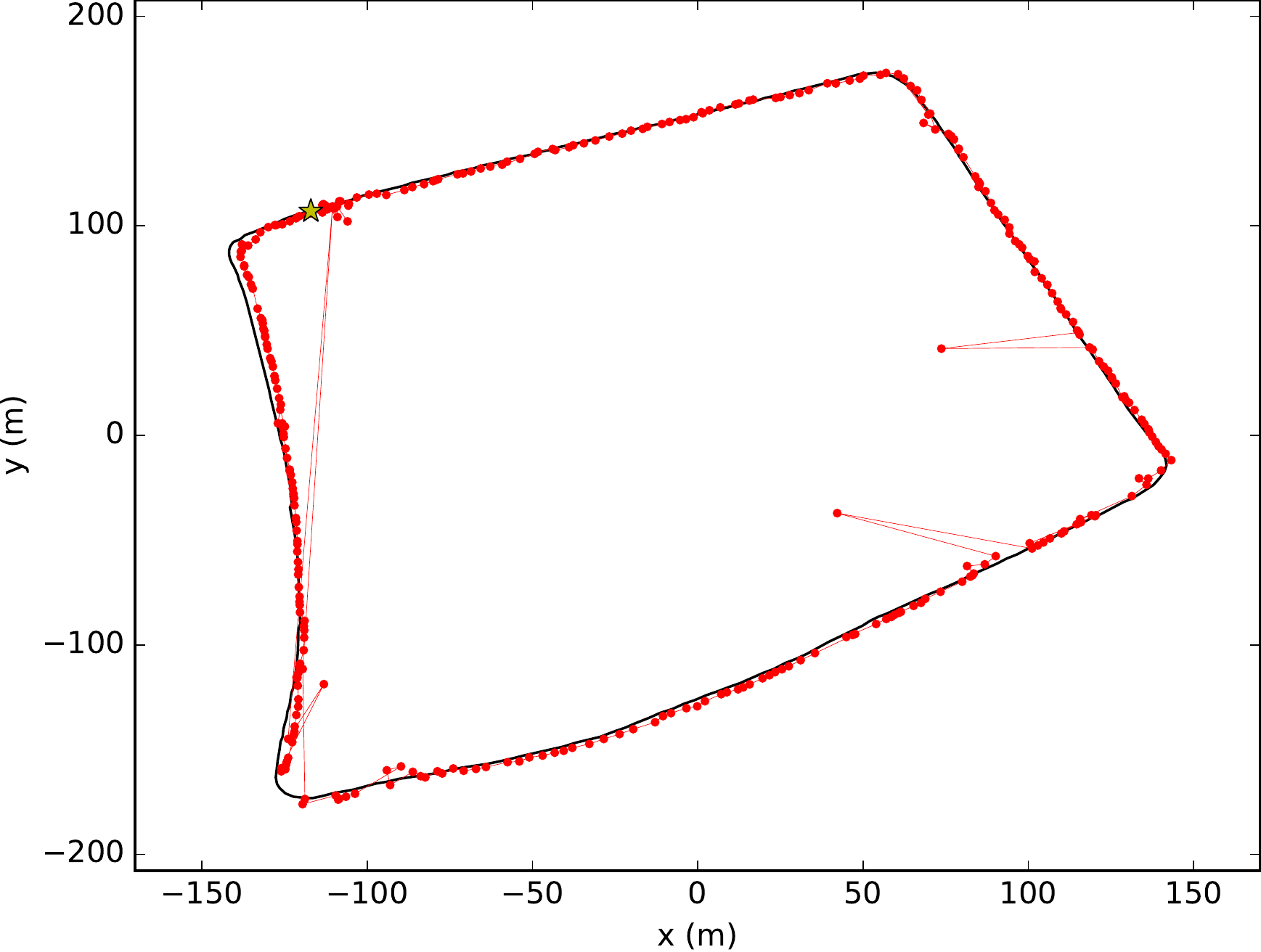}
        \vspace{-1.5em}
        \caption{\small MapNet+(GPS) ({\bf 6.78m, 2.72\degree})}
    \end{subfigure}
    \vspace{-.5em} 
    \caption{\small \textbf{Camera localization results on the LOOP scene (1120m long) of the
    Oxford RobotCar dataset~\cite{RobotCarDatasetIJRR}}. The ground truth camera trajectory
    is the black line, the star indicates the first frame, and the red lines show the
    camera pose predictions. The caption of each figure shows the mean translation error (m) and mean rotation error (\degree).
    MapNet+(1seq) uses one unlabeled sequence, while MapNet+(2seq) uses two unlabeled sequences.
    {\bf Left}: MapNet+ trained with unlabeled images and stereo VO. {\bf Right}: MapNet+ trained with unlabeled images and GPS data.}
    \label{fig:map_compare_robotcar}
\end{figure*}

\begin{figure*}
    \captionsetup[subfigure]{labelformat=empty}
    \centering
    \begin{subfigure}{0.32\linewidth}
        \centering
        \includegraphics[width=\linewidth]{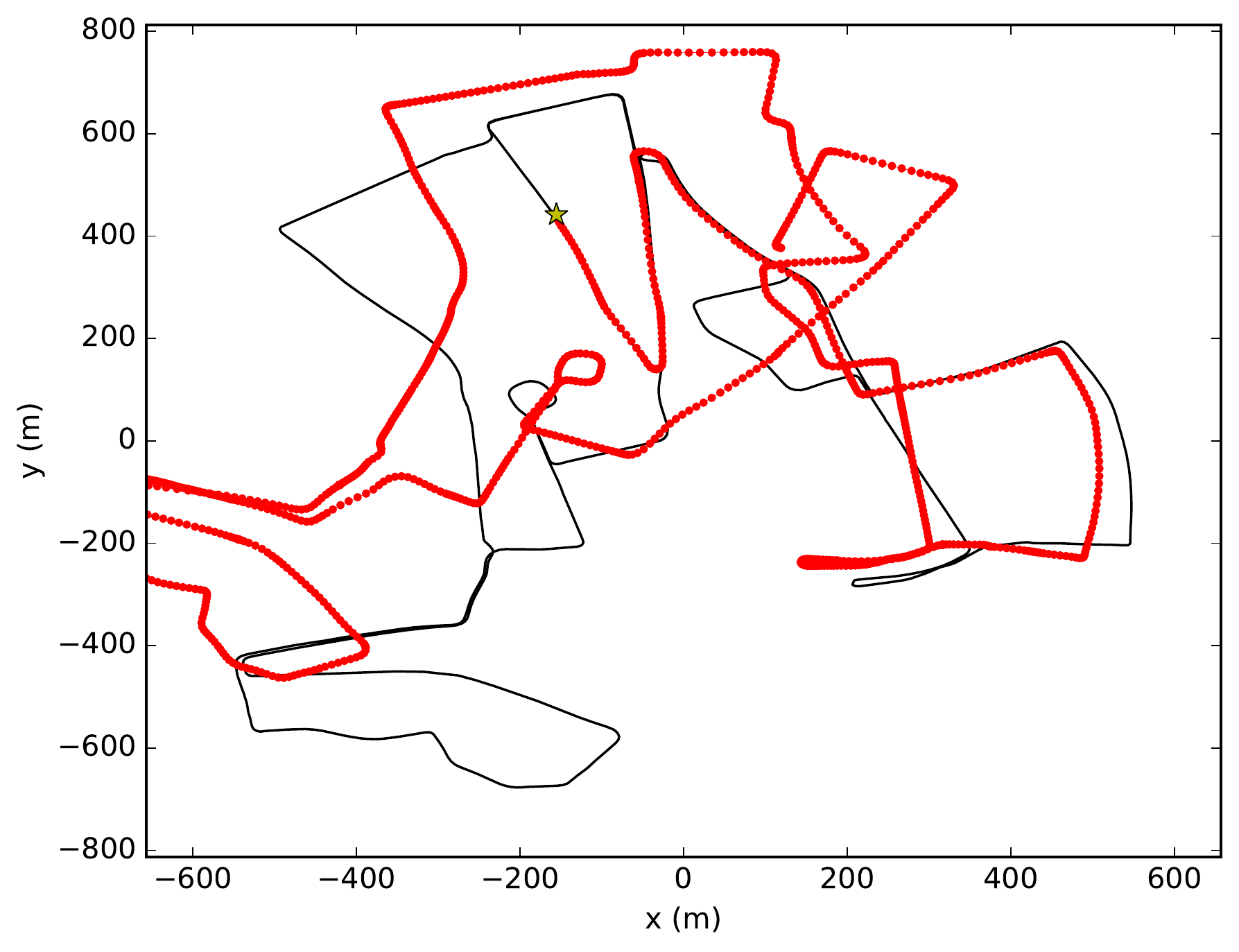}
        \vspace{-1.5em}
        \caption{\small Stereo VO (222.0m, 13.7\degree)}
    \end{subfigure}
    \hfill
    \begin{subfigure}{0.32\linewidth}
        \centering
        \includegraphics[width=\linewidth]{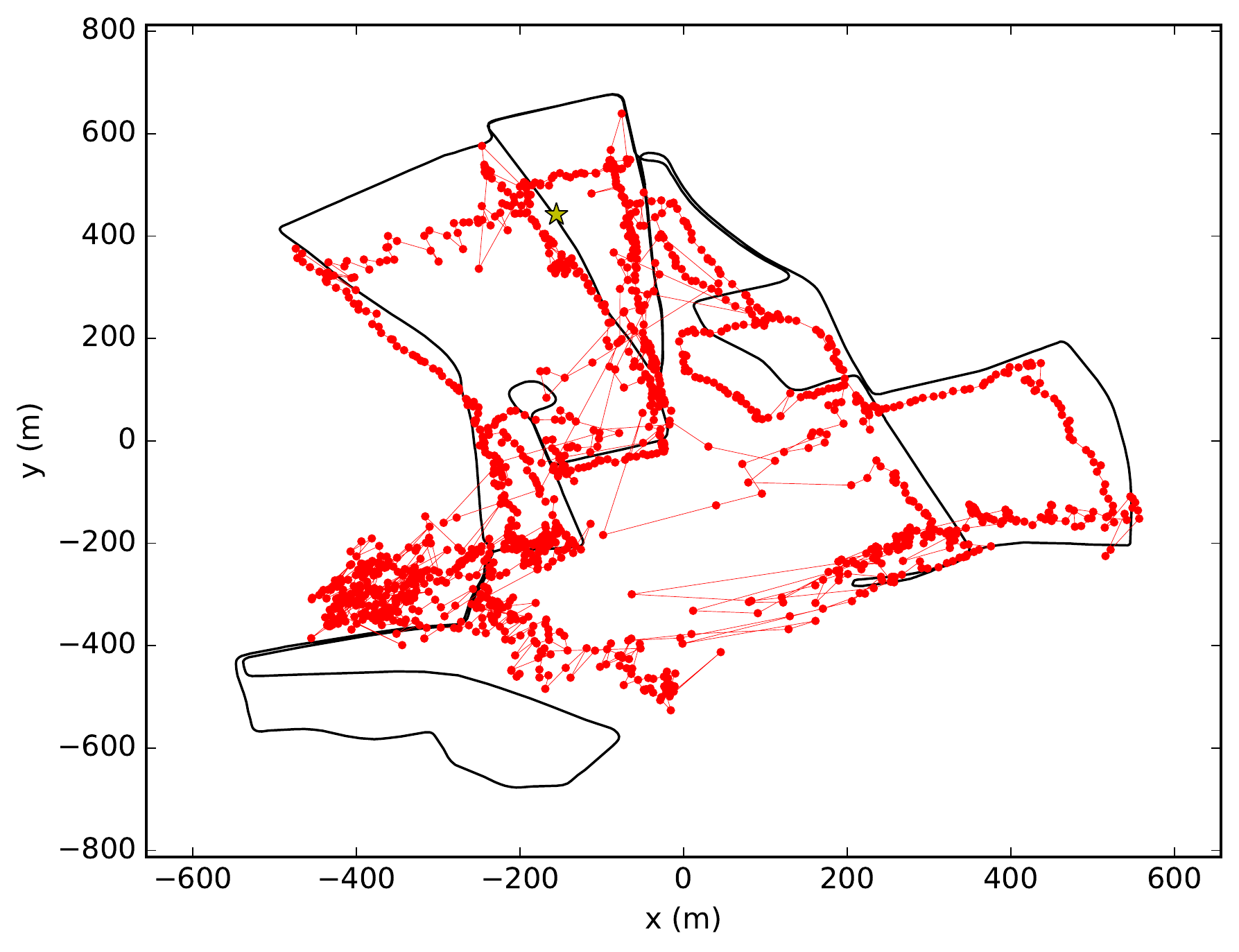}
        \vspace{-1.5em}
        \caption{\small PoseNet (125.6m, 27.1\degree)}
    \end{subfigure}
    \hfill
    \begin{subfigure}{0.32\linewidth}
        \centering
        \includegraphics[width=\linewidth]{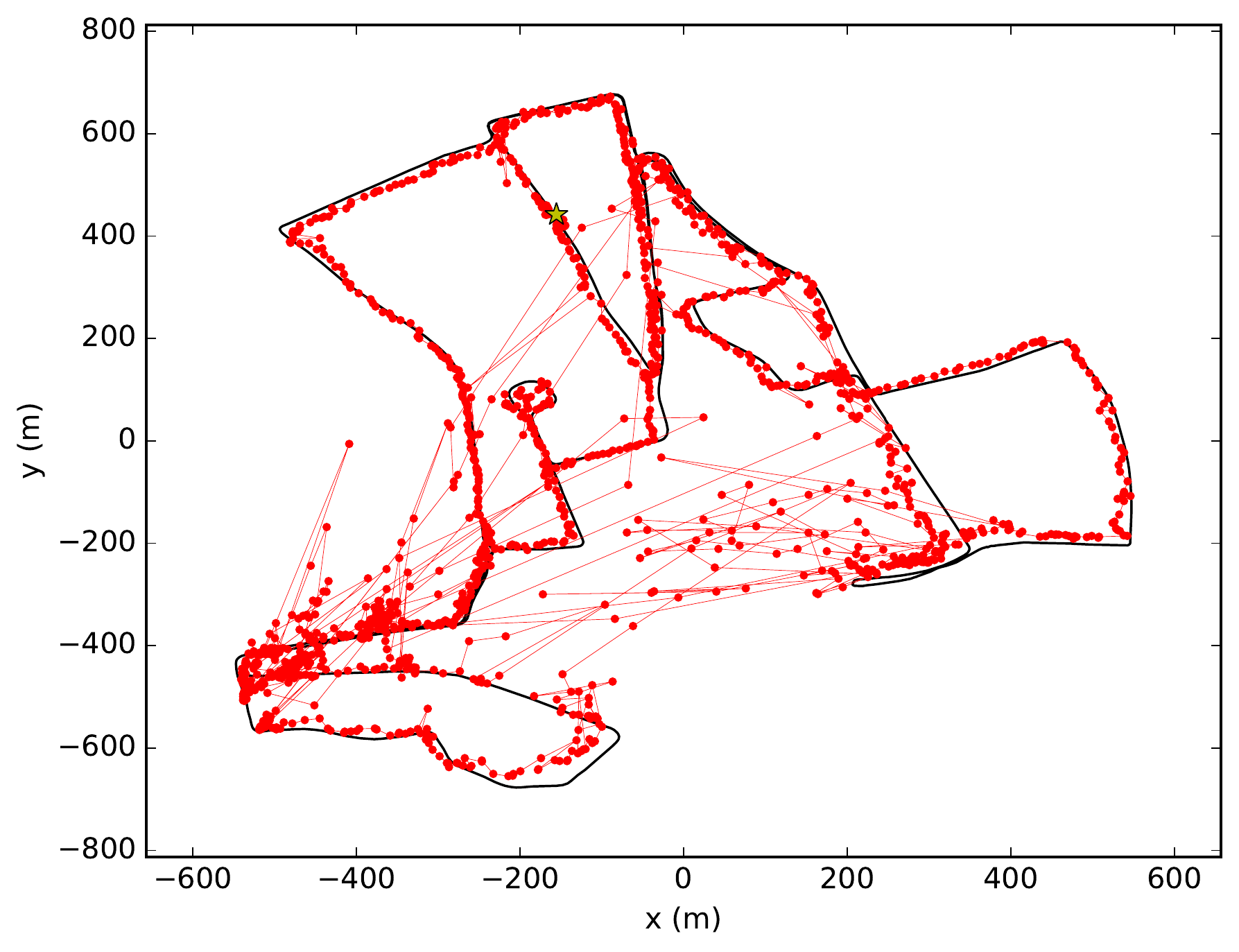}
        \vspace{-1.5em}
        \caption{\small MapNet (41.4m, 12.5\degree)}
    \end{subfigure}

    \begin{subfigure}{0.32\linewidth}
        \centering
        \includegraphics[width=0.95\linewidth]{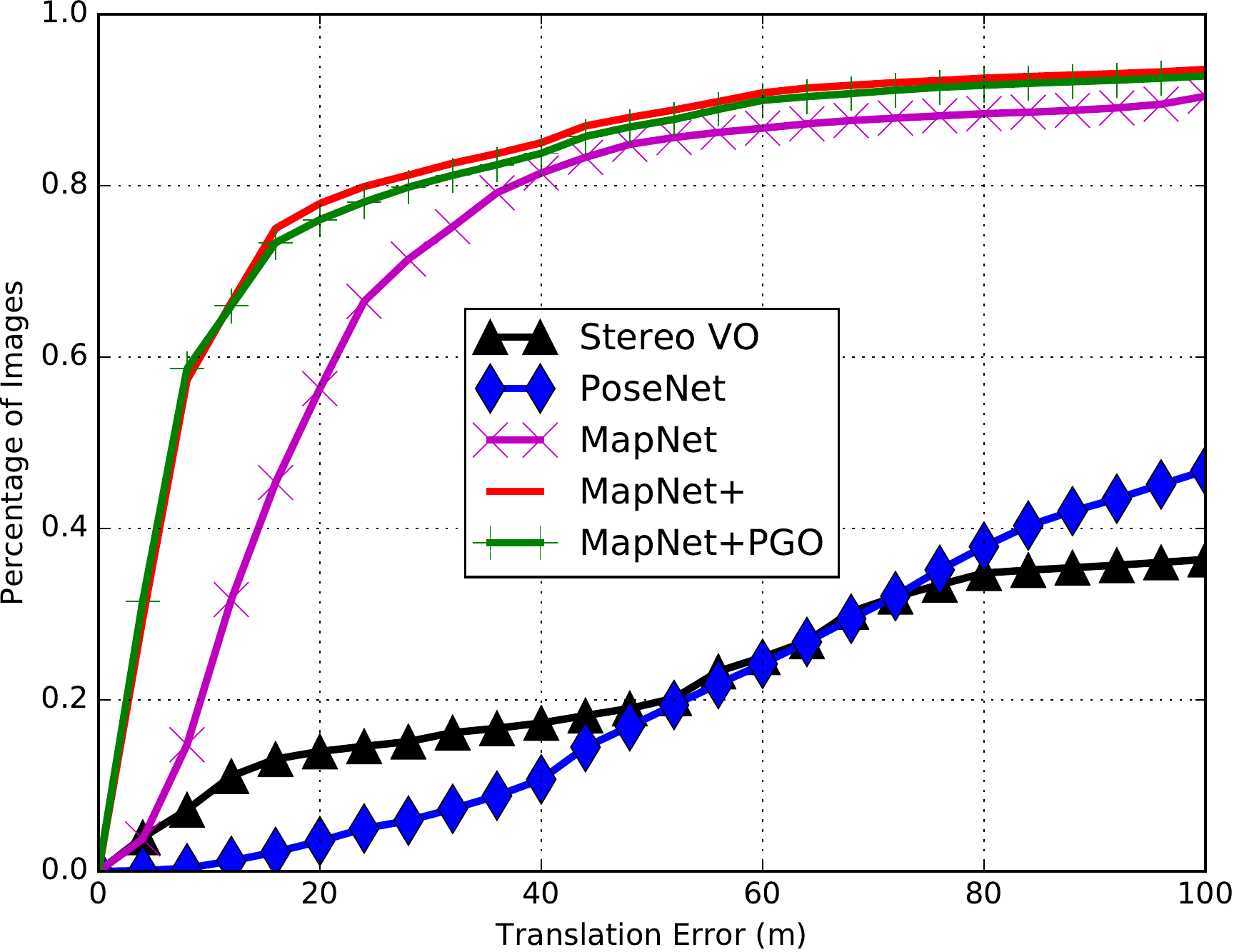}
        \vspace{-.5em}
        \caption{\small Cumulative Distribution Translation Error}
    \end{subfigure}
    \hfill
    \begin{subfigure}{0.32\linewidth}
        \centering
        \includegraphics[width=\linewidth]{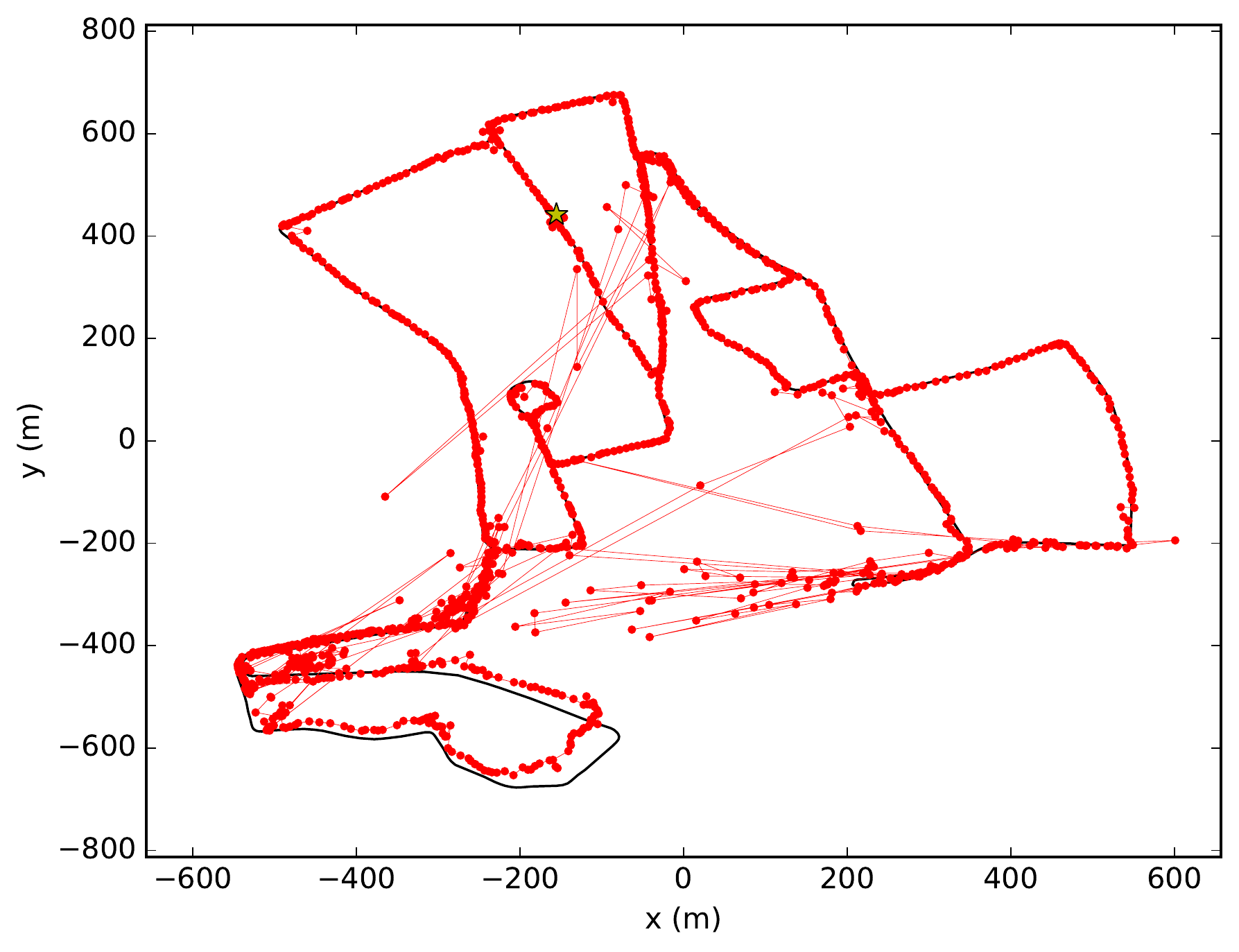}
        \vspace{-1.5em}
        \caption{\small MapNet+ (30.3m, 7.8\degree)}
    \end{subfigure}
    \hfill
    \begin{subfigure}{0.32\linewidth}
        \centering
        \includegraphics[width=\linewidth]{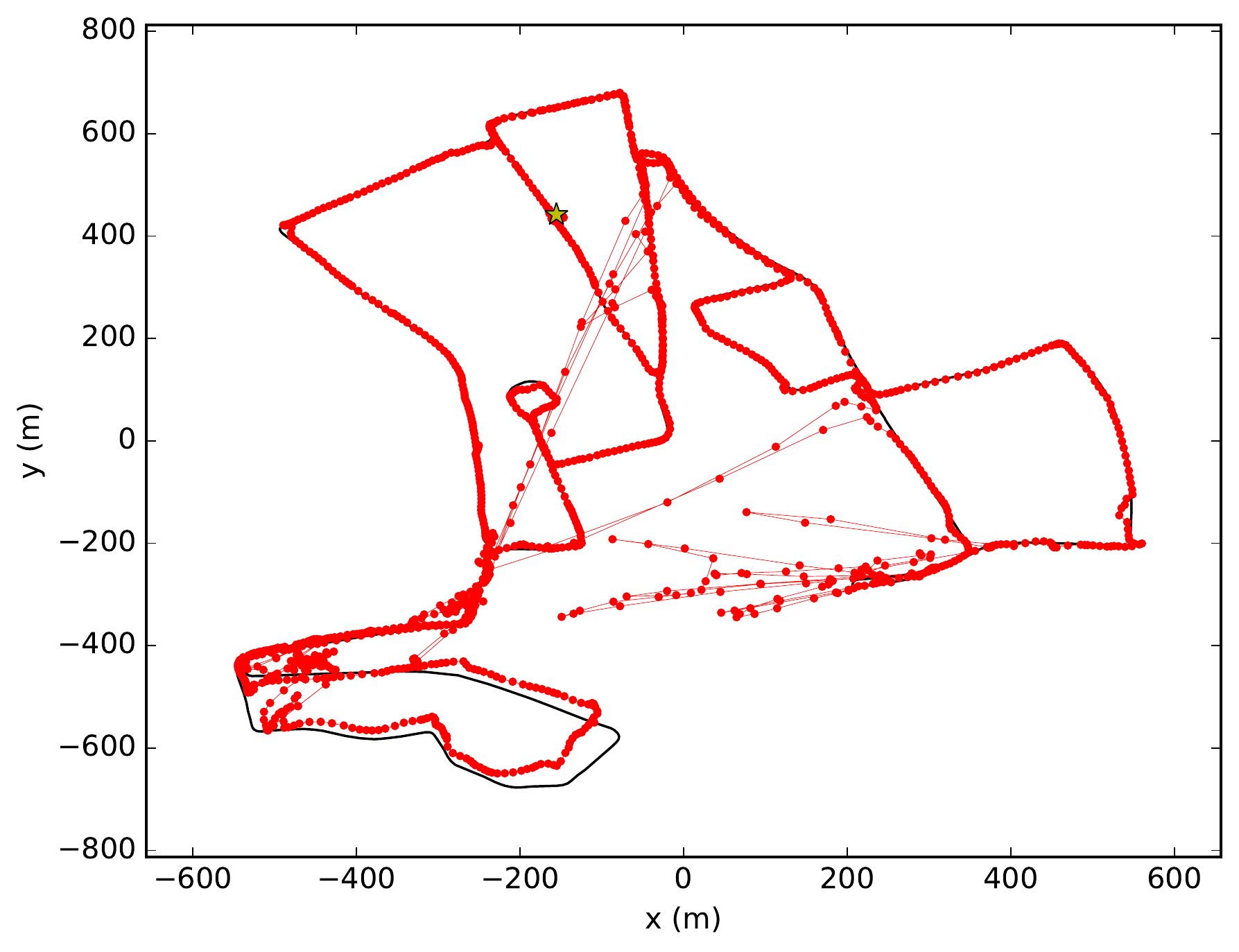}
        \vspace{-1.5em}
        \caption{\small MapNet+PGO ({\bf 29.5m, 7.8\degree})}
    \end{subfigure}
    \vspace{-1em}
    \caption{\small Comparison of camera localization results on the FULL scene (9562m long) of the
    Oxford RobotCar dataset~\cite{RobotCarDatasetIJRR}. The ground truth camera trajectory
    is the black line, and the star indicates the first frame. The red lines show the
    results of stereo VO (provided by the dataset),
    our version of PoseNet+$\log\mathbf{q}$, MapNet, and its variations. The caption
    of each figure shows the mean translation error (m) and mean rotation error (\degree).
    A plot of the cumulative distribution of the translation error is also included.}
    \label{fig:map_compare_robotcar_full}
\end{figure*}

\begin{figure}
    \centering
    \begin{subfigure}{0.48\linewidth}
        \centering
        \includegraphics[width=\linewidth]{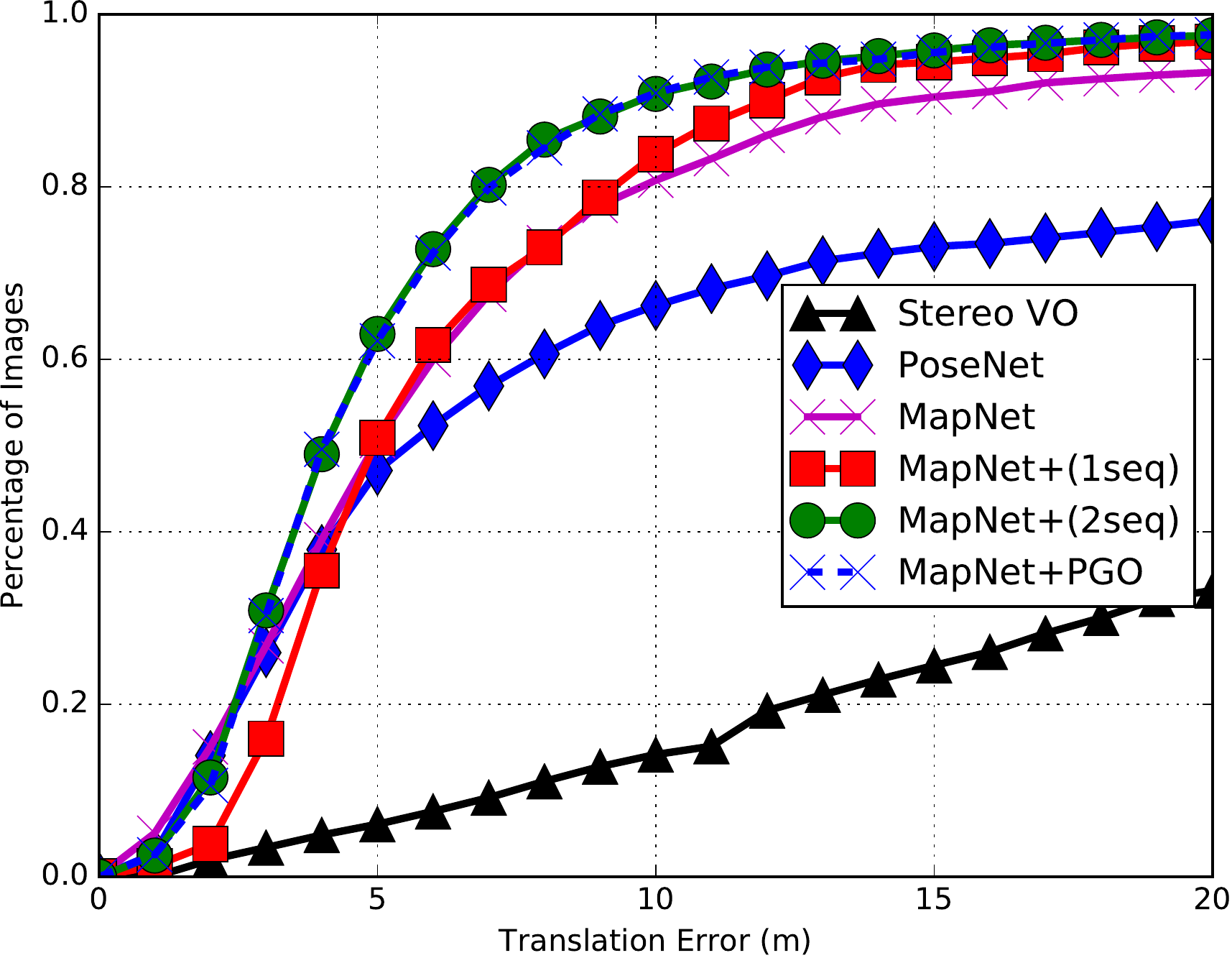}
    \end{subfigure}
    \hfill
    \begin{subfigure}{0.48\linewidth}
        \centering
        \includegraphics[width=\linewidth]{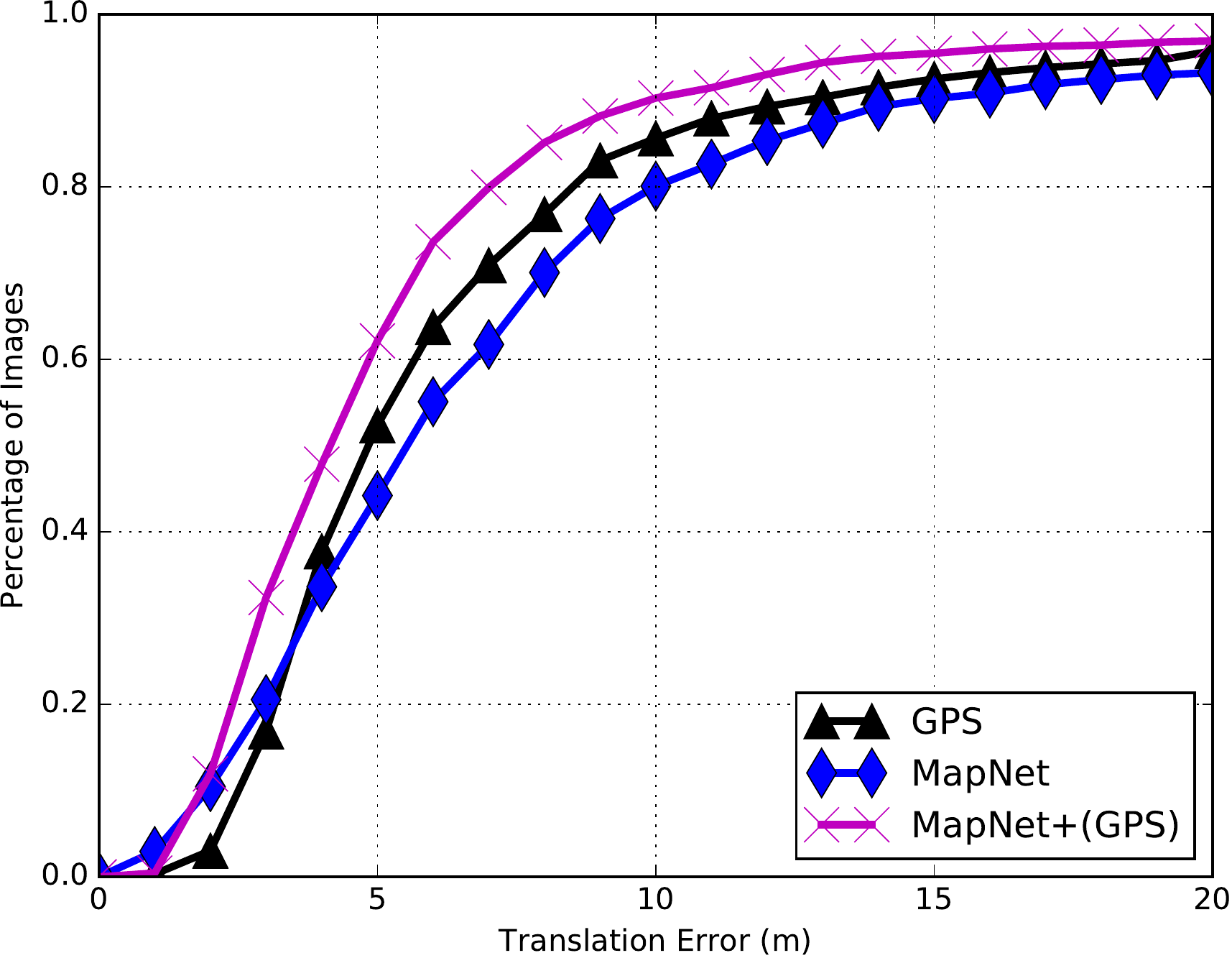}
    \end{subfigure}
    \vspace{-1em} 
    \caption{\small Cumulative distributions of the translation errors (m) for all the methods 
    evaluated on Oxford RobotCar LOOP. $x$-axis is the translation error and $y$-axis is the percentage of frames with error less than
    the value.} 
    \label{fig:res_err_robotcar}
\end{figure}

\subsection{Experiments on the Oxford RobotCar Dataset}

\paragraph{Results on the LOOP Route}
We first train a baseline PoseNet (with the $\log\mathbf{q}$ parameterization for rotation) and a MapNet model using 
two labelled sequences captured on the LOOP route under cloudy weather,
while the testing sequence is captured under sunny weather. We then perform two experiments with different auxiliary data for MapNet+.

\begin{figure}
    \centering
    \begin{subfigure}{0.49\linewidth}
        \centering
        \includegraphics[width=\linewidth]{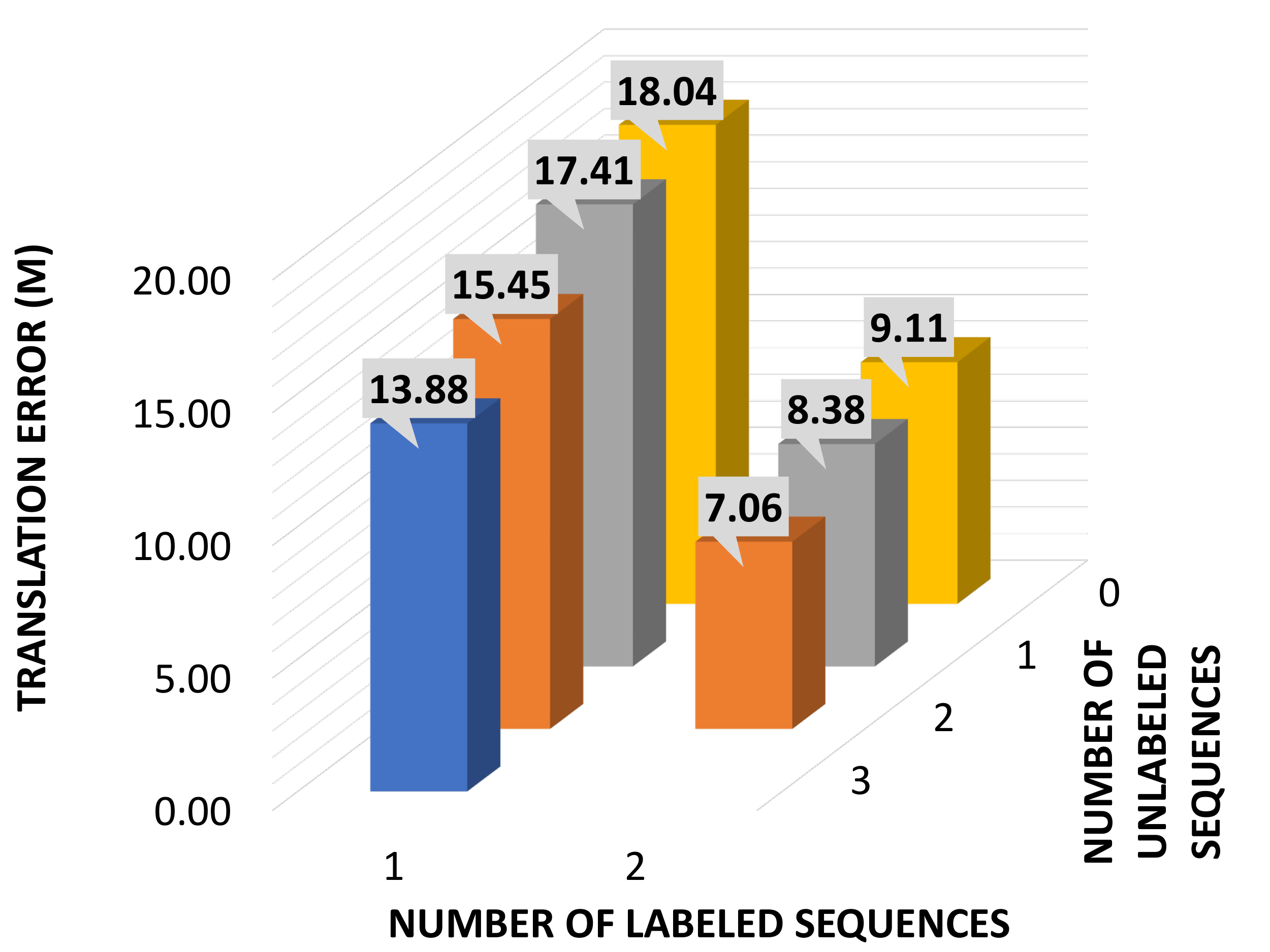}
    \end{subfigure}
    \hfill
    \begin{subfigure}{0.49\linewidth}
        \centering
        \includegraphics[width=\linewidth]{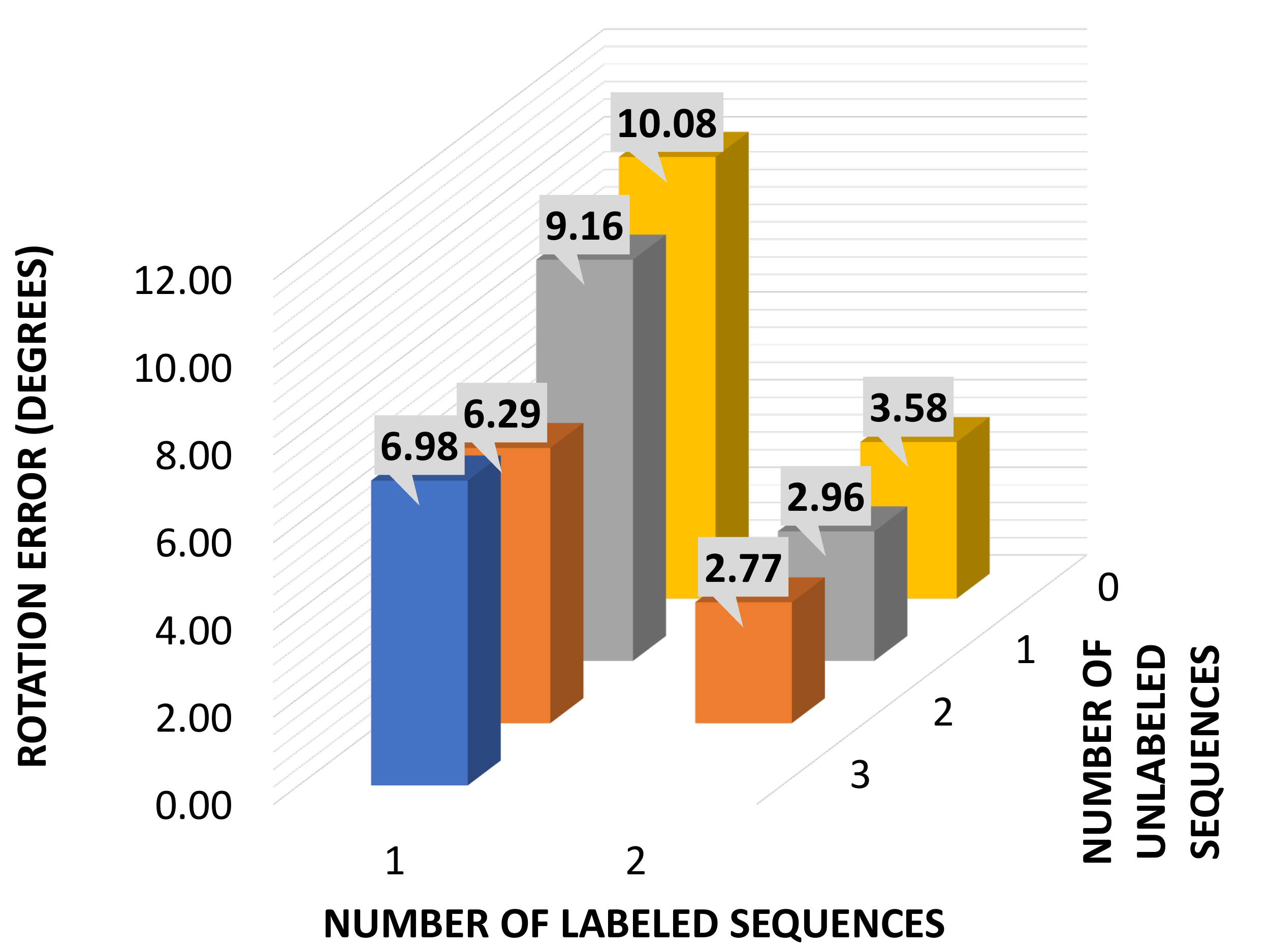}
    \end{subfigure}
    \vspace{-1em} 
    \caption{\small \textbf{Left}: Mean translation error (m) and \textbf{right}: mean rotation error (\degree) for MapNet+ models trained with 
    varying amounts of labeled and unlabeled data on the Oxford RobotCar LOOP sequence. 
    X-axis indicates the number of labeled sequences (1-2), Y-axis indicates the number of unlabeled sequences (0-3) and Z-axis indicates the error.}
    \label{fig:loop_ablation_robotcar}
\end{figure}

In the first experiment, MapNet+ is trained on additional unlabeled LOOP sequences separate from the testing sequence,
with stereo VO provided with the dataset. To tease apart the influence of labeled and unlabeled data in the effectiveness of our MapNet+ models,
we train them with varying amounts of labeled (one to two sequences)
and unlabeled data (zero to three sequences). Figure~\ref{fig:loop_ablation_robotcar} shows the mean
translation and rotation errors of these models on the testing sequence.
While labeled data is clearly more important than an equal amount of unlabeled data,
we show that unlabeled data does consistantly improve performance as more becomes available. This trend bodes well
for real-world scenarios, where the amount of unlabeled data available far exceeds the amount of labeled data.

In the second experiment, MapNet+ is trained with GPS \ie, 
the dataset $\mathcal{T}$ contains two sequences of images and their GPS locations, which are separate from the testing sequence.
Since GPS measurements are sparse
(less than 10\% of images have corresponding GPS measurements), we
first linearly interpolate GPS measurements for entire sequence. We
define the loss of auxillary data $\mathcal{T}$ in Equation~(\ref{eq:mapnet+})
as $L_{\mathcal{T}}(\Theta) = \sum_{i=1}^{\vert\mathcal{T}\vert} h(\mathbf{p_i},
\mathbf{\hat{p}_i})$, where $\mathbf{\hat{p}}_i$ is the linearly interpolated GPS measurement of the 2D
camera location. 

Figure~\ref{fig:map_compare_robotcar} shows the estimated camera poses for all
the methods in these two experiments along with the mean translation error (m) and rotation
error (\degree). 
Figure~\ref{fig:res_err_robotcar} shows the cumulative distributions of
the translation errors for all the methods on the LOOP route. In both figures, 
the left part shows that MapNet significantly improves the estimation compared to PoseNet and stereo VO.
MapNet+ and MapNet+PGO further improve the pose predictions. 
The right part shows that by fusing GPS signals,
MapNet+(GPS) obtains better results compared to MapNet and GPS alone.

\vspace{-.5em}
\paragraph{Results on the FULL Route}


We also evaluated our approach on the challenging 9562 m long FULL route of the Oxford RobotCar
dataset. Figure~\ref{fig:map_compare_robotcar_full} shows the results of all the
models with mean translation error (m) and rotation error (\degree). 
MapNet significantly outperforms the baseline PoseNet (both trained for 100 epochs) and the stereo VO (provided by the dataset).
By fusing the stereo VO information, MapNet+ and MapNet+PGO further improve the result.
The cumulative distributions of the translation errors also show the large improvement over
the baselines. 

Note that there are some outlier predictions in both LOOP (Fig.~\ref{fig:map_compare_robotcar}) and FULL (Fig.~\ref{fig:map_compare_robotcar_full}).
These often correspond to images with large over-exposed regions, and can be filtered out with simple post-processing (e.g. temporal median filtering)
as shown in the supplementary material. We also computed saliency maps $s(x,y) = \frac{1}{6}|\sum_{i=1}^6 \frac{\partial p_i}{\partial I(x,y)}|$ (magnitude gradient of the mean of the 6-element output w.r.t. input image, maxed over the 3 color channels) for PoseNet
and MapNet+ on both the 7-scenes and RobotCar dataset. We find that compared to PoseNet, MapNet+ focuses more on geometrically meaningful regions
and its saliency map is more consistent over time. Examples are shown in the supplementary material.

\section{Conclusions and Discussions}
\label{sec:conclusion}

In summary, MapNet learns a
\emph{general, data-driven} map representation for camera localization.
Our models bring geometric constraints widely used in visual SLAM
and SfM into DNN-based learning, which allow us to learn from
unlabeled data and to easily fuse other input sources (\eg,
visual odometry, GPS, IMU).  We evaluate our approach on both indoor and outdoor datasets and
show significantly better performance than baselines.

Unlike the mapping in traditional visual SLAM systems, MapNet and
MapNet+ cannot expand maps to unknown space. In future work, a tighter integration with visual
SLAM systems may enable mapping of unknown regions. Leveraging the recent success in
extracting high-level semantic information (\eg, objects and scene
composition) may also improve camera localization.

%
%

\begin{abstract}
    In this supplementary document, we provide more implementation details of our method,
    names of the sequences used in the experiments on the Oxford RobotCar dataset,
    more analysis and visualization of the experimental results presented in the main paper, 
    and the detailed derivation of pose-graph optimization (PGO) in MapNet+PGO.
    Please also refer to the supplementary video for more visualizations of the camera localization results.
\end{abstract}

\section*{Pose Graph Optimization in MapNet+PGO}

The purpose of pose-graph optimization (PGO) is to refine the input poses such that the refined poses are close to the input poses (from MapNet+), and the relative transforms between the refined poses agree with the input visual odometries. It is an iterative optimization process~\cite{grimes10pgo, grisetti2010tutorial}.

\paragraph{Inputs}
Pose predictions $\left\lbrace\mathbf{p_i}\right\rbrace_{i=1}^{T}$ and visual odometry (VOs) $\mathbf{\hat{v}}_{ij}$ between consecutive poses. Both poses and VOs are 6-dimensional (3d translation $\mathbf{t}$ + 3d log quaternion $\mathbf{w}$). For the rest of the algorithm, the log quaternions are converted to unit quaternion using the exponential map~\cite{Hertzberg08quaternion}:
\begin{equation}\label{eq:qexp}
\mathbf{q} = (\cos \Vert\mathbf{w}\Vert, \frac{\mathbf{w}}{\Vert\mathbf{w}\Vert}\sin \Vert\mathbf{w}\Vert)
\end{equation}

\paragraph{Objective Function}
State vector $z$ is the concatenation of all $T$ pose vectors. The total objective function is the sum of the costs of all constraints. The constraints can be either for the absolute pose or for the relative pose between a pair of poses. For both of these categories, there are separate constraints for translation and rotation.
\begin{align}
E(z) &= \sum_c E_c(z)\nonumber\\
&= \sum_c \bar{h}(f_c(c), k_c)
\end{align}
where $\bar{h}(\cdot)$ is the pose distance function from Equation~(7) of the main paper. $f_c$ is a function that maps the state vector to the quantity relevant for the constraint $c$. For example, it selects $\mathbf{p}_i$ from the state vector for a constraint on the absolute pose, or computes the VO between poses for $\mathbf{p}_i$ and $\mathbf{p}_j$ for a constraint on the relative pose. $k_c$ is the observation for that constraint, and remains constant throughout the optimization process. For example:
\begin{itemize}
\item For the absolute pose constraints, $k_c$ is the MapNet+ prediction.
\item For the relative pose constraints, $k_c$ is the input VO $\mathbf{\hat{v}}_{ij}$.
\end{itemize}
Following~\cite{grimes10pgo, grisetti2010tutorial}, we define $\bar{h}(\cdot)$ as:
\begin{equation}
\bar{h}(f_c(c), k_c) = (f_c(z) - k_c)^T S_c (f_c(z) - k_c)
\end{equation}
where $S_c$ the covariance matrix for the constraint.

\paragraph{Optimization}
Following \cite{grimes10pgo}, We first linearize $f_c$ around $\bar{z}$, the current value of $z$:
\begin{equation}
f_c(\bar{z} + \Delta z) \approx f_c(\bar{z}) + \frac{\partial f_c}{\partial z}\bigg\rvert_{z=\bar{z}}\Delta z
\end{equation}
and take the Cholesky decomposition of $S_c$: $S_c = L_cL_c^T$. Hence the linearized objective function becomes:
\begin{align}
E(\Delta z) &= \sum_c (f_c(\bar{z} + \Delta z) - k_c)^T S_c (f_c(\bar{z} + \Delta z) - k_c)\nonumber\\
&\approx \sum_c \left(f_c(\bar{z}) + \frac{\partial f_c}{\partial z}\bigg\rvert_{z=\bar{z}}\Delta z - k_c\right)^T L_c\nonumber\\
&L_c^T \left(f_c(\bar{z}) + \frac{\partial f_c}{\partial z}\bigg\rvert_{z=\bar{z}}\Delta z - k_c\right)\nonumber\\
&= \sum_c \bigg\rvert\bigg\rvert L_c^T \left( f_c(\bar{z}) + \frac{\partial f_c}{\partial z}\bigg\rvert_{z=\bar{z}}\Delta z - k_c \right) \bigg\lvert\bigg\lvert^2\nonumber\\
&= \sum_c ||J_c \Delta z - r_c||^2
\end{align}
where \textbf{Jacobian} $J_c = L_c^T \frac{\partial f_c}{\partial z}\bigg\rvert_{z=\bar{z}}$ and \textbf{residue} $r_c = L_c^T (k_c - f_c(\bar{z}))$. We will solve for $\Delta z$.\par

Stacking the individual Jacobians and residuals vertically, we arrive at the least squares problem:
\begin{equation}
\Delta z^* = \min_{\Delta z} || J \Delta z - r ||^2
\end{equation}
This can be solved by $\Delta z^* = (J^TJ)^{-1}J^Tr$.\par
Finally, we update the state vector:
\begin{equation}\label{eq:update}
z = z \boxplus \Delta z
\end{equation}
where $\boxplus$ is the manifold update operation, needed because of the quaternions (more details below).

\paragraph{Detour: Manifolds for Quaternion Update}\label{sec:manifold}
As mentioned in~\cite{grisetti2010tutorial}, if we had used a simple addition in the update Equation~(\ref{eq:update}), it would have broken the constraints introduced by the over-parameterization of quaternions. So we use manifolds. According to~\cite{grisetti2010tutorial}, ``A mainfold is a space that is not necessarily Euclidean in a global scale, but can be seen as Euclidean on a local scale". The idea is to calculate the update for quaternion in a minimal 3d representation, and then apply this update to the 4d representation of quaterinion in $z$ using $\boxplus$. We use the ``exponential map''~\cite{Hertzberg08quaternion} to implement $\boxplus$.
For this, we re-cast the objective function as a function of the update on the manifold, $\Delta \breve{z}$:
\begin{equation}
E(\Delta \breve{z}) = \sum_c (f_c(\bar{z} \boxplus \Delta \breve{z}) - k_c)^T S_c (f_c(\bar{z} \boxplus \Delta \breve{z}) - k_c)
\end{equation}
The linearization step is:
\begin{equation}
f_c(\bar{z} \boxplus \Delta \breve{z})
\approx f_c(\bar{z}) + \frac{\partial f_c}{\partial z}\bigg\rvert_{z=\bar{z}} \frac{\partial \bar{z} \boxplus \Delta \breve{z}}{\partial \Delta \breve{z}}\bigg\rvert_{\Delta\breve{z}=0}\Delta \breve{z}\\
\end{equation}
So the Jacobian in this case is:
\begin{equation}\label{eq:qj1}
\breve{J_c} = J_c \frac{\partial \bar{z} \boxplus \Delta \breve{z}}{\partial \Delta \breve{z}}\bigg\rvert_{\Delta\breve{z}=0}
\end{equation}
Let us see how $\bar{z} \boxplus \Delta \breve{z}$ is implemented.
\begin{equation}
\bar{z} \boxplus \Delta \breve{z} = \bar{z} \cdot \Delta\tilde{z}
\end{equation} where $\Delta\tilde{z}$ is the normal 4d quaternion that has been created from the 3d minimal representation $\Delta\breve{z}$ using the exponential map (Equation~(\ref{eq:qexp})).
So the derivative of the exponential map at $\Delta\breve{z} = 0$ is $M_e = \begin{bmatrix}0 & 0 & 0 \\ 1 & 0 & 0 \\ 0 & 1 & 0 \\ 0 & 0 & 1\end{bmatrix}$.
Hence,
\begin{align}
\frac{\partial \bar{z} \boxplus \Delta \breve{z}}{\partial \Delta \breve{z}} \bigg\rvert_{\Delta\breve{z}=0}
&= \frac{\partial \bar{z} \cdot \Delta\tilde{z}}{\partial \Delta\tilde{z}} \frac{\partial \Delta\tilde{z}}{\partial \Delta\breve{z}} \bigg\rvert_{\Delta\breve{z}=0}\nonumber\\
&= \frac{\partial \bar{z} \cdot \Delta\tilde{z}}{\partial \Delta\tilde{z}} M_e\label{eq:qj2}
\end{align}
For the first term, we use the formula for derivative of quaternion product from~\cite{quaternion_math}.

\paragraph{Jacobian of Absolute Translation Constraint}
$f_c$ just selects the appropriate 3 translation elements of a pose from the state vector $z$, so $\breve{J_c} = L_c^T [\mathbf{0}, \ldots, I_3, \ldots, \mathbf{0}]$.

\paragraph{Jacobian of Absolute Rotation Constraint}
$f_c$ selects the appropriate 4 quaternion elements of a pose from the state vector $z$. However, since the update is on the manifold, the Jacobian $\breve{J_c}$ is computed as shown in Equations~(\ref{eq:qj1}) and~(\ref{eq:qj2}) with $J_c = L_c^T \cdot I_4$.

\paragraph{Jacobian of Relative Translation Constraint}
$f_c$ computes the translation component of the VO $\mathbf{v}_{ij}$ between $\mathbf{p}_{i}$ and $\mathbf{p}_{j}$, which is $\mathbf{q}_j (\mathbf{t}_i-\mathbf{t}_j)\mathbf{q}_j^{-1}$ according to Equation~(6) in the main paper. Hence $J_c$ has $\frac{\partial \mathbf{q}_j\mathbf{t}_i\mathbf{q}_j^{-1}}{\partial \mathbf{t}_i}$ in the block corresponding to $\mathbf{t}_i$ and $-\frac{\partial \mathbf{q}_j\mathbf{t}_j\mathbf{q}_j^{-1}}{\partial \mathbf{t}_j}$ in the block corresponding to $\mathbf{t}_j$. Both these formulae can be found in~\cite{quaternion_math}.

\paragraph{Jacobian of Relative Rotation Constraint}
$f_c$ computes the rotation component of the VO $\mathbf{v}_{ij}$ between $\mathbf{p}_{i}$ and $\mathbf{p}_{j}$, which is $q_j^{-1} \cdot q_i$ according to Equation~(6) in the main paper. Hence $J_c$ has $\frac{\partial \mathbf{q}_j^{-1} \cdot \mathbf{q}_i}{\partial \mathbf{q}_i}$ in the Jacobian block corresponding to $\mathbf{q}_i$ and $\frac{\partial \mathbf{q}_j^{-1} \cdot \mathbf{q}_i}{\partial \mathbf{q}_j}$ in the Jacobian block corresponding to $\mathbf{q}_j$. Both these formulae can be found in~\cite{quaternion_math}.

\paragraph{Update on the Manifold}\label{sec:update}
The updates for translation parts of the state vector are performed by simply adding the update vector to the state vector. For the quaternion parts, the minimal representations in the update need to be converted back to the 4d representation using the exponential map in Equation~(\ref{eq:qexp}), and then quaternion-multiplied to the state vector quaternions.

\paragraph{Implementation Details}
The covariance matrix $S_c$ is set to identity for all the translation constraints and tuned to $\sigma I_3$ ($\sigma$=10 to 35) for different scenes in the 7-Scenes dataset. For the RobotCar dataset, we use $\sigma=20$ for LOOP and $\sigma=10$ for FULL.

\section*{Details of Image Pair Sampling}

In both MapNet and MapNet+ (Sections 3.2 and 3.3 of the main paper) training, we need to
sample image pairs $(\mathbf{I}_i,\mathbf{I}_j)$ from each input image
sequence. This sampling is done within each tuple of $s$ images sampled with a
gap $k$ frames. More specifically, suppose we have $N$ images in an
input sequence, $\mathbf{I}_1,\cdots,\mathbf{I}_N$. Each entry in each minibatch during the training
of MapNet and MapNet+ consists of a tuple of $s$ conseutive images that are $k$ frames apart from each other, \ie, 
$(\mathbf{I}_{i}, \mathbf{I}_{i+k}, \cdots, \mathbf{I}_{i+k(s-2)}, \mathbf{I}_{i+k(s-1)})$.

\begin{table*}[!ht]
    \small
    \centering
    \caption{Statistics of state-of-the-art methods on the 7-Scenes dataset.}
    \label{tab:res_7scenes_stats}
    \begin{tabular}{lll|lll}
        \toprule
        Scene   & PoseNet+$\log\mathbf{q}$ & DSO~\cite{Engel2017DSO} & MapNet & MapNet+ & MapNet+PGO \\
        \midrule
        Avg Median (Scene)  &  0.23m, 8.49   & 0.51m, 29.44 & 0.21m, 7.77 & 0.19m, 7.29  & {\bf 0.18m,  6.55} \\
        Avg Median (Seq)    &  0.24m, 7.40   & 0.93m, 39.20 & 0.22m, 6.88 & {\bf 0.20m}, 6.18 & 0.21m, {\bf 6.16} \\ 
        Avg Mean (Scene)    &  0.28m, 10.43  & 1.27m, 46.48 & 0.27m, 10.08 & 0.23m, 8.27 & {\bf 0.22m, 7.89}  \\
        Avg Mean (Seq)      &  0.30m, 9.84   & 1.62m, 40.28 & 0.28m, 9.12 & 0.24m, 7.42  & {\bf 0.23m, 7.29} \\
        \bottomrule
    \end{tabular}
\end{table*}
\FloatBarrier
Within this tuple
of $s$ images, each two neighboring elements will form an image pair for training. For example,
both $(\mathbf{I}_{i}, \mathbf{I}_{i+k})$ and 
$(\mathbf{I}_{i+k(s-2)}, \mathbf{I}_{i+k(s-1)})$ are valid image pairs.

\section*{Details of the Sequences used in the Experiments on the RobotCar Dataset}
Sequences in RobotCar are named by the date and time of their capture.
\paragraph{Experiments on the LOOP Scene} 

To train the baseline PoseNet and MapNet, we used the following two sequences as the dataset $\mathcal{D}$ with ground truth supervision.

\begin{itemize}
    \item 2014-06-26-09-24-58
    \item 2014-06-26-08-53-56
\end{itemize}
We used the following two sequences as the unlabeled dataset $\mathcal{T}$ to train MapNet+
\begin{itemize}
    \item 2014-05-14-13-50-20
    \item 2014-05-14-13-46-12
\end{itemize}
MapNet+(1seq) used the first sequence in $\mathcal{T}$, and MapNet+(2seq) used both sequences in $\mathcal{T}$.
These two sequences are also used in MapNet+(GPS) for updating the MapNet with GPS measurements. 

We used the following sequences for testing, which are completely separated from all the sequences in $\mathcal{D}$ and $\mathcal{T}$.
\begin{itemize}
    \item 2014-06-23-15-36-04
    \item 2014-06-23-15-41-25
\end{itemize}
Figure 5 in our main paper showed the testing results on 2014-06-23-15-41-25 for visualization (we obtained similar results on the other testing sequence).\\

\noindent \textbf{Figure 8 of the main paper}: The MapNet+ model trained with one sequence of labeled data used $\mathcal{D}=\lbrace$2014-06-26-09-24-58$\rbrace$ and
increasingly larger subsets of unlabeled data $\mathcal{T}=\lbrace$2014-06-26-08-53-56, 2014-05-14-13-50-20, 2014-05-14-13-46-12$\rbrace$.
The MapNet+ model trained with 2 sequences of labeled data used $\mathcal{D}=\lbrace$2014-06-26-09-24-58, 2014-06-26-08-53-56$\rbrace$ and
increasingly larger subsets of unlabeled data $\mathcal{T}=\lbrace$2014-05-14-13-50-20, 2014-05-14-13-46-12$\rbrace$. All these models were tested on 2014-06-23-15-36-04.

\paragraph{Experiments on the FULL Scene} 
To train the baseline PoseNet and MapNet, we used the following two sequences as the labeled dataset $\mathcal{D}$
\begin{itemize}
    \item 2014-11-28-12-07-13
    \item 2014-12-02-15-30-08
\end{itemize}
We used the following sequence as the unlabeled dataset $\mathcal{T}$
\begin{itemize}
    \item 2014-12-12-10-45-15
\end{itemize}
We used the following sequence for testing, which is completely separated from all the learning methods 
\begin{itemize}
    \item 2014-12-09-13-21-02
\end{itemize}

\section*{Experiments on the 7-Scenes Dataset}
\begin{figure*}
    \captionsetup[subfigure]{labelformat=empty}
    \centering
    \begin{subfigure}{0.15\linewidth}
        \centering
        \includegraphics[width=\linewidth]{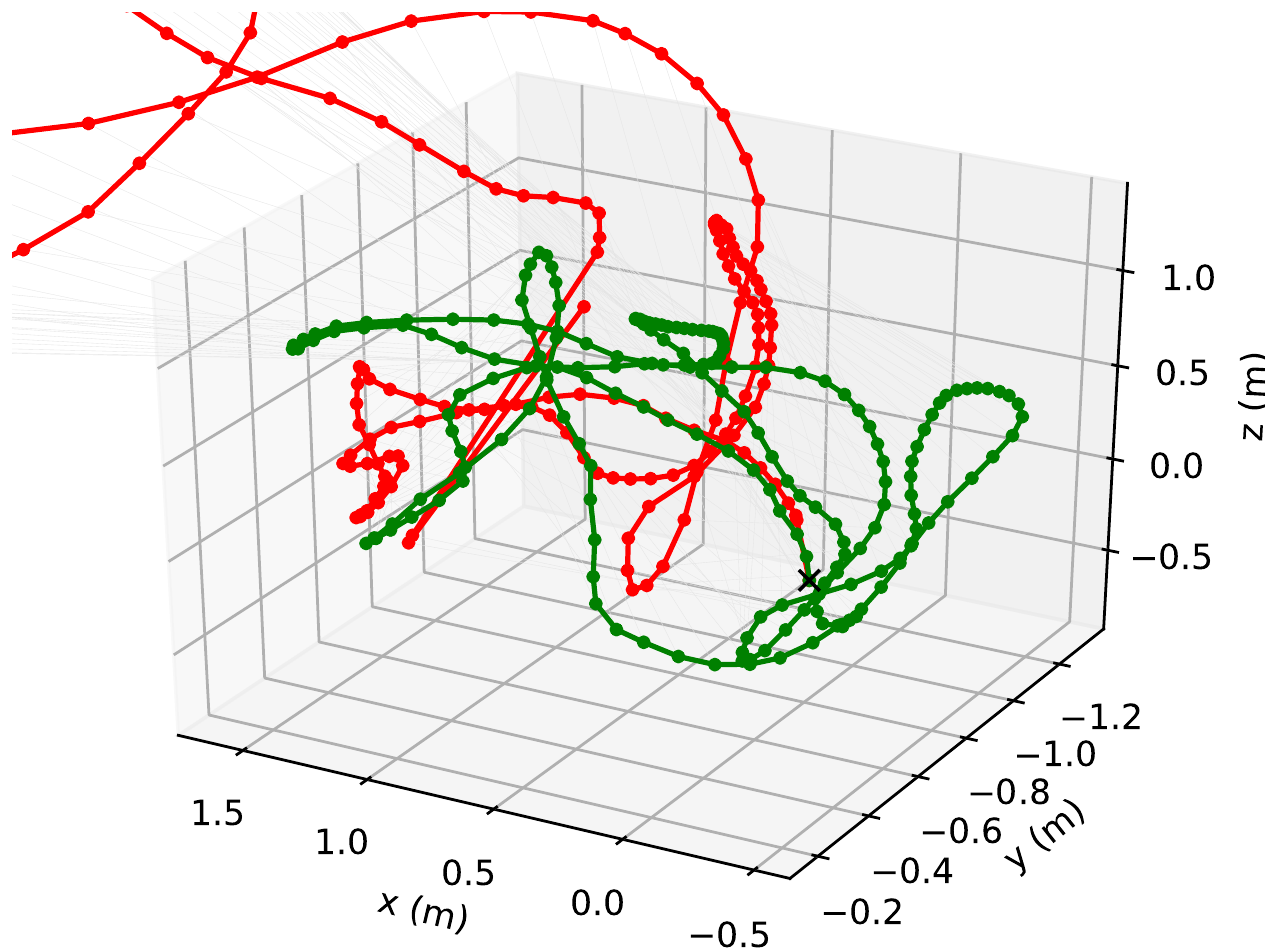}
        \includegraphics[width=\linewidth]{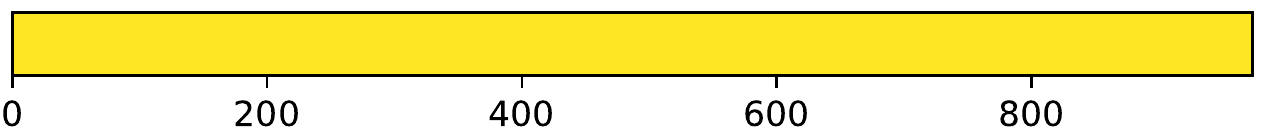}
    \end{subfigure}
    \hfill
    \begin{subfigure}{0.15\linewidth}
        \centering
        \includegraphics[width=\linewidth]{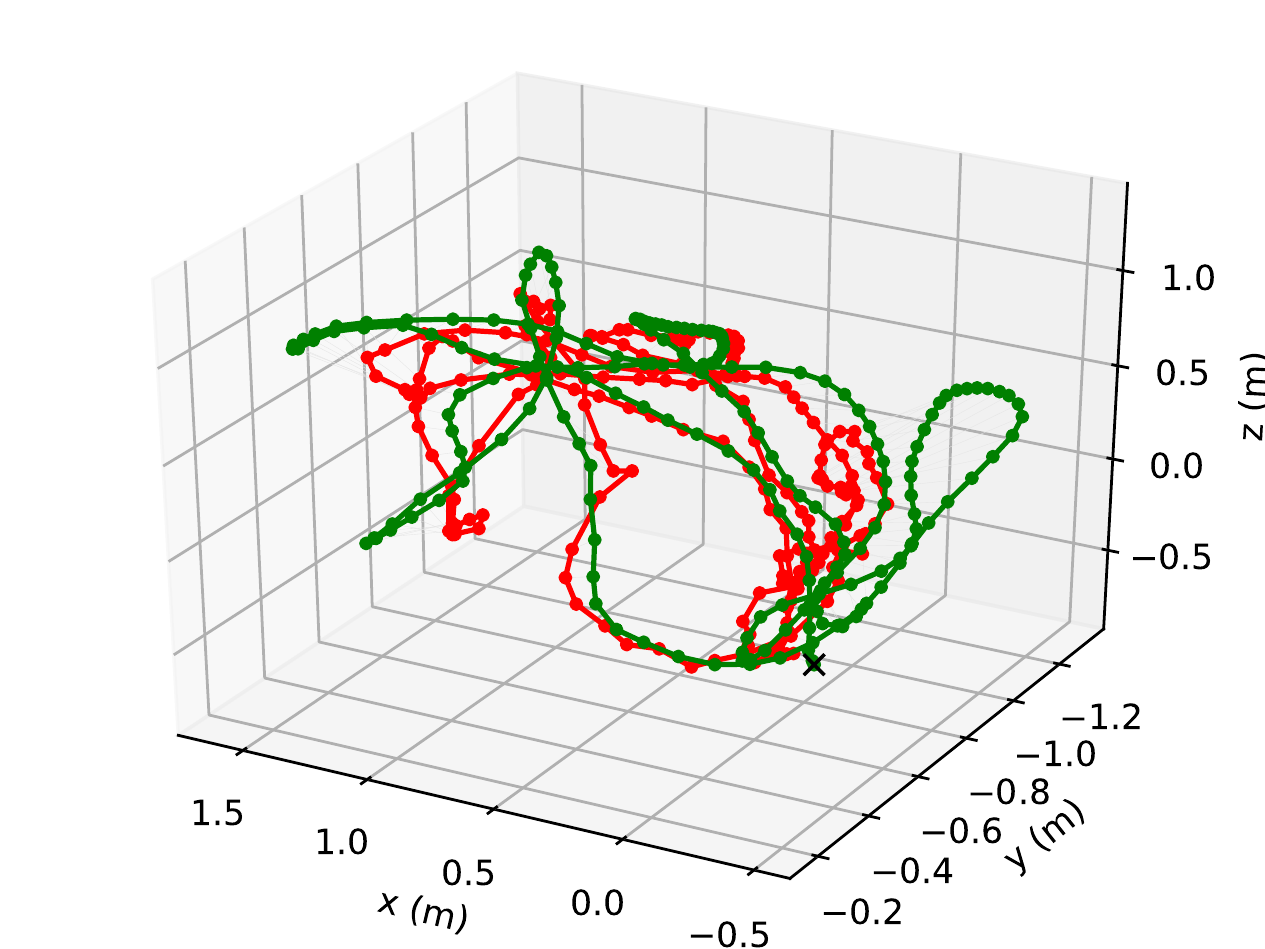}
        \includegraphics[width=\linewidth]{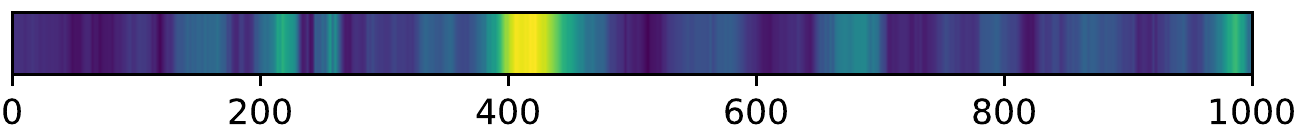}
    \end{subfigure}
    \hfill
    \begin{subfigure}{0.15\linewidth}
        \centering
        \includegraphics[width=\linewidth]{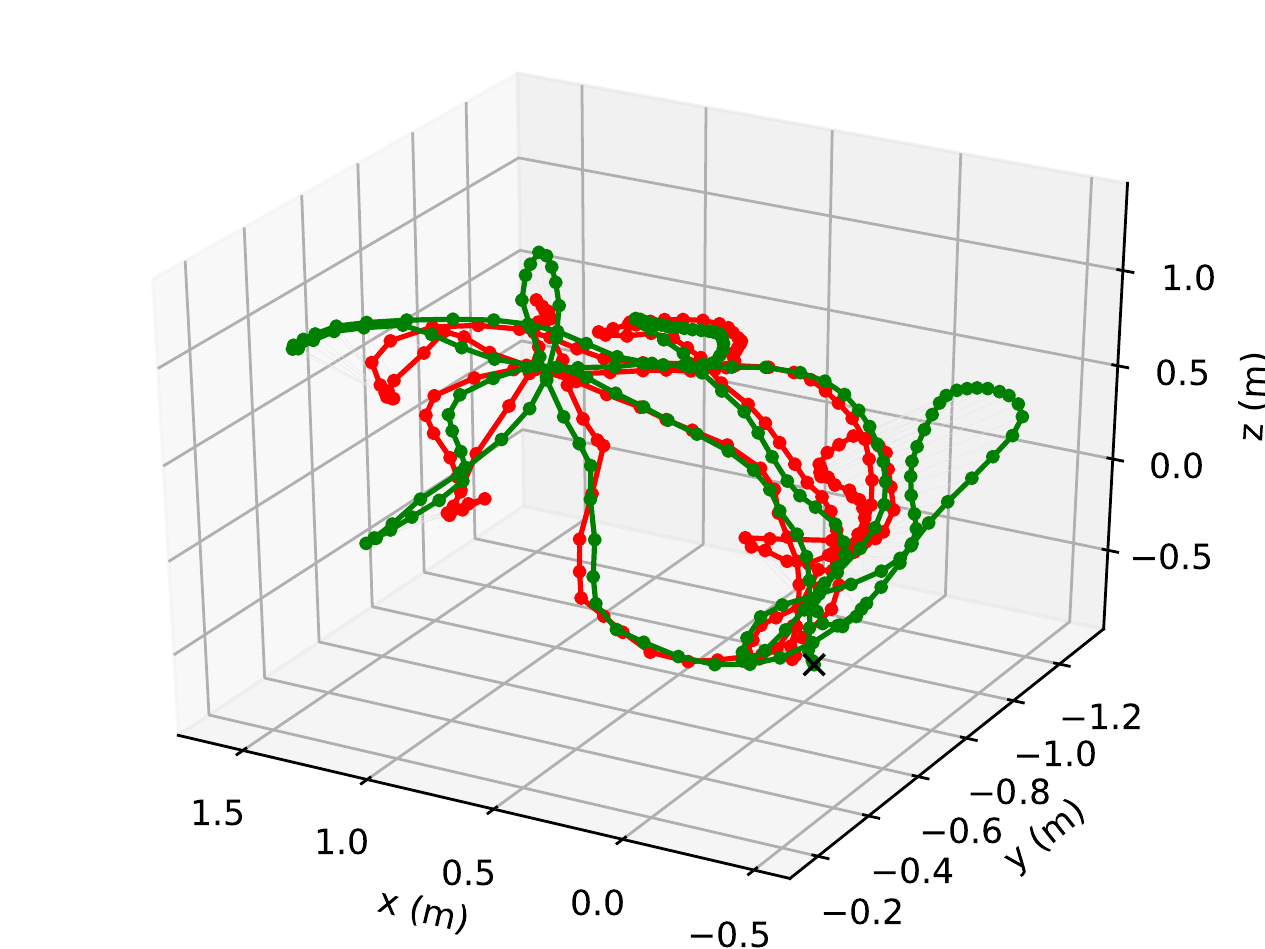}
        \includegraphics[width=\linewidth]{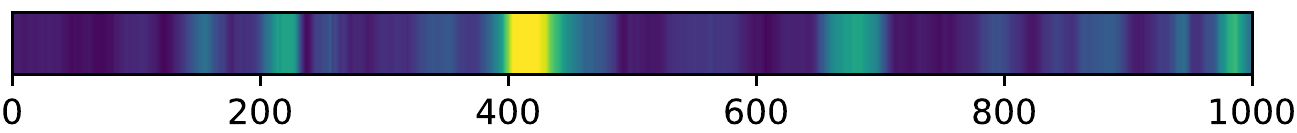}
    \end{subfigure}
    \hfill
    \begin{subfigure}{0.15\linewidth}
        \centering
        \includegraphics[width=\linewidth]{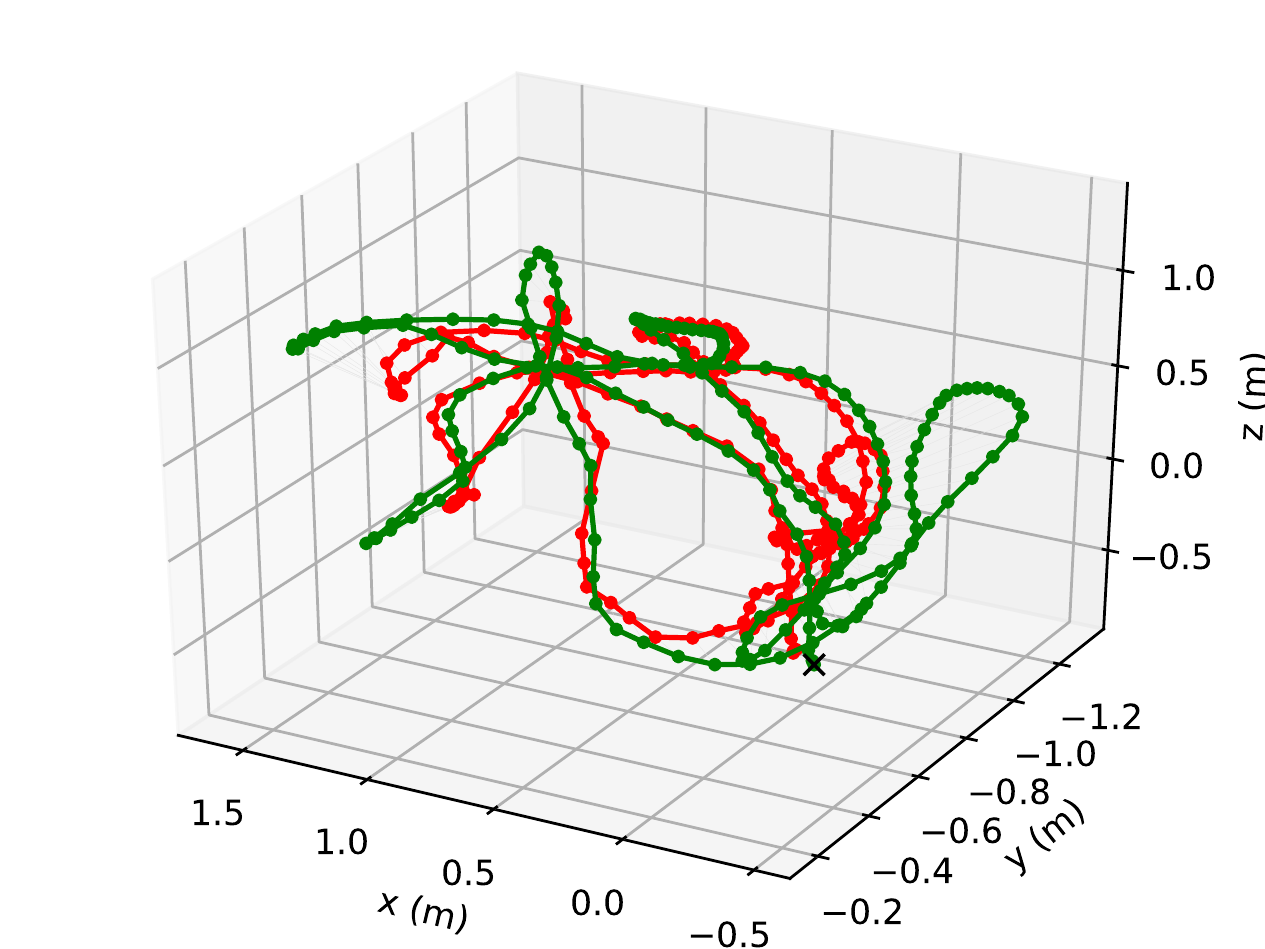}
        \includegraphics[width=\linewidth]{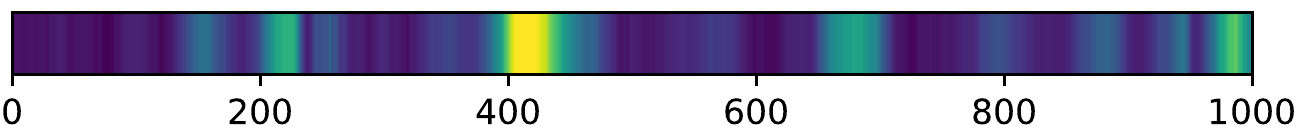}
    \end{subfigure}
    \hfill
    \begin{subfigure}{0.15\linewidth}
        \centering
        \includegraphics[width=\linewidth]{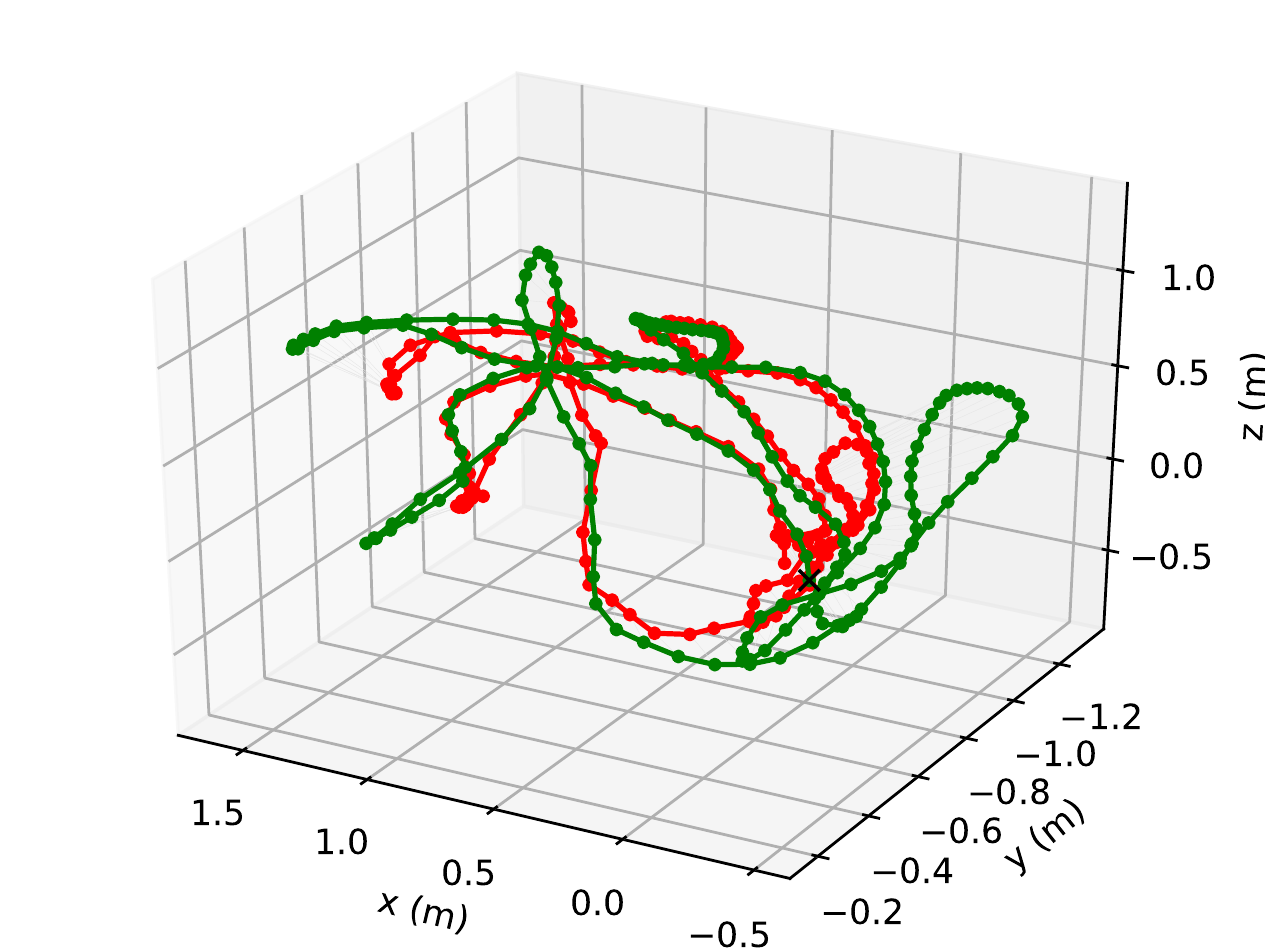}
        \includegraphics[width=\linewidth]{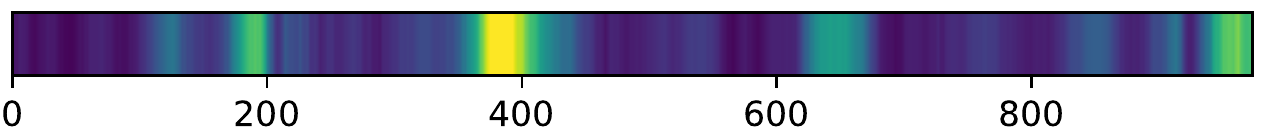}
    \end{subfigure}

    \begin{subfigure}{0.15\linewidth}
        \centering
        \includegraphics[width=\linewidth]{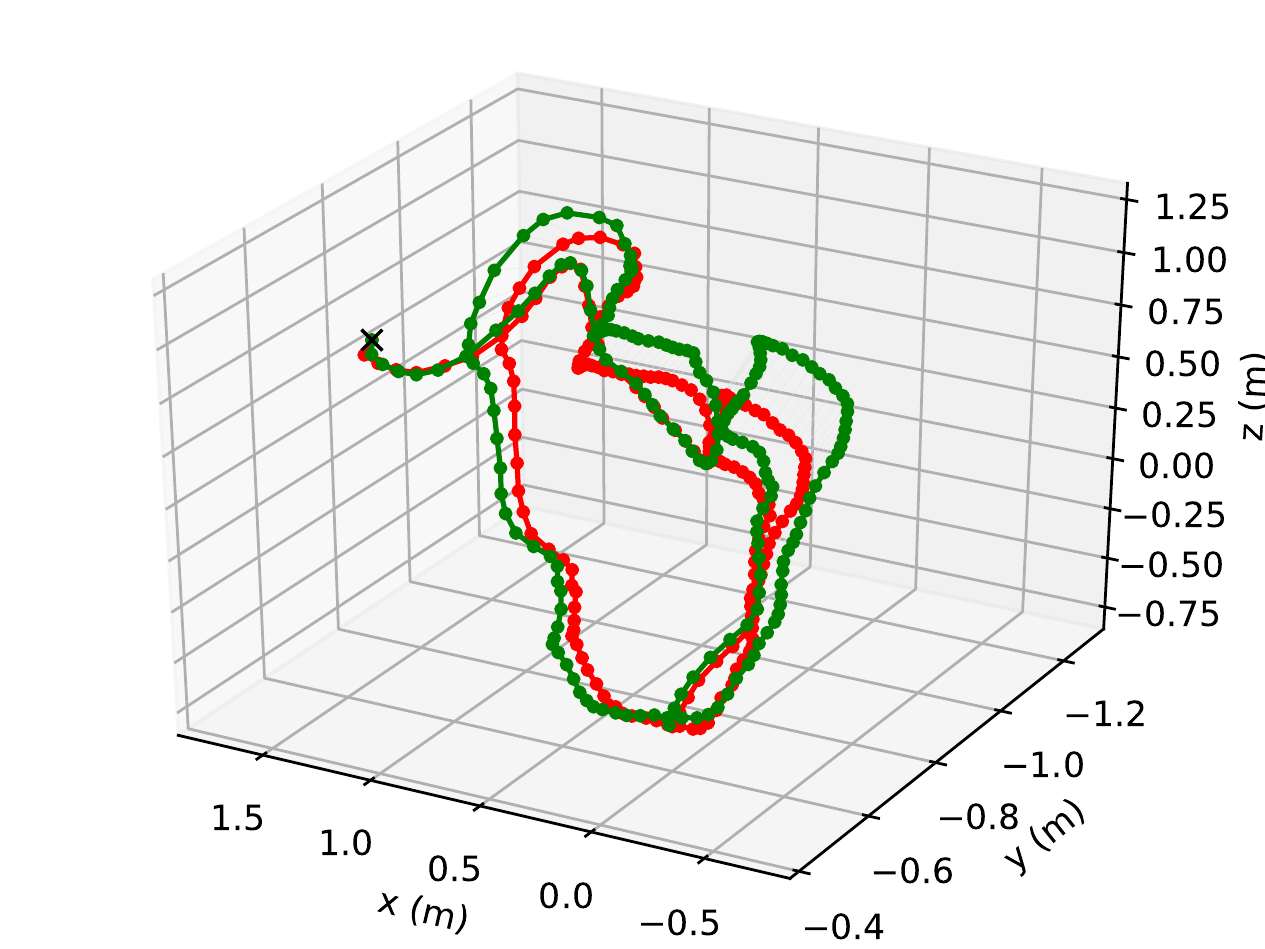}
        \includegraphics[width=\linewidth]{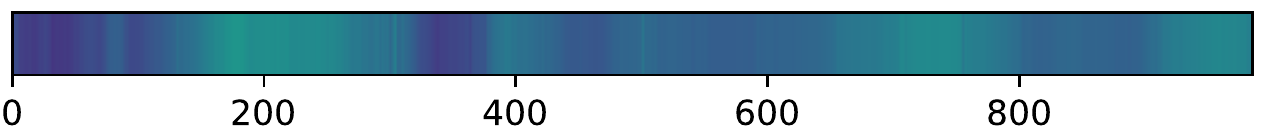}
    \end{subfigure}
    \hfill
    \begin{subfigure}{0.15\linewidth}
        \centering
        \includegraphics[width=\linewidth]{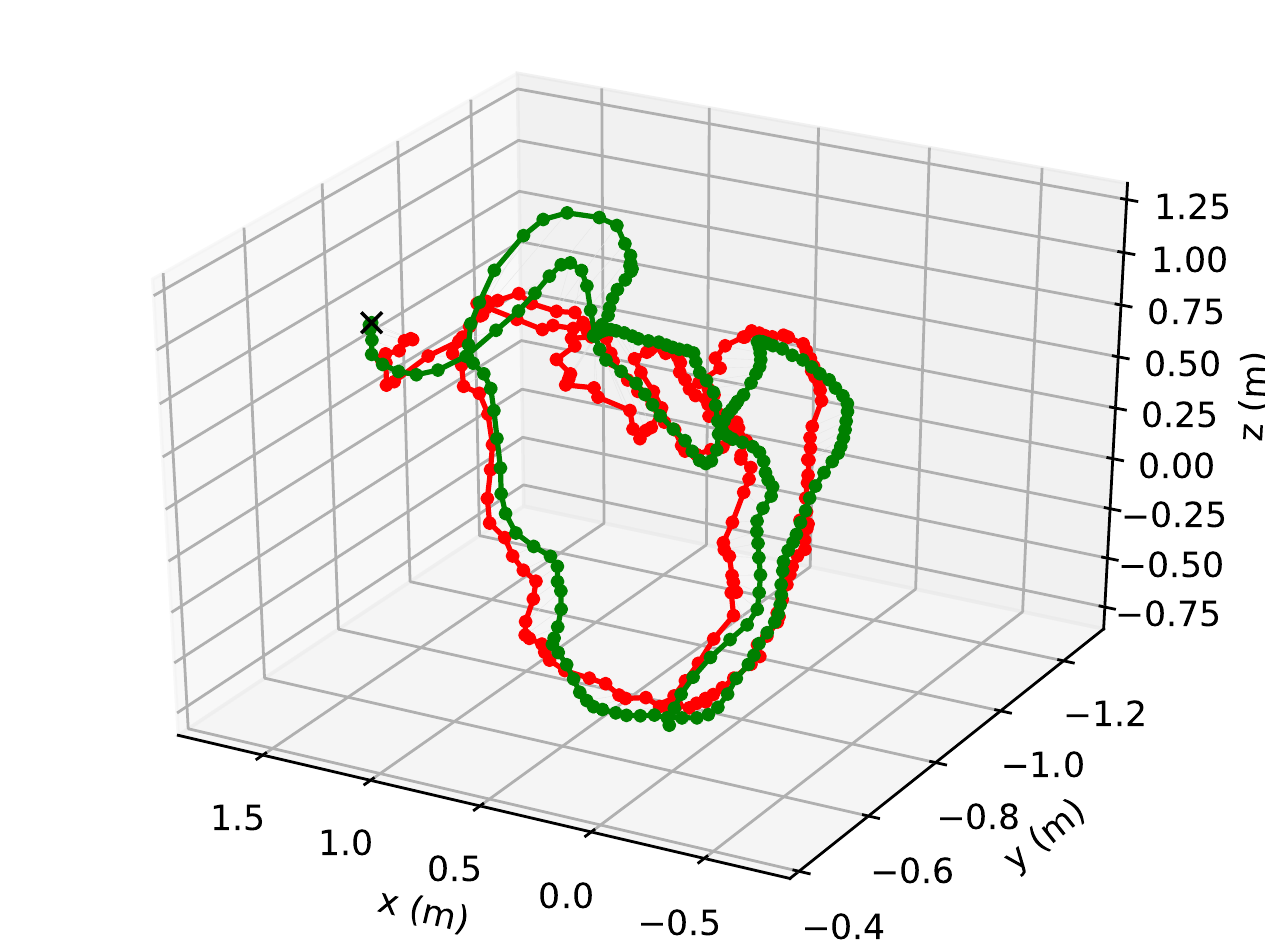}
        \includegraphics[width=\linewidth]{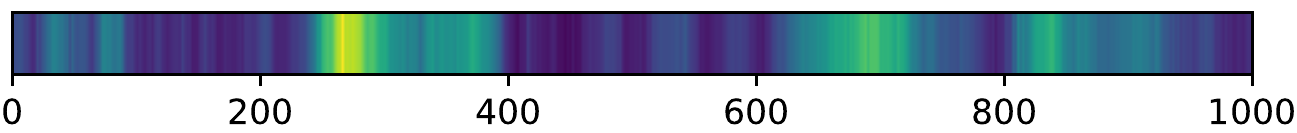}
    \end{subfigure}
    \hfill
    \begin{subfigure}{0.15\linewidth}
        \centering
        \includegraphics[width=\linewidth]{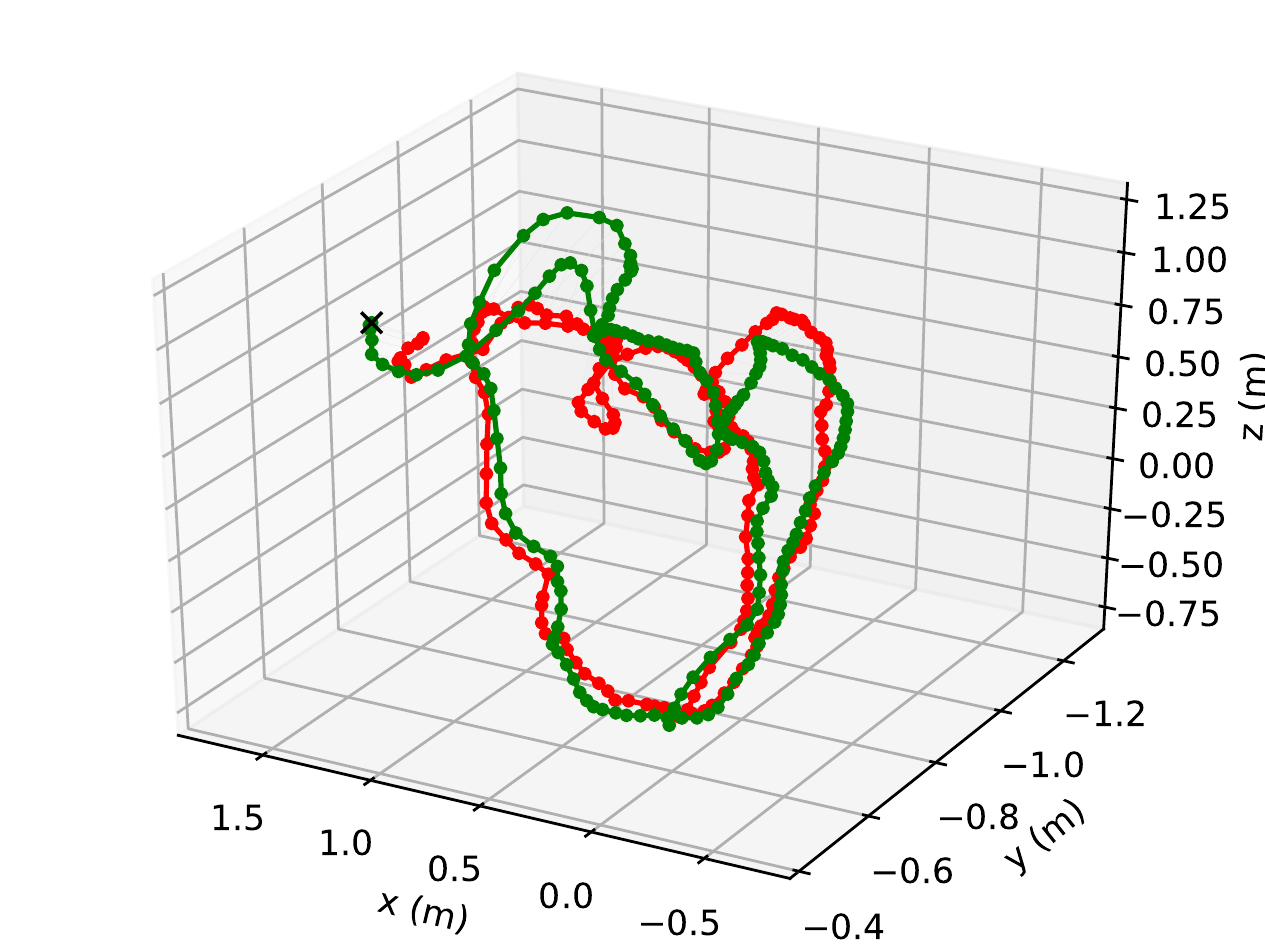}
        \includegraphics[width=\linewidth]{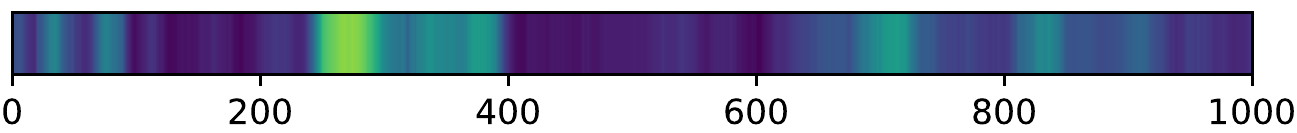}
    \end{subfigure}
    \hfill
    \begin{subfigure}{0.15\linewidth}
        \centering
        \includegraphics[width=\linewidth]{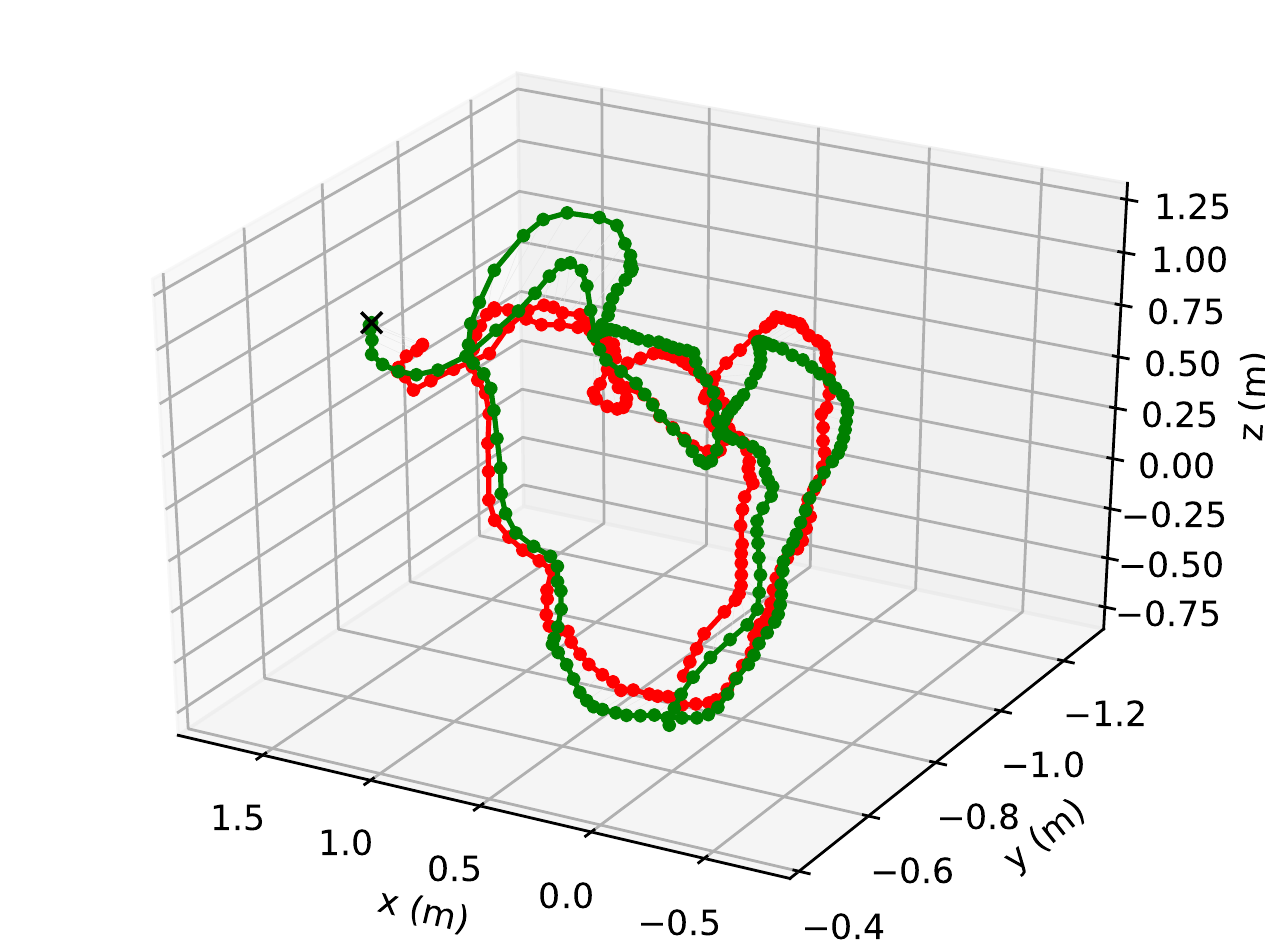}
        \includegraphics[width=\linewidth]{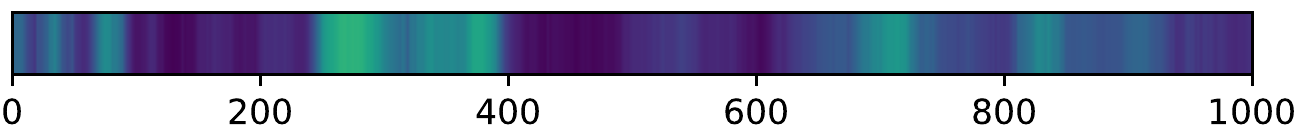}
    \end{subfigure}
    \hfill
    \begin{subfigure}{0.15\linewidth}
        \centering
        \includegraphics[width=\linewidth]{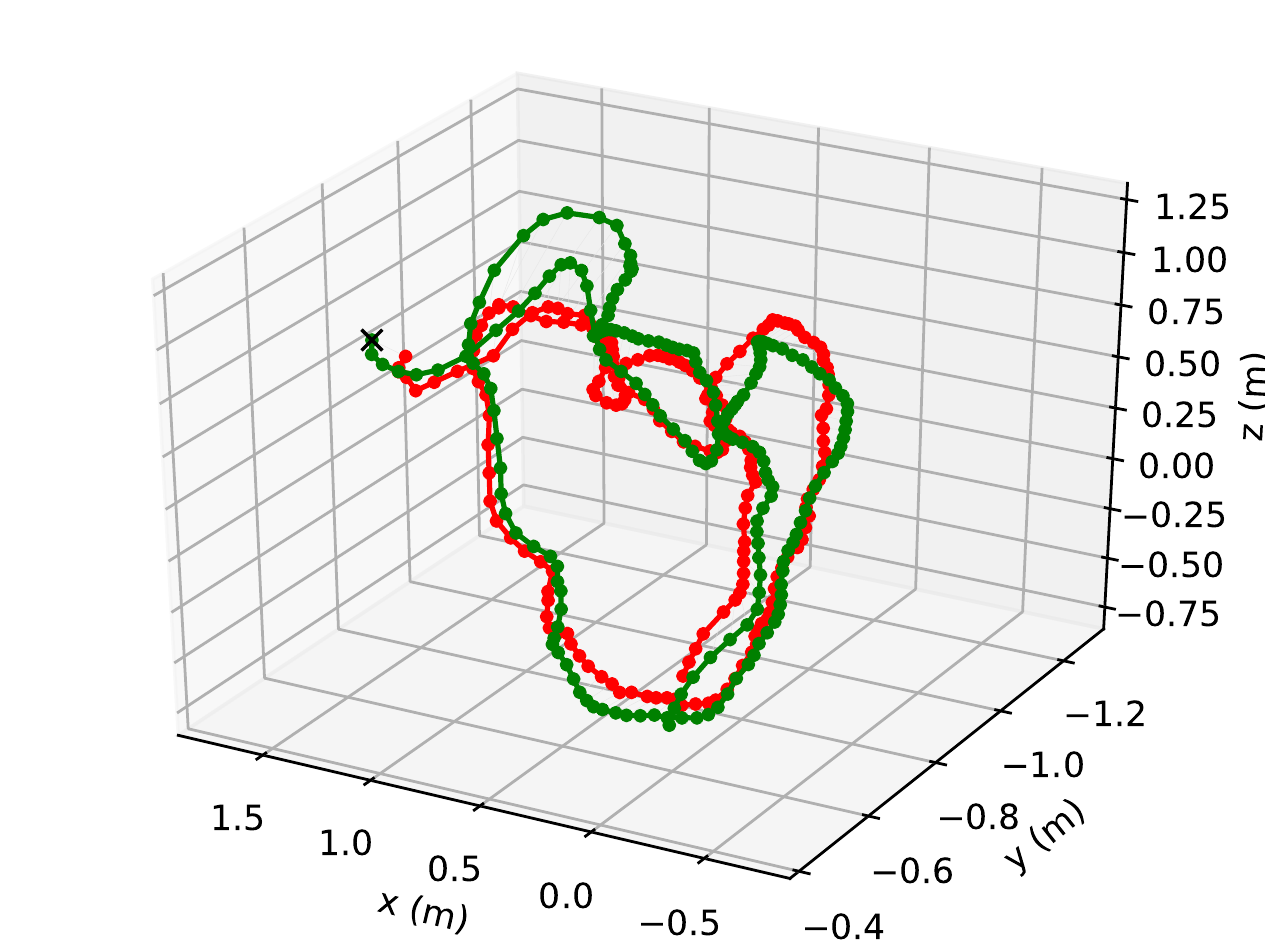}
        \includegraphics[width=\linewidth]{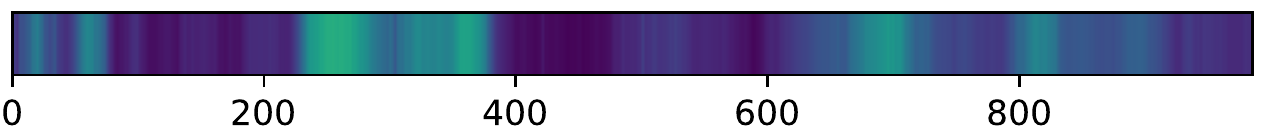}
    \end{subfigure}

    \begin{subfigure}{0.15\linewidth}
        \centering
        \includegraphics[width=\linewidth]{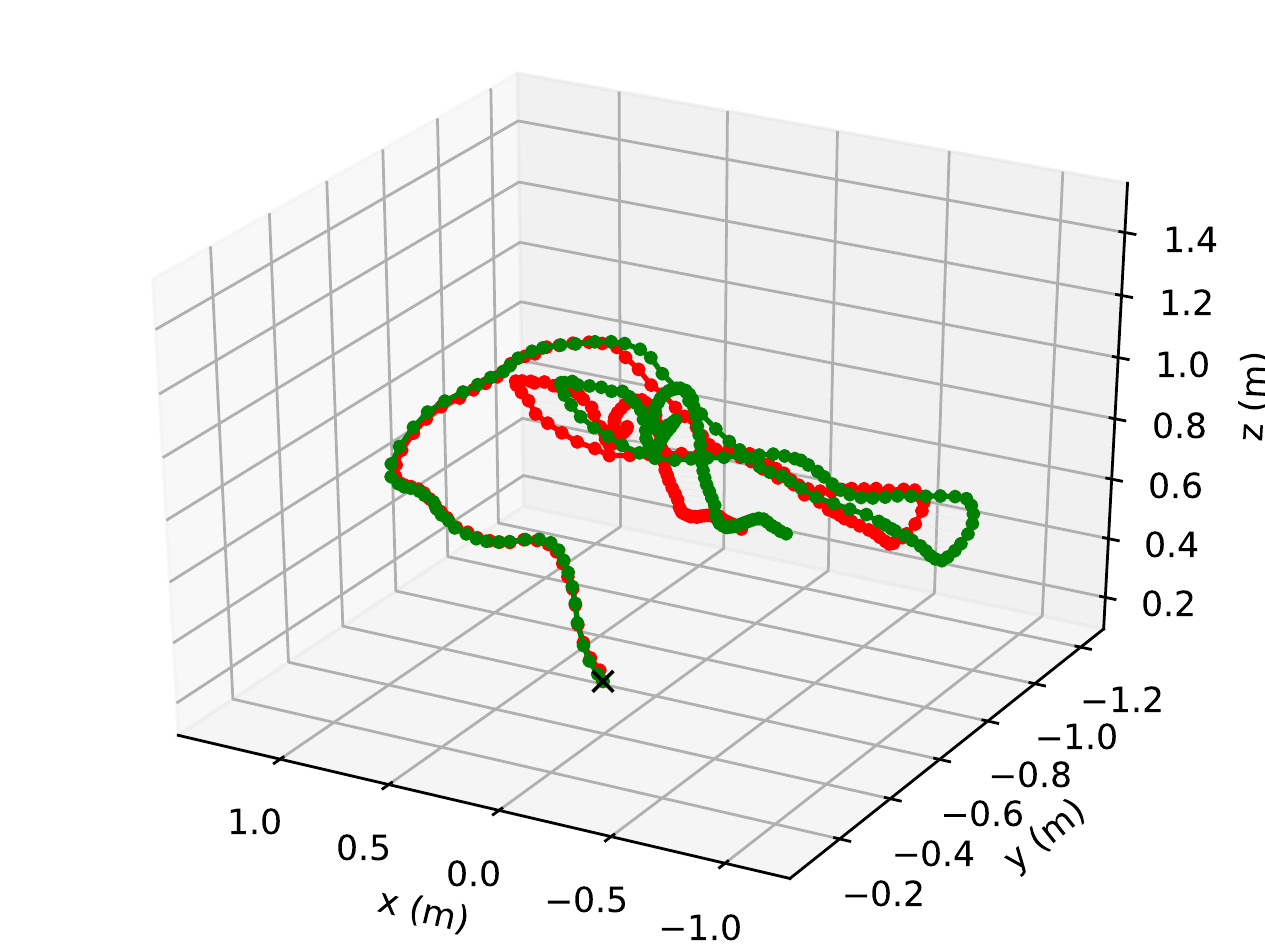}
        \includegraphics[width=\linewidth]{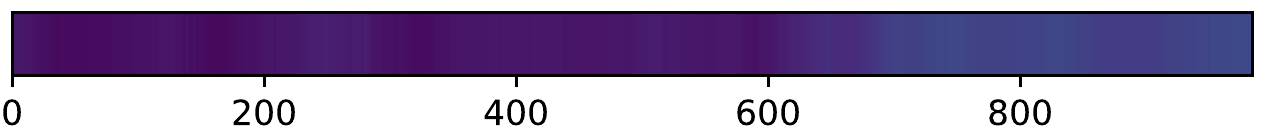}
    \end{subfigure}
    \hfill
    \begin{subfigure}{0.15\linewidth}
        \centering
        \includegraphics[width=\linewidth]{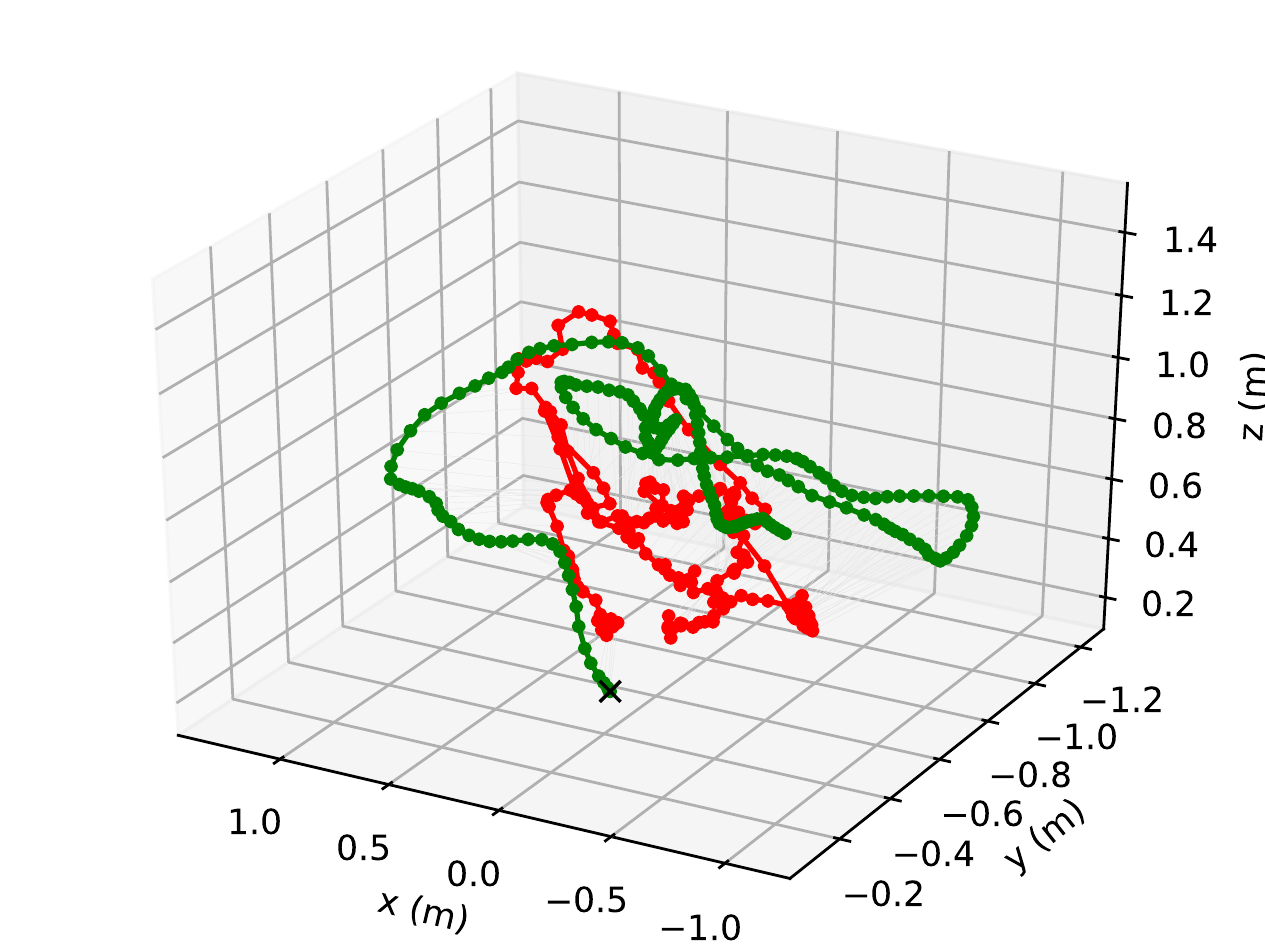}
        \includegraphics[width=\linewidth]{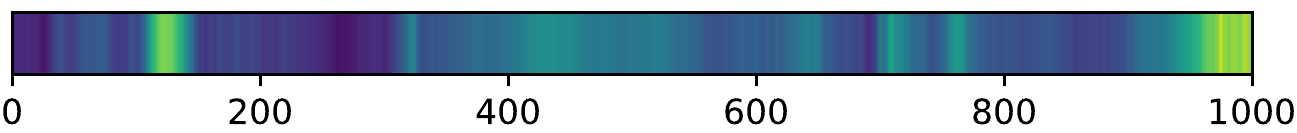}
    \end{subfigure}
    \hfill
    \begin{subfigure}{0.15\linewidth}
        \centering
        \includegraphics[width=\linewidth]{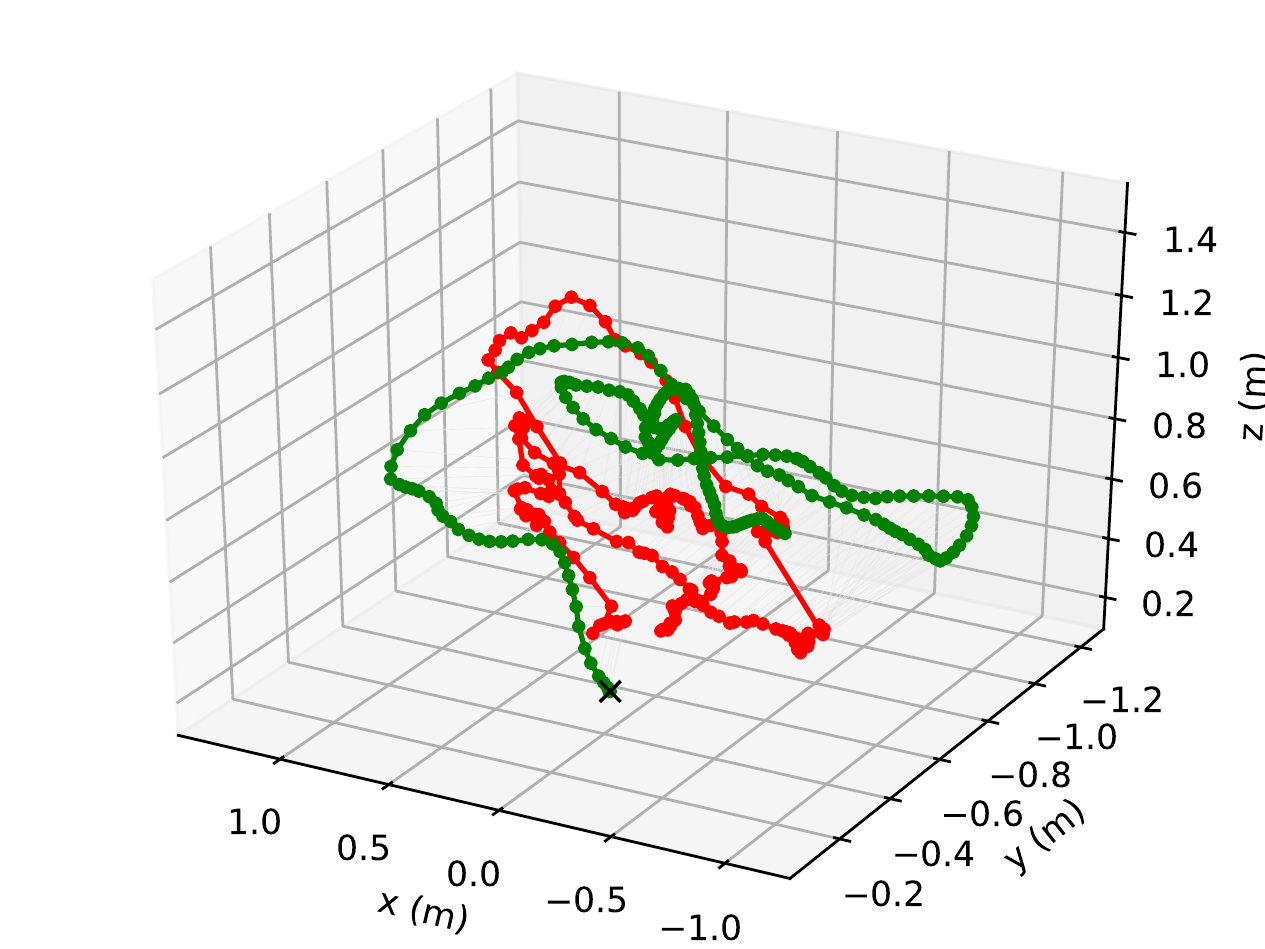}
        \includegraphics[width=\linewidth]{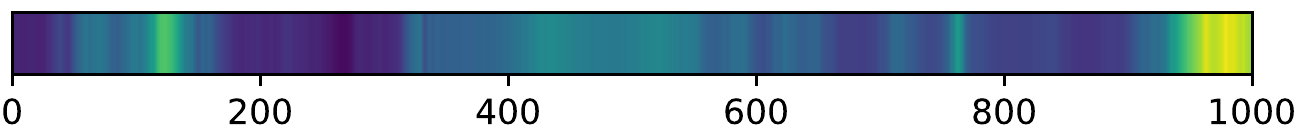}
    \end{subfigure}
    \hfill
    \begin{subfigure}{0.15\linewidth}
        \centering
        \includegraphics[width=\linewidth]{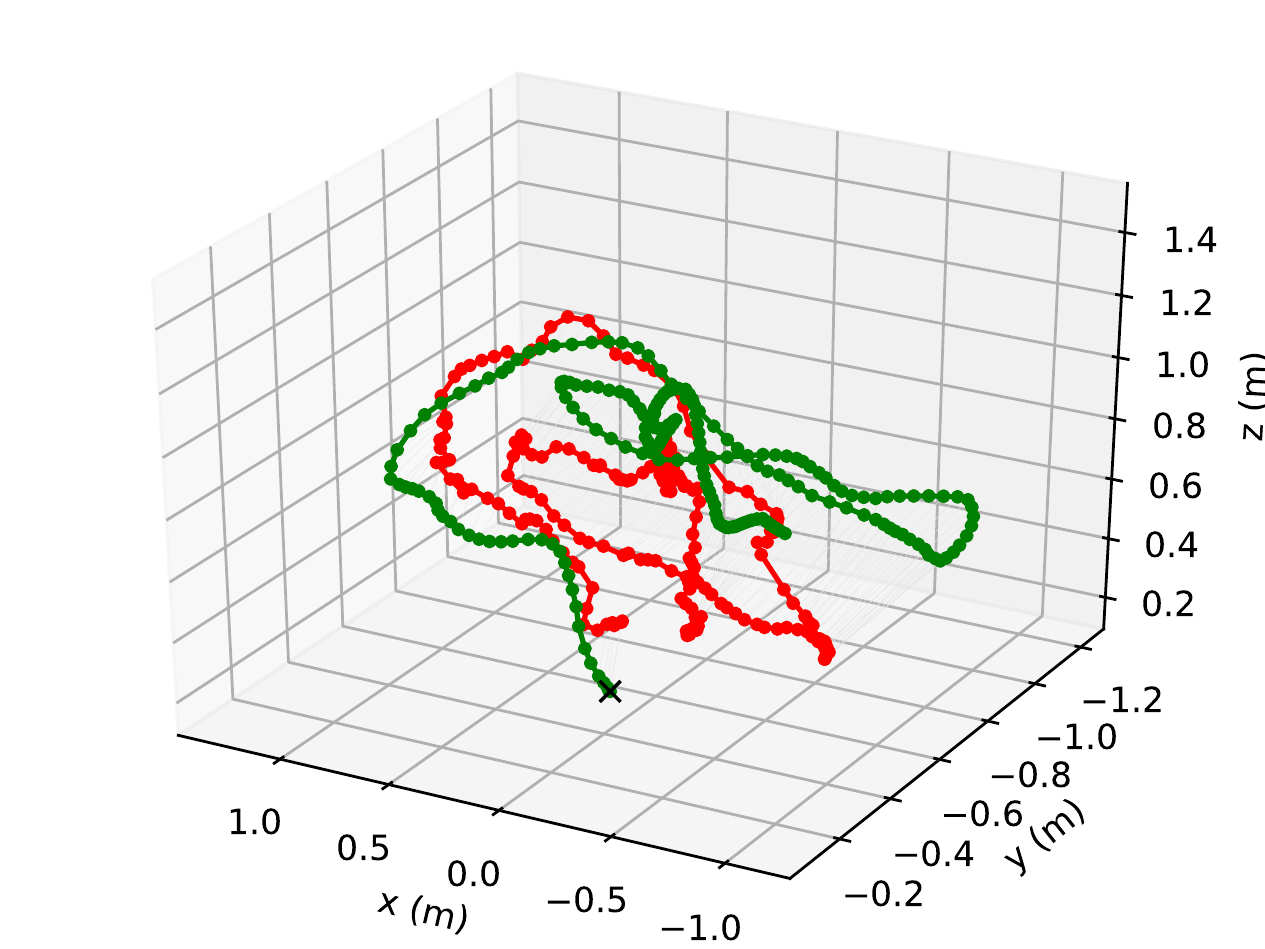}
        \includegraphics[width=\linewidth]{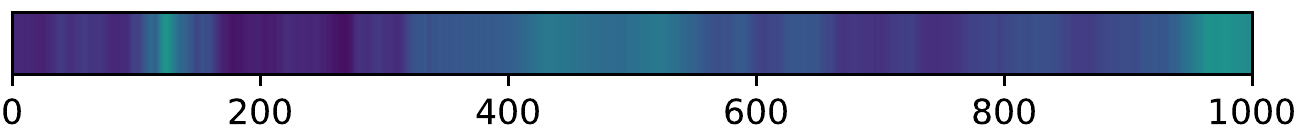}
    \end{subfigure}
    \hfill
    \begin{subfigure}{0.15\linewidth}
        \centering
        \includegraphics[width=\linewidth]{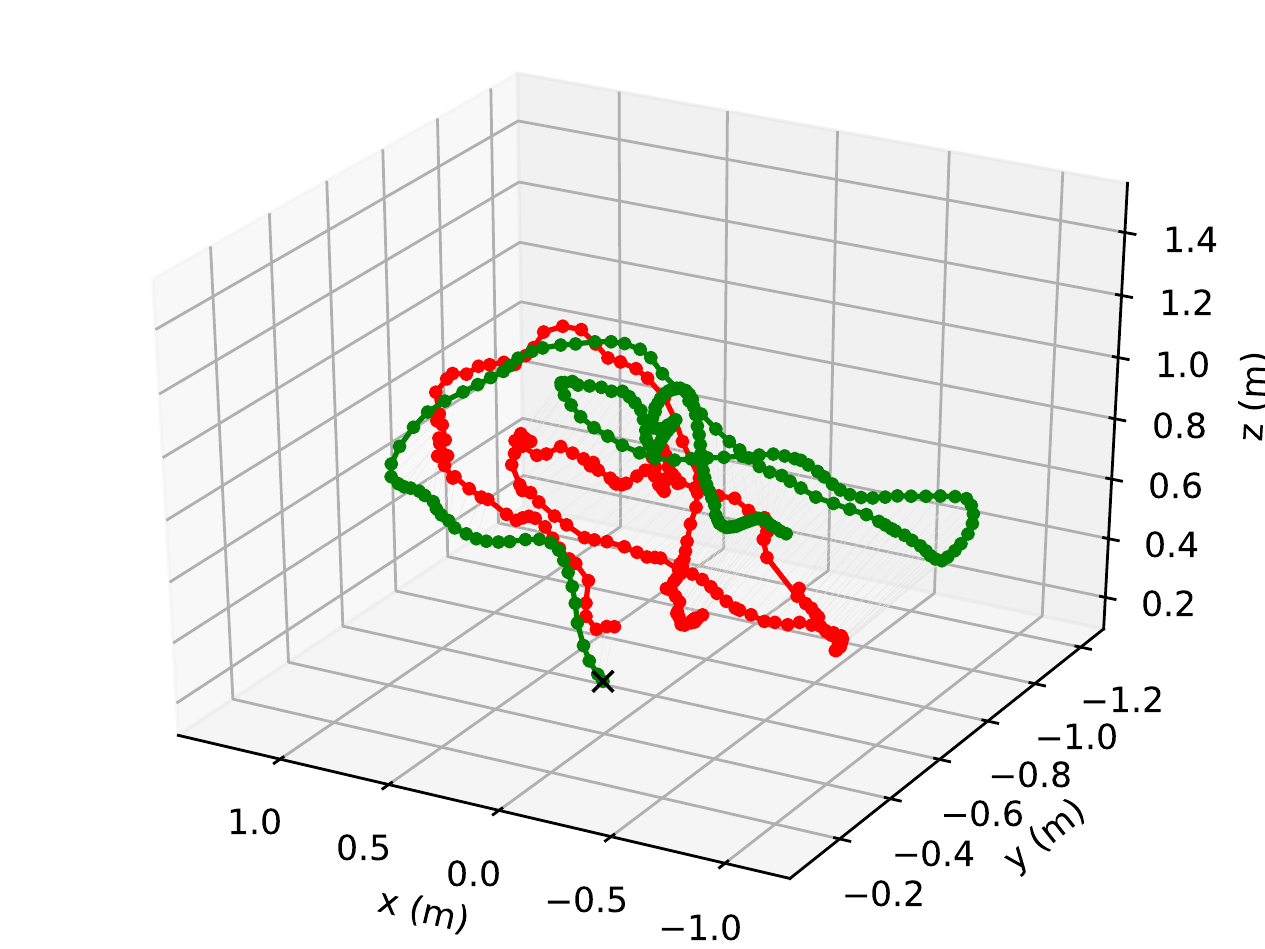}
        \includegraphics[width=\linewidth]{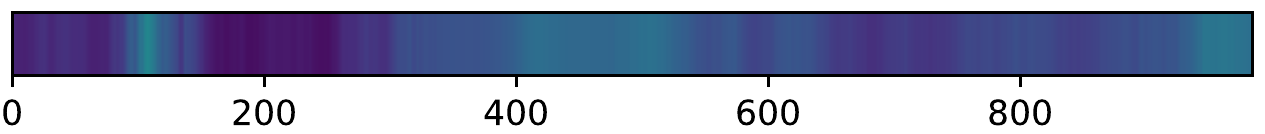}
    \end{subfigure}

    \begin{subfigure}{0.15\linewidth}
        \centering
        \includegraphics[width=\linewidth]{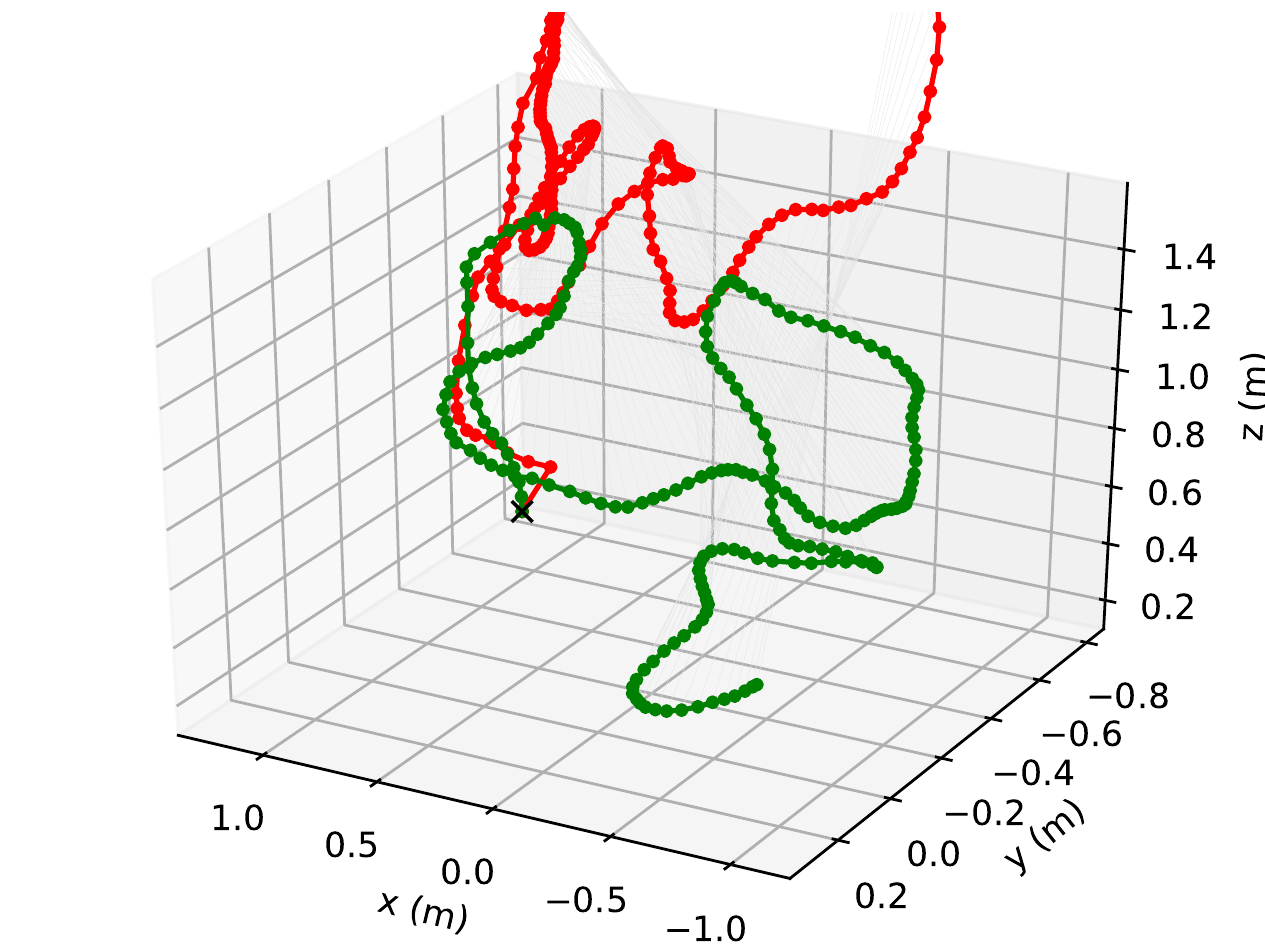}
        \includegraphics[width=\linewidth]{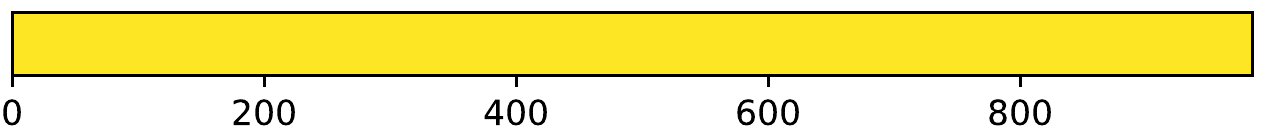}
    \end{subfigure}
    \hfill
    \begin{subfigure}{0.15\linewidth}
        \centering
        \includegraphics[width=\linewidth]{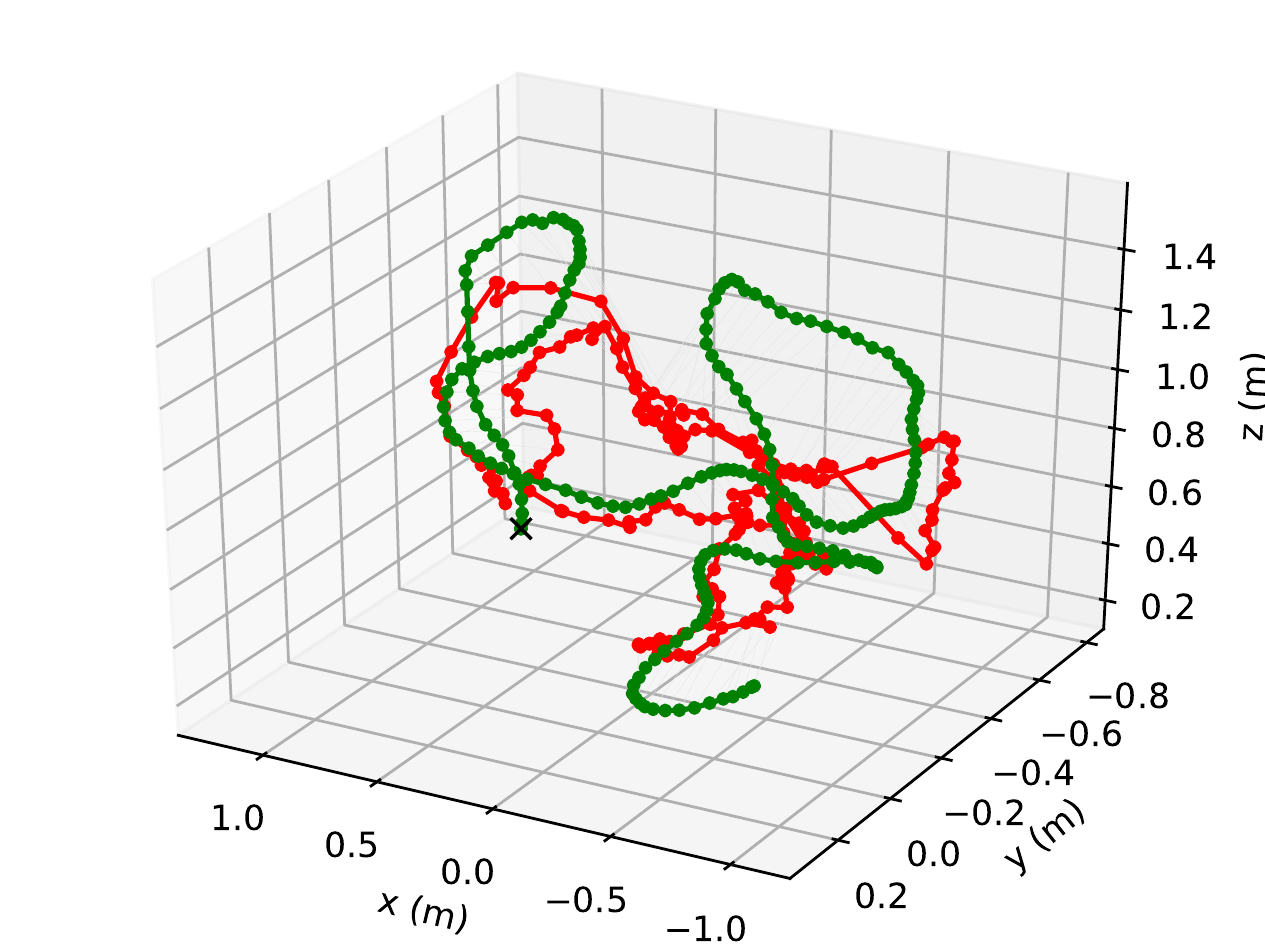}
        \includegraphics[width=\linewidth]{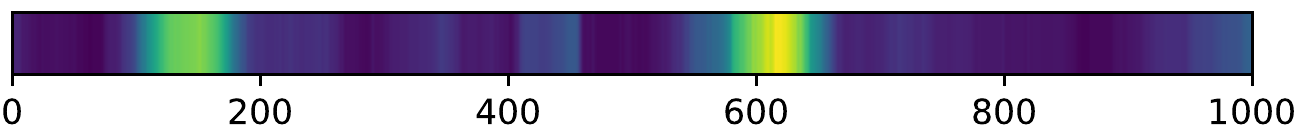}
    \end{subfigure}
    \hfill
    \begin{subfigure}{0.15\linewidth}
        \centering
        \includegraphics[width=\linewidth]{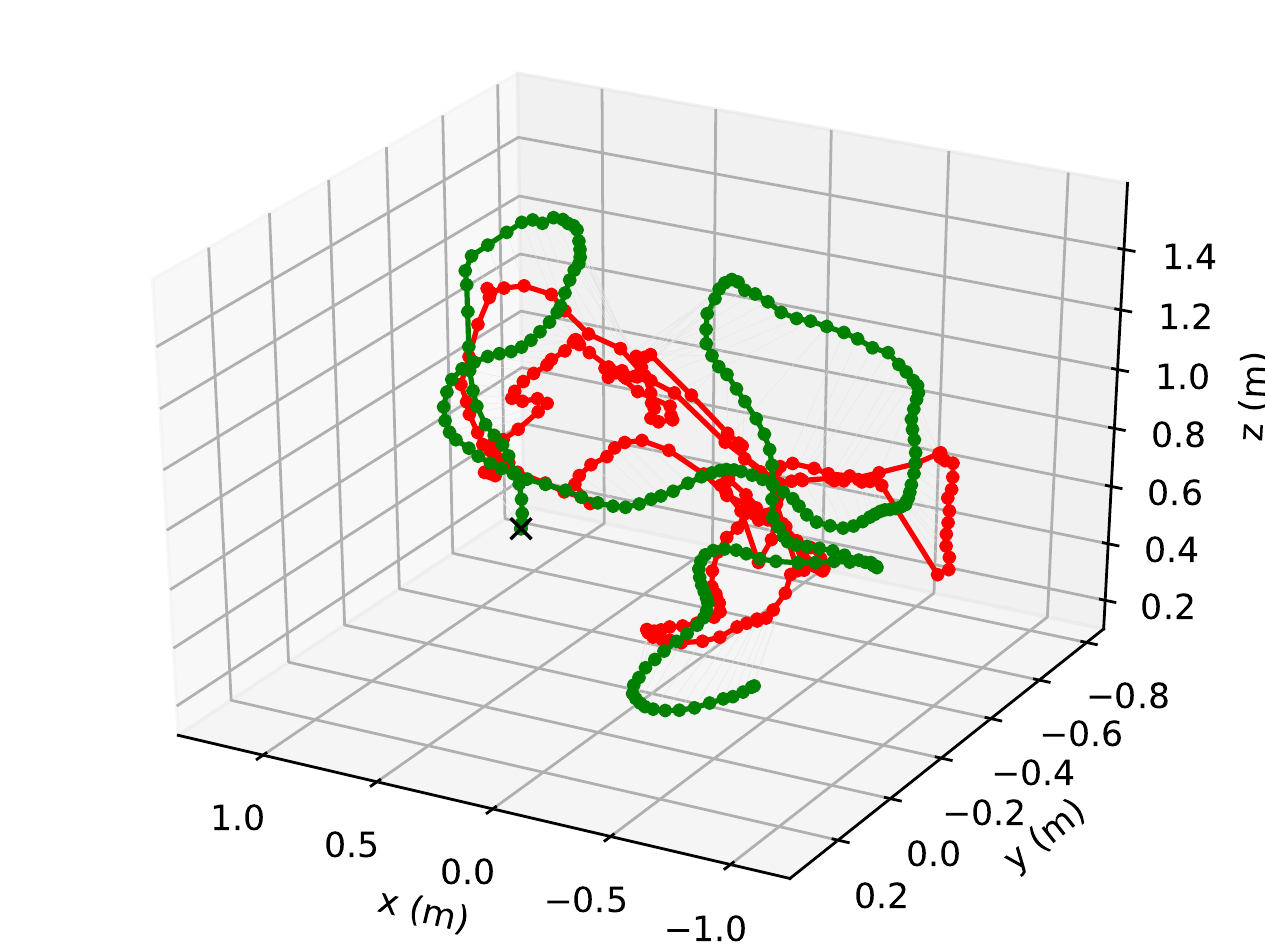}
        \includegraphics[width=\linewidth]{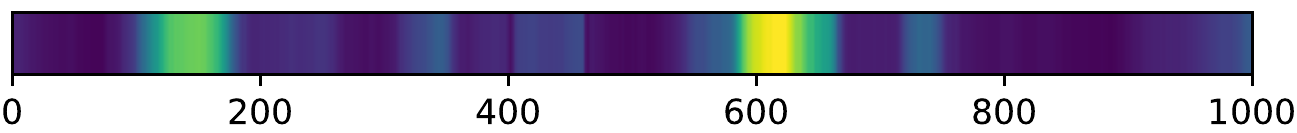}
    \end{subfigure}
    \hfill
    \begin{subfigure}{0.15\linewidth}
        \centering
        \includegraphics[width=\linewidth]{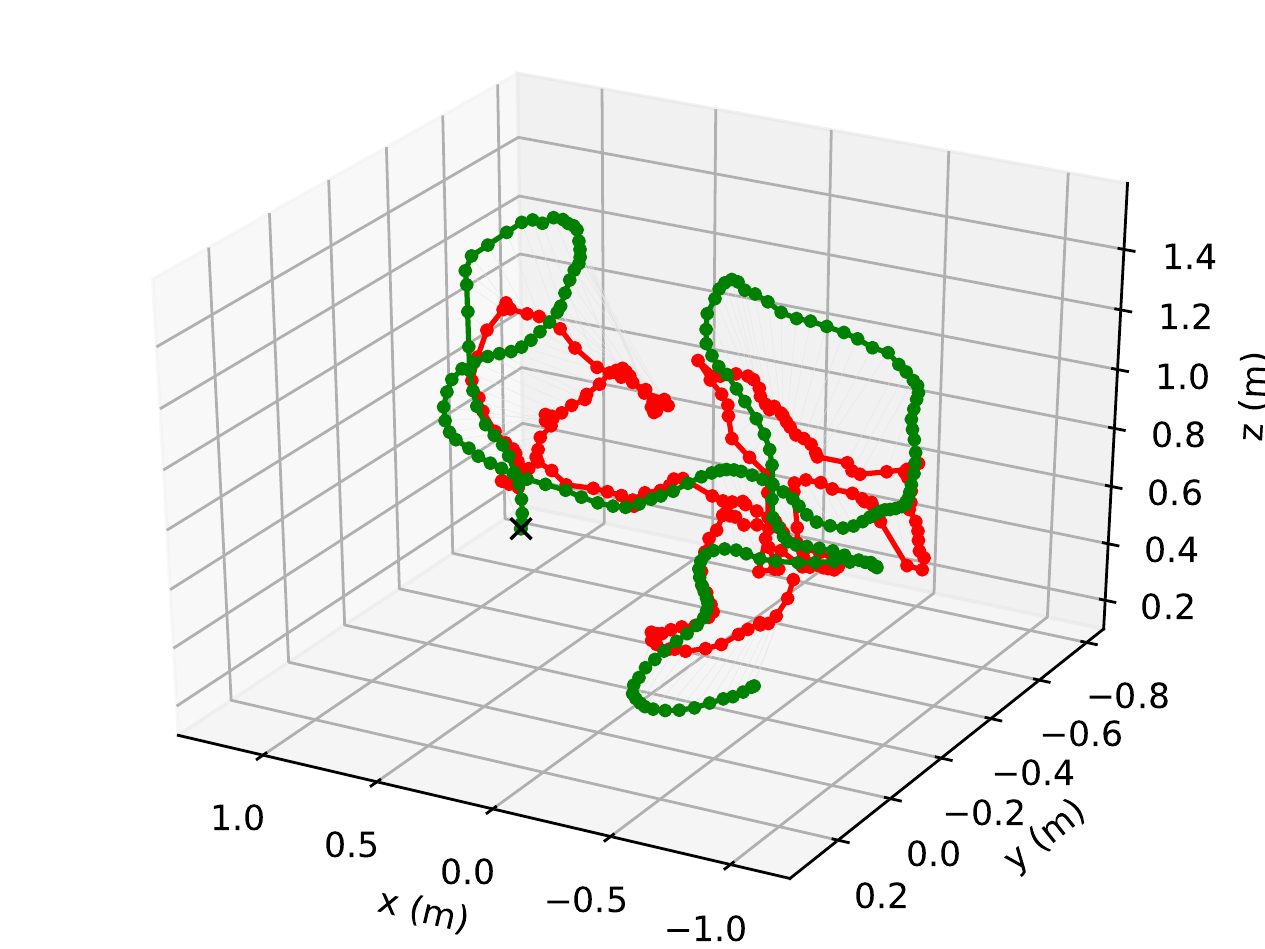}
        \includegraphics[width=\linewidth]{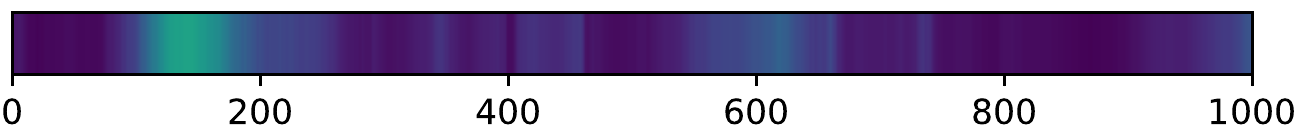}
    \end{subfigure}
    \hfill
    \begin{subfigure}{0.15\linewidth}
        \centering
        \includegraphics[width=\linewidth]{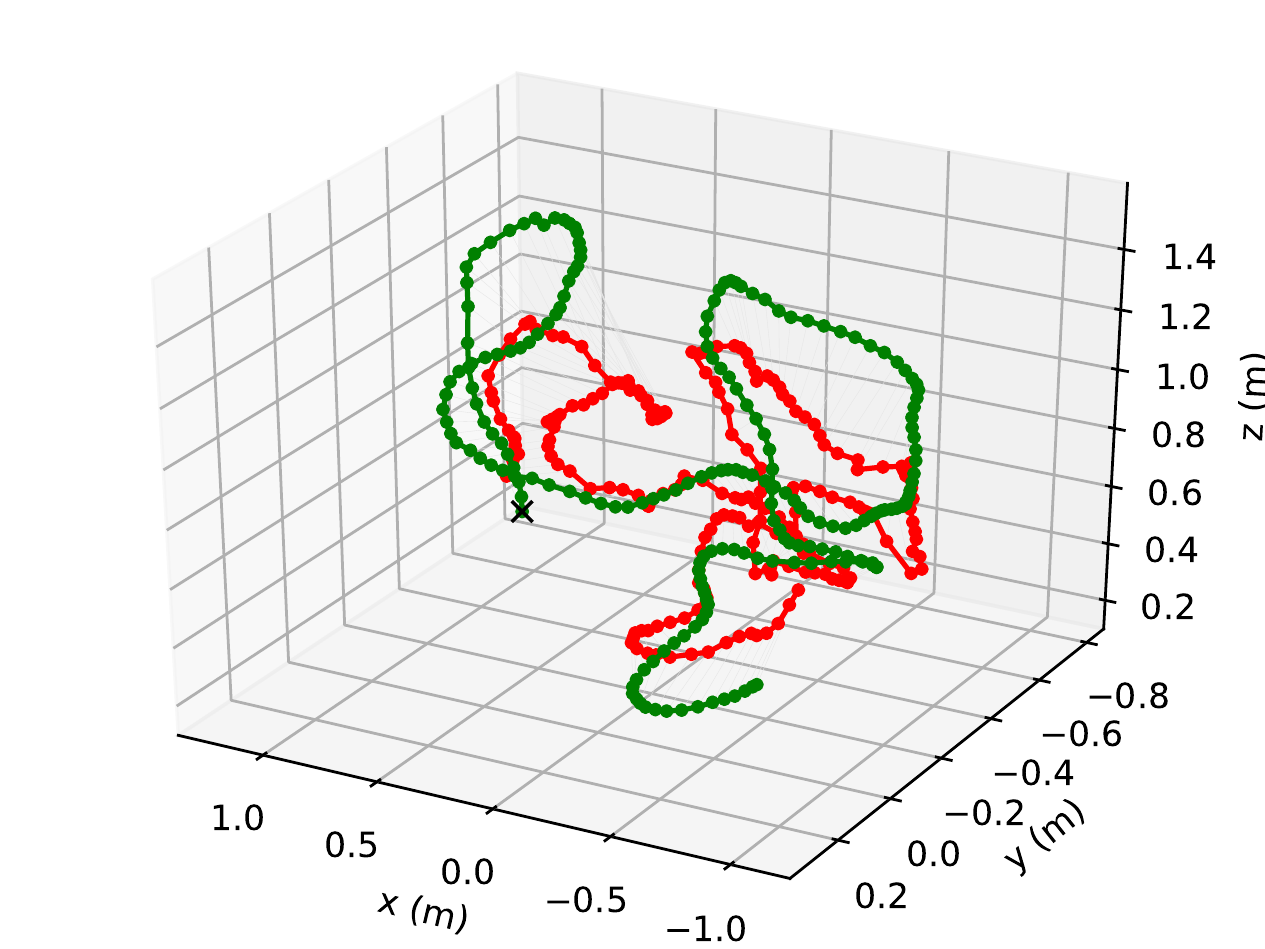}
        \includegraphics[width=\linewidth]{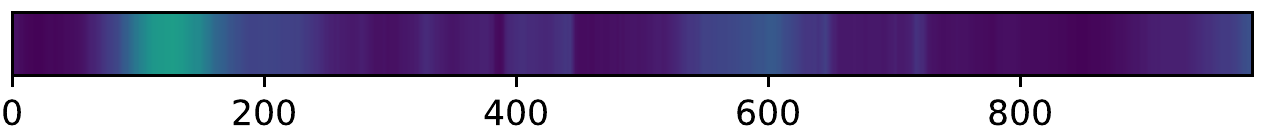}
    \end{subfigure}

    \begin{subfigure}{0.15\linewidth}
        \centering
        \includegraphics[width=\linewidth]{figures/7scenes/heads_dsoA_0_t.pdf}
        \includegraphics[width=\linewidth]{figures/7scenes/heads_dsoA_0_q.pdf}
    \end{subfigure}
    \hfill
    \begin{subfigure}{0.15\linewidth}
        \centering
        \includegraphics[width=\linewidth]{figures/7scenes/heads_posenet_0_t.pdf}
        \includegraphics[width=\linewidth]{figures/7scenes/heads_posenet_0_q.pdf}
    \end{subfigure}
    \hfill
    \begin{subfigure}{0.15\linewidth}
        \centering
        \includegraphics[width=\linewidth]{figures/7scenes/heads_vidvo_0_t.pdf}
        \includegraphics[width=\linewidth]{figures/7scenes/heads_vidvo_0_q.pdf}
    \end{subfigure}
    \hfill
    \begin{subfigure}{0.15\linewidth}
        \centering
        \includegraphics[width=\linewidth]{figures/7scenes/heads_vidvo_online_0_t.pdf}
        \includegraphics[width=\linewidth]{figures/7scenes/heads_vidvo_online_0_q.pdf}
    \end{subfigure}
    \hfill
    \begin{subfigure}{0.15\linewidth}
        \centering
        \includegraphics[width=\linewidth]{figures/7scenes/heads_vidvo_online_pgo_dso_0_t.pdf}
        \includegraphics[width=\linewidth]{figures/7scenes/heads_vidvo_online_pgo_dso_0_q.pdf}
    \end{subfigure}

    \begin{subfigure}{0.15\linewidth}
        \centering
        \includegraphics[width=\linewidth]{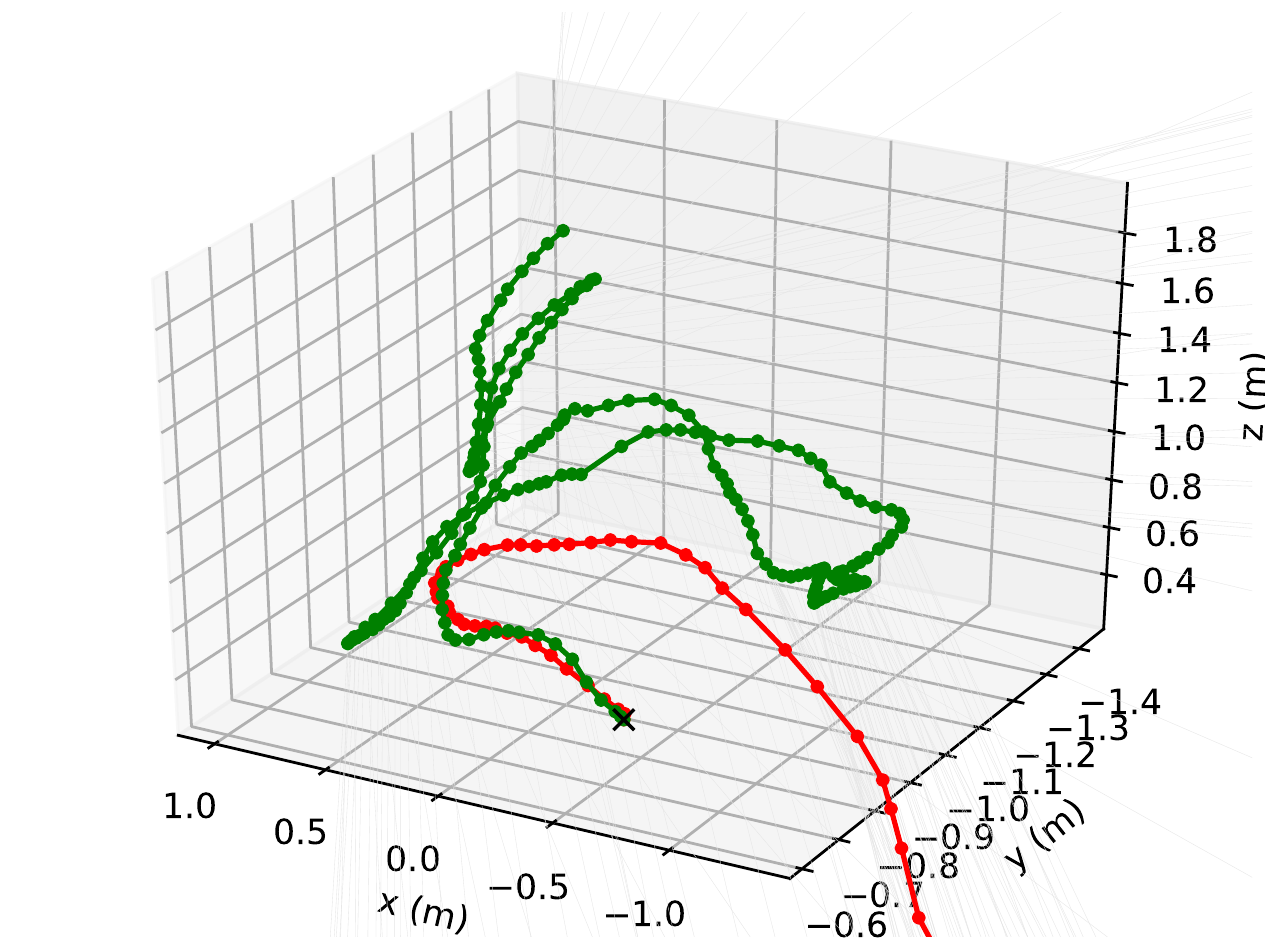}
        \includegraphics[width=\linewidth]{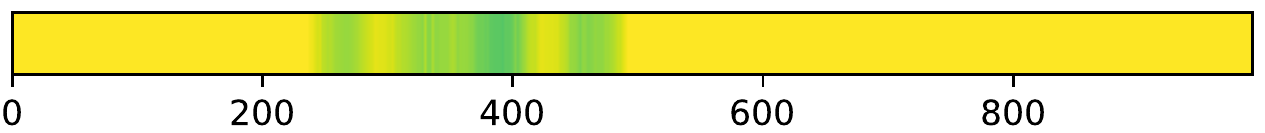}
    \end{subfigure}
    \hfill
    \begin{subfigure}{0.15\linewidth}
        \centering
        \includegraphics[width=\linewidth]{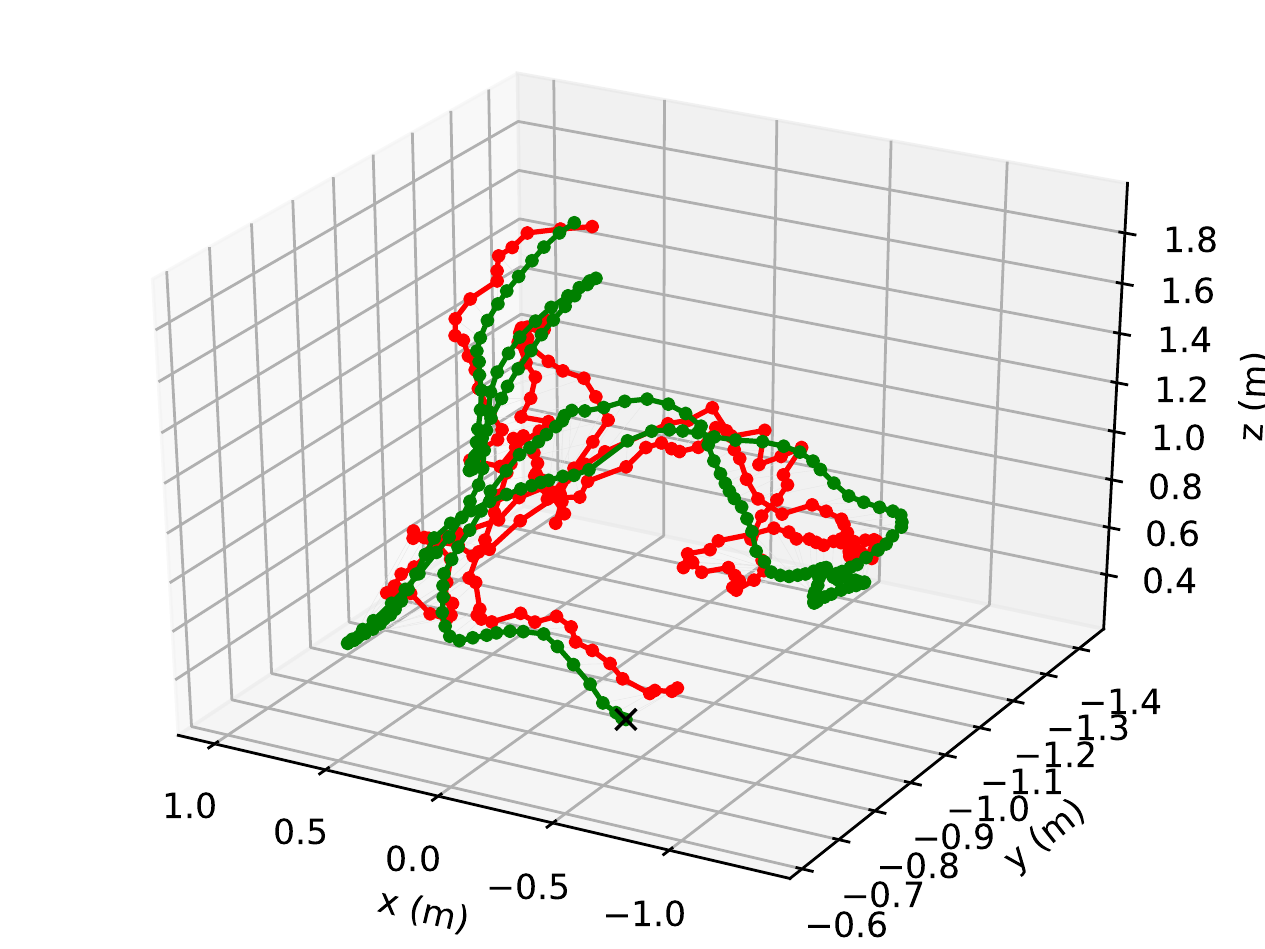}
        \includegraphics[width=\linewidth]{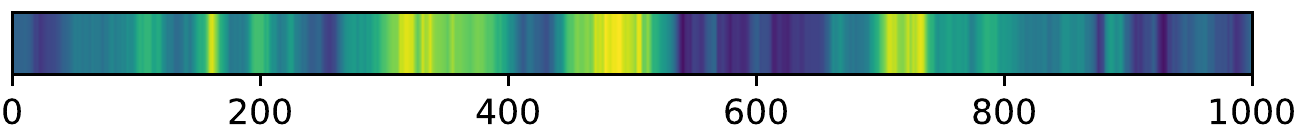}
    \end{subfigure}
    \hfill
    \begin{subfigure}{0.15\linewidth}
        \centering
        \includegraphics[width=\linewidth]{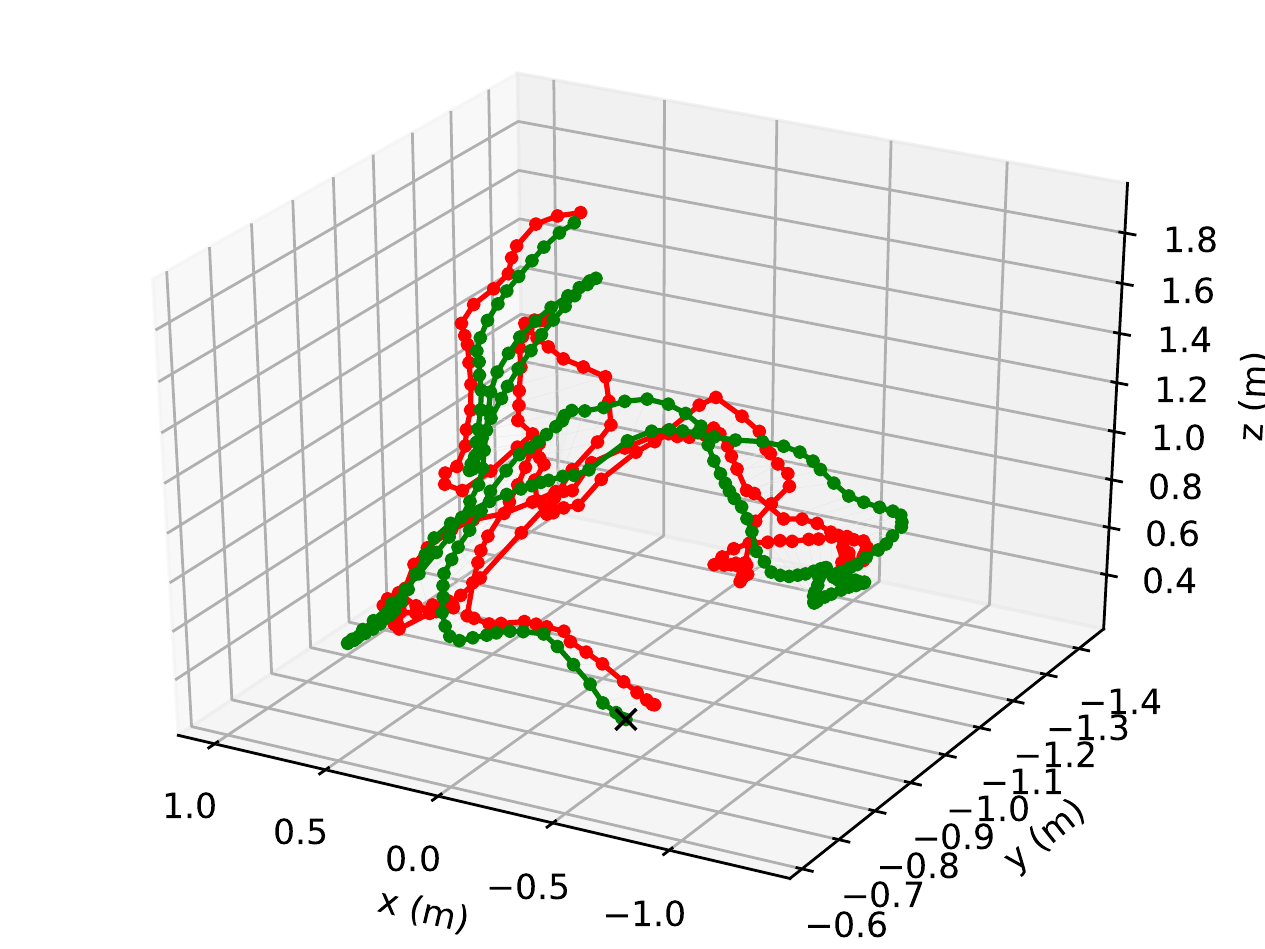}
        \includegraphics[width=\linewidth]{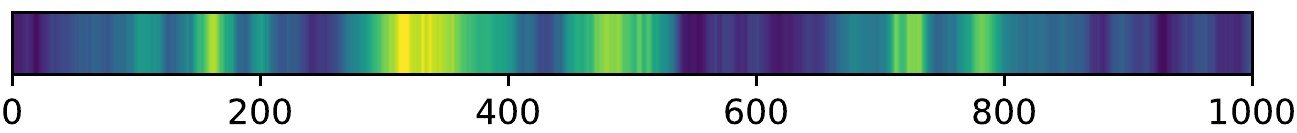}
    \end{subfigure}
    \hfill
    \begin{subfigure}{0.15\linewidth}
        \centering
        \includegraphics[width=\linewidth]{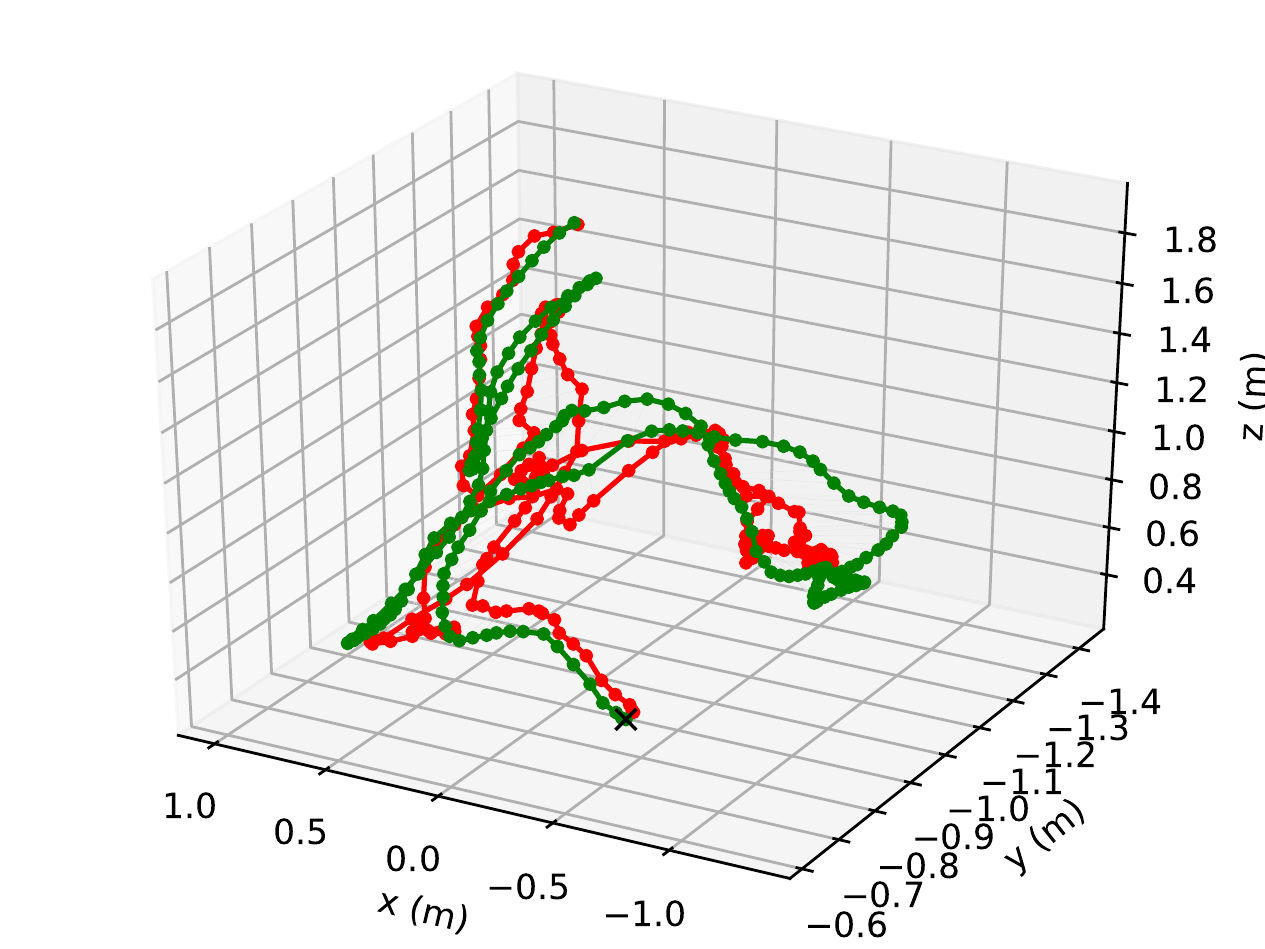}
        \includegraphics[width=\linewidth]{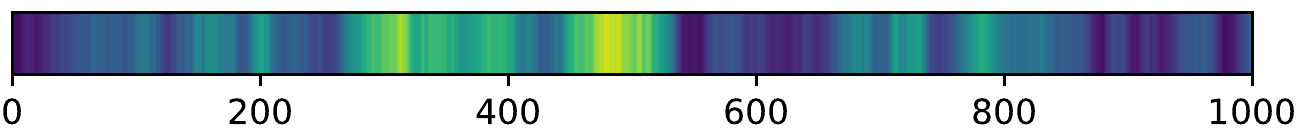}
    \end{subfigure}
    \hfill
    \begin{subfigure}{0.15\linewidth}
        \centering
        \includegraphics[width=\linewidth]{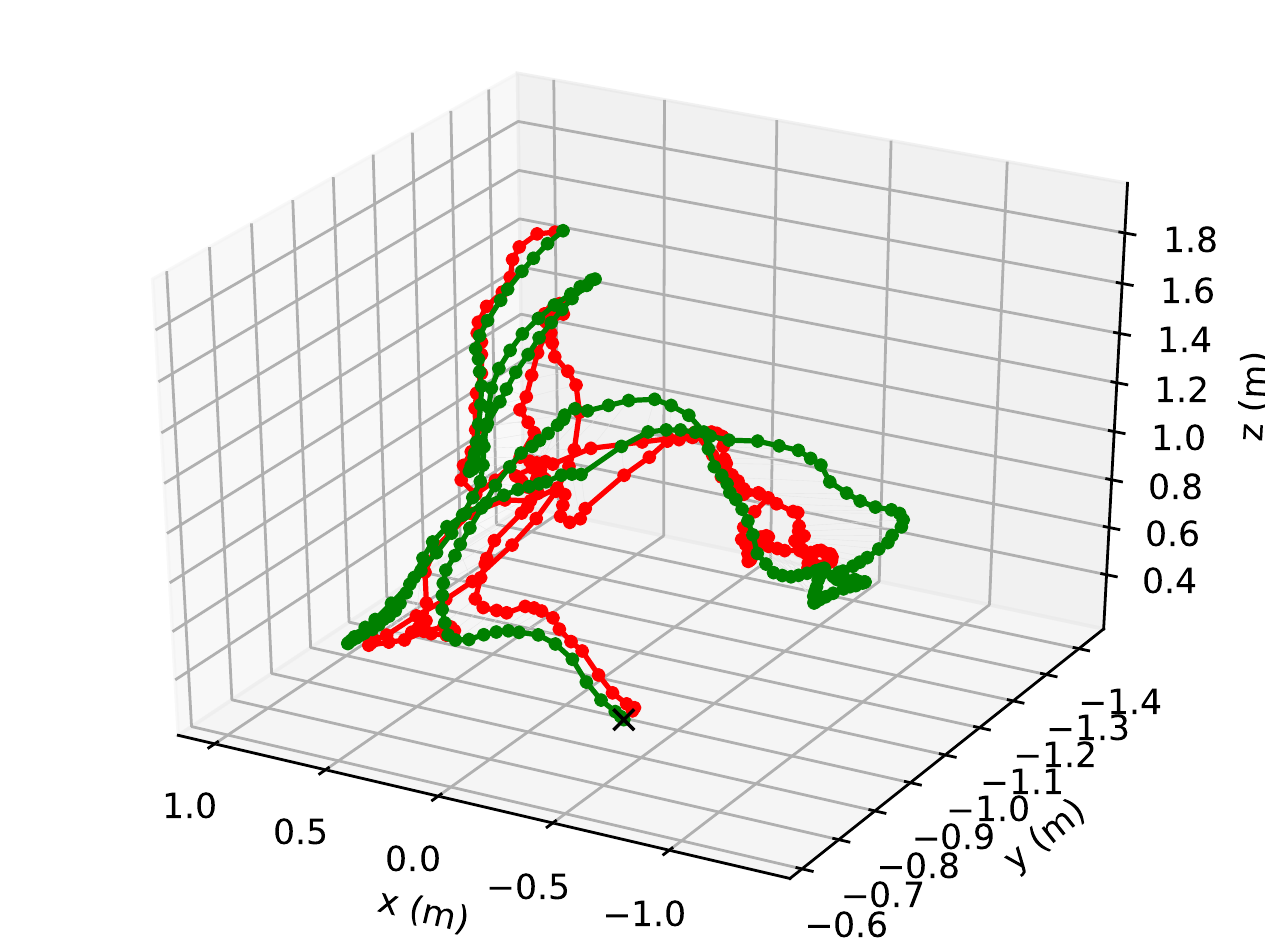}
        \includegraphics[width=\linewidth]{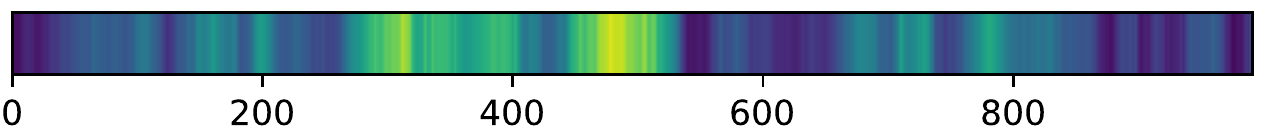}
    \end{subfigure}

    \begin{subfigure}{0.15\linewidth}
        \centering
        \includegraphics[width=\linewidth]{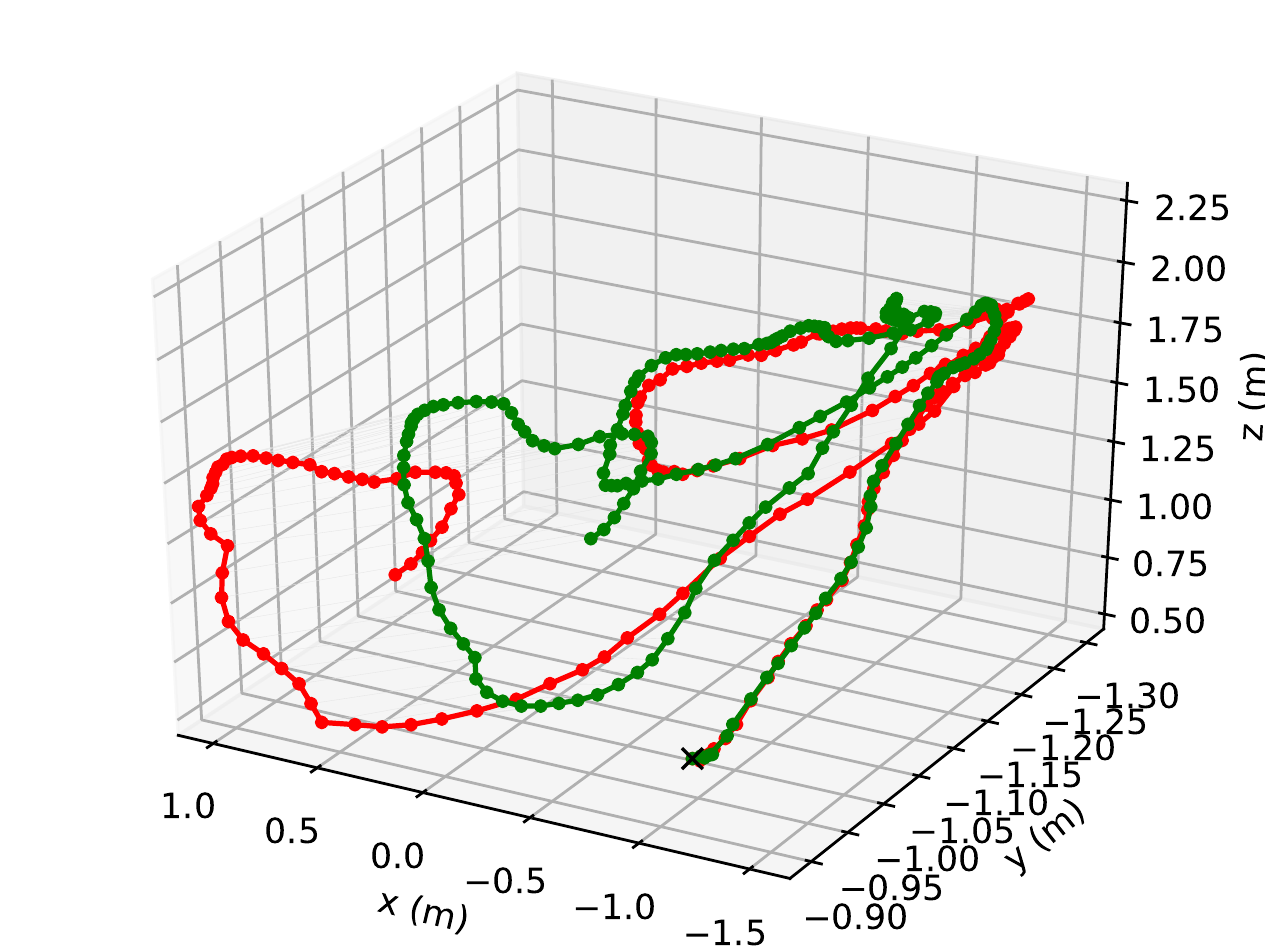}
        \includegraphics[width=\linewidth]{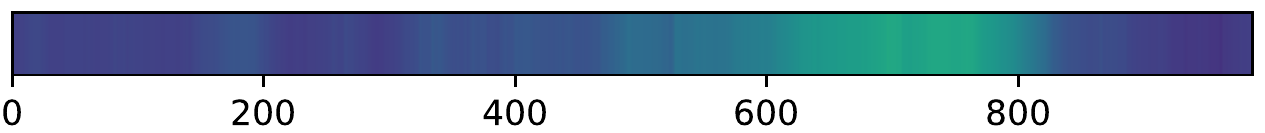}
    \end{subfigure}
    \hfill
    \begin{subfigure}{0.15\linewidth}
        \centering
        \includegraphics[width=\linewidth]{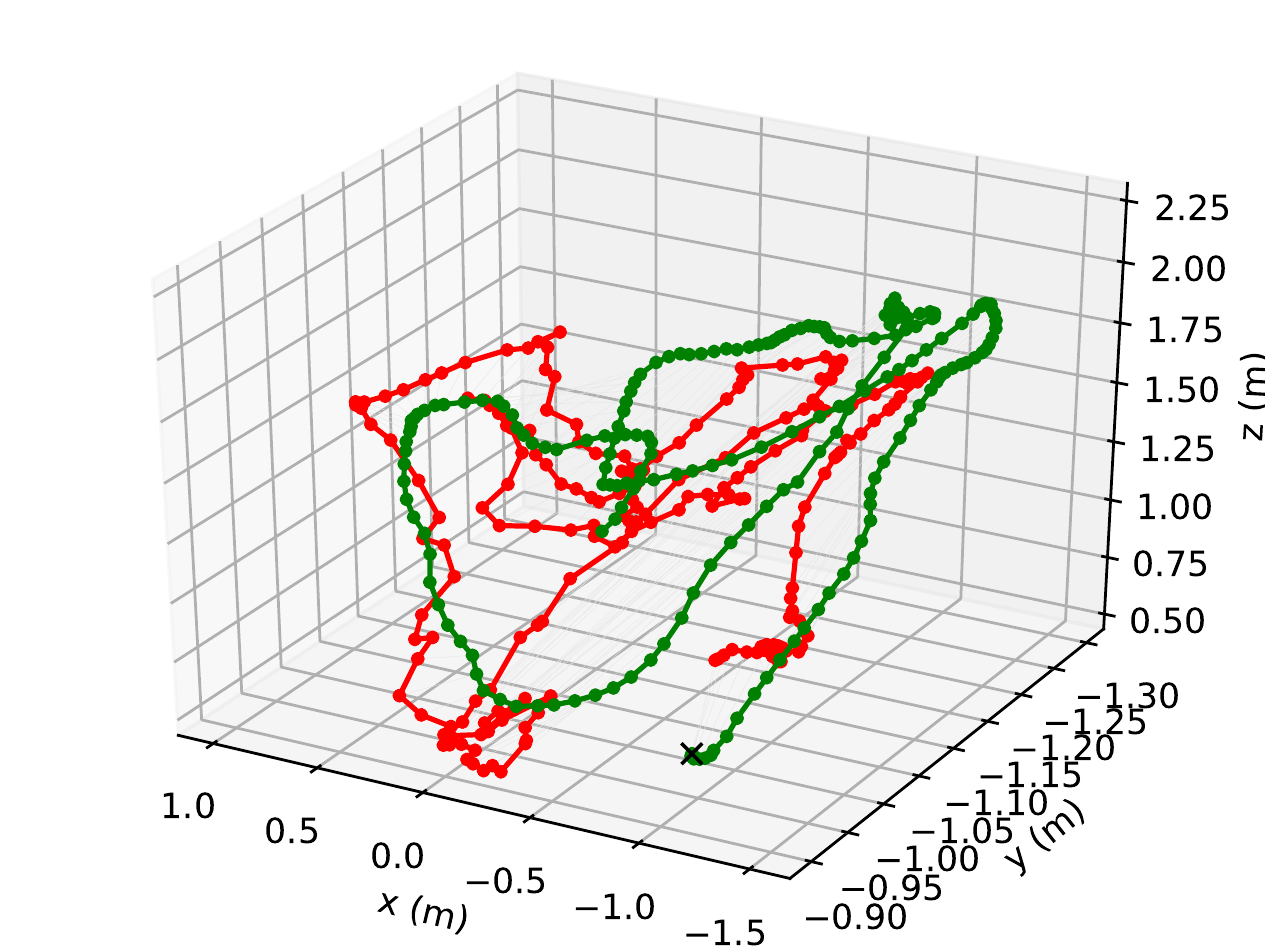}
        \includegraphics[width=\linewidth]{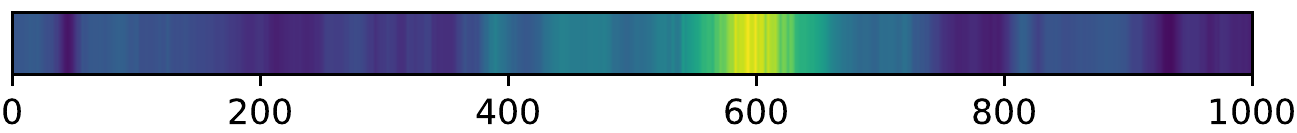}
    \end{subfigure}
    \hfill
    \begin{subfigure}{0.15\linewidth}
        \centering
        \includegraphics[width=\linewidth]{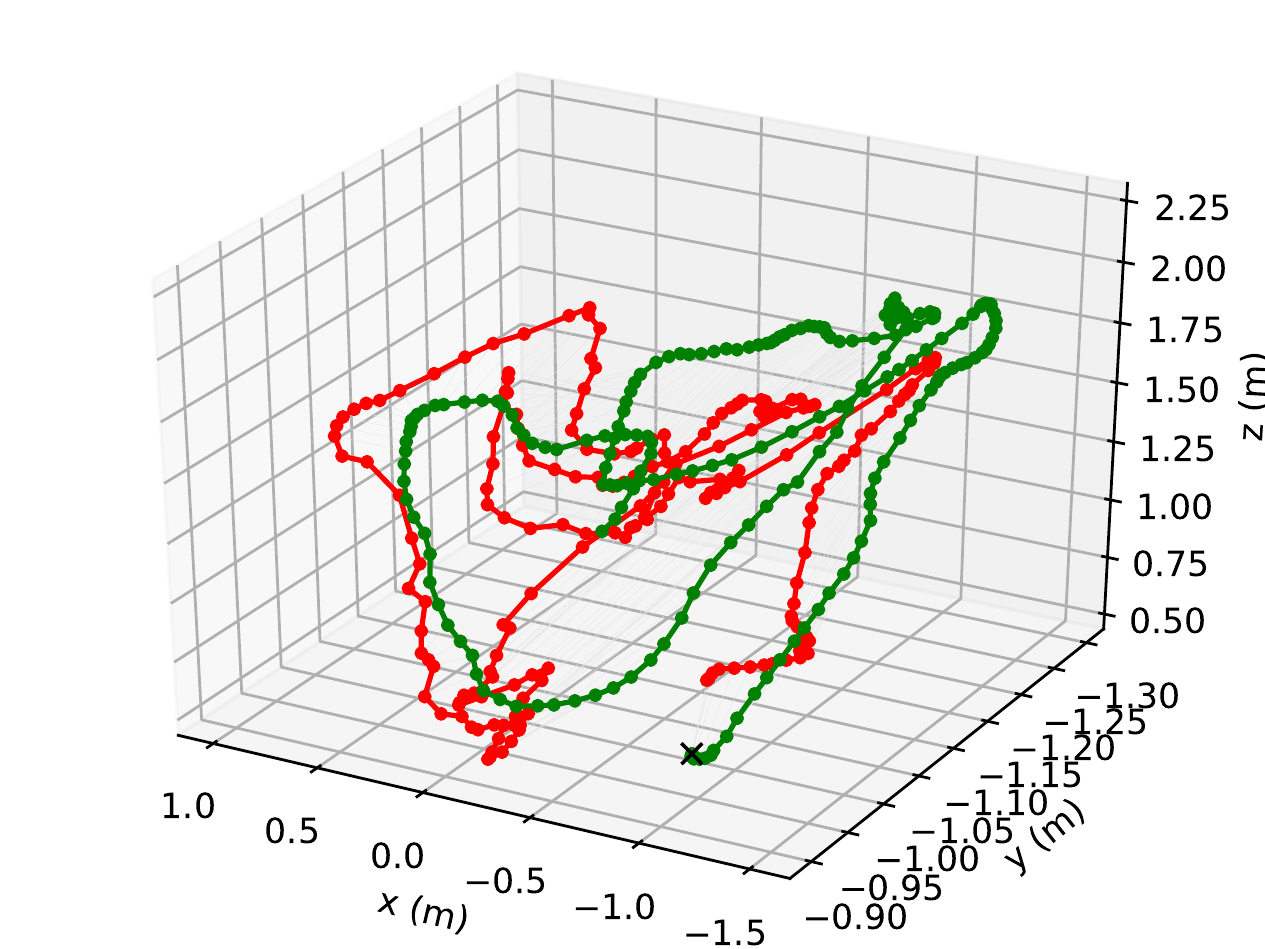}
        \includegraphics[width=\linewidth]{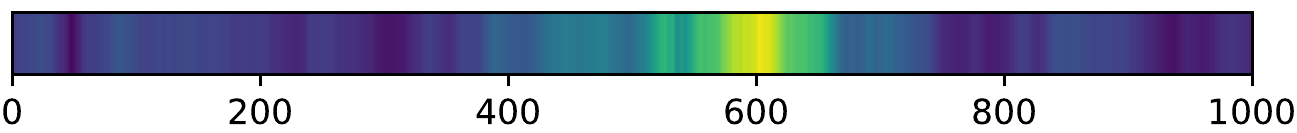}
    \end{subfigure}
    \hfill
    \begin{subfigure}{0.15\linewidth}
        \centering
        \includegraphics[width=\linewidth]{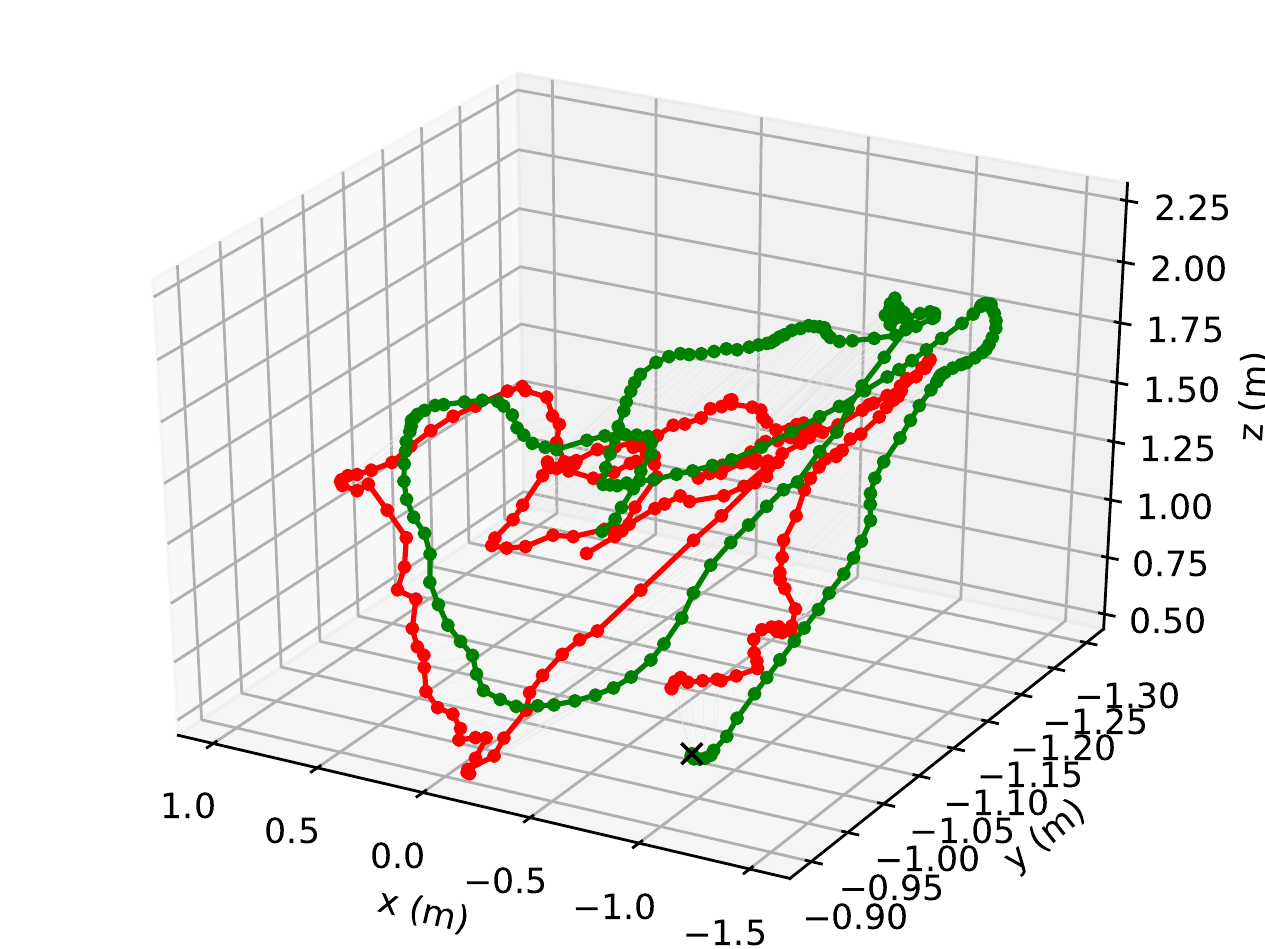}
        \includegraphics[width=\linewidth]{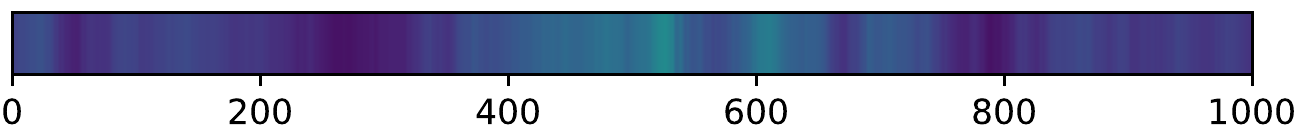}
    \end{subfigure}
    \hfill
    \begin{subfigure}{0.15\linewidth}
        \centering
        \includegraphics[width=\linewidth]{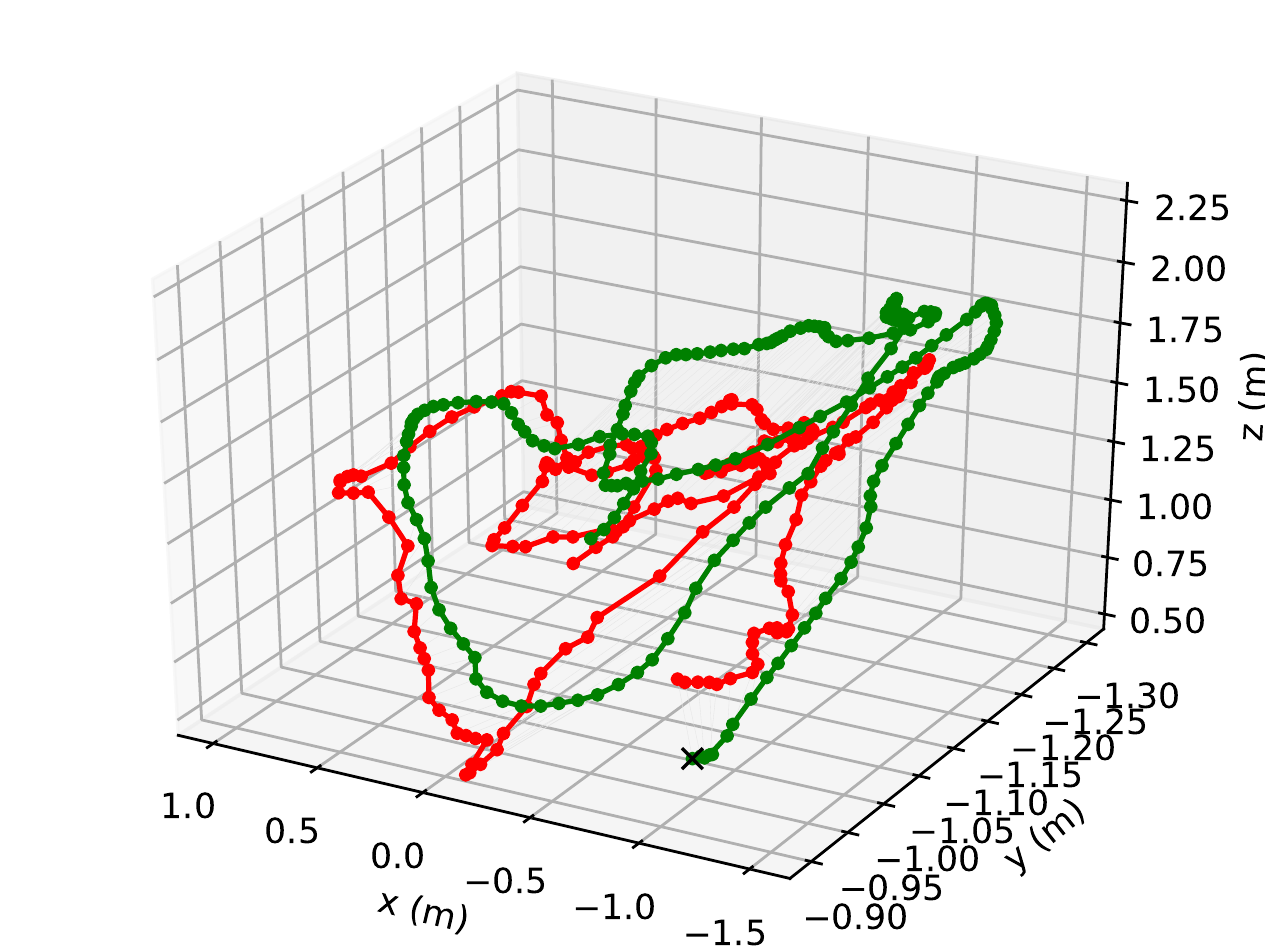}
        \includegraphics[width=\linewidth]{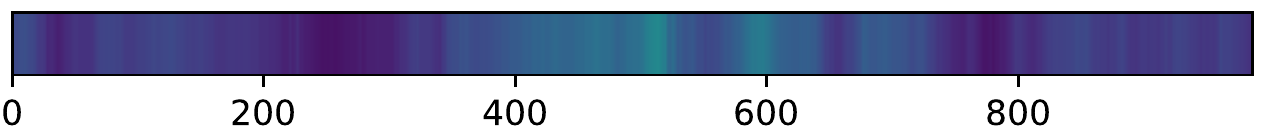}
    \end{subfigure}

    \begin{subfigure}{0.15\linewidth}
        \centering
        \includegraphics[width=\linewidth]{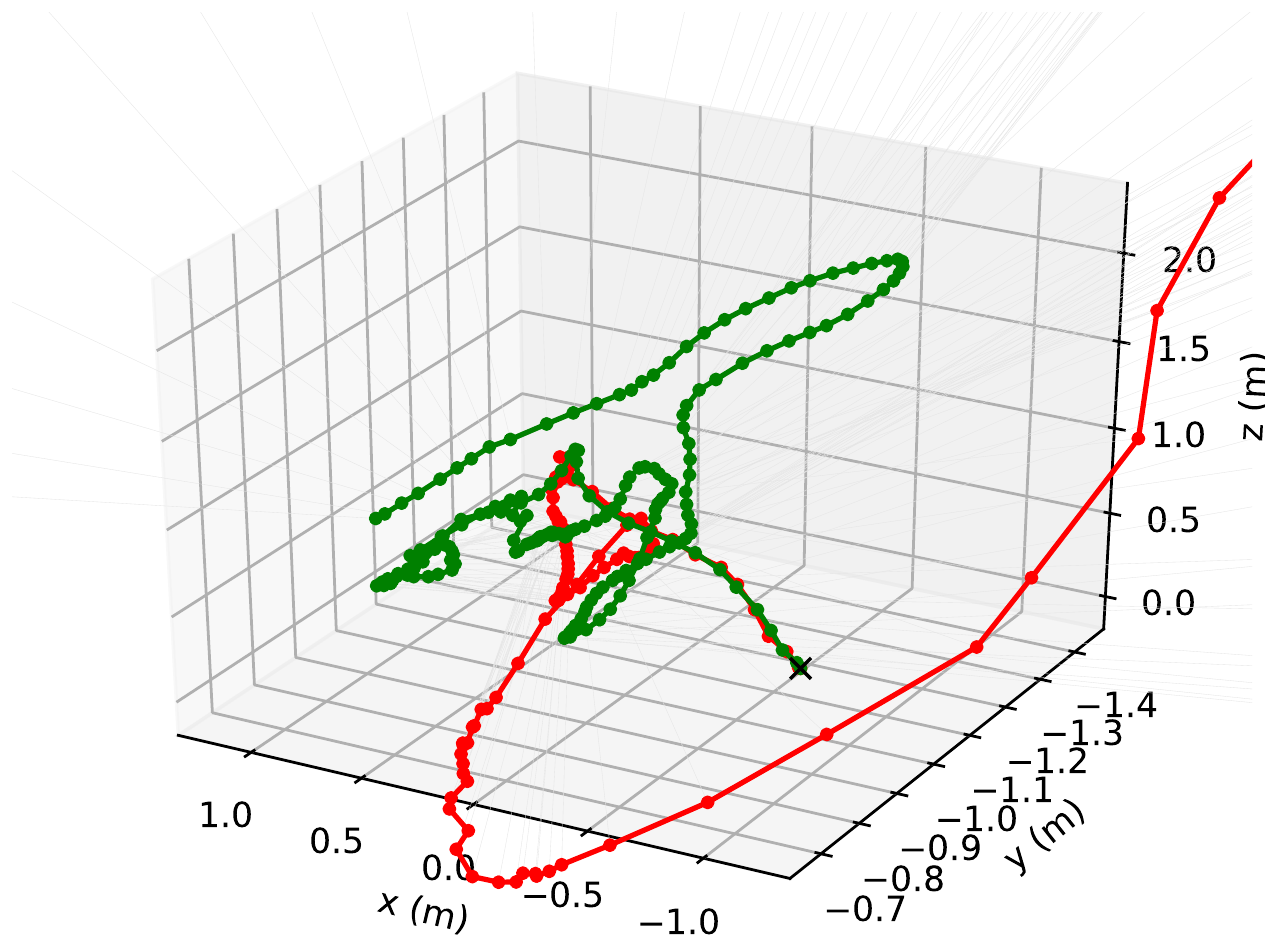}
        \includegraphics[width=\linewidth]{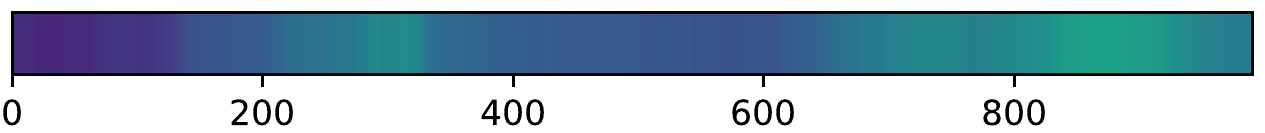}
    \end{subfigure}
    \hfill
    \begin{subfigure}{0.15\linewidth}
        \centering
        \includegraphics[width=\linewidth]{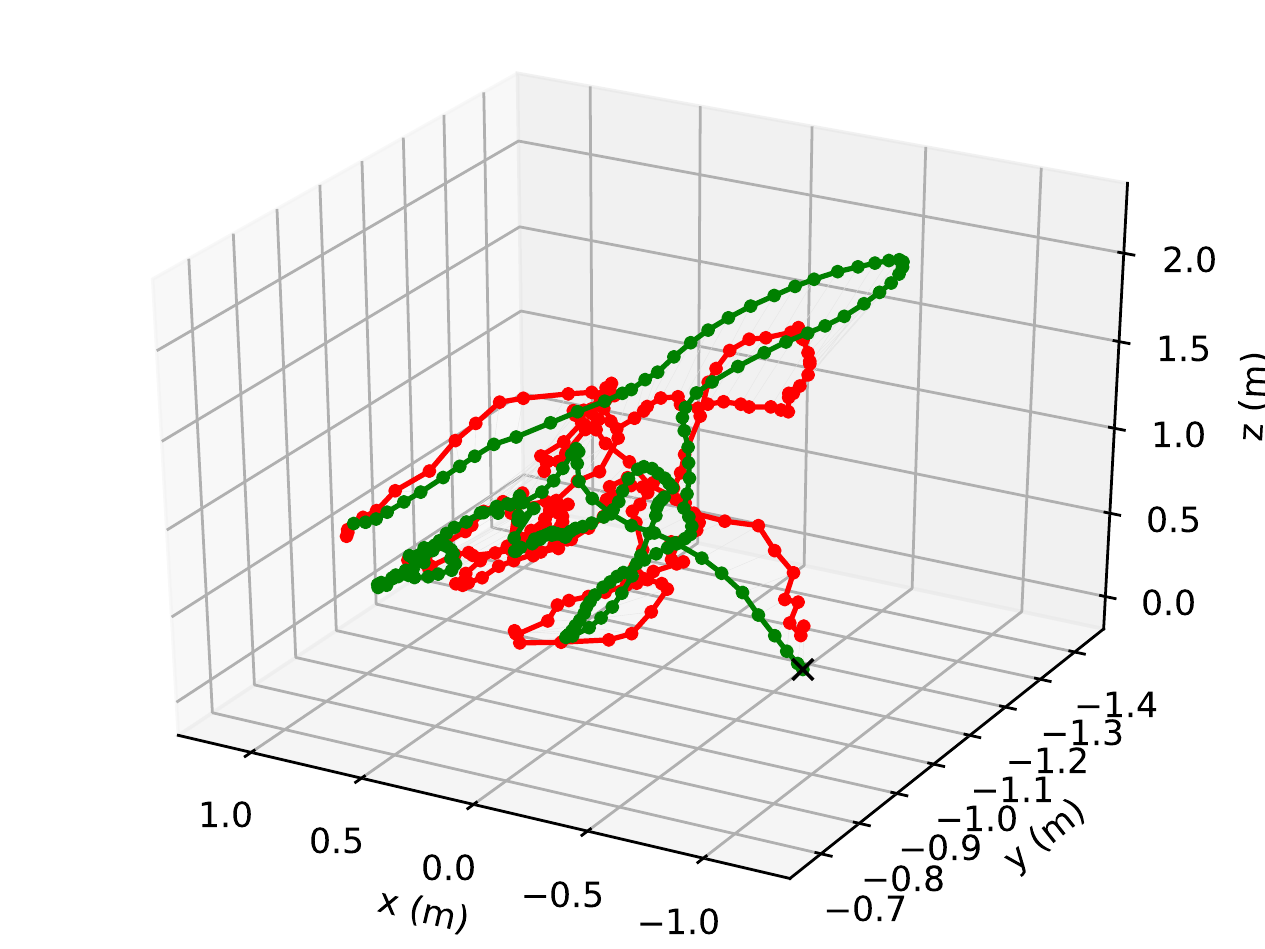}
        \includegraphics[width=\linewidth]{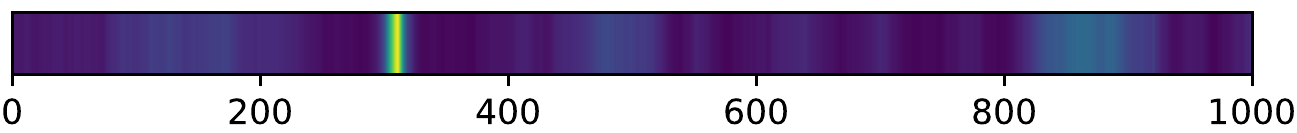}
    \end{subfigure}
    \hfill
    \begin{subfigure}{0.15\linewidth}
        \centering
        \includegraphics[width=\linewidth]{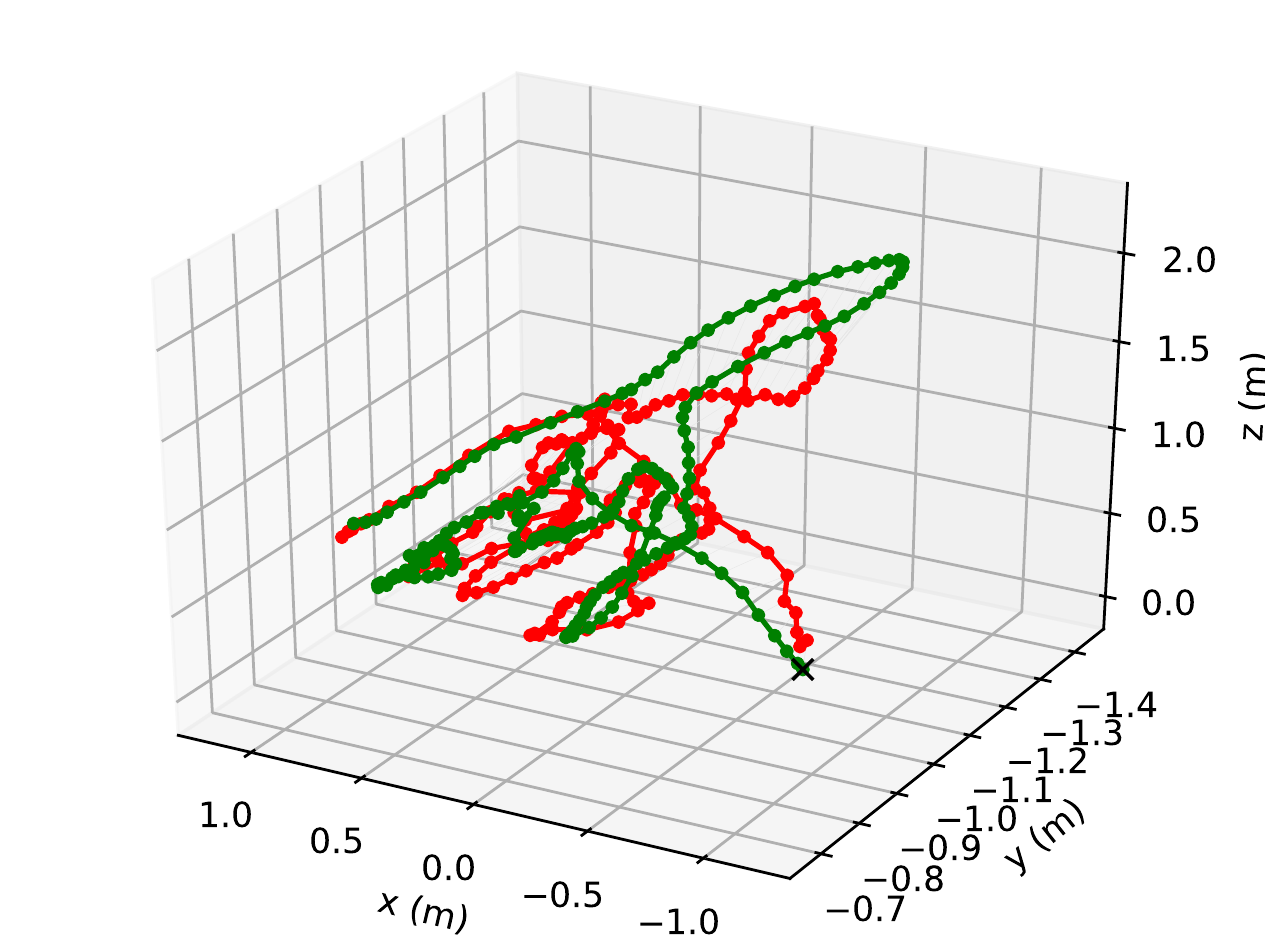}
        \includegraphics[width=\linewidth]{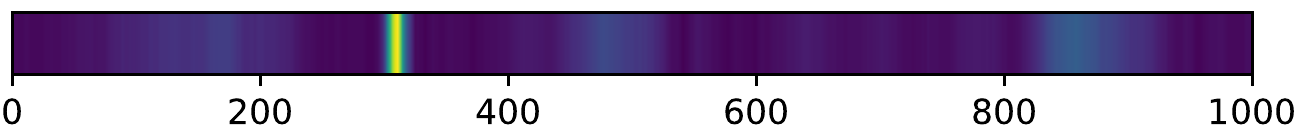}
    \end{subfigure}
    \hfill
    \begin{subfigure}{0.15\linewidth}
        \centering
        \includegraphics[width=\linewidth]{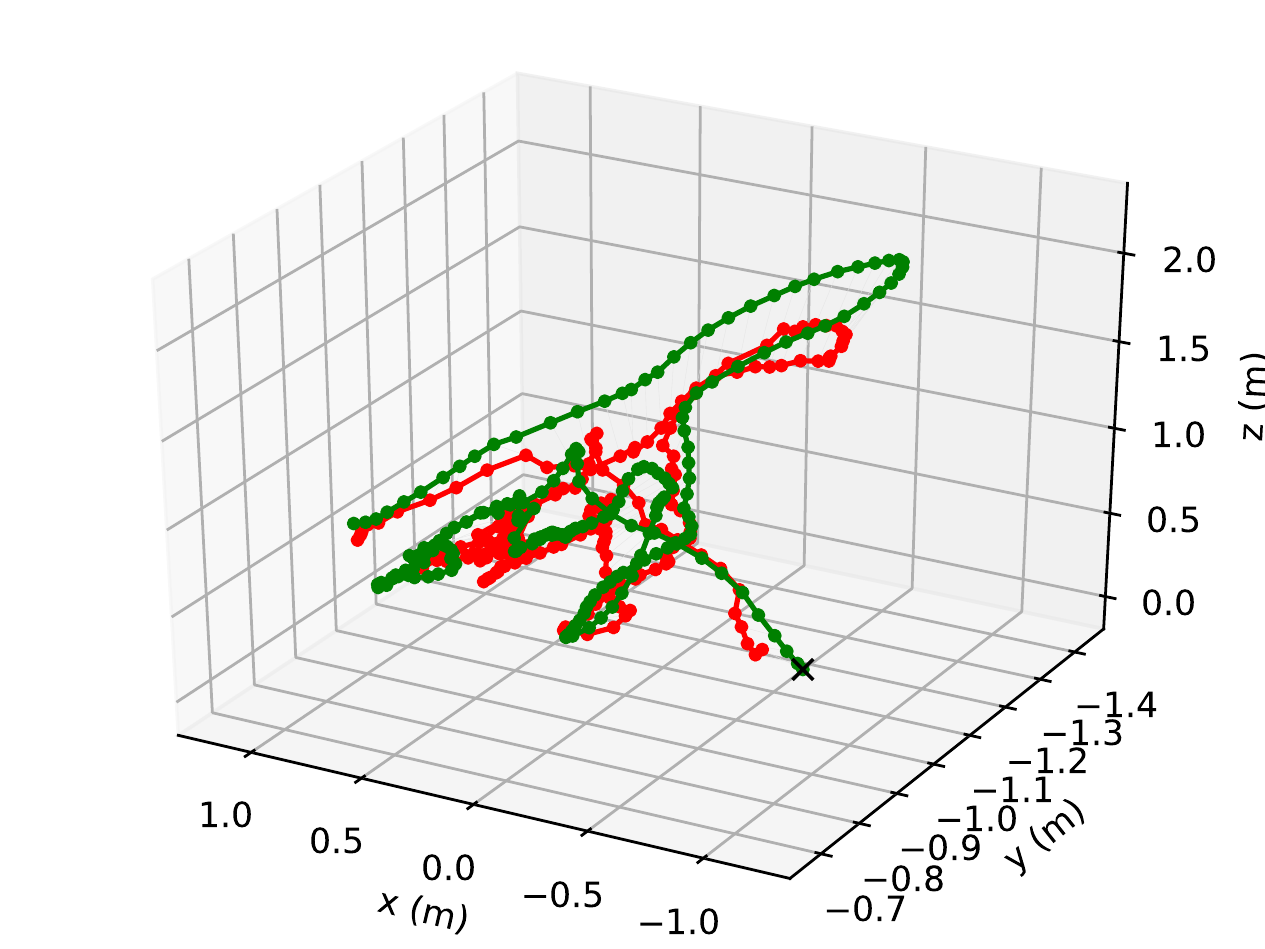}
        \includegraphics[width=\linewidth]{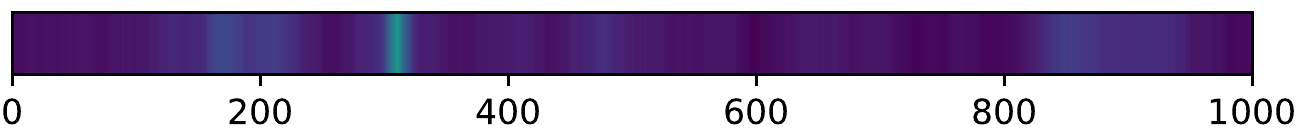}
    \end{subfigure}
    \hfill
    \begin{subfigure}{0.15\linewidth}
        \centering
        \includegraphics[width=\linewidth]{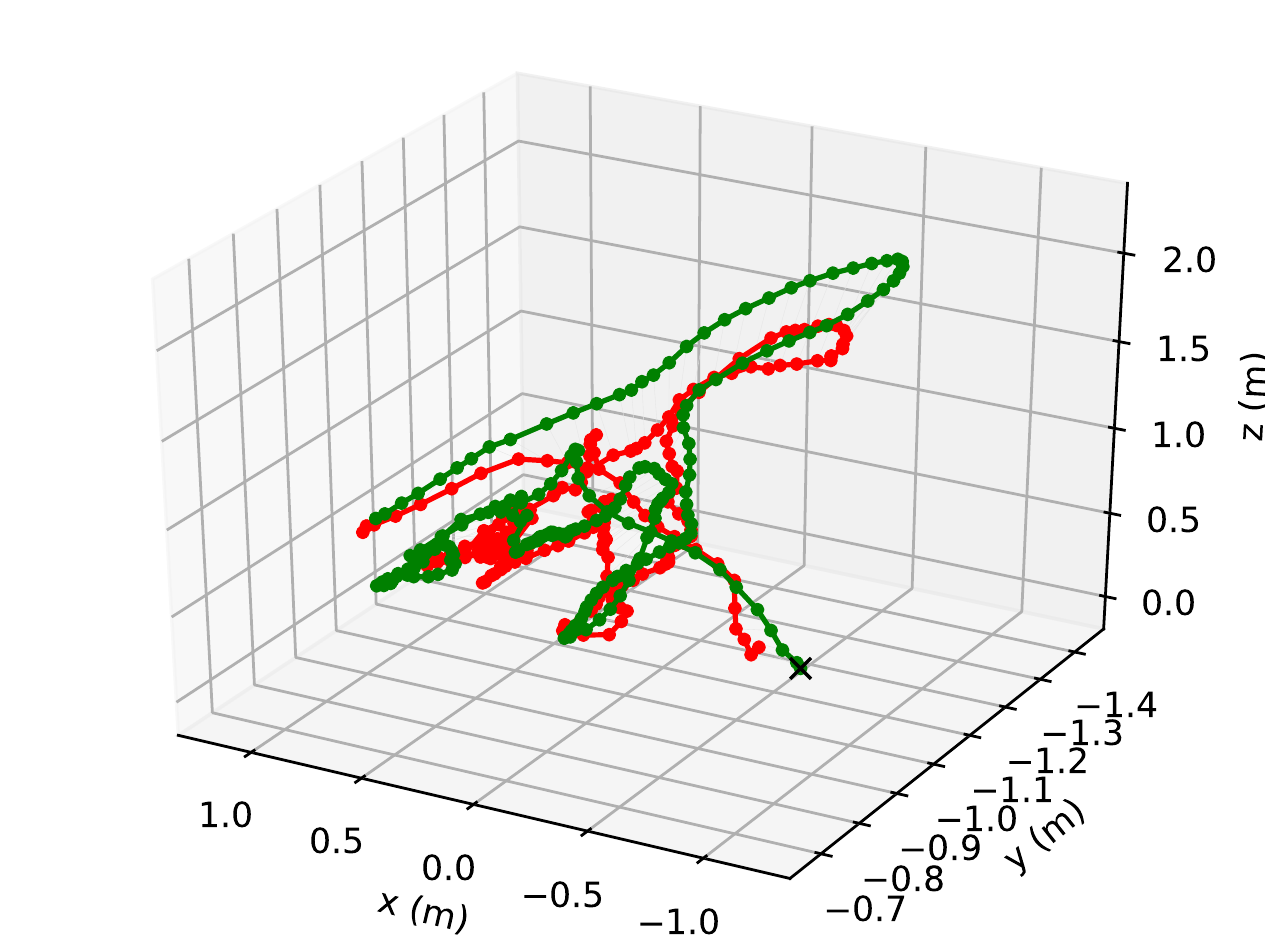}
        \includegraphics[width=\linewidth]{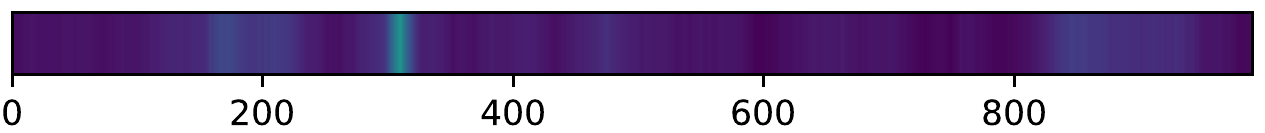}
    \end{subfigure}

    \begin{subfigure}{0.15\linewidth}
        \centering
        \includegraphics[width=\linewidth]{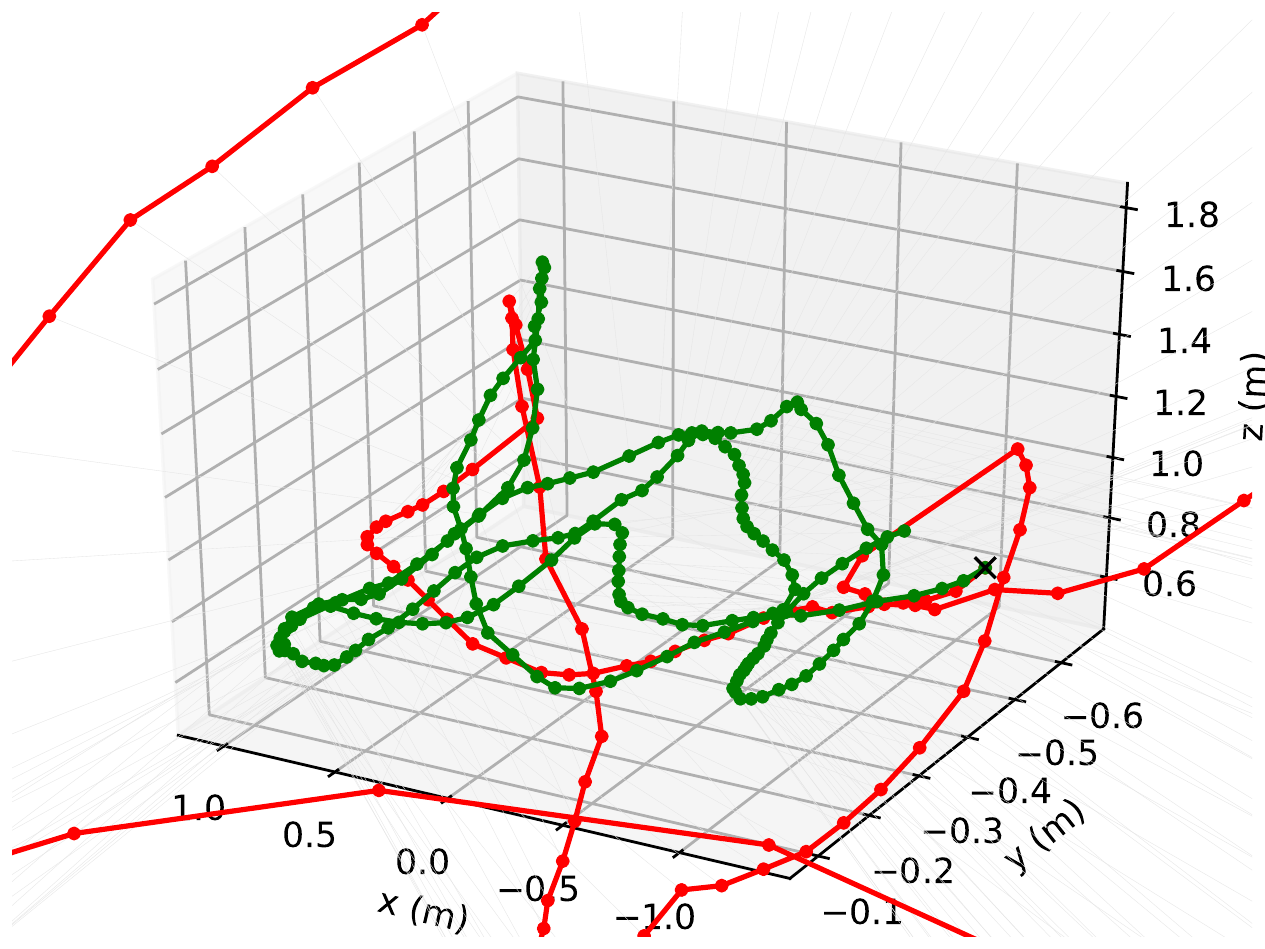}
        \includegraphics[width=\linewidth]{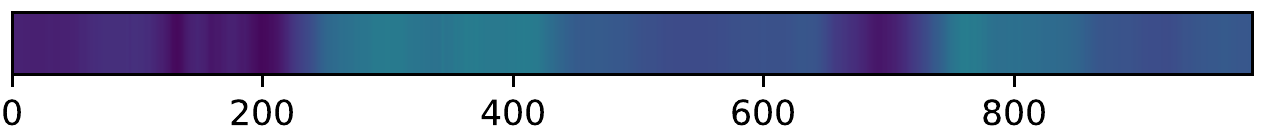}
        \caption{DSO~\cite{Engel2017DSO}}
    \end{subfigure}
    \hfill
    \begin{subfigure}{0.15\linewidth}
        \centering
        \includegraphics[width=\linewidth]{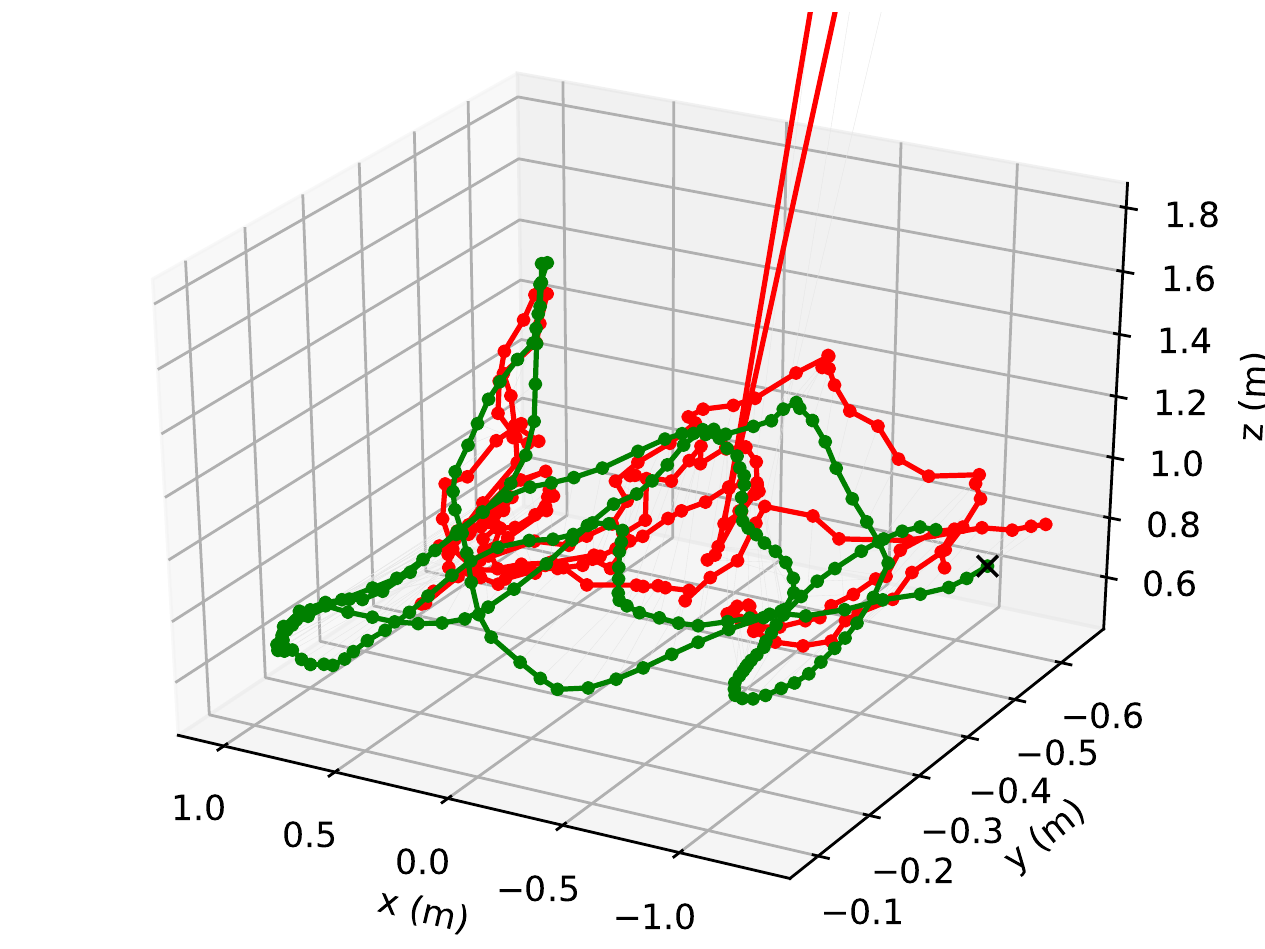}
        \includegraphics[width=\linewidth]{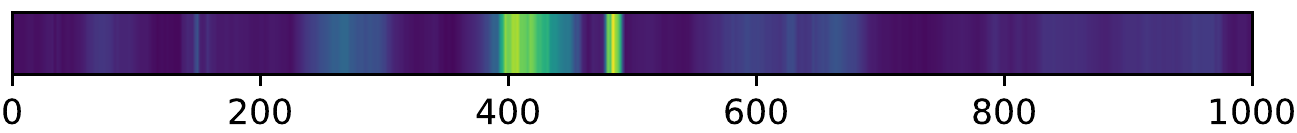}
        \caption{PoseNet~\cite{Kendall17cvpr, Kendall15iccv, Kendall16icra}}
    \end{subfigure}
    \hfill
    \begin{subfigure}{0.15\linewidth}
        \centering
        \includegraphics[width=\linewidth]{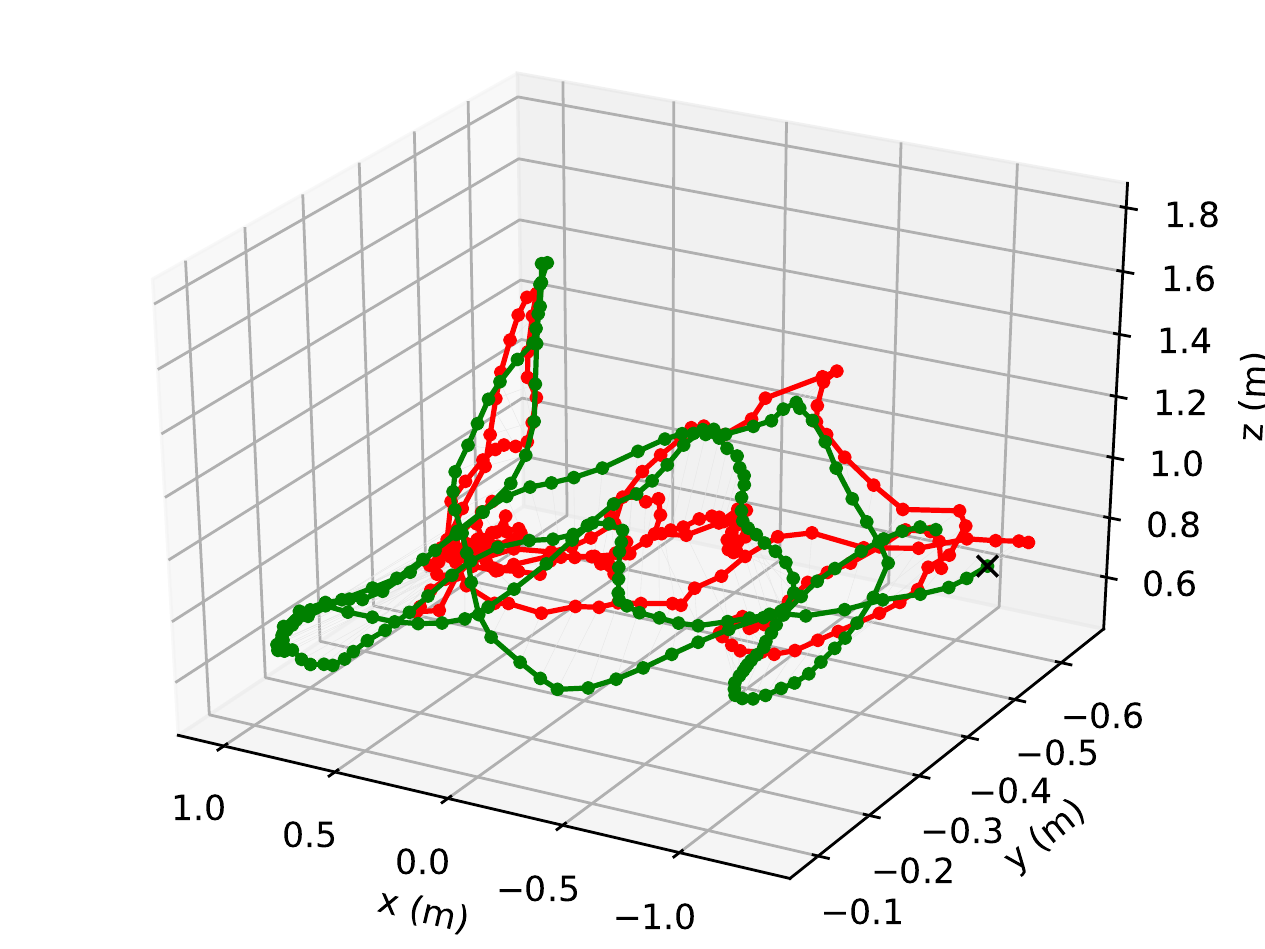}
        \includegraphics[width=\linewidth]{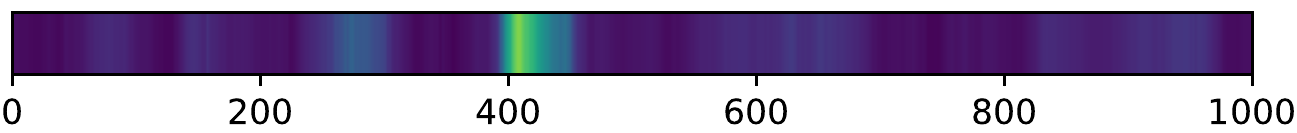}
        \caption{MapNet}
    \end{subfigure}
    \hfill
    \begin{subfigure}{0.15\linewidth}
        \centering
        \includegraphics[width=\linewidth]{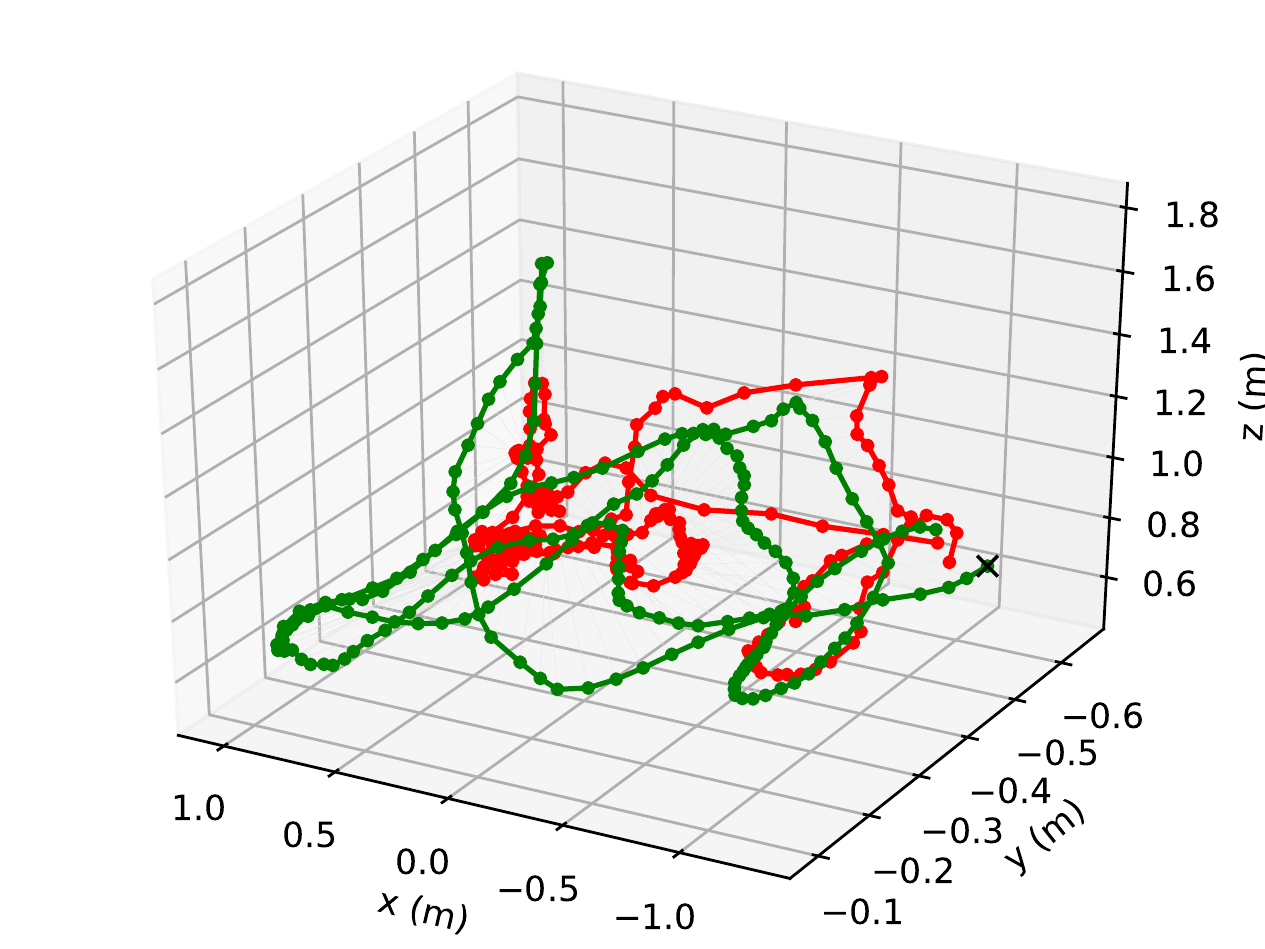}
        \includegraphics[width=\linewidth]{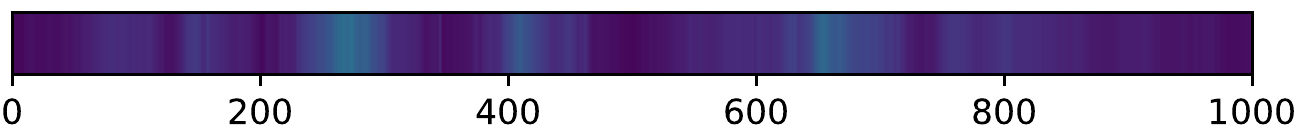}
        \caption{MapNet+}
    \end{subfigure}
    \hfill
    \begin{subfigure}{0.15\linewidth}
        \centering
        \includegraphics[width=\linewidth]{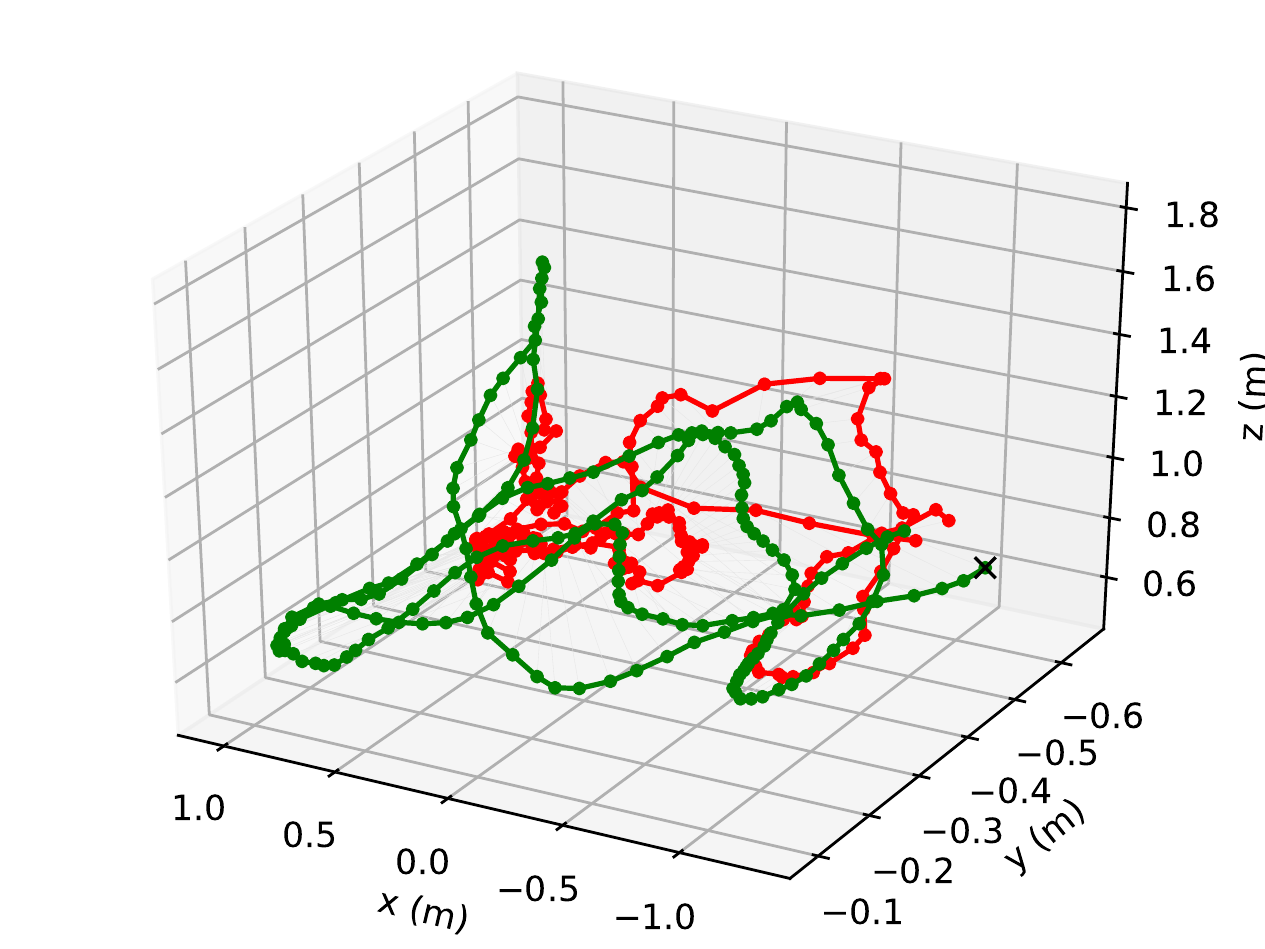}
        \includegraphics[width=\linewidth]{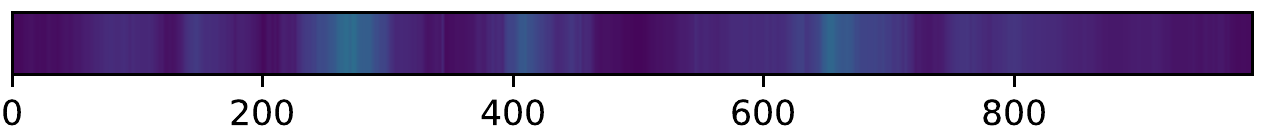}
        \caption{MapNet+PGO}
    \end{subfigure}
    \vspace{-1em}
    \caption{\small \textbf{Results on the 7-Scenes dataset.} 
    The 3d plots show the camera position (green for ground truth and red for predictions). The colorbars below 
    show the errors of the predicted camera orientation (blue for small error and yellow for large error) with
    frame number on the X axis. 
    From top to bottom are testing sequences: Chess-Seq-03, Chess-Seq-05, Fire-Seq-03, Fire-Seq-04, 
    Head-Seq-01, Office-Seq-02, Office-Seq-06, Office-Seq-07, and Office-Seq-09.} 
    \label{fig:res_7scenes}
\end{figure*}

\begin{figure*}
    \captionsetup[subfigure]{labelformat=empty}
    \centering
    \begin{subfigure}{0.15\linewidth}
        \centering
        \includegraphics[width=\linewidth]{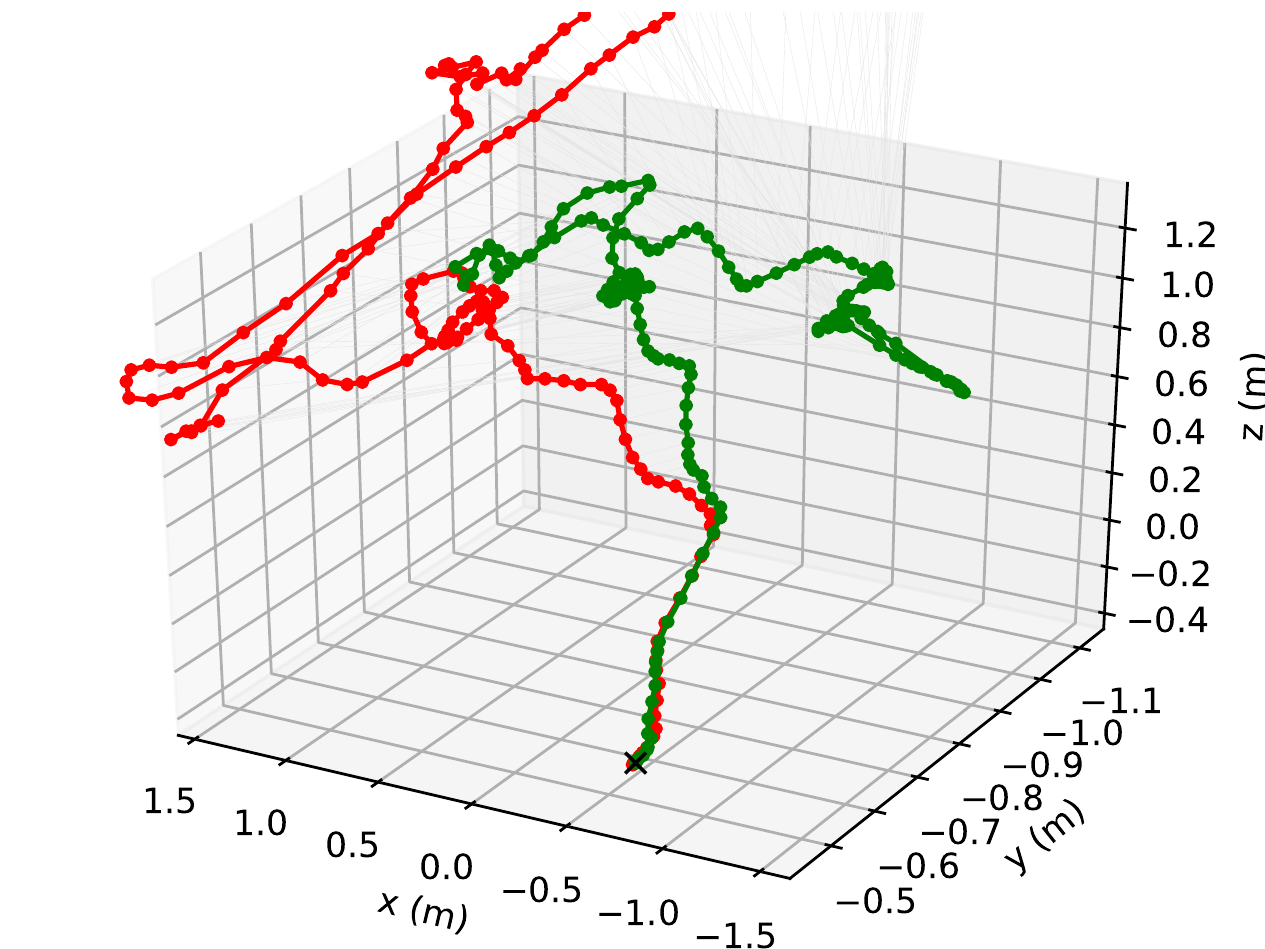}
        \includegraphics[width=\linewidth]{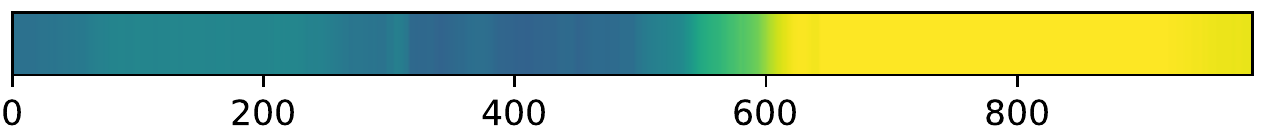}
    \end{subfigure}
    \hfill
    \begin{subfigure}{0.15\linewidth}
        \centering
        \includegraphics[width=\linewidth]{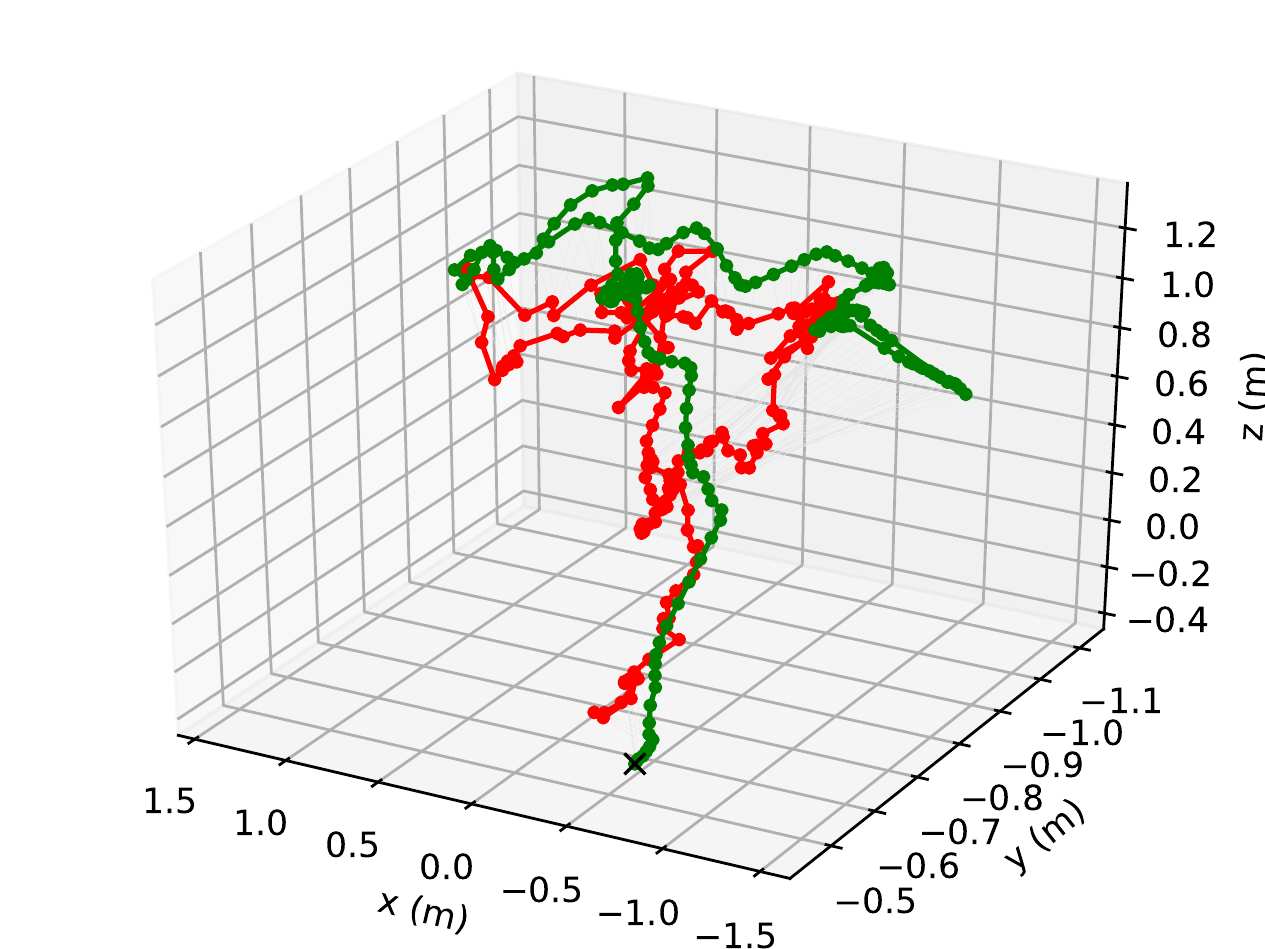}
        \includegraphics[width=\linewidth]{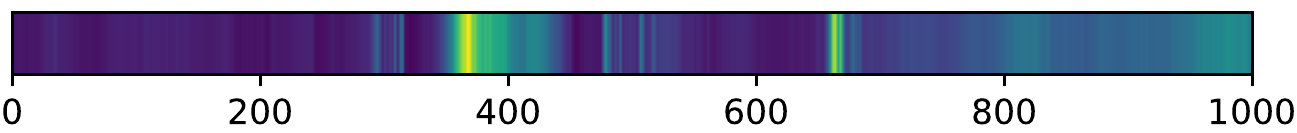}
    \end{subfigure}
    \hfill
    \begin{subfigure}{0.15\linewidth}
        \centering
        \includegraphics[width=\linewidth]{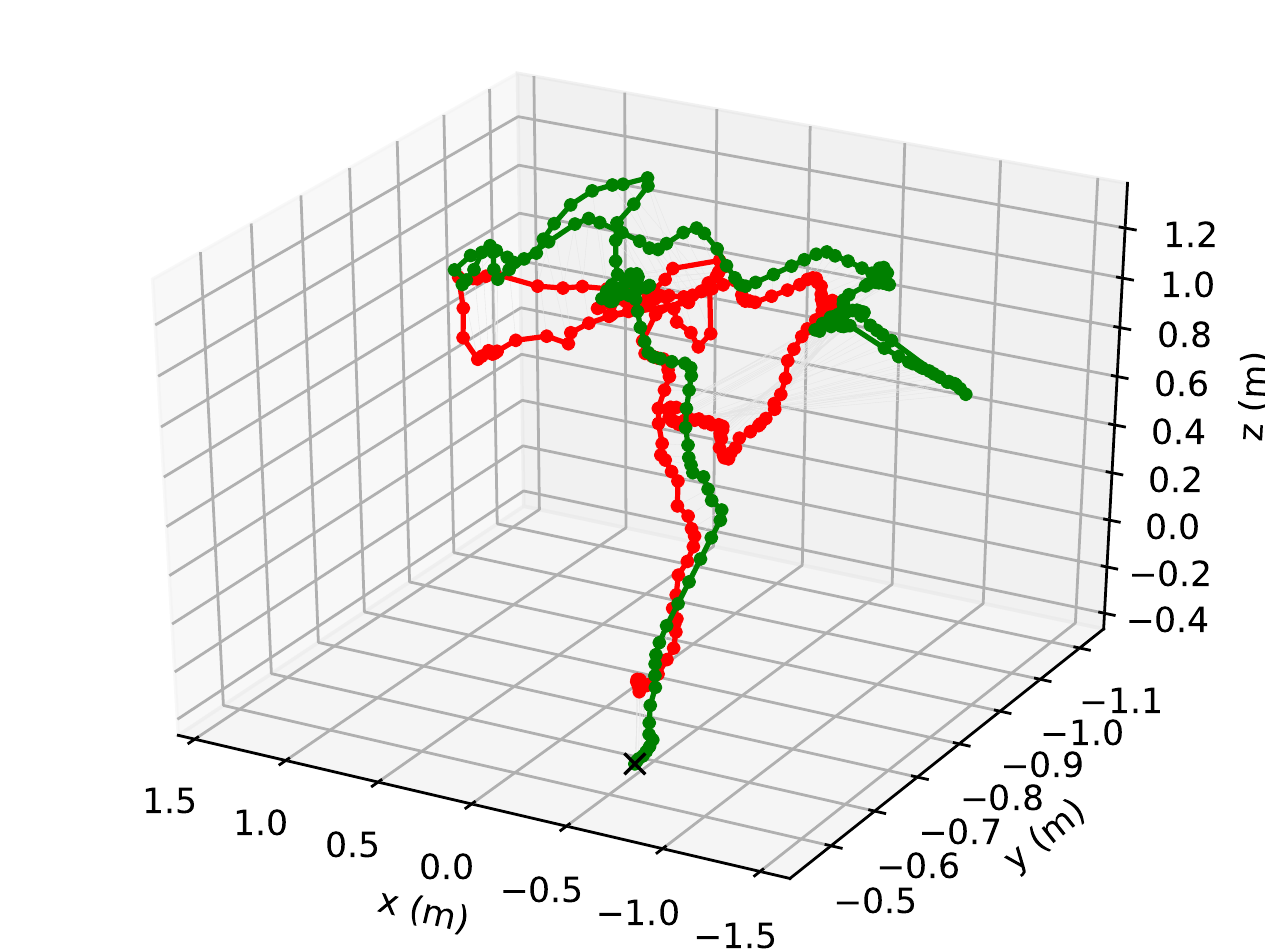}
        \includegraphics[width=\linewidth]{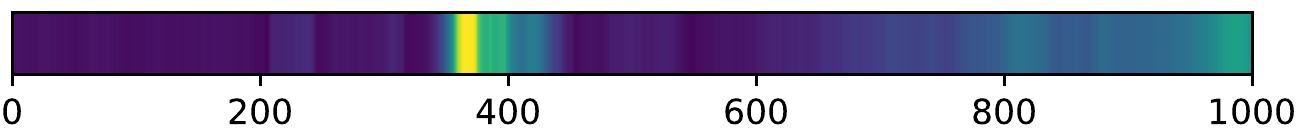}
    \end{subfigure}
    \hfill
    \begin{subfigure}{0.15\linewidth}
        \centering
        \includegraphics[width=\linewidth]{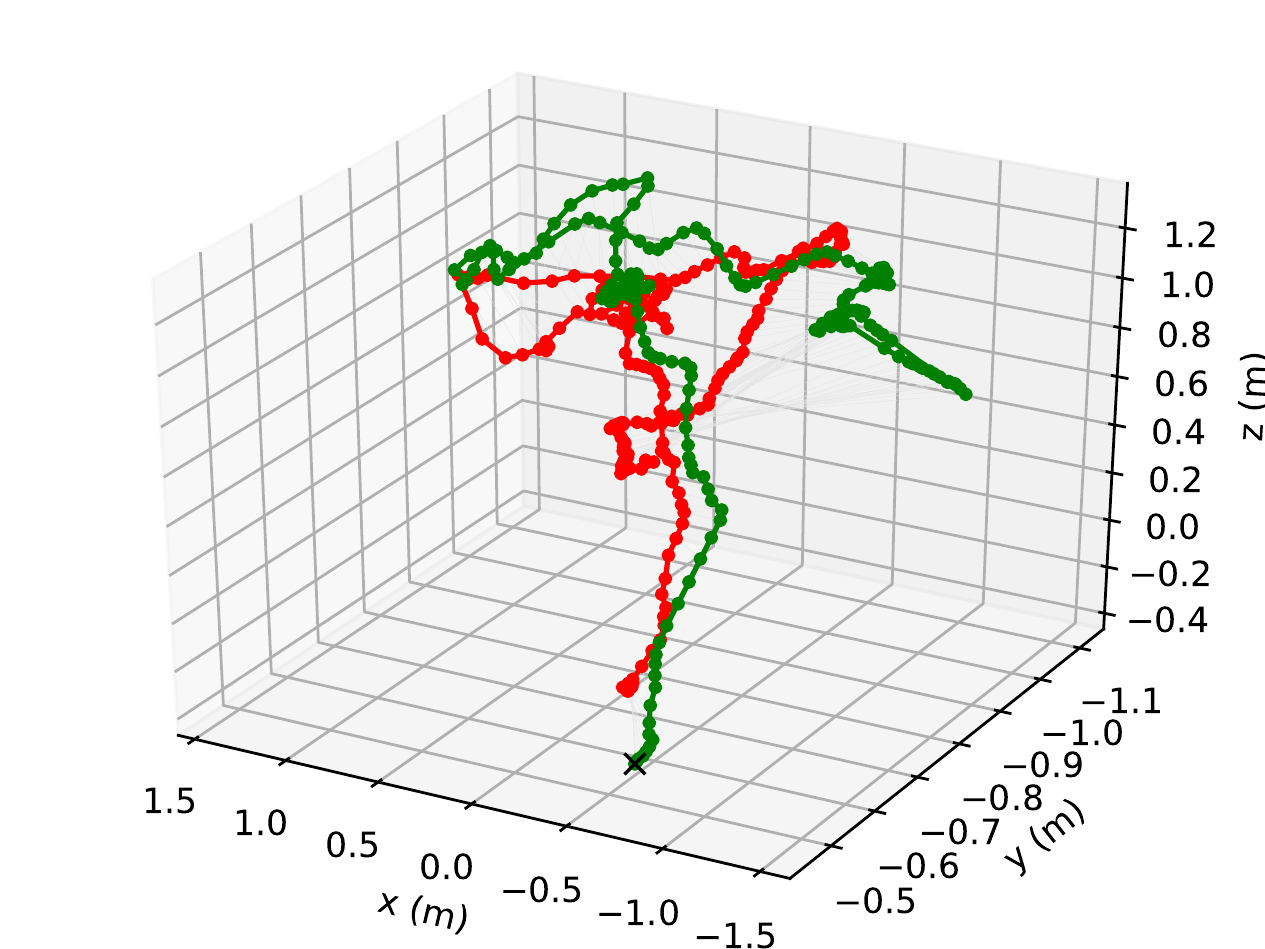}
        \includegraphics[width=\linewidth]{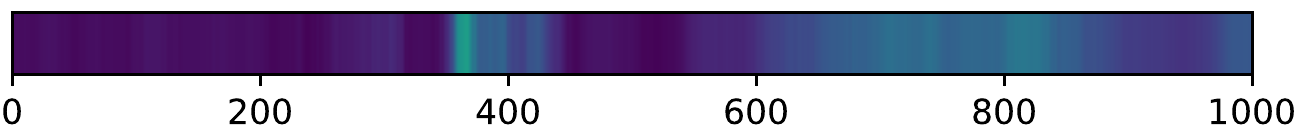}
    \end{subfigure}
    \hfill
    \begin{subfigure}{0.15\linewidth}
        \centering
        \includegraphics[width=\linewidth]{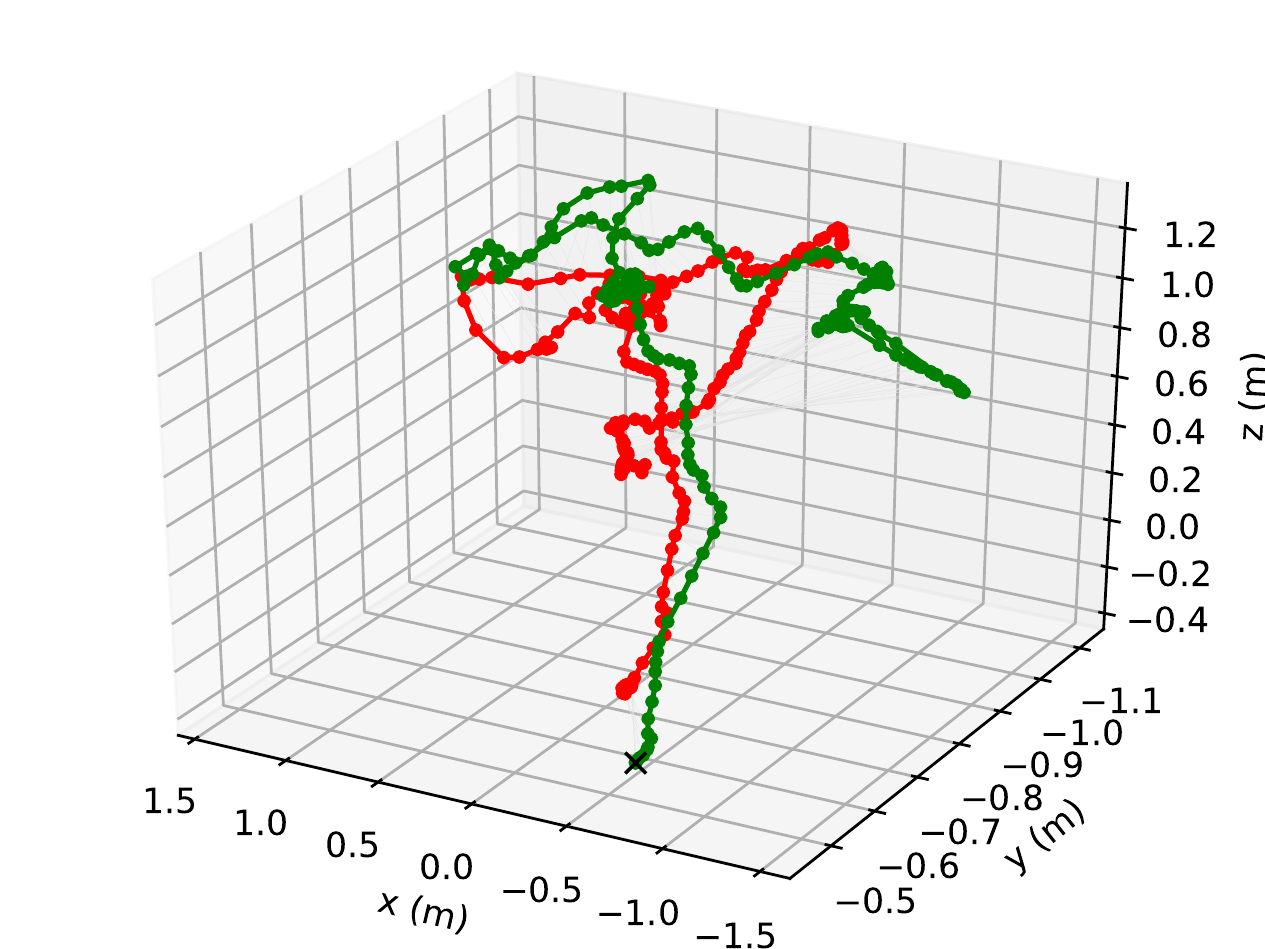}
        \includegraphics[width=\linewidth]{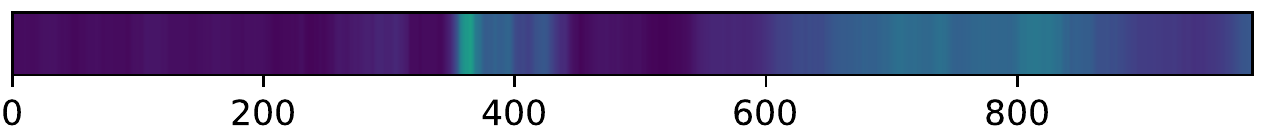}
    \end{subfigure}

    \begin{subfigure}{0.15\linewidth}
        \centering
        \includegraphics[width=\linewidth]{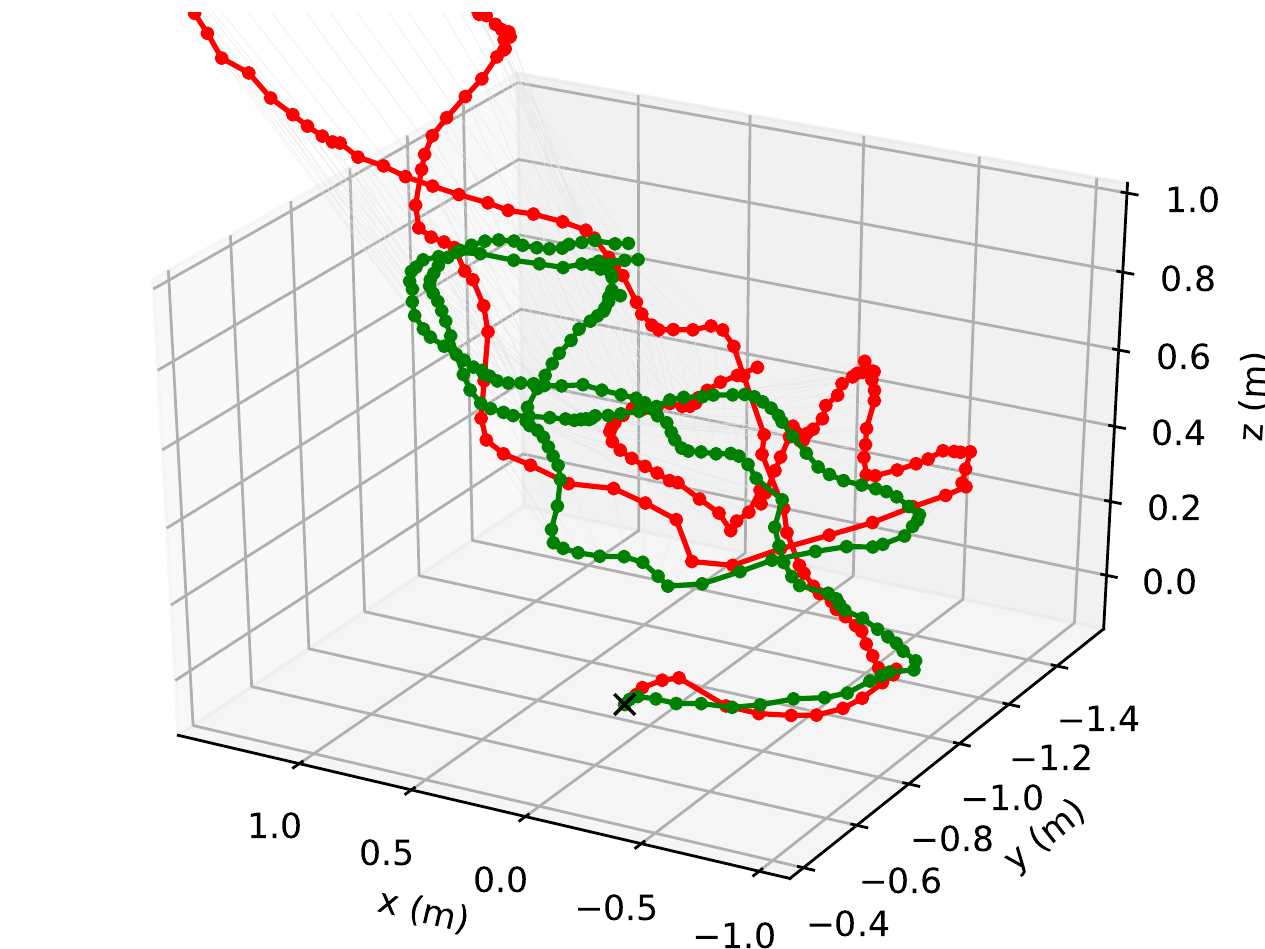}
        \includegraphics[width=\linewidth]{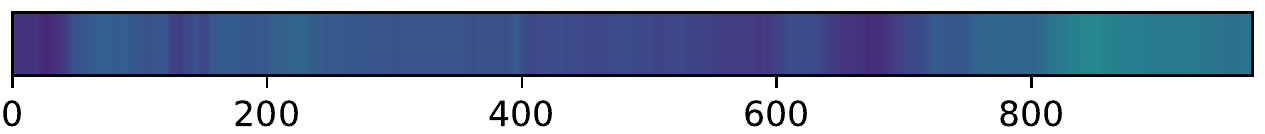}
    \end{subfigure}
    \hfill
    \begin{subfigure}{0.15\linewidth}
        \centering
        \includegraphics[width=\linewidth]{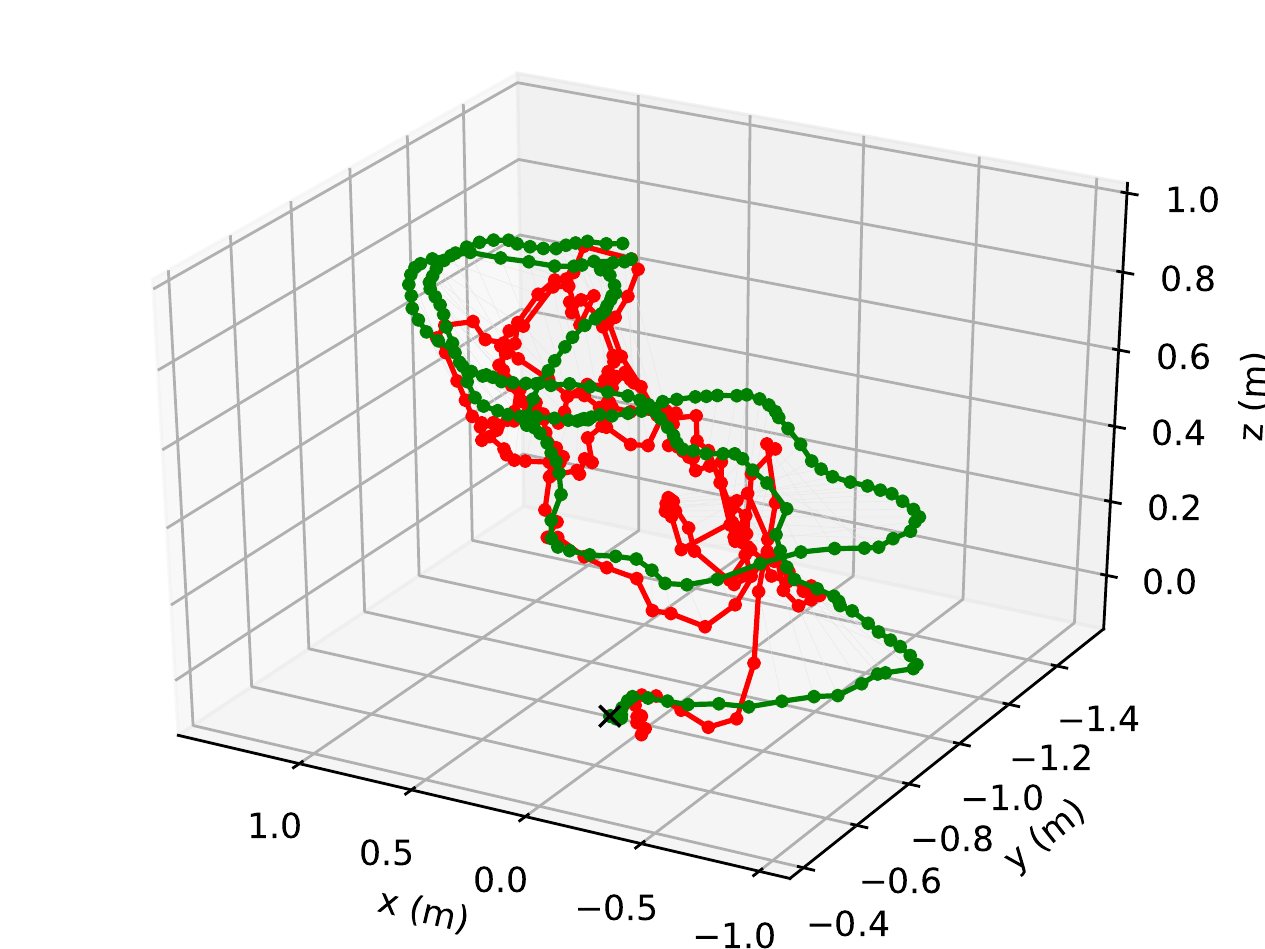}
        \includegraphics[width=\linewidth]{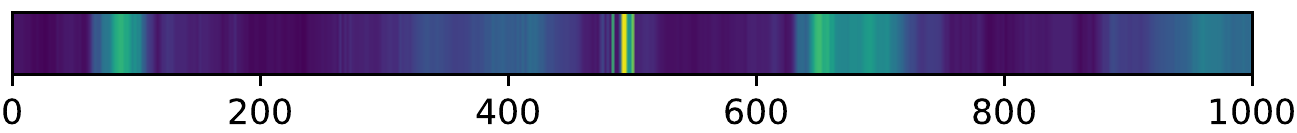}
    \end{subfigure}
    \hfill
    \begin{subfigure}{0.15\linewidth}
        \centering
        \includegraphics[width=\linewidth]{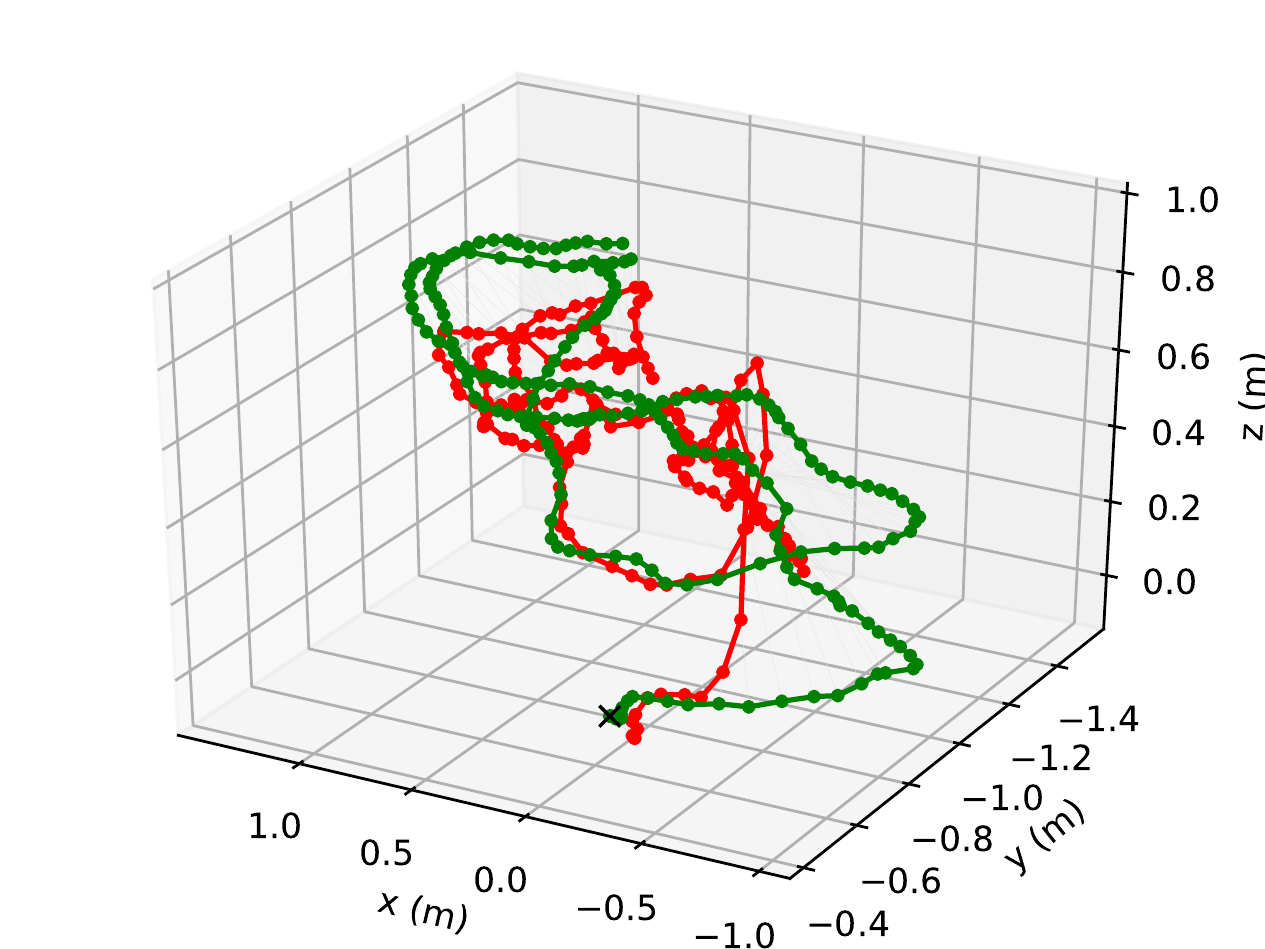}
        \includegraphics[width=\linewidth]{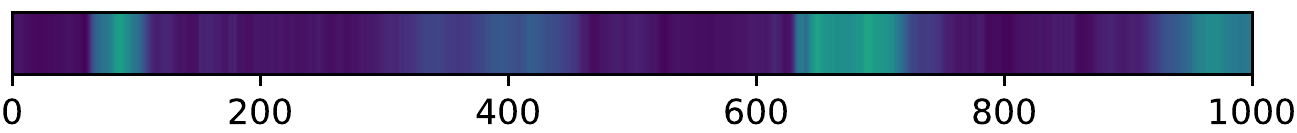}
    \end{subfigure}
    \hfill
    \begin{subfigure}{0.15\linewidth}
        \centering
        \includegraphics[width=\linewidth]{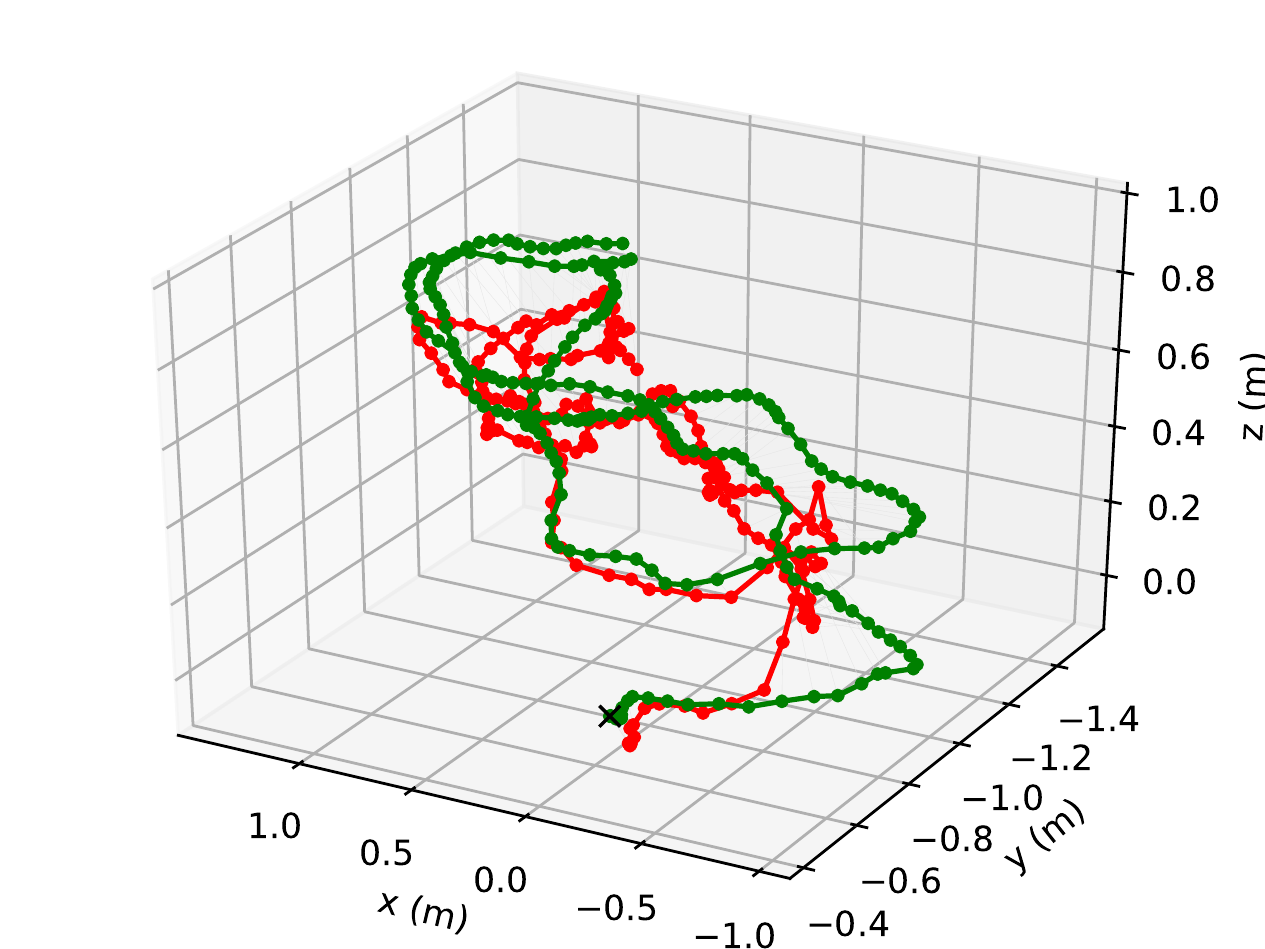}
        \includegraphics[width=\linewidth]{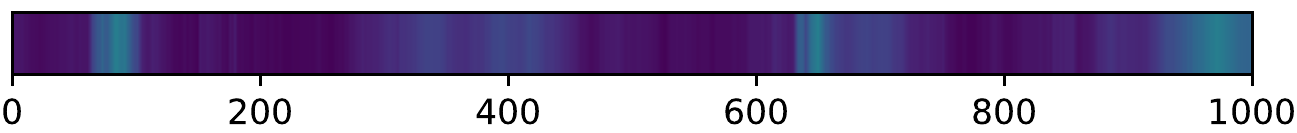}
    \end{subfigure}
    \hfill
    \begin{subfigure}{0.15\linewidth}
        \centering
        \includegraphics[width=\linewidth]{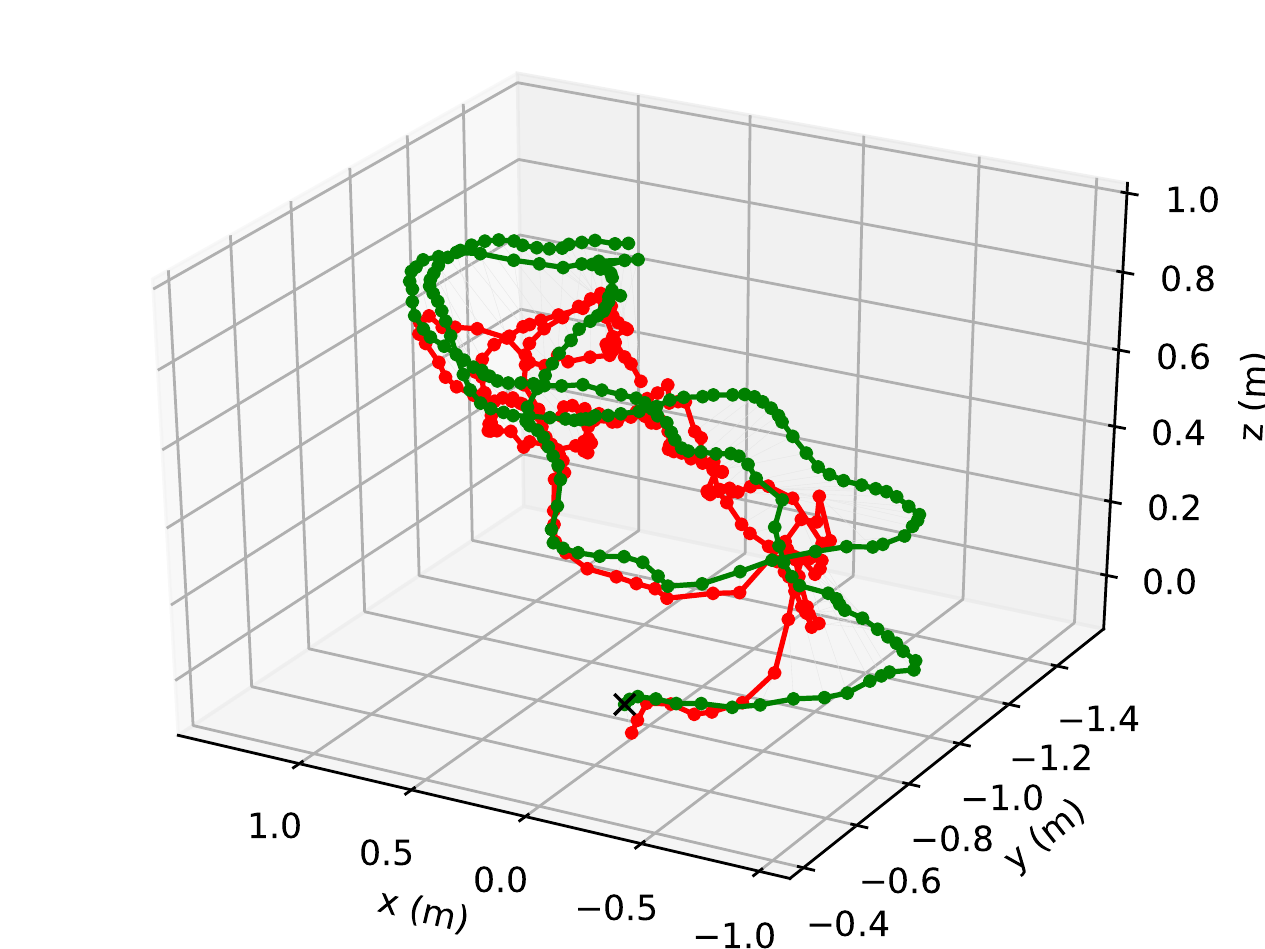}
        \includegraphics[width=\linewidth]{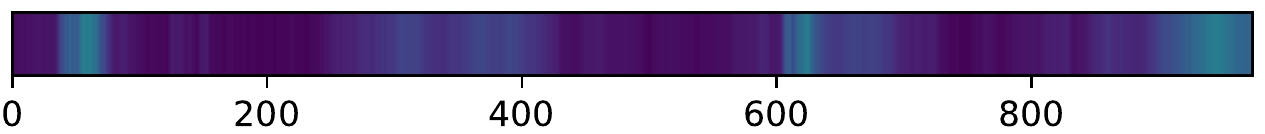}
    \end{subfigure}

    \begin{subfigure}{0.15\linewidth}
        \centering
        \includegraphics[width=\linewidth]{figures/7scenes/redkitchen_dsoA_0_t.pdf}
        \includegraphics[width=\linewidth]{figures/7scenes/redkitchen_dsoA_0_q.pdf}
    \end{subfigure}
    \hfill
    \begin{subfigure}{0.15\linewidth}
        \centering
        \includegraphics[width=\linewidth]{figures/7scenes/redkitchen_posenet_0_t.pdf}
        \includegraphics[width=\linewidth]{figures/7scenes/redkitchen_posenet_0_q.pdf}
    \end{subfigure}
    \hfill
    \begin{subfigure}{0.15\linewidth}
        \centering
        \includegraphics[width=\linewidth]{figures/7scenes/redkitchen_vidvo_0_t.pdf}
        \includegraphics[width=\linewidth]{figures/7scenes/redkitchen_vidvo_0_q.pdf}
    \end{subfigure}
    \hfill
    \begin{subfigure}{0.15\linewidth}
        \centering
        \includegraphics[width=\linewidth]{figures/7scenes/redkitchen_vidvo_online_0_t.pdf}
        \includegraphics[width=\linewidth]{figures/7scenes/redkitchen_vidvo_online_0_q.pdf}
    \end{subfigure}
    \hfill
    \begin{subfigure}{0.15\linewidth}
        \centering
        \includegraphics[width=\linewidth]{figures/7scenes/redkitchen_vidvo_online_pgo_dso_0_t.pdf}
        \includegraphics[width=\linewidth]{figures/7scenes/redkitchen_vidvo_online_pgo_dso_0_q.pdf}
    \end{subfigure}

    \begin{subfigure}{0.15\linewidth}
        \centering
        \includegraphics[width=\linewidth]{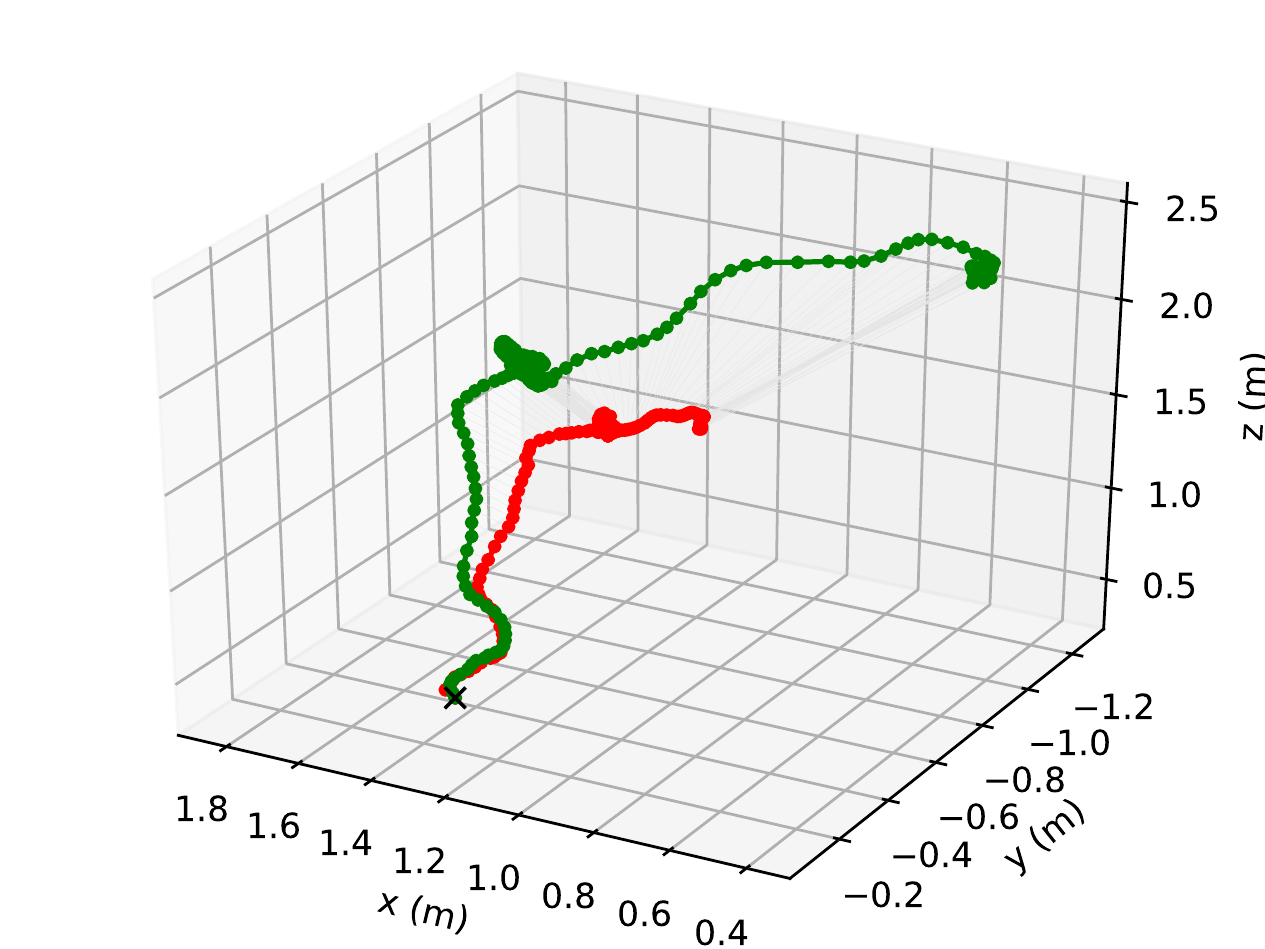}
        \includegraphics[width=\linewidth]{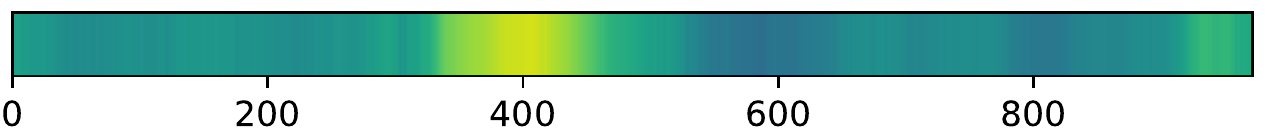}
    \end{subfigure}
    \hfill
    \begin{subfigure}{0.15\linewidth}
        \centering
        \includegraphics[width=\linewidth]{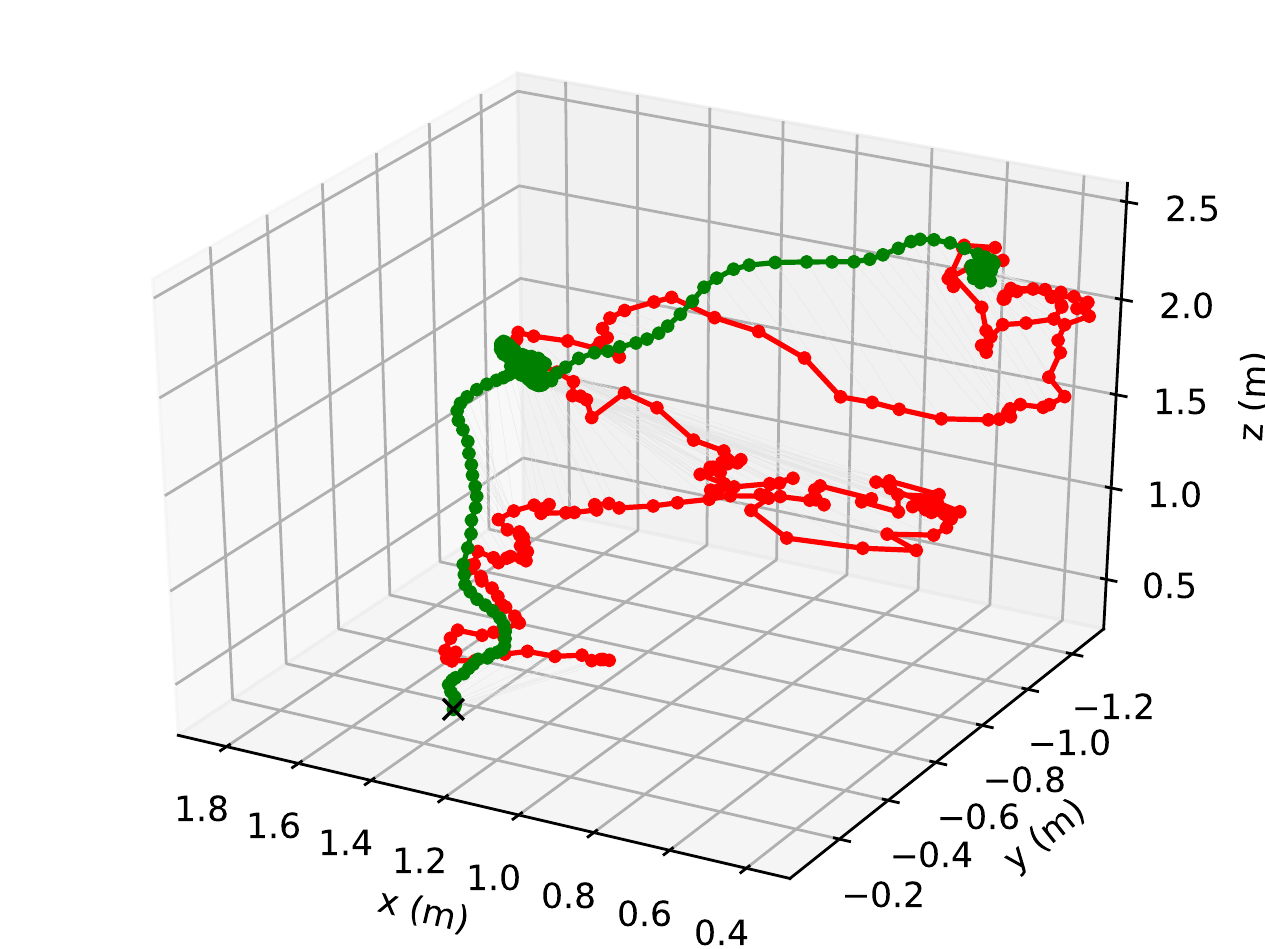}
        \includegraphics[width=\linewidth]{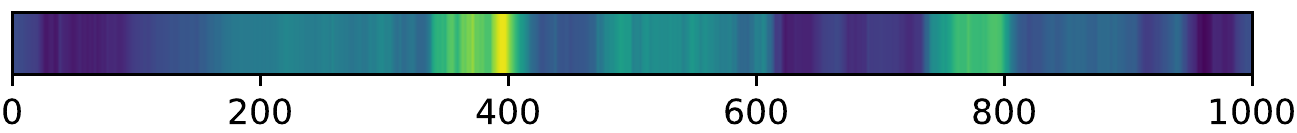}
    \end{subfigure}
    \hfill
    \begin{subfigure}{0.15\linewidth}
        \centering
        \includegraphics[width=\linewidth]{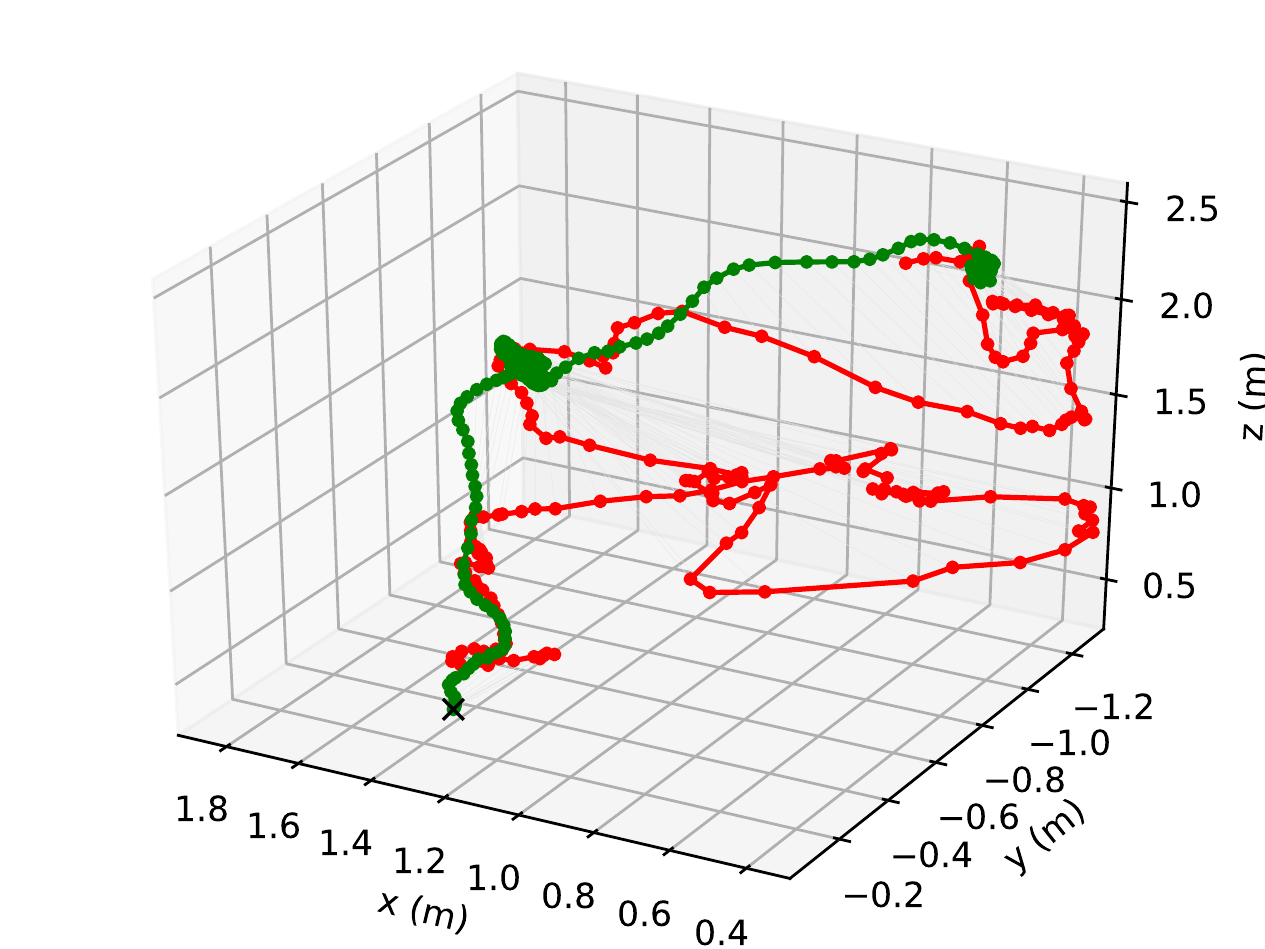}
        \includegraphics[width=\linewidth]{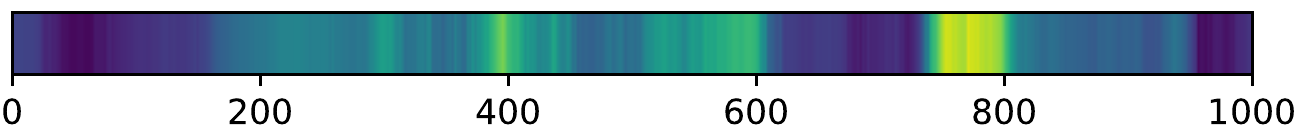}
    \end{subfigure}
    \hfill
    \begin{subfigure}{0.15\linewidth}
        \centering
        \includegraphics[width=\linewidth]{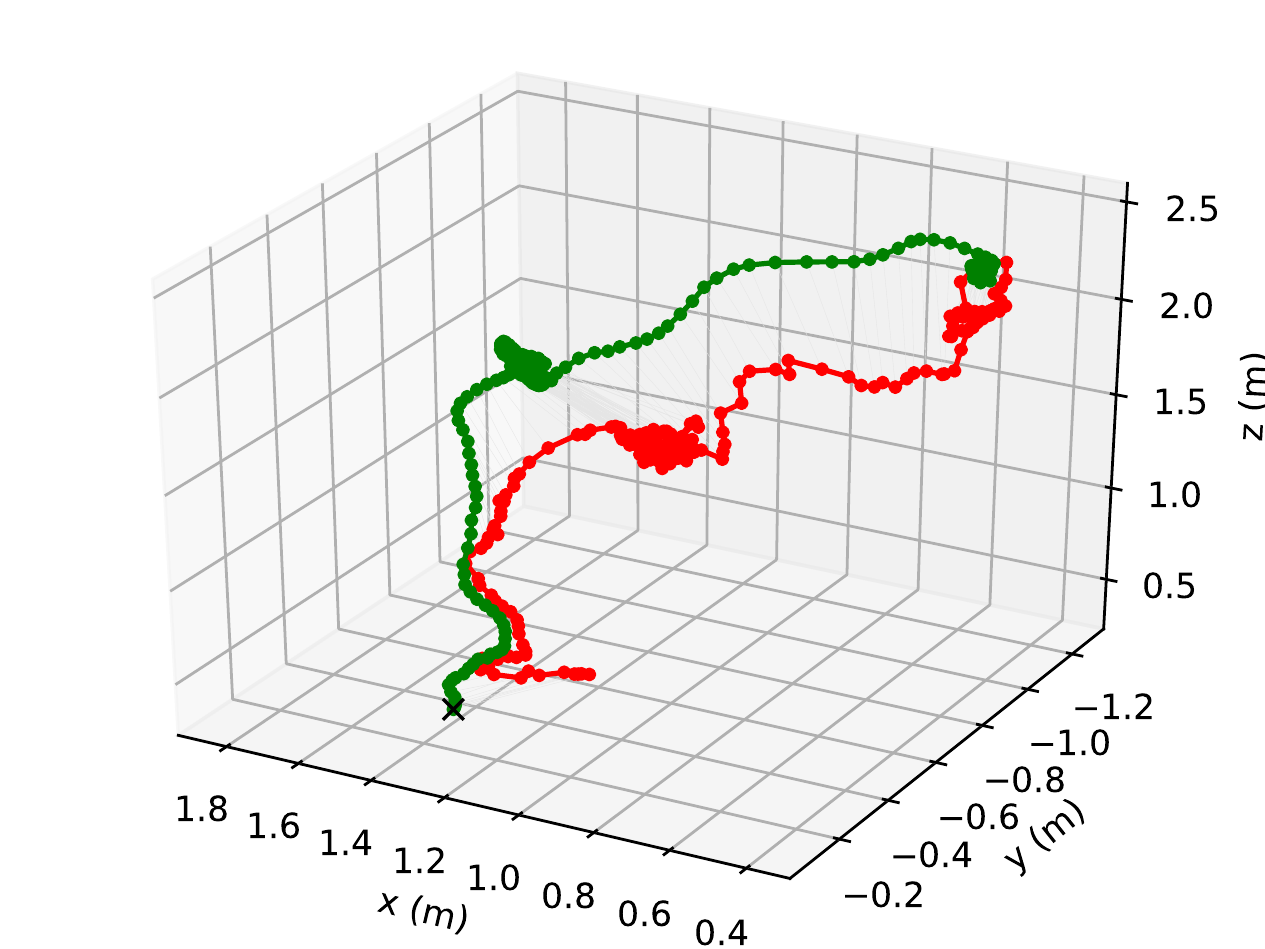}
        \includegraphics[width=\linewidth]{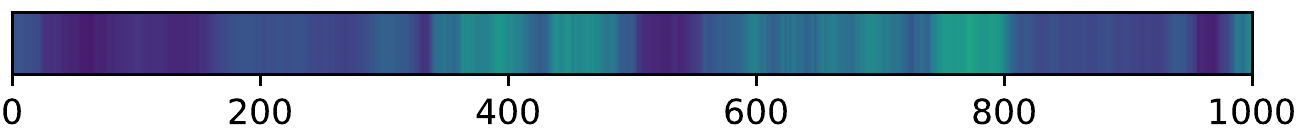}
    \end{subfigure}
    \hfill
    \begin{subfigure}{0.15\linewidth}
        \centering
        \includegraphics[width=\linewidth]{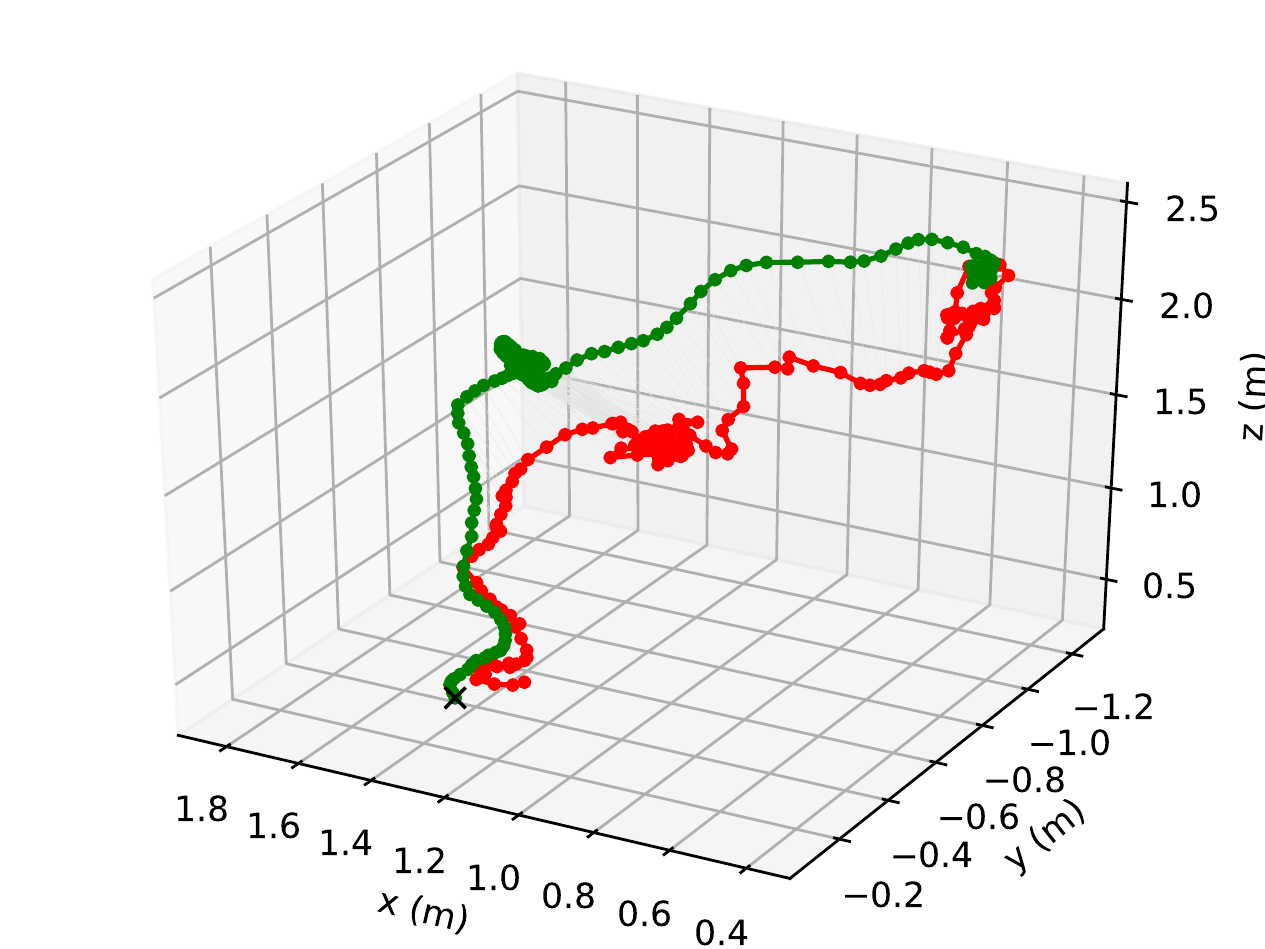}
        \includegraphics[width=\linewidth]{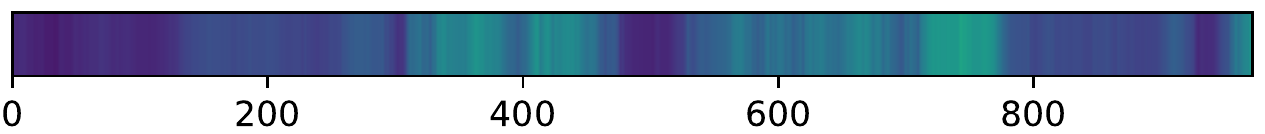}
    \end{subfigure}

    \begin{subfigure}{0.15\linewidth}
        \centering
        \includegraphics[width=\linewidth]{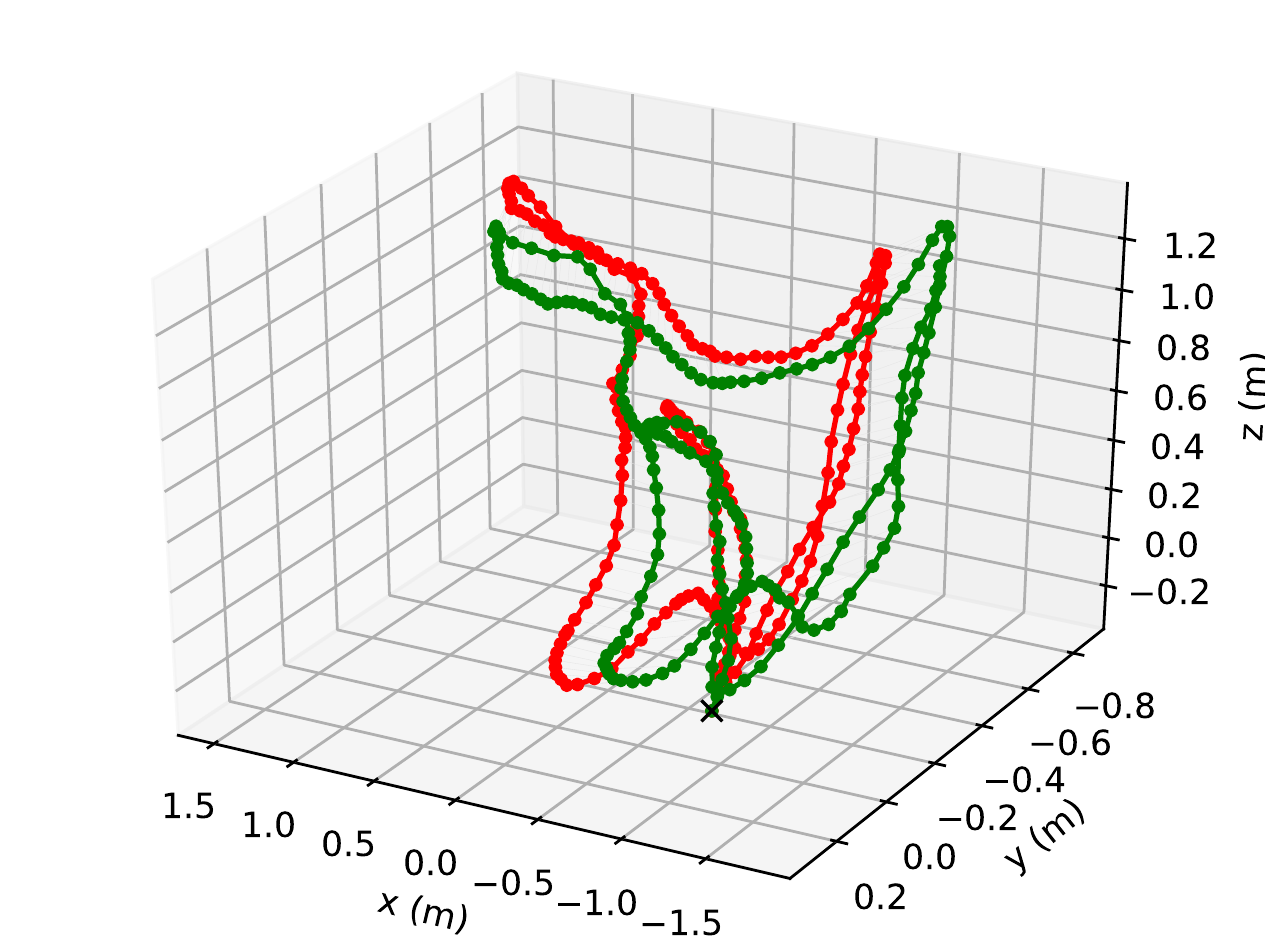}
        \includegraphics[width=\linewidth]{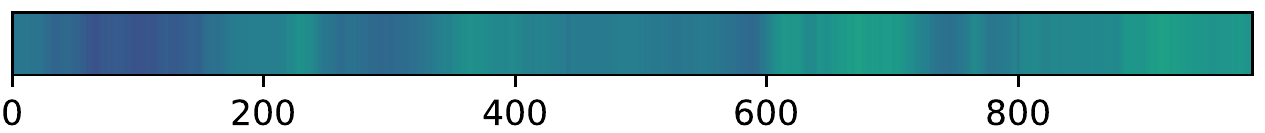}
    \end{subfigure}
    \hfill
    \begin{subfigure}{0.15\linewidth}
        \centering
        \includegraphics[width=\linewidth]{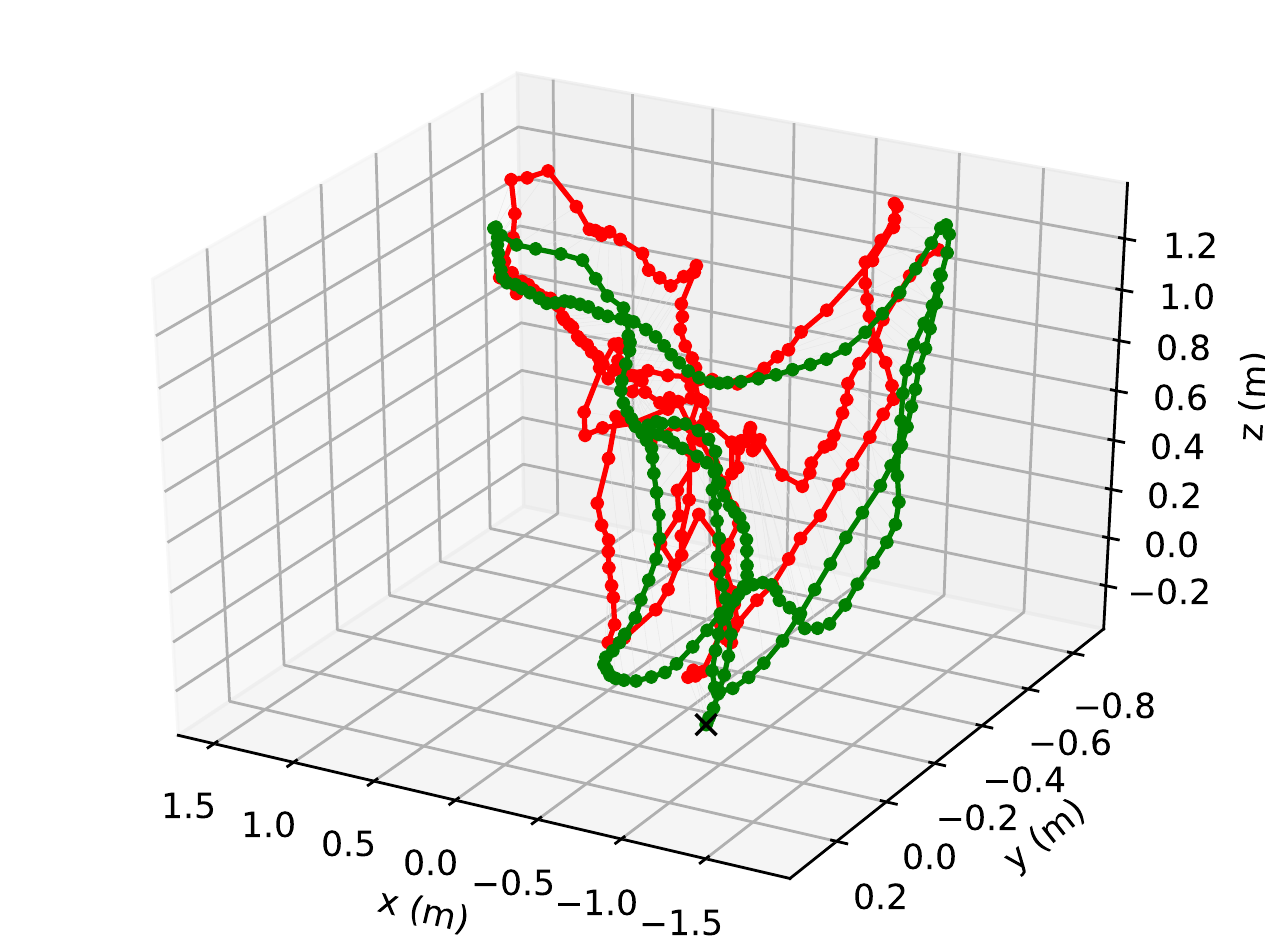}
        \includegraphics[width=\linewidth]{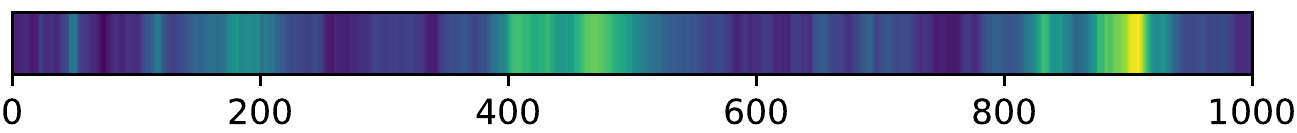}
    \end{subfigure}
    \hfill
    \begin{subfigure}{0.15\linewidth}
        \centering
        \includegraphics[width=\linewidth]{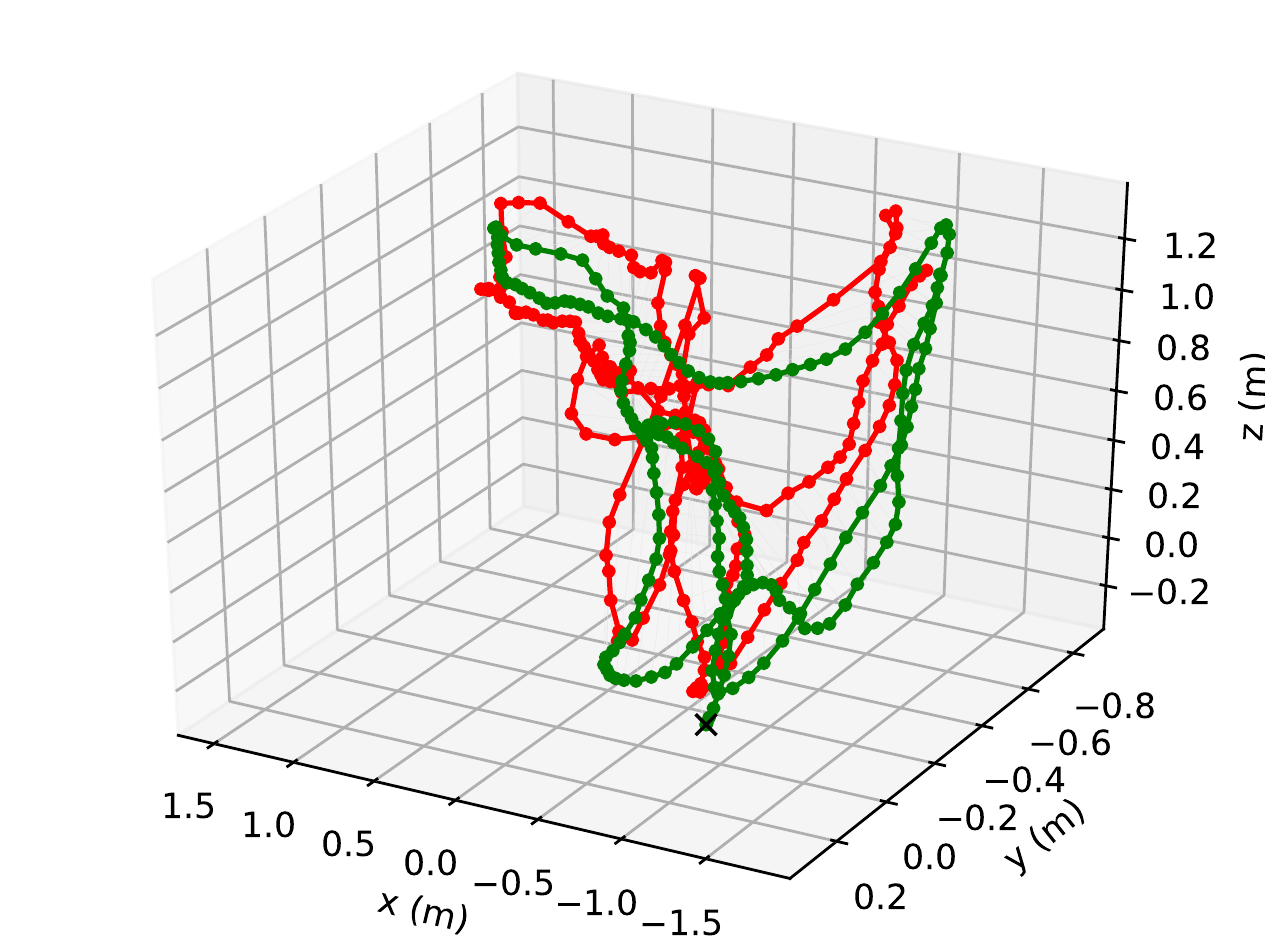}
        \includegraphics[width=\linewidth]{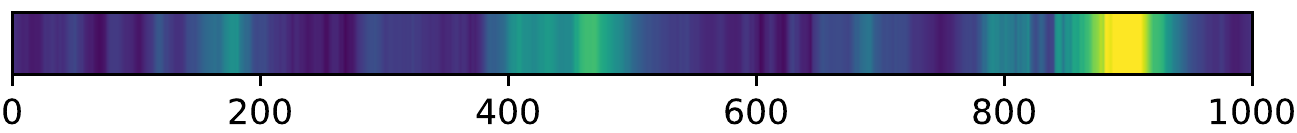}
    \end{subfigure}
    \hfill
    \begin{subfigure}{0.15\linewidth}
        \centering
        \includegraphics[width=\linewidth]{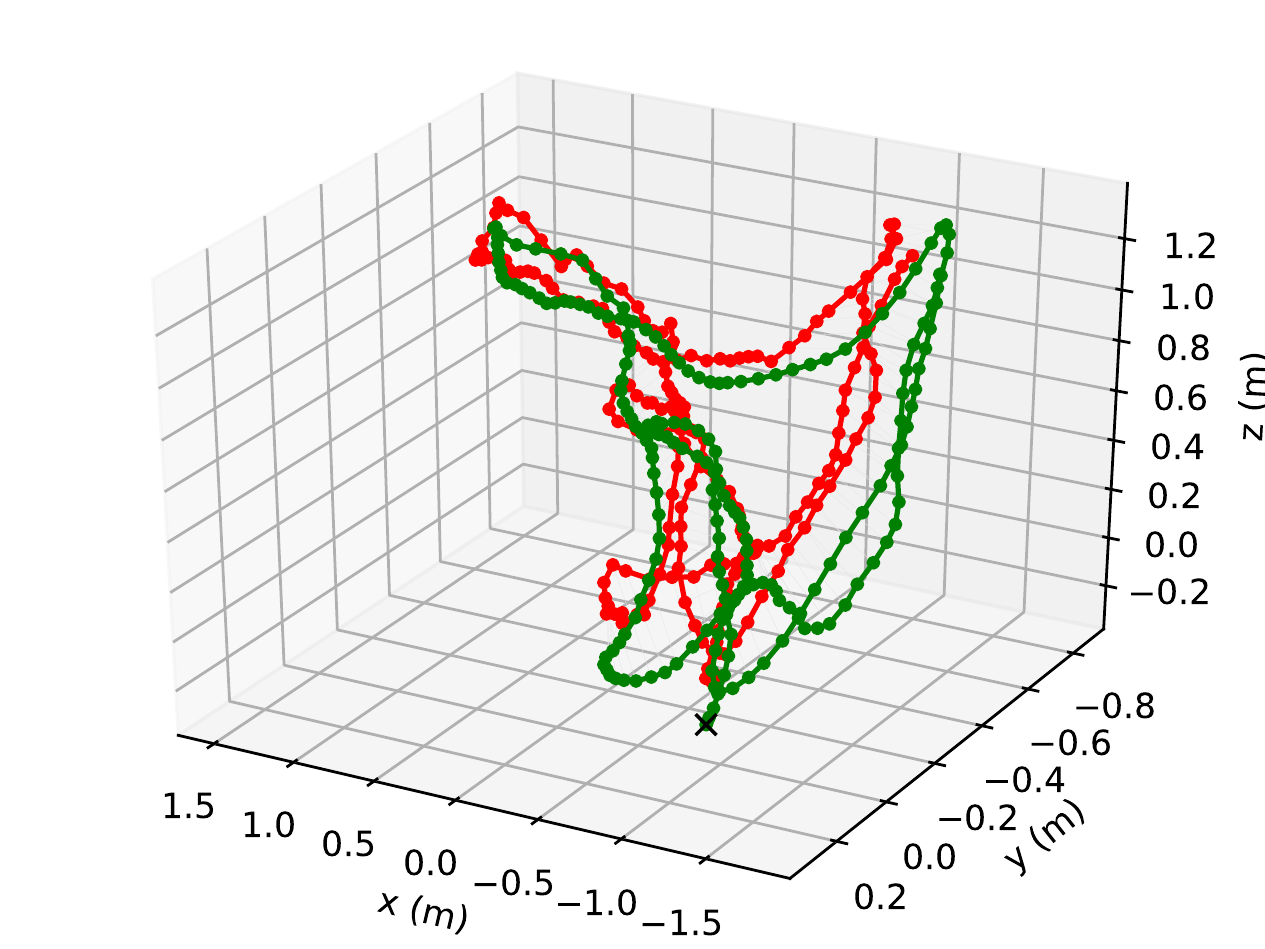}
        \includegraphics[width=\linewidth]{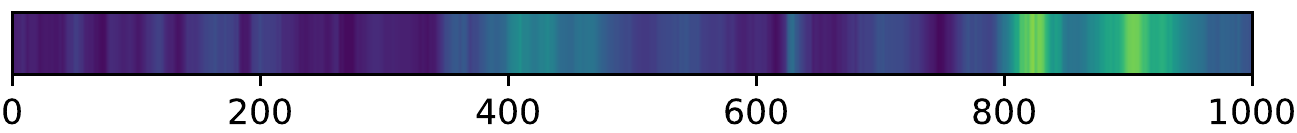}
    \end{subfigure}
    \hfill
    \begin{subfigure}{0.15\linewidth}
        \centering
        \includegraphics[width=\linewidth]{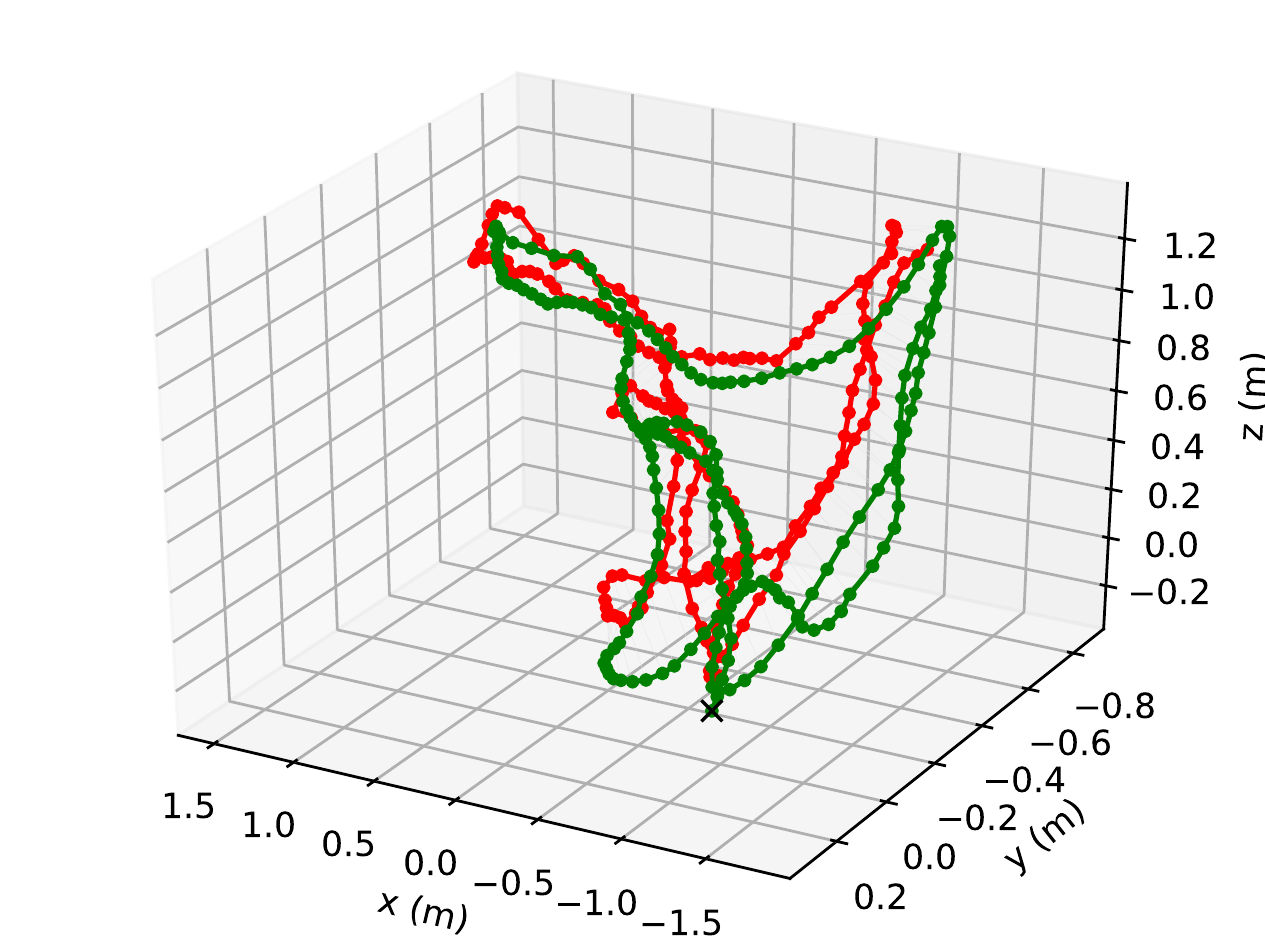}
        \includegraphics[width=\linewidth]{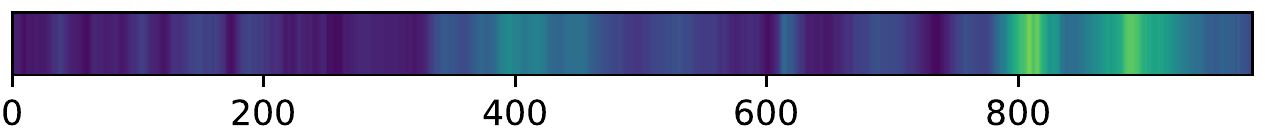}
    \end{subfigure}

    \begin{subfigure}{0.15\linewidth}
        \centering
        \includegraphics[width=\linewidth]{figures/7scenes/redkitchen_dsoA_3_t.pdf}
        \includegraphics[width=\linewidth]{figures/7scenes/redkitchen_dsoA_3_q.pdf}
    \end{subfigure}
    \hfill
    \begin{subfigure}{0.15\linewidth}
        \centering
        \includegraphics[width=\linewidth]{figures/7scenes/redkitchen_posenet_3_t.pdf}
        \includegraphics[width=\linewidth]{figures/7scenes/redkitchen_posenet_3_q.pdf}
    \end{subfigure}
    \hfill
    \begin{subfigure}{0.15\linewidth}
        \centering
        \includegraphics[width=\linewidth]{figures/7scenes/redkitchen_vidvo_3_t.pdf}
        \includegraphics[width=\linewidth]{figures/7scenes/redkitchen_vidvo_3_q.pdf}
    \end{subfigure}
    \hfill
    \begin{subfigure}{0.15\linewidth}
        \centering
        \includegraphics[width=\linewidth]{figures/7scenes/redkitchen_vidvo_online_3_t.pdf}
        \includegraphics[width=\linewidth]{figures/7scenes/redkitchen_vidvo_online_3_q.pdf}
    \end{subfigure}
    \hfill
    \begin{subfigure}{0.15\linewidth}
        \centering
        \includegraphics[width=\linewidth]{figures/7scenes/redkitchen_vidvo_online_pgo_dso_3_t.pdf}
        \includegraphics[width=\linewidth]{figures/7scenes/redkitchen_vidvo_online_pgo_dso_3_q.pdf}
    \end{subfigure}

    \begin{subfigure}{0.15\linewidth}
        \centering
        \includegraphics[width=\linewidth]{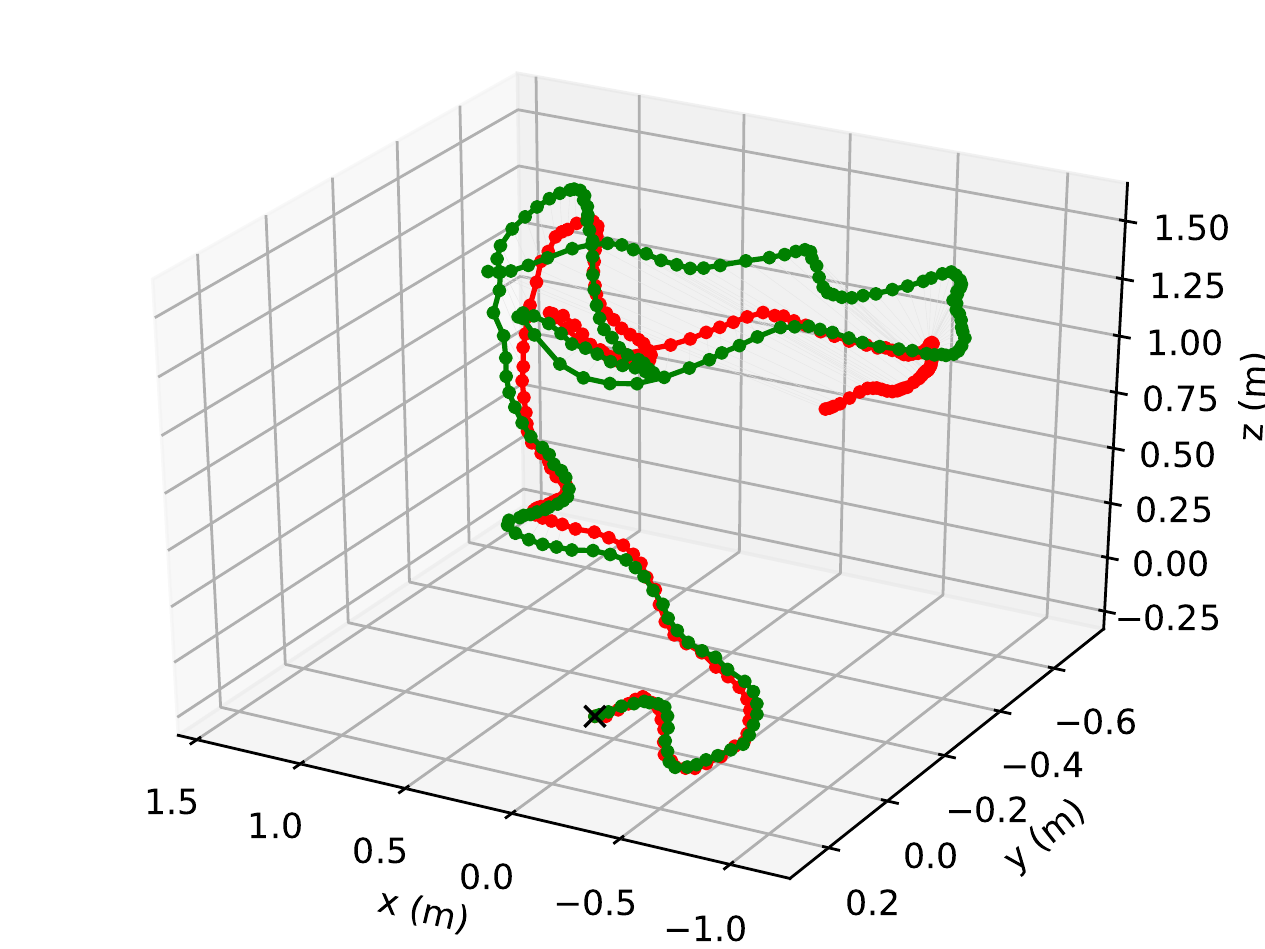}
        \includegraphics[width=\linewidth]{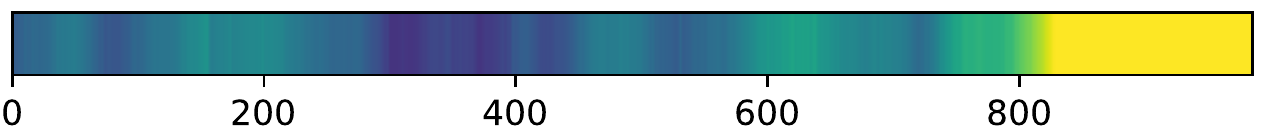}
    \end{subfigure}
    \hfill
    \begin{subfigure}{0.15\linewidth}
        \centering
        \includegraphics[width=\linewidth]{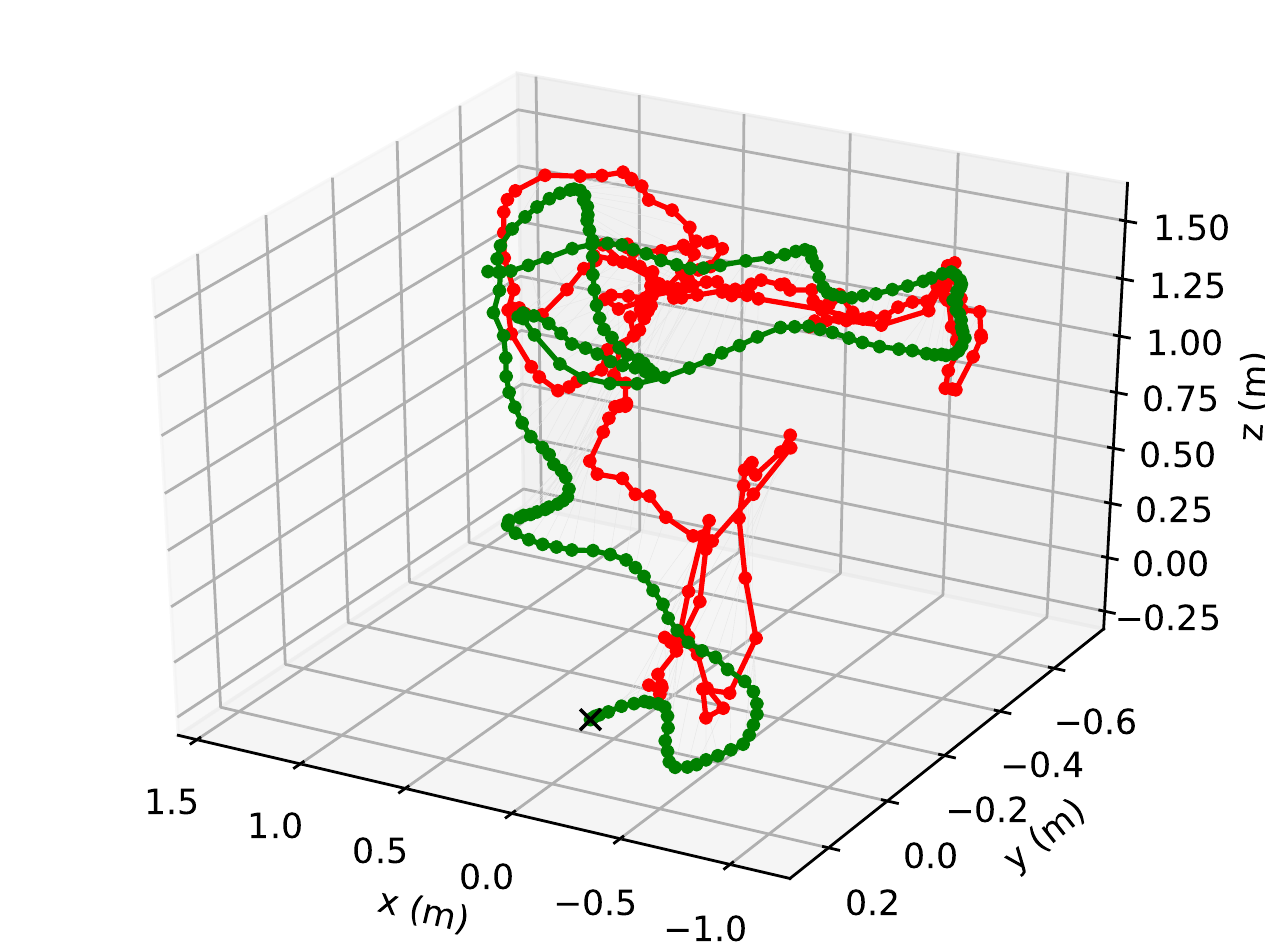}
        \includegraphics[width=\linewidth]{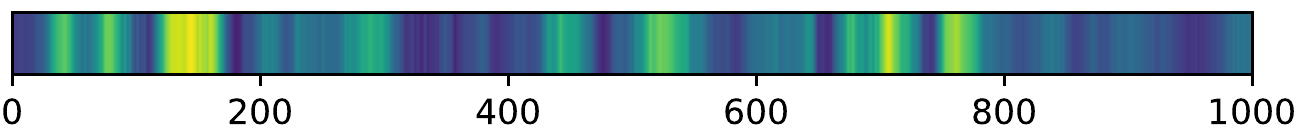}
    \end{subfigure}
    \hfill
    \begin{subfigure}{0.15\linewidth}
        \centering
        \includegraphics[width=\linewidth]{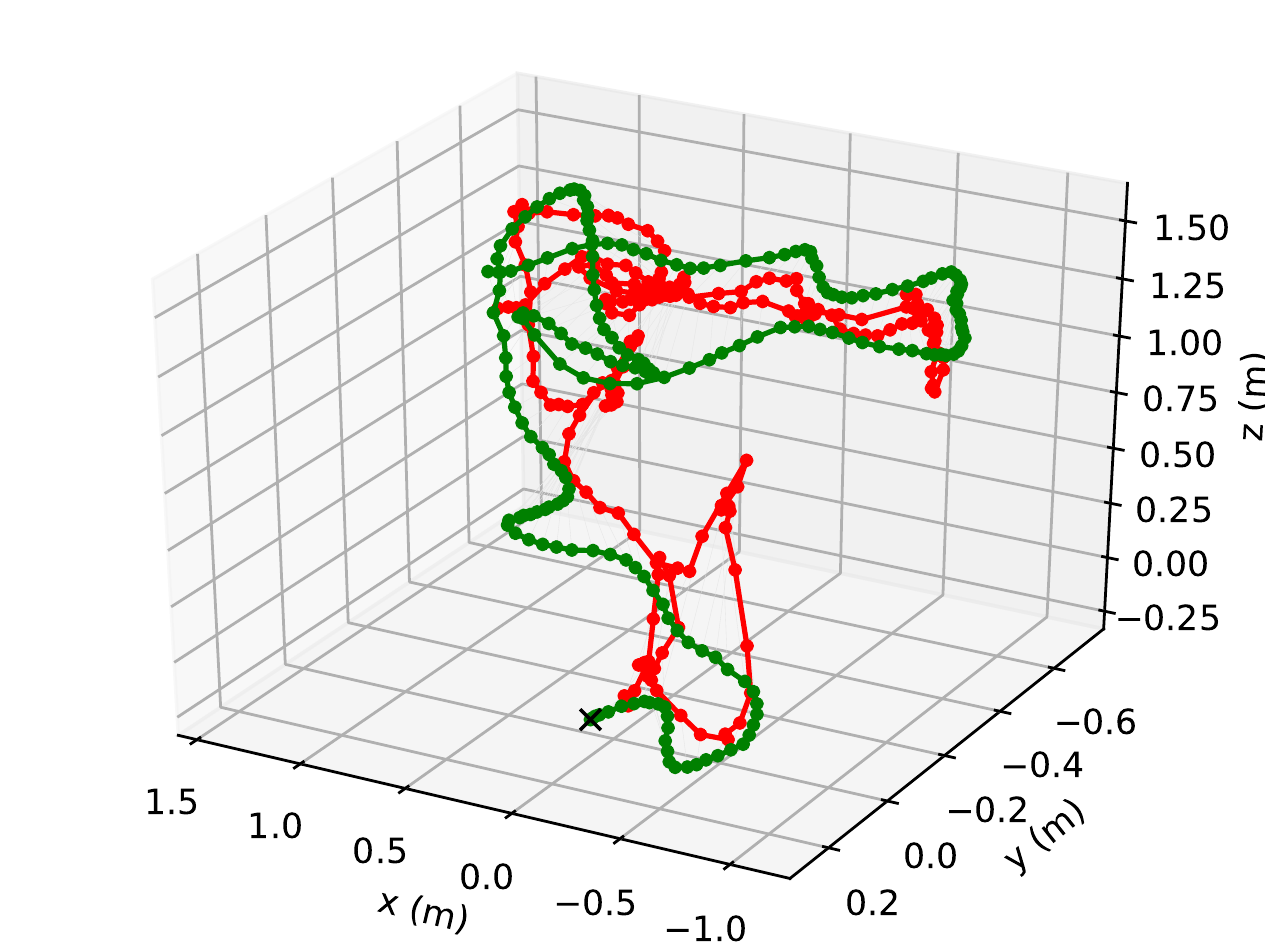}
        \includegraphics[width=\linewidth]{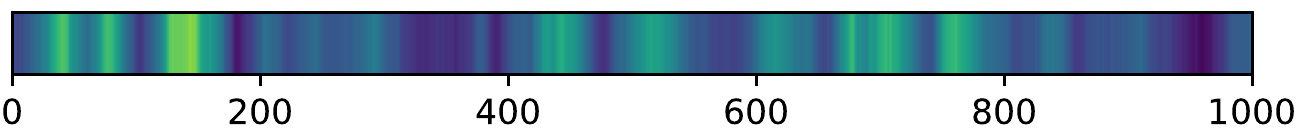}
    \end{subfigure}
    \hfill
    \begin{subfigure}{0.15\linewidth}
        \centering
        \includegraphics[width=\linewidth]{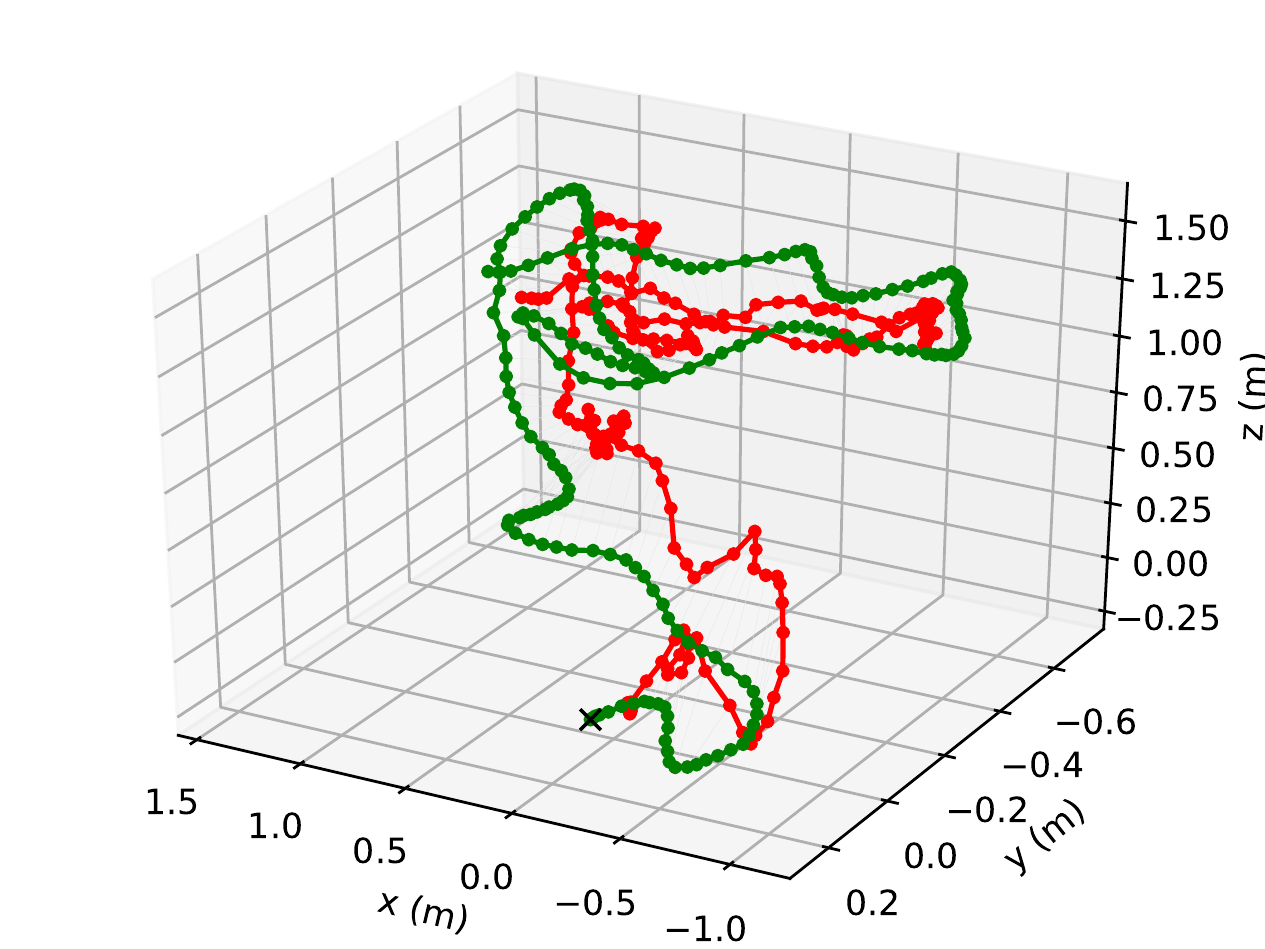}
        \includegraphics[width=\linewidth]{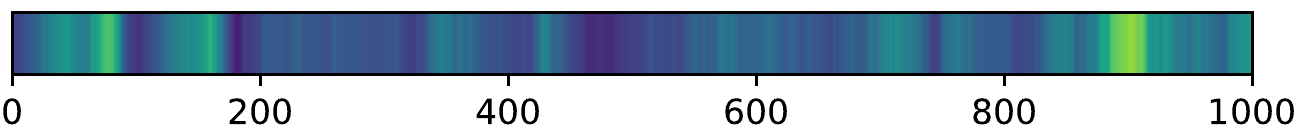}
    \end{subfigure}
    \hfill
    \begin{subfigure}{0.15\linewidth}
        \centering
        \includegraphics[width=\linewidth]{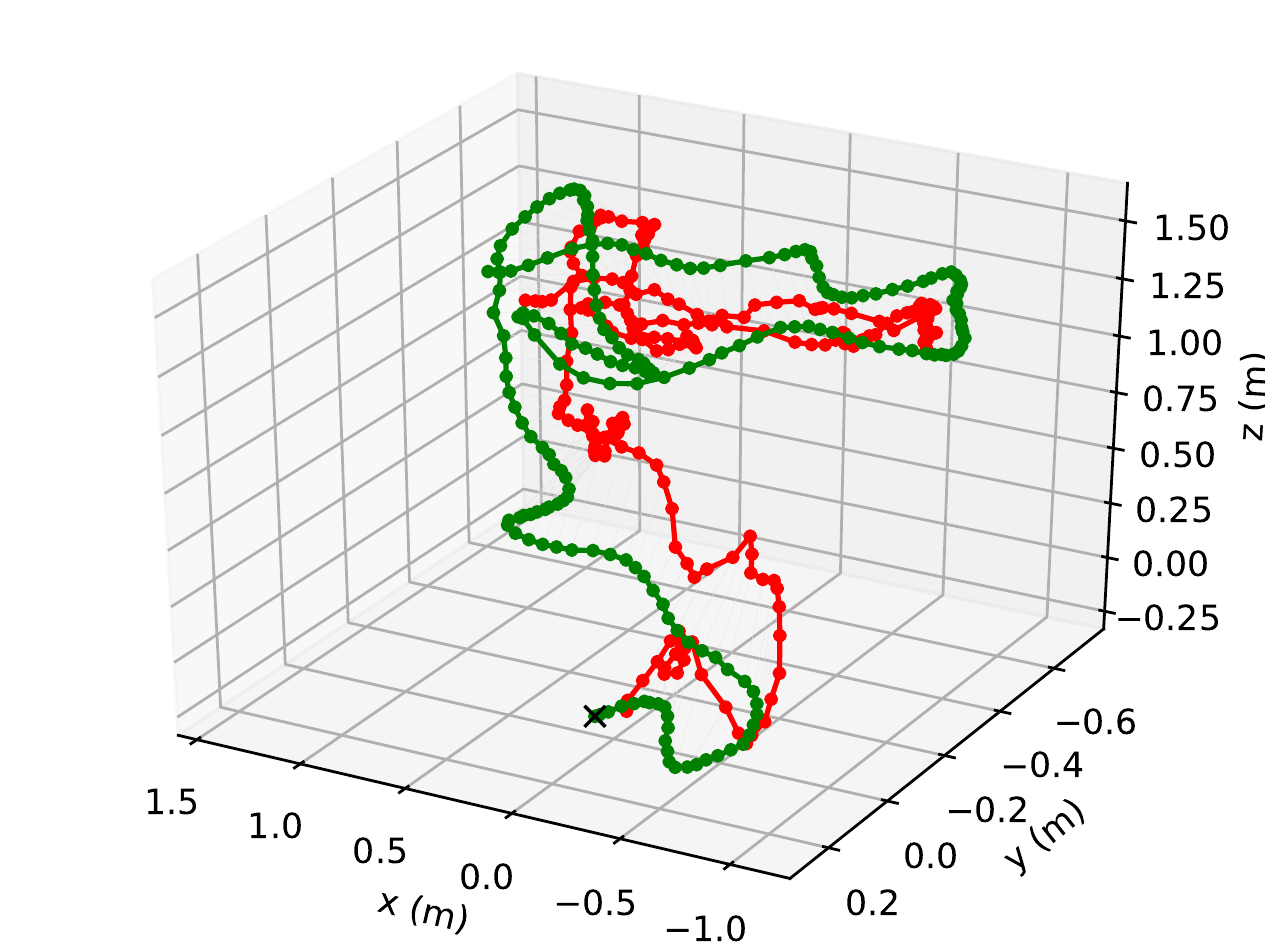}
        \includegraphics[width=\linewidth]{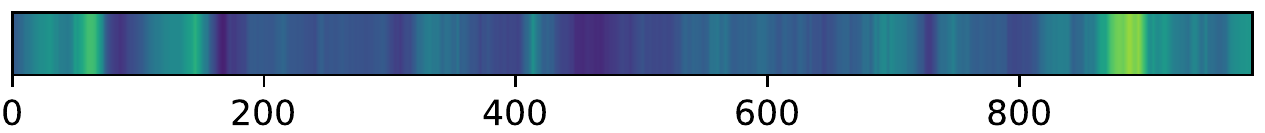}
    \end{subfigure}

    \begin{subfigure}{0.15\linewidth}
        \centering
        \includegraphics[width=\linewidth]{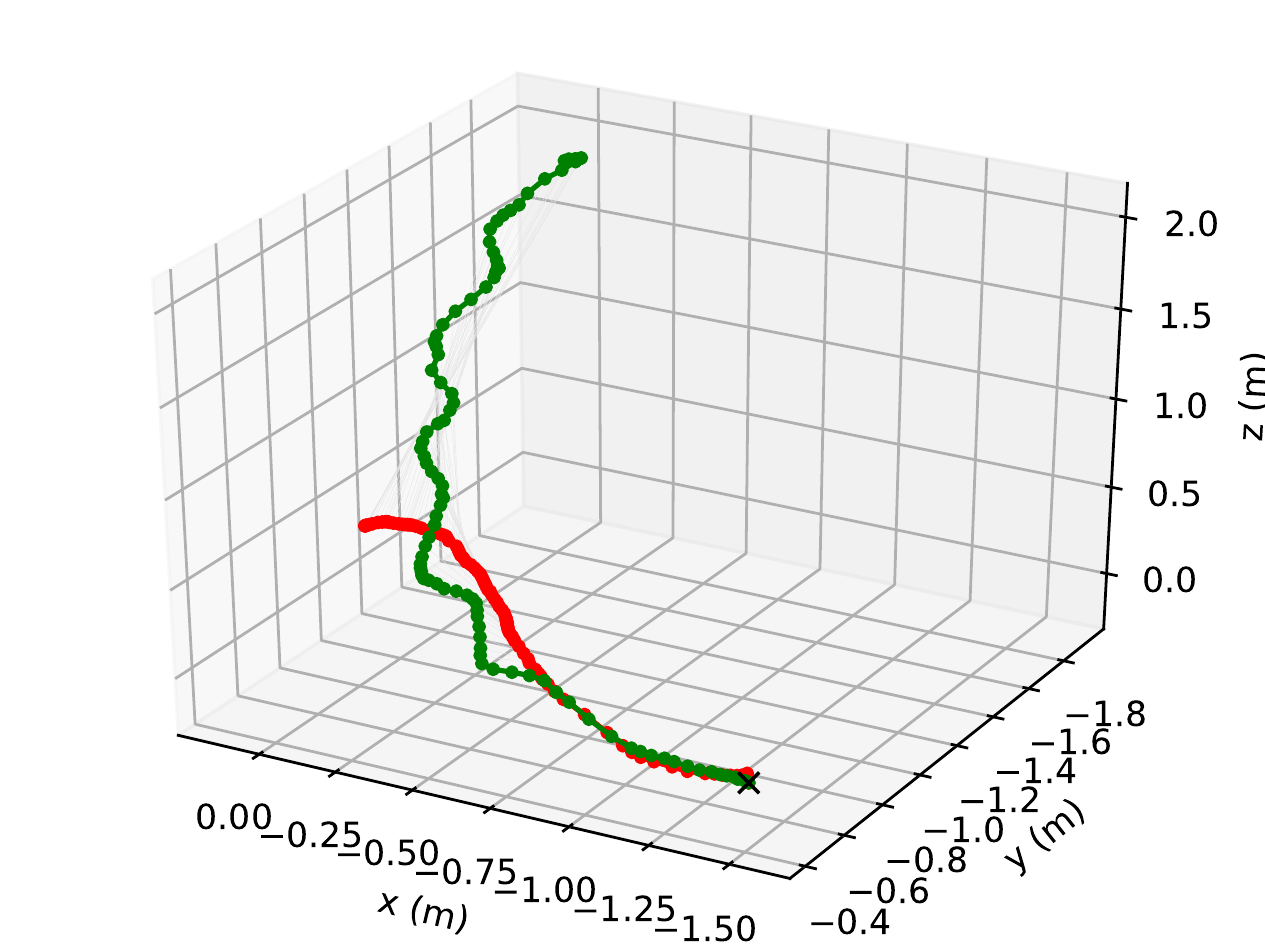}
        \includegraphics[width=\linewidth]{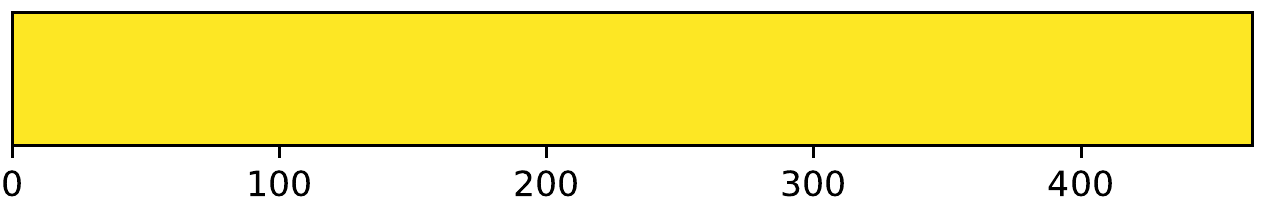}
    \end{subfigure}
    \hfill
    \begin{subfigure}{0.15\linewidth}
        \centering
        \includegraphics[width=\linewidth]{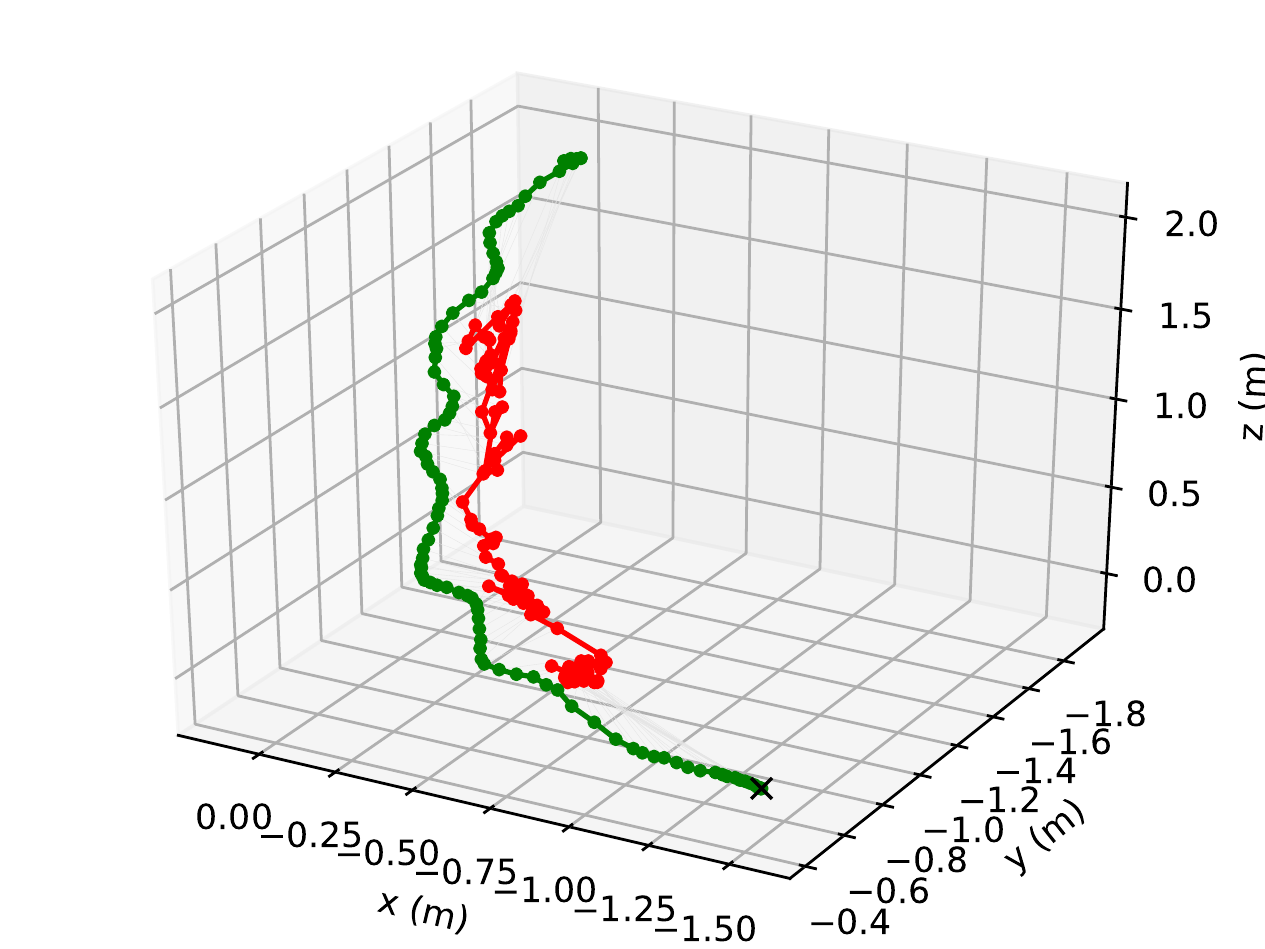}
        \includegraphics[width=\linewidth]{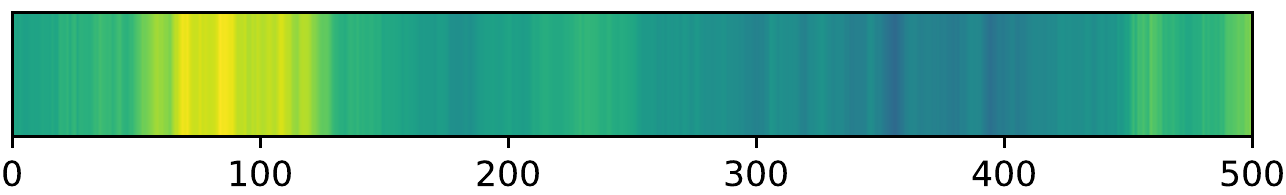}
    \end{subfigure}
    \hfill
    \begin{subfigure}{0.15\linewidth}
        \centering
        \includegraphics[width=\linewidth]{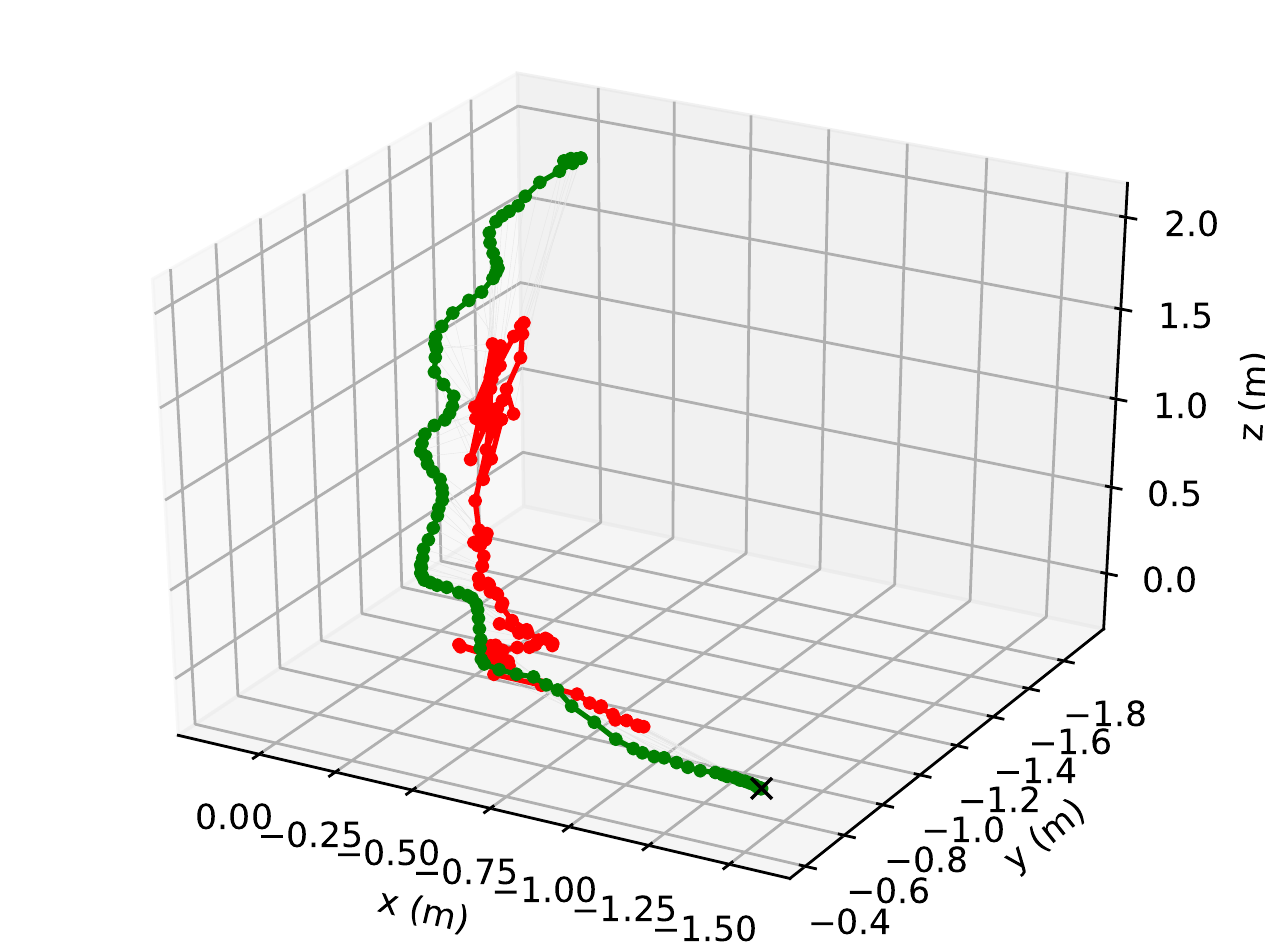}
        \includegraphics[width=\linewidth]{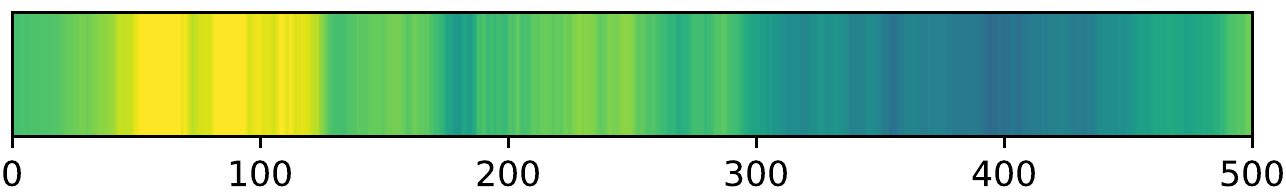}
    \end{subfigure}
    \hfill
    \begin{subfigure}{0.15\linewidth}
        \centering
        \includegraphics[width=\linewidth]{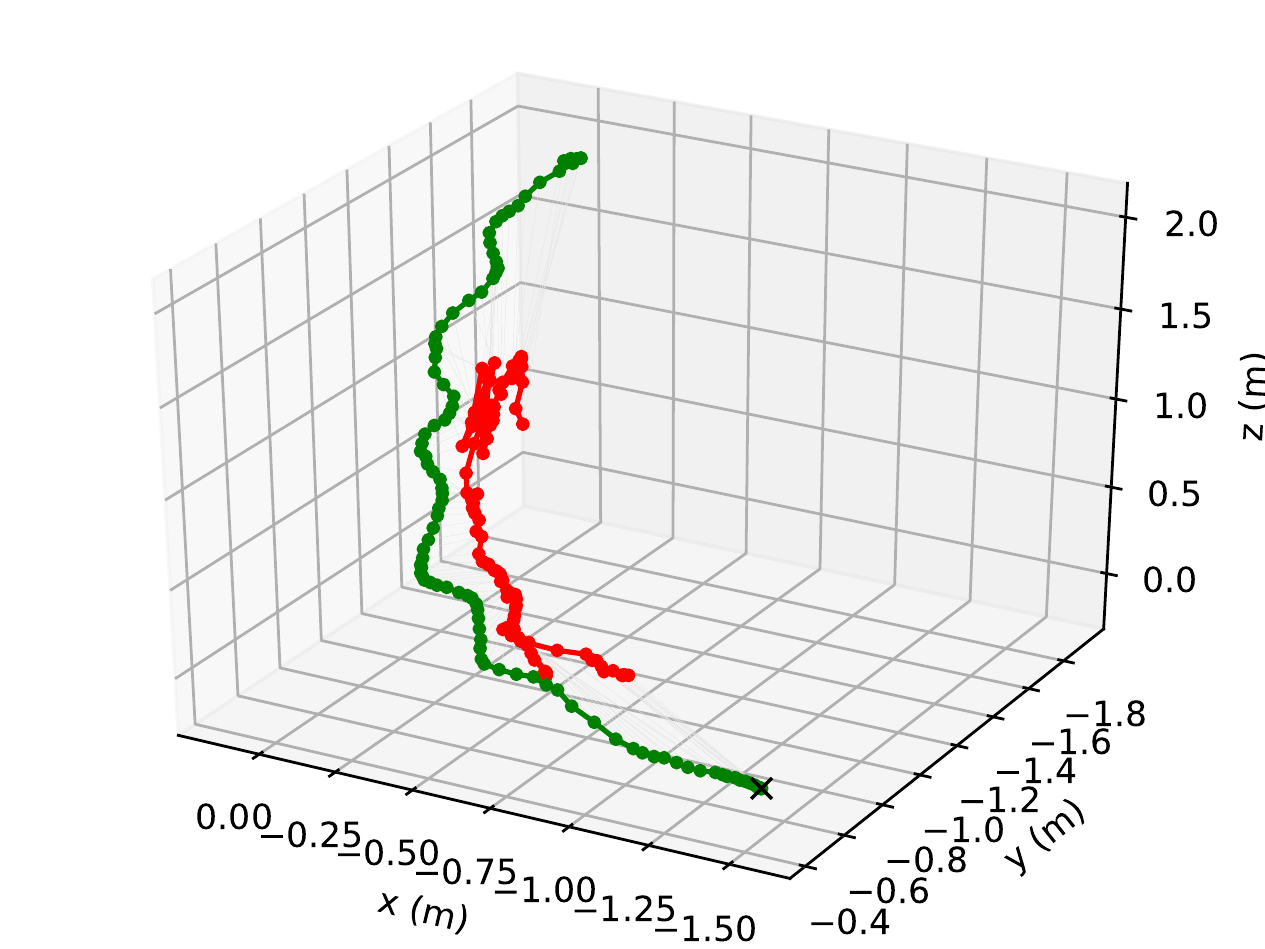}
        \includegraphics[width=\linewidth]{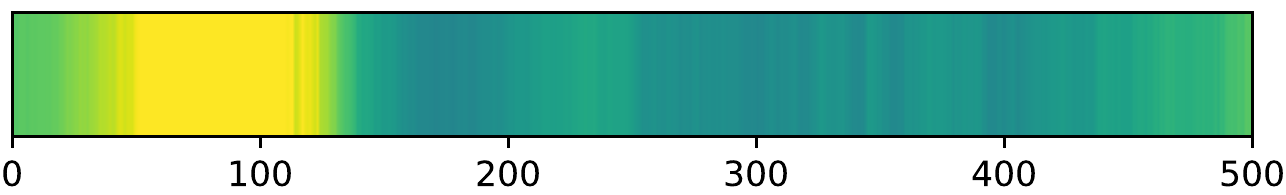}
    \end{subfigure}
    \hfill
    \begin{subfigure}{0.15\linewidth}
        \centering
        \includegraphics[width=\linewidth]{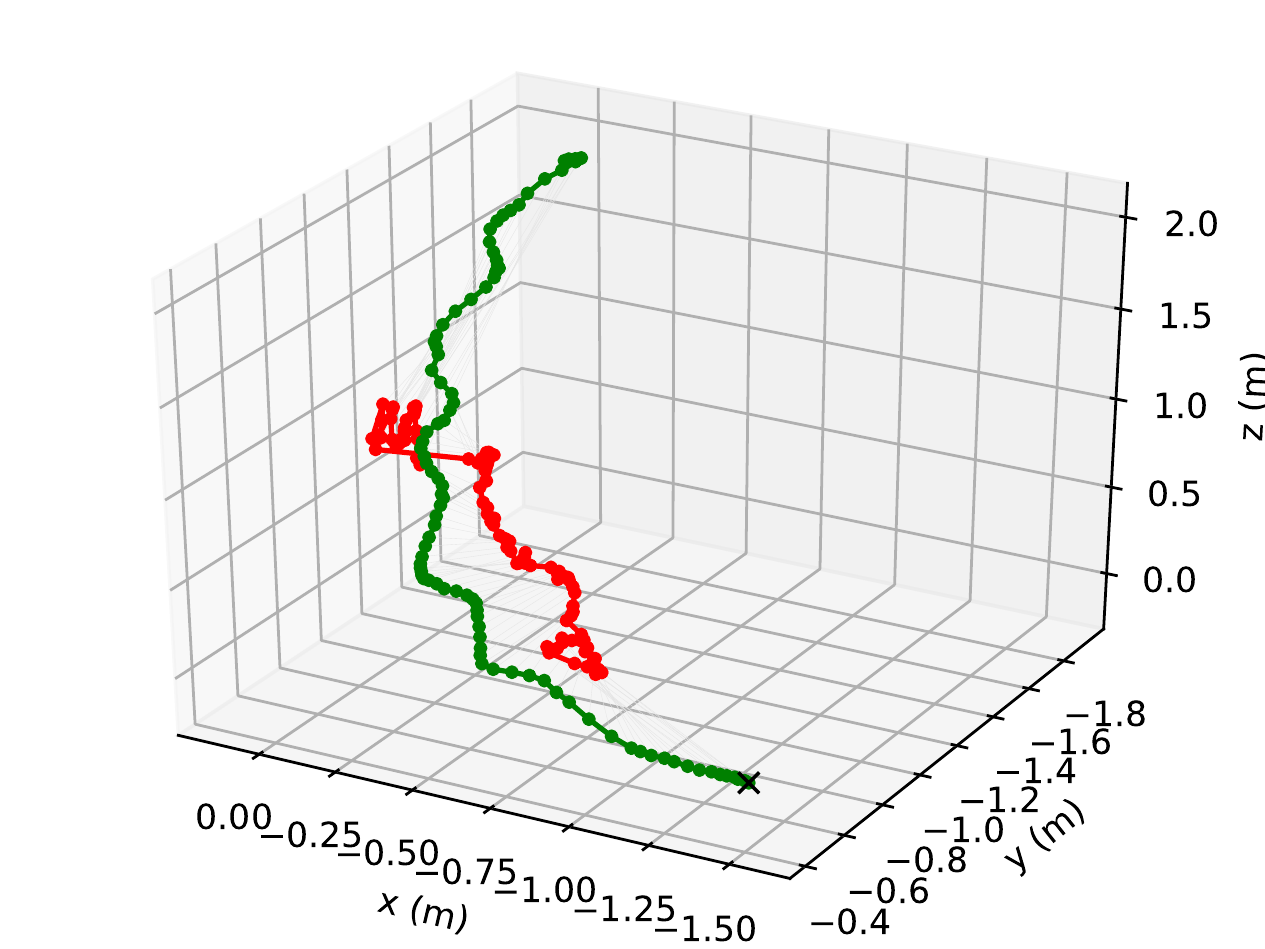}
        \includegraphics[width=\linewidth]{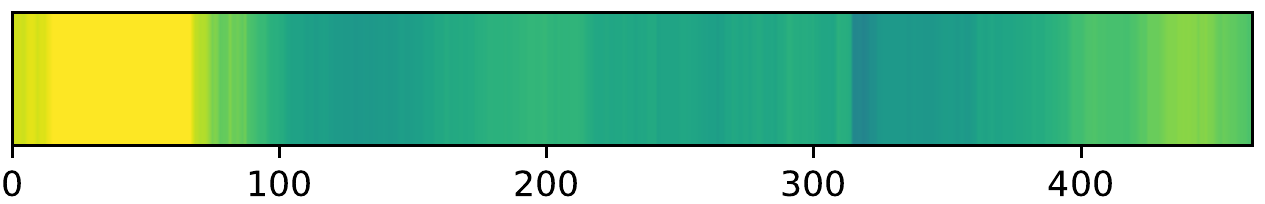}
    \end{subfigure}

    \begin{subfigure}{0.15\linewidth}
        \centering
        \includegraphics[width=\linewidth]{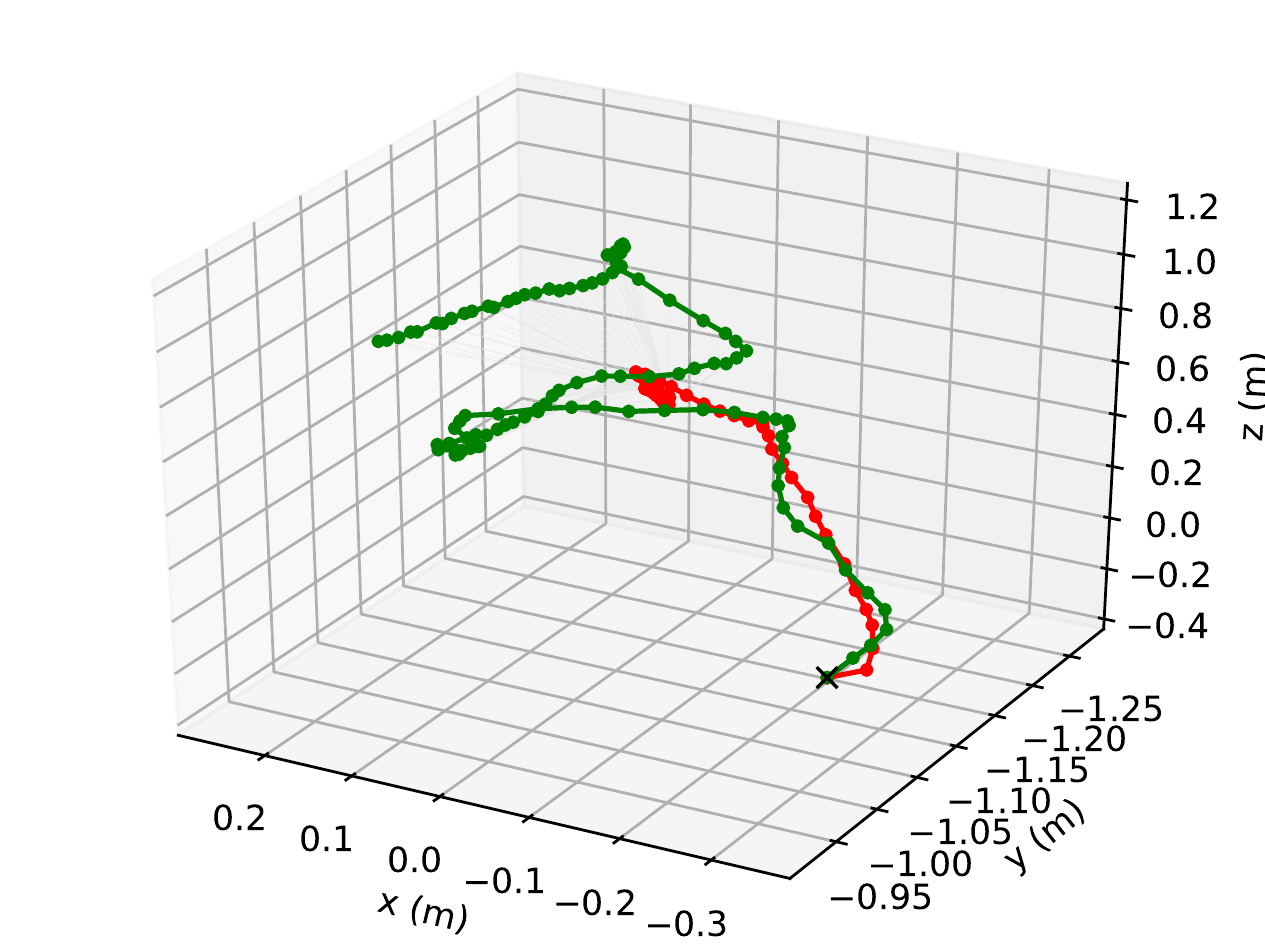}
        \includegraphics[width=\linewidth]{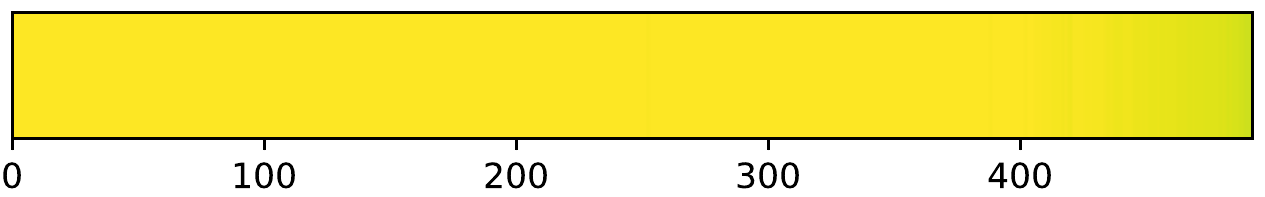}
        \caption{DSO~\cite{Engel2017DSO}}
    \end{subfigure}
    \hfill
    \begin{subfigure}{0.15\linewidth}
        \centering
        \includegraphics[width=\linewidth]{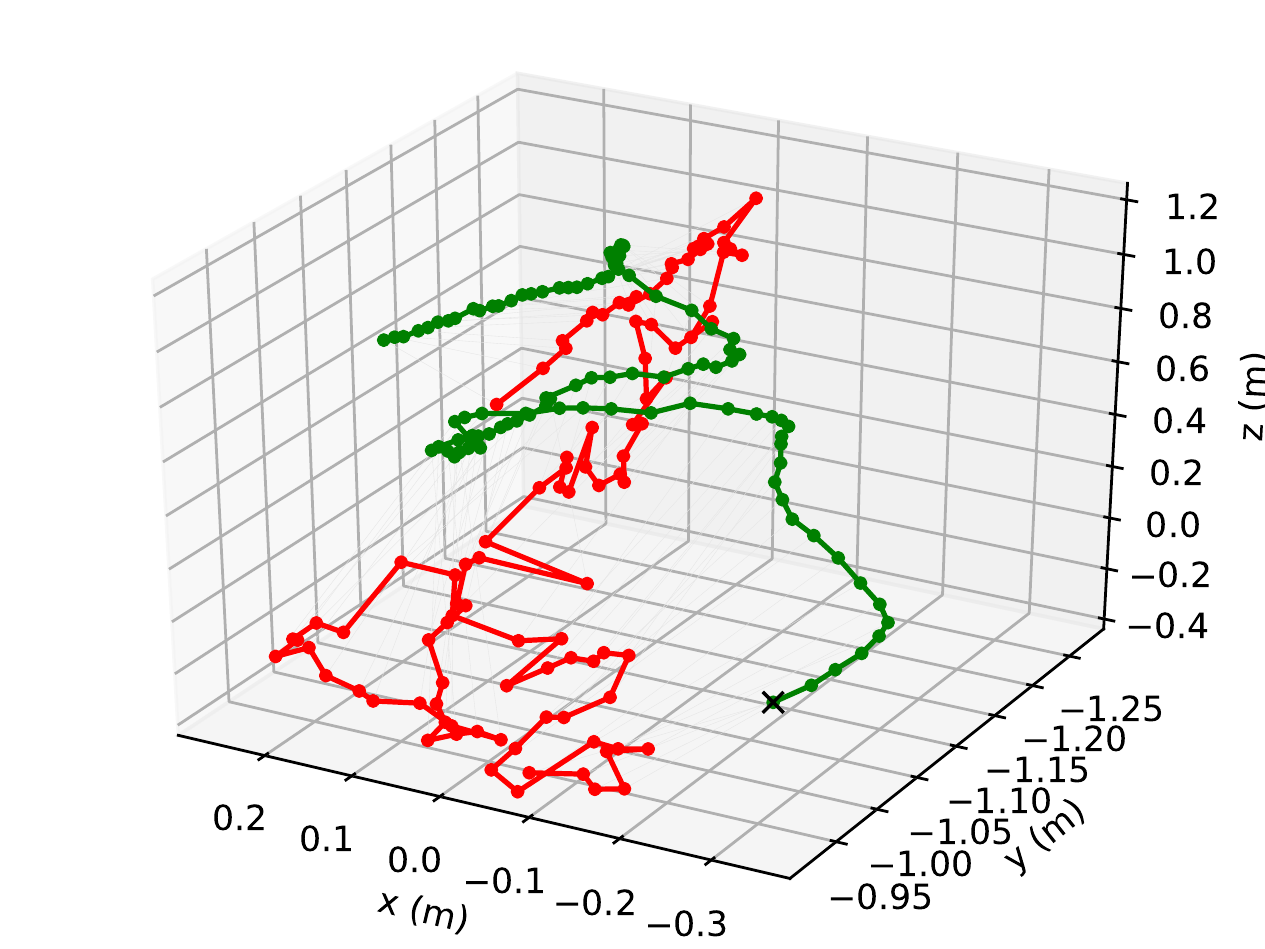}
        \includegraphics[width=\linewidth]{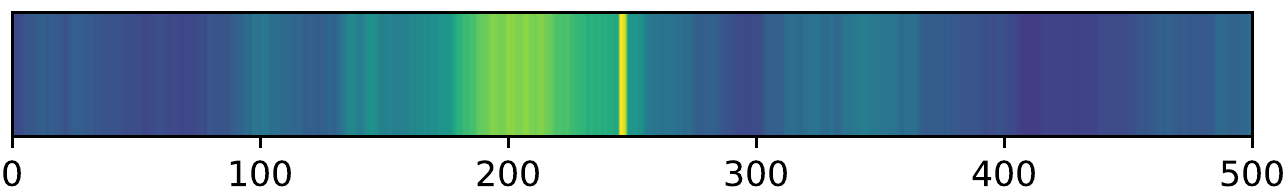}
        \caption{PoseNet~\cite{Kendall17cvpr, Kendall15iccv, Kendall16icra}}
    \end{subfigure}
    \hfill
    \begin{subfigure}{0.15\linewidth}
        \centering
        \includegraphics[width=\linewidth]{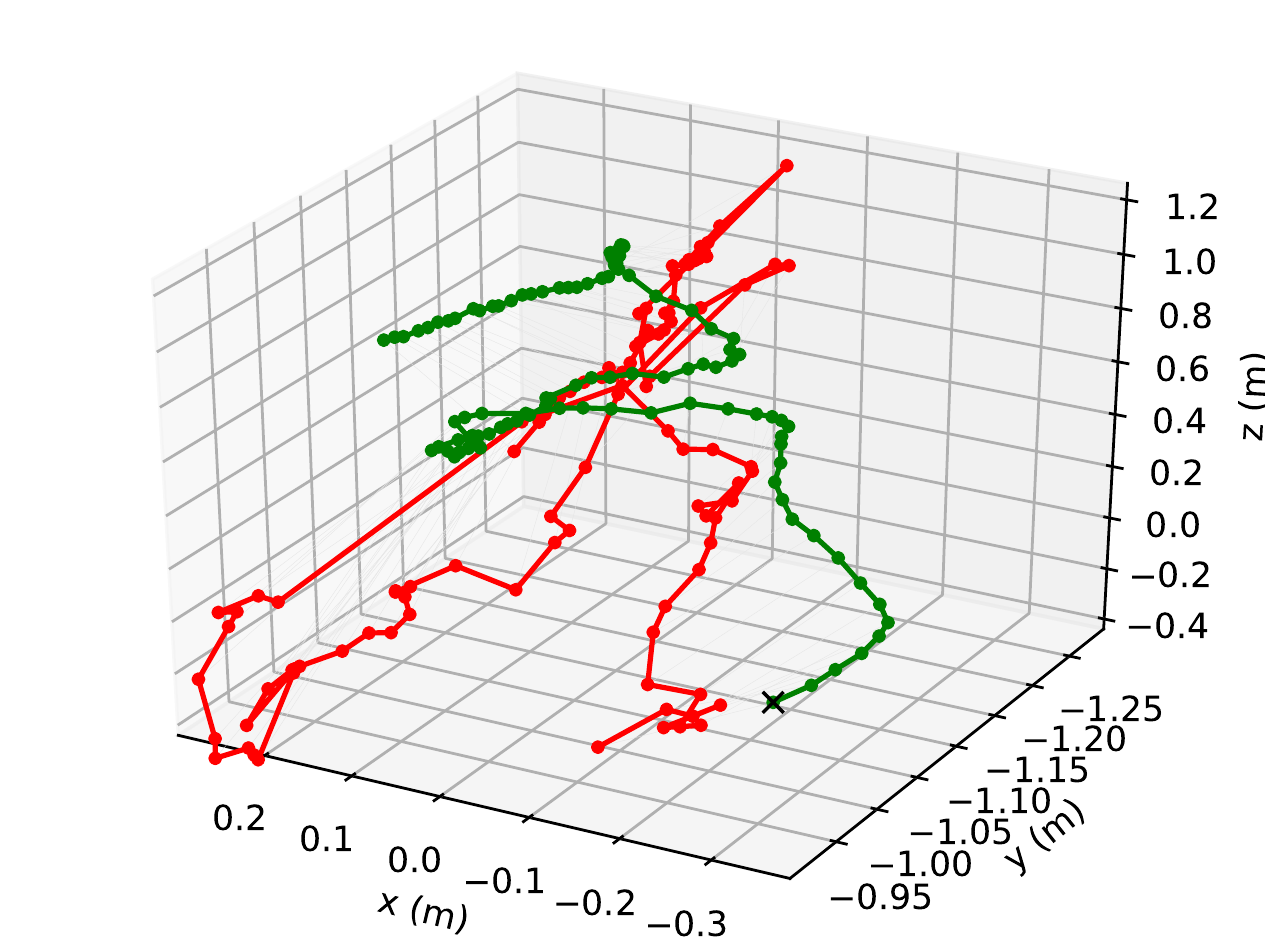}
        \includegraphics[width=\linewidth]{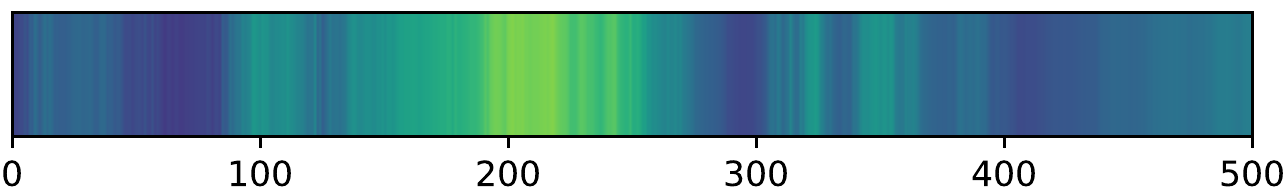}
        \caption{MapNet}
    \end{subfigure}
    \hfill
    \begin{subfigure}{0.15\linewidth}
        \centering
        \includegraphics[width=\linewidth]{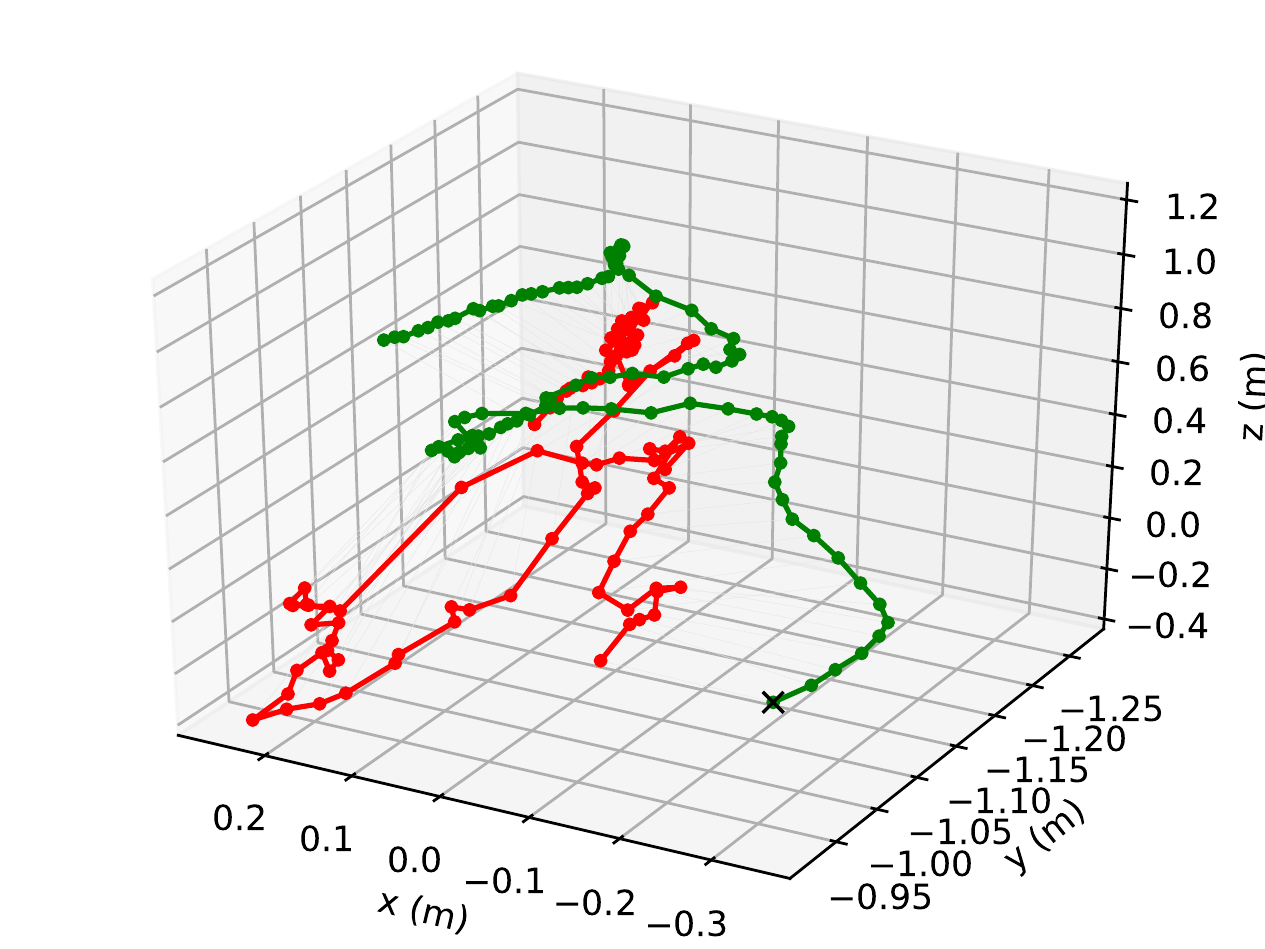}
        \includegraphics[width=\linewidth]{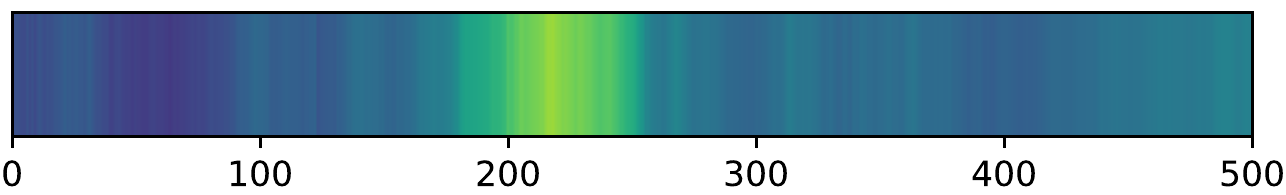}
        \caption{MapNet+}
    \end{subfigure}
    \hfill
    \begin{subfigure}{0.15\linewidth}
        \centering
        \includegraphics[width=\linewidth]{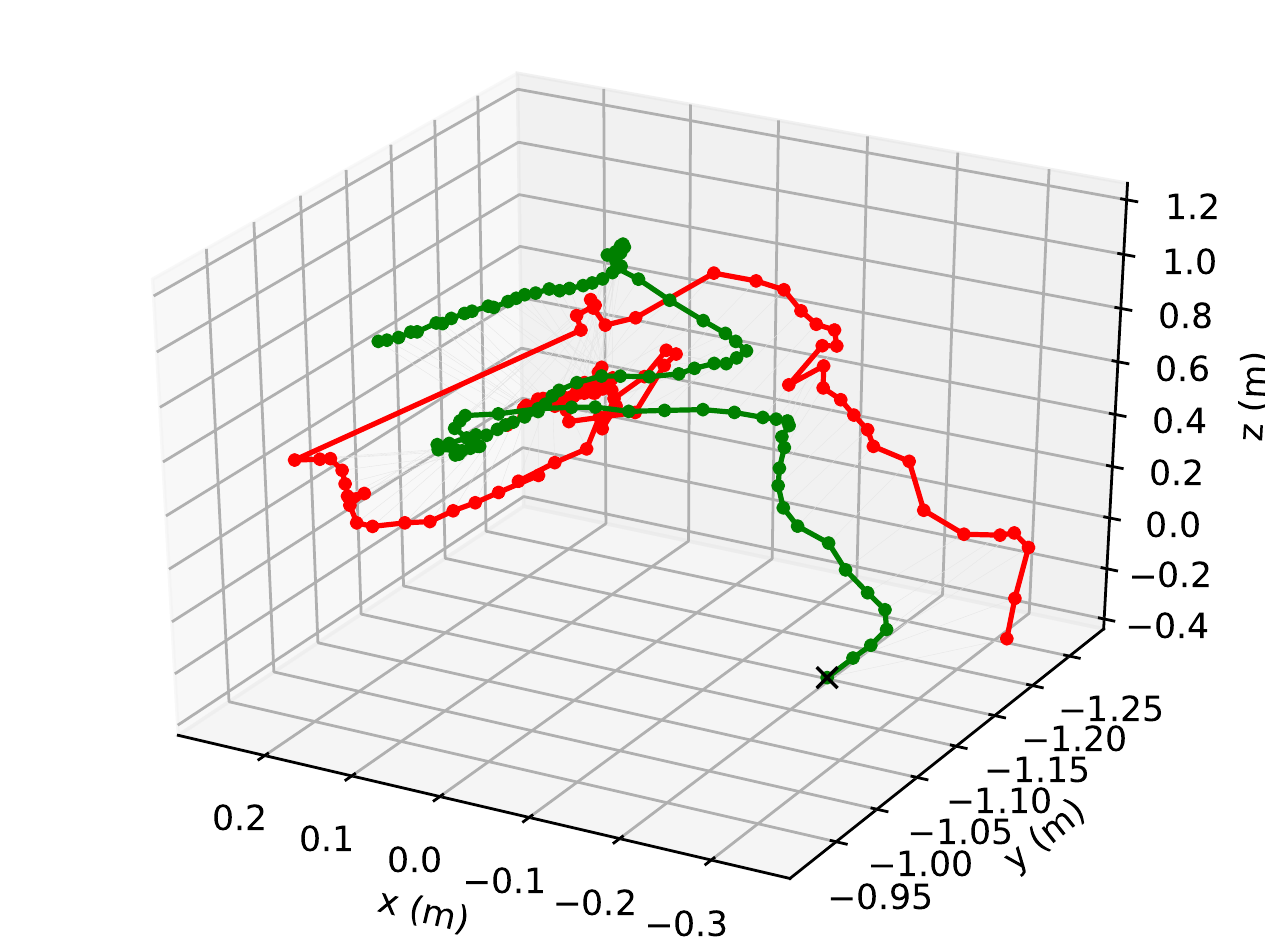}
        \includegraphics[width=\linewidth]{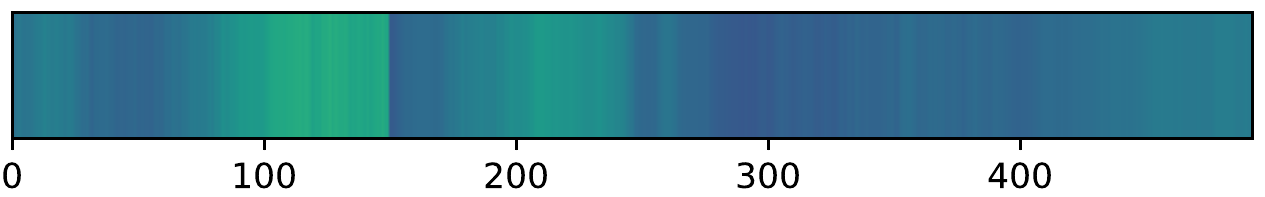}
        \caption{MapNet+PGO}
    \end{subfigure}
    \vspace{-1em}
    \caption{\small \textbf{Results on the 7-Scenes dataset (continued).} 
     The 3d plots show the camera position (green for ground truth and red for predictions). The colorbars below 
    show the errors of the predicted camera orientation (blue for small error and yellow for large error) with frame
    number on the X axis. 
    From top to bottom are testing sequences: 
    Pumpkin-Seq-01, Pumpkin-Seq-07, Redkitchen-Seq-03, Redkitchen-Seq-04, Redkitchen-Seq-06, Redkitchen-Seq-12, Redkitchen-Seq-14,
    Stairs-Seq-01, and Stairs-Seq-04.}
    \label{fig:res_7scenes_2}
\end{figure*}

Figure~\ref{fig:res_7scenes} and Figure~\ref{fig:res_7scenes_2} show the results for all the 18 testing sequences
on the 7-Scenes dataset. Table~\ref{tab:res_7scenes_stats} lists a variety of statistics computed on all the 18 testing sequences,
where Avg Median (Scene) means the averaged values of the median error over each scene,
and Avg Median (Seq) means the averaged values of the median error over each sequence in the scene.
As shown, both these two figures and the table support the same conclusion as described in the main paper.

\section*{Experiments on the RobotCar Dataset}
\begin{figure*}
    \centering
    \includegraphics[width=\linewidth]{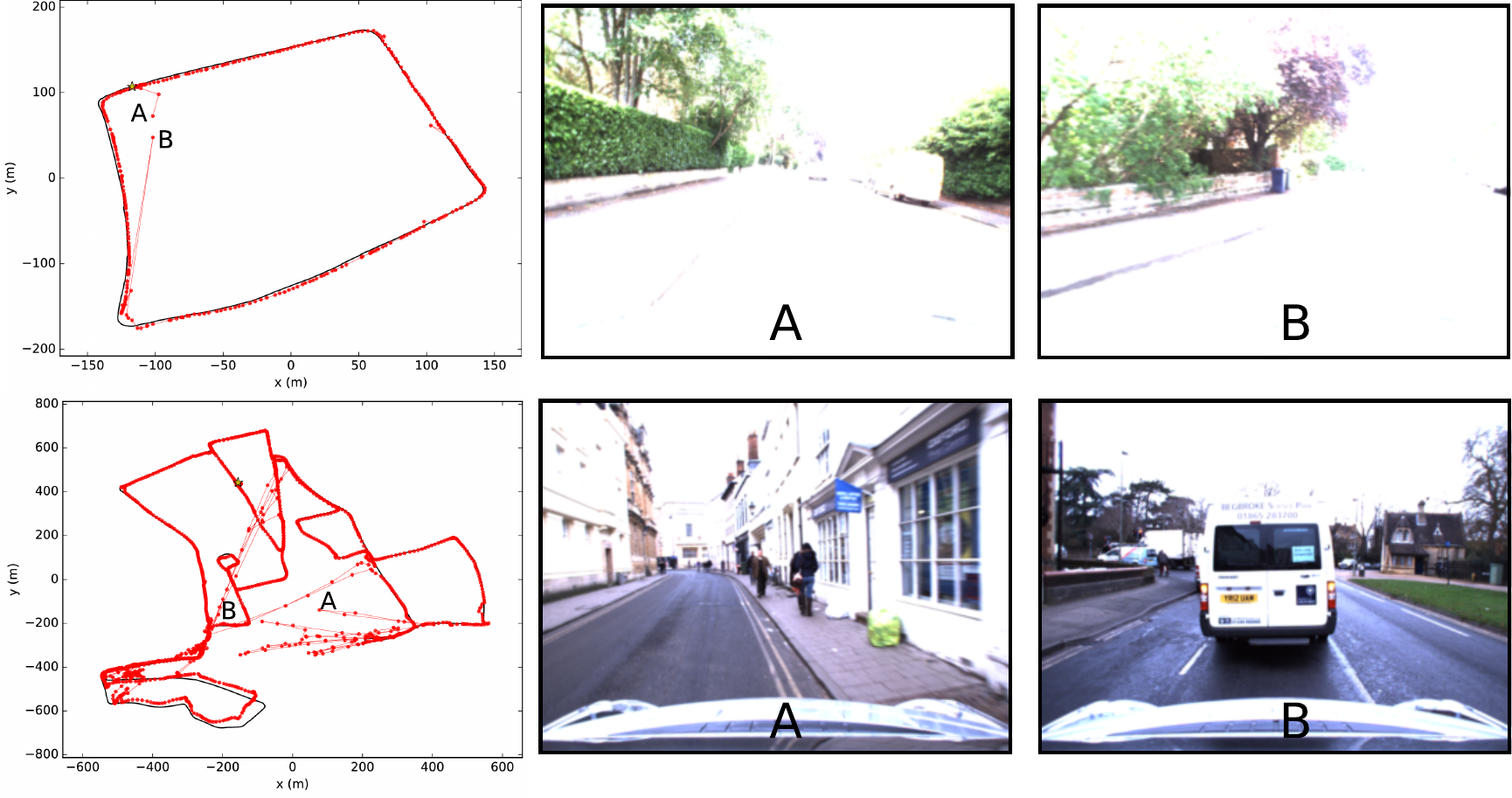}
    \caption{\textbf{Images corresponding to the spurious estimation of MapNet+PGO for the LOOP scene (top) and the FULL scene (bottom)}. These outliers
    usually corresponds to images with large over-exposed regions, or large regions on moving objects (\eg, truck), which often can be filtered out
    with simple temporal median filtering (see Figure~\ref{fig:res_robotcar_filter}).}
    \label{fig:res_outlier}
\end{figure*}

\begin{figure*}
    \small
    \captionsetup[subfigure]{labelformat=empty}
    \centering
    \begin{subfigure}{0.32\linewidth}
        \centering
        \includegraphics[width=\linewidth]{figures/robotcar/vidvo_online_pgo_stereo_t.pdf}
        \vspace{-1.5em}
        \caption{\small MapNet+PGO (6.73m, 2.23\degree)}
    \end{subfigure}
    \hfill
    \begin{subfigure}{0.32\linewidth}
        \centering
        \includegraphics[width=\linewidth]{figures/robotcar/vidvo_online_gps_t.pdf}
        \vspace{-1.5em}
        \caption{\small MapNet+(GPS) (6.78m, 2.72\degree)}
    \end{subfigure}
    \hfill
    \begin{subfigure}{0.32\linewidth}
        \centering
        \includegraphics[width=\linewidth]{figures/robotcar_full/vidvo_online_pgo_stereo_v2_t.pdf}
        \vspace{-1.5em}
        \caption{\small MapNet+PGO (29.5m, 7.8\degree)}
    \end{subfigure}
 
    \vspace{1em}
    \line(1,0){300}
    \vspace{1em}

    \begin{subfigure}{0.32\linewidth}
        \centering
        \includegraphics[width=\linewidth]{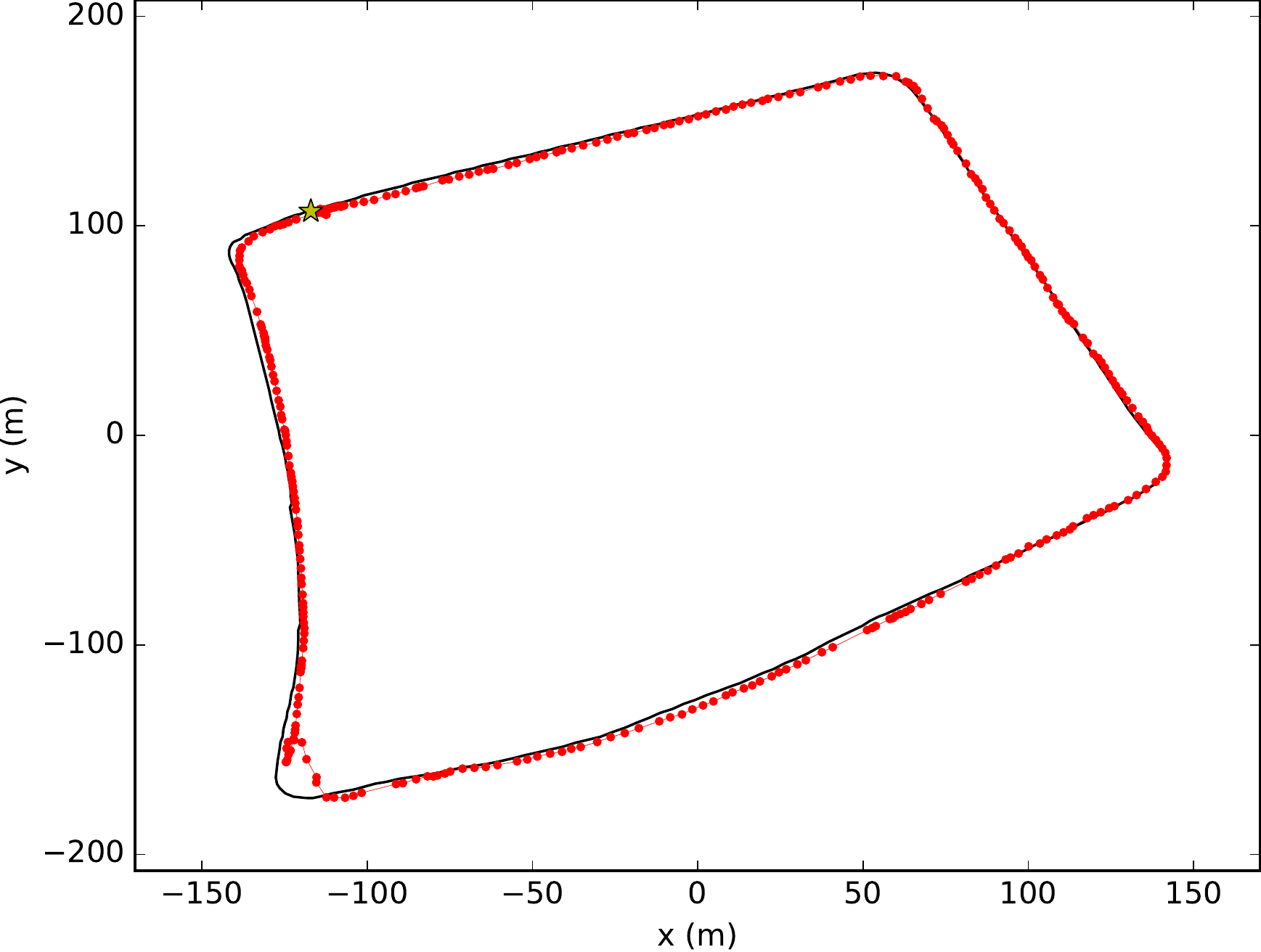}
        \vspace{-1.5em}
        \caption{\small MapNet+PGO (5.74m, 2.23\degree)}
    \end{subfigure}
    \hfill
    \begin{subfigure}{0.32\linewidth}
        \centering
        \includegraphics[width=\linewidth]{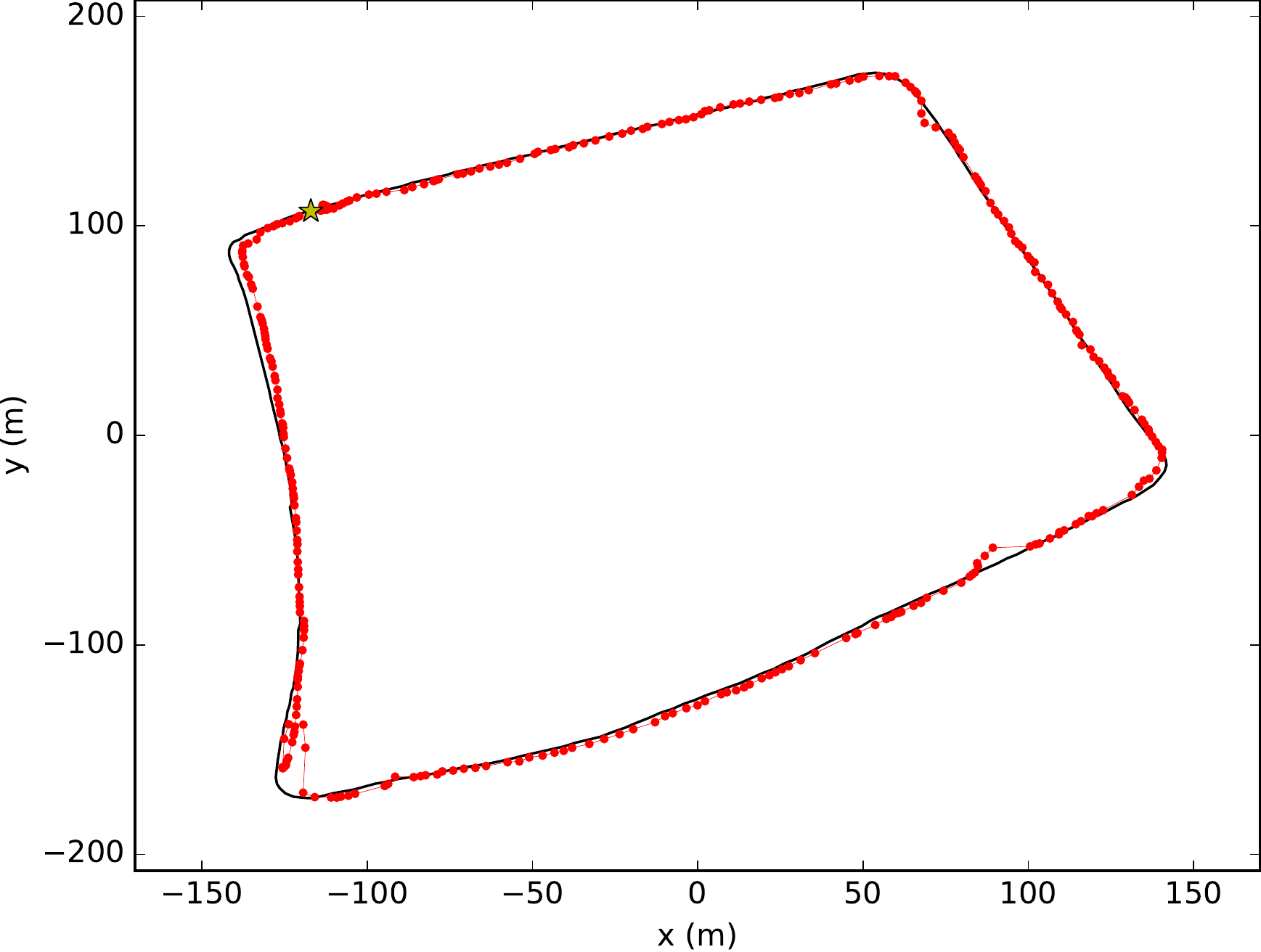}
        \vspace{-1.5em}
        \caption{\small MapNet+(GPS) (4.95m, 2.72\degree)}
    \end{subfigure}
    \begin{subfigure}{0.32\linewidth}
        \centering
        \includegraphics[width=\linewidth]{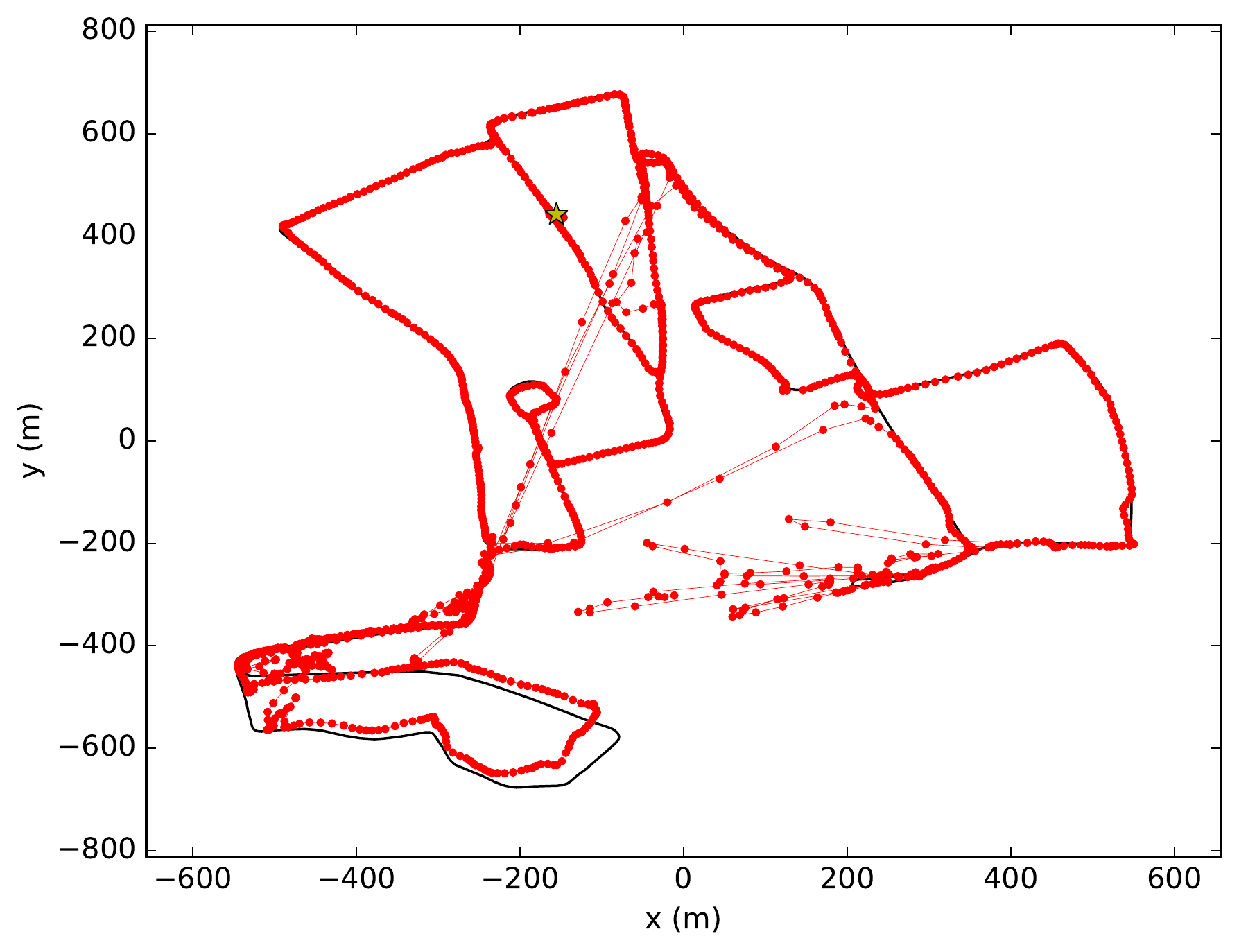}
        \vspace{-1.5em}
        \caption{\small MapNet+PGO (29.2m, 7.8\degree)}
    \end{subfigure}
 
    \vspace{-.5em} 
    \caption{\small \textbf{Camera localization results before (TOP) and after (BOTTOM) temporal median filtering.}
    The spurious estimations can be effectively removed with a simple median filtering (with the window size of 51 frames).}
    \label{fig:res_robotcar_filter}
\end{figure*}

Figure~\ref{fig:res_outlier} shows the images corresponding to the outliers in camera localization results of MapNet+PGO
for both the LOOP scene and the FULL scene. As shown, these outliers often correspond to images with large over-exposed regions,
or large regions covered with moving objects (\eg, truck). Some of these outliers can be filtered out simply with temporal median filtering,
as shown in Figure~\ref{fig:res_robotcar_filter}.


\begin{figure*}
    \centering
    \begin{subfigure}{0.45\linewidth}
        \centering
        \includegraphics[width=\linewidth]{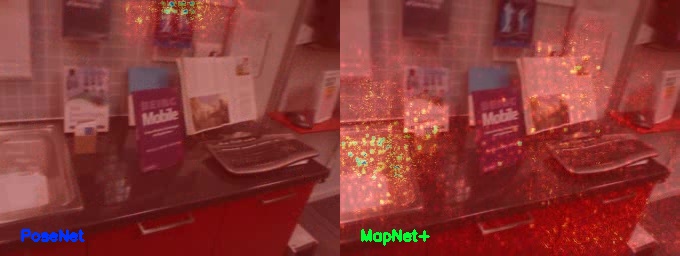}
    \end{subfigure}
    \hfill
    \begin{subfigure}{0.45\linewidth}
        \centering
        \includegraphics[width=\linewidth]{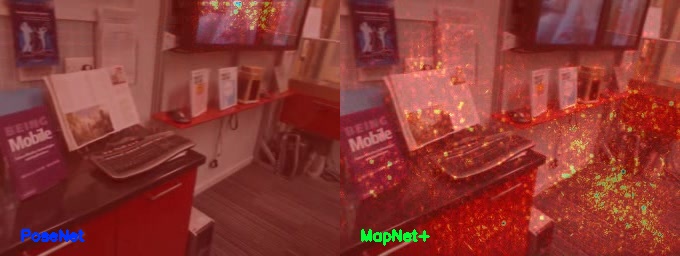}
    \end{subfigure}
    \begin{subfigure}{0.45\linewidth}
        \centering
        \includegraphics[width=\linewidth]{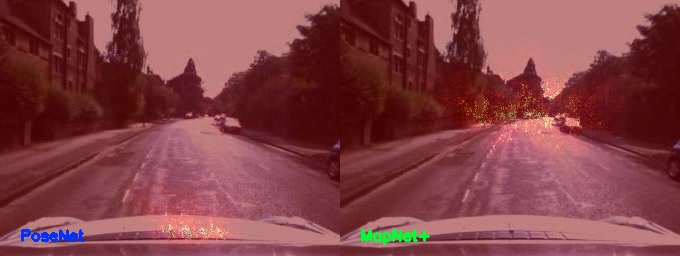}
    \end{subfigure}
    \hfill
    \begin{subfigure}{0.45\linewidth}
        \centering
        \includegraphics[width=\linewidth]{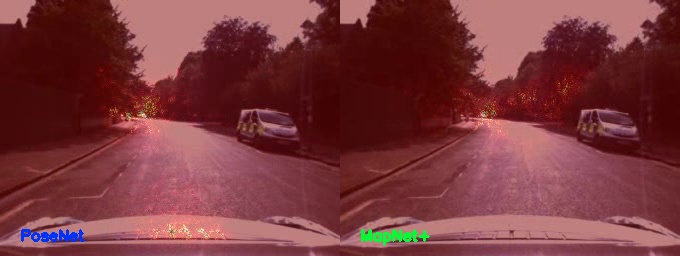}
    \end{subfigure}
    \caption{Attention maps for example images from the 7 Scenes dataset (top) and RobotCar dataset (bottom). In all 4 examples, we observe that
    MapNet+ focuses more on geometrically meaningful regions compared to PoseNet, and its saliency map is more consistent over time. Please see
    videos at \href{http://youtu.be/197N30A9RdE}{\url{http://youtu.be/197N30A9RdE}} to observe temporal consistency and more example frames.}
    \label{fig:attention_maps}
\end{figure*}

We also computed saliency maps $s(x,y) = \frac{1}{6}|\sum_{i=1}^6 \frac{\partial p_i}{\partial I(x,y)}|$ (magnitude gradient of the mean of the 6-element output w.r.t. input image, maxed over the 3 color channels) of PoseNet and MapNet+ on both the 7-scenes and RobotCar dataset (\texttt{redkitchen} and \texttt{loop} sequences). As shown in Fig~\ref{fig:attention_maps}, compared to PoseNet, MapNet+ focuses more on geometrically meaningful regions and its saliency map is more consistent over time. 


\FloatBarrier

{\small
\bibliographystyle{ieee}
\bibliography{references}

\begin{thebibliography}{10}\itemsep=-1pt

\bibitem{pytorch}
{PyTorch}, 2017.

\bibitem{quatbook}
S.~Altmann.
\newblock {\em Rotations, Quaternions, and Double Groups}.
\newblock Dover Publications, 2005.

\bibitem{NetVLAD}
R.~Arandjelovic, P.~Gronat, A.~Torii, T.~Pajdla, and J.~Sivic.
\newblock {NetVLAD}: {CNN} architecture for weakly supervised place
  recognition.
\newblock In {\em Proceedings of IEEE Conference on Computer Vision and Pattern
  Recognition (CVPR)}, 2016.

\bibitem{VLAD}
R.~Arandjelovic and A.~Zisserman.
\newblock All about {VLAD}.
\newblock In {\em Proceedings of IEEE Conference on Computer Vision and Pattern
  Recognition (CVPR)}, 2013.

\bibitem{bergmann17calibration}
P.~Bergmann, R.~Wang, and D.~Cremers.
\newblock Online photometric calibration of auto exposure video for real-time
  visual odometry and slam.
\newblock In {\em arXiv}, 2017.

\bibitem{Borg05mds}
I.~Borg and P.~J. Groenen.
\newblock {\em Modern multidimensional scaling: Theory and applications}.
\newblock Springer Science \& Business Media, 2005.

\bibitem{Brachmann17RANSAC}
E.~Brachmann, A.~Krull, S.~Nowozin, J.~Shotton, F.~Michel, S.~Gumhold, and
  C.~Rother.
\newblock {DSAC}: Differential {RANSAC} for camera localization.
\newblock In {\em Proceedings of IEEE Conference on Computer Vision and Pattern
  Recognition (CVPR)}, 2017.

\bibitem{Carlone16PGO}
L.~Carlone, G.~Calafiore, and F.~Dellaert.
\newblock Pose graph optimization in the complex domain: Duality, optimal
  solutions, and verification.
\newblock {\em IEEE Transactions on Robotics}, 32(3):545--565, 2016.

\bibitem{Chapelle06book}
O.~Chapelle, B.~Scholkopf, and A.~Zien.
\newblock {\em Semi-Supervised Learning}.
\newblock MIT Press, 2006.

\bibitem{Civera11Semantic}
J.~Civera, D.~Gaivez-Lopez, L.~Riazuelo, J.~Tardos, and J.~Montiel.
\newblock Towards semantic {SLAM} using a monocular camera.
\newblock In {\em IEEE/RSJ International Conference on Intelligent Robots and
  Systems (IROS)}, 2011.

\bibitem{Clark17VidLoc}
R.~Clark, S.~Wang, A.~Markham, N.~Trigoni, and H.~Wen.
\newblock {VidLoc}: A deep spatio-temporal model for {6-DoF} videoclip
  relocalization.
\newblock In {\em Proceedings of IEEE Conference on Computer Vision and Pattern
  Recognition (CVPR)}, 2017.

\bibitem{Cummins2008BOW}
M.~Cummins and P.~Newman.
\newblock {FAB-MAP}: Probabilistic localization and mapping in the space of
  appearance.
\newblock {\em The International Journal of Robotics Research}, 27(6):647--665,
  2008.

\bibitem{dam1998quaternions}
E.~B. Dam, M.~Koch, and M.~Lillholm.
\newblock {\em Quaternions, interpolation and animation}, volume~2.
\newblock 1998.

\bibitem{Davison03}
A.~J. Davison.
\newblock Real-time simultaneous localisation and mapping with a single camera.
\newblock In {\em Proceedings of IEEE International Conference on Computer
  Vision (ICCV)}, 2003.

\bibitem{Dellaert06ijrr}
F.~Dellaert and M.~Kaess.
\newblock Square {Root} {SAM}: Simultaneous localization and mapping via square
  root information smoothing.
\newblock {\em Intl. J. of Robotics Research (IJRR)}, 25(12):1181--1204, Dec.
  2006.

\bibitem{Segonne08transduction}
O.~Duchenne, J.~Y. Audibert, R.~Keriven, J.~Ponce, and F.~Segonne.
\newblock Segmentation by transduction.
\newblock In {\em 2008 IEEE Conference on Computer Vision and Pattern
  Recognition}, pages 1--8, June 2008.

\bibitem{Duckett02PGO}
T.~Duckett, S.~Marsland, and J.~Shapiro.
\newblock Fast, online learning of globally consistent maps.
\newblock {\em Autonomous Robots}, 12(3):287--300, 2002.

\bibitem{Engel2017DSO}
J.~Engel, V.~Koltun, and D.~Cremers.
\newblock {DSO}: Direct sparse odometry.
\newblock {\em IEEE Transactions on Pattern Analysis and Machine Intelligence
  (TPAMI)}, 2017.

\bibitem{Engel2013SVO}
J.~Engel, T.~Schops, and D.~Cremers.
\newblock Semi-dense visual odometry for a monocular camera.
\newblock In {\em Proceedings of IEEE International Conference on Computer
  Vision (ICCV)}, 2013.

\bibitem{Engel2014LSD}
J.~Engel, T.~Schops, and D.~Cremers.
\newblock {LSD-SLAM}: Large-scale direct monocular {SLAM}.
\newblock In {\em Proceedings of European Conference on Computer Vision
  (ECCV)}, 2014.

\bibitem{gomez2017pl}
R.~Gomez-Ojeda, F.-A. Moreno, D.~Scaramuzza, and J.~Gonzalez-Jimenez.
\newblock {PL-SLAM: a Stereo SLAM System through the Combination of Points and
  Line Segments}.
\newblock {\em arXiv preprint arXiv:1705.09479}, 2017.

\bibitem{Govindu01BA}
V.~Govindu.
\newblock Combining two-view constraints for motion estimation.
\newblock In {\em Proceedings of IEEE Conference on Computer Vision and Pattern
  Recognition (CVPR)}, 2001.

\bibitem{Govindu04BA}
V.~Govindu.
\newblock Lie-algebraic averaging for globally consistent motion estimation.
\newblock In {\em Proceedings of IEEE Conference on Computer Vision and Pattern
  Recognition (CVPR)}, 2004.

\bibitem{grimes10pgo}
M.~K. Grimes, D.~Anguelov, and Y.~LeCun.
\newblock {Hybrid hessians for flexible optimization of pose graphs}.
\newblock In {\em Intelligent Robots and Systems (IROS), 2010 IEEE/RSJ
  International Conference on}, pages 2997--3004. IEEE, 2010.

\bibitem{grisetti2010tutorial}
G.~Grisetti, R.~Kummerle, C.~Stachniss, and W.~Burgard.
\newblock A tutorial on graph-based slam.
\newblock {\em IEEE Intelligent Transportation Systems Magazine}, 2(4):31--43,
  2010.

\bibitem{He16ResNet}
K.~He, X.~Zhang, S.~Ren, and J.~Sun.
\newblock Deep residual learning for image recognition.
\newblock In {\em Proceedings of IEEE Conference on Computer Vision and Pattern
  Recognition (CVPR)}, 2016.

\bibitem{Hertzberg08quaternion}
C.~Hertzberg.
\newblock A framework for sparse, non-linear least squares problems on
  manifolds, 2008.

\bibitem{Huynh09}
D.~Huynh.
\newblock Metrics for {3D} rotations: Comparison and analysis.
\newblock {\em Journal of Mathematical Imaging and Vision}, 35(2):155--164,
  2009.

\bibitem{Stuhmer10}
S.~G. J.~Stuhmer and D.~Cremers.
\newblock Real-time dense geometry from a handheld camera.
\newblock In {\em Joint Pattern Recognition Symposium}, pages 11--20, 2010.

\bibitem{Jegou12}
H.~Jegou, F.~Perronnin, M.~Douez, J.~Sanchez, P.~Perez, and C.~Schmid.
\newblock Aggregating local image descriptors into compact codes.
\newblock {\em IEEE Transactions on Pattern Analysis and Machine Intelligence
  (TPAMI)}, 34(9):1704--1716, 2012.

\bibitem{quaternion_math}
Y.-B. Jia.
\newblock {Quaternion and Rotation}.
\newblock Com S 477/577 Lecture Notes at
  \url{http://web.cs.iastate.edu/~cs577/handouts/quaternion.pdf}, 2008.

\bibitem{Kendall16icra}
A.~Kendall and R.~Cipolla.
\newblock Modeling uncertainty in deep learning for camera relocalization.
\newblock In {\em IEEE International Conference on Robotics and Automation
  (ICRA)}, 2016.

\bibitem{Kendall17cvpr}
A.~Kendall and R.~Cipolla.
\newblock Geometric loss functions for camera pose regression with deep
  learning.
\newblock In {\em Proceedings of IEEE Conference on Computer Vision and Pattern
  Recognition (CVPR)}, 2017.

\bibitem{Kendall15iccv}
A.~Kendall, M.~Grimes, and R.~Cipolla.
\newblock {PoseNet}: A convolutional network for real-time {6-DOF} camera
  relocalization.
\newblock In {\em Proceedings of IEEE International Conference on Computer
  Vision (ICCV)}, 2015.

\bibitem{kingma14adam}
D.~Kingma and J.~Ba.
\newblock Adam: A method for stochastic optimization.
\newblock {\em arXiv preprint arXiv:1412.6980}, 2014.

\bibitem{Klein2007PTAM}
G.~Klein and D.~Murray.
\newblock Parallel tracking and maping for small {AR} workspaces.
\newblock In {\em IEEE and ACM International Symposium on Mixed and Augmented
  Reality (ISMAR)}, 2007.

\bibitem{Li2012Point}
Y.~Li, N.~Snavely, D.~Huttenlocher, and P.~Fua.
\newblock Worldwide pose estimation using {3D} point clouds.
\newblock In {\em Proceedings of European Conference on Computer Vision
  (ECCV)}, 2012.

\bibitem{lour09}
M.~A. Lourakis and A.~Argyros.
\newblock {SBA: A Software Package for Generic Sparse Bundle Adjustment}.
\newblock {\em ACM Trans. Math. Software}, 36(1):1--30, 2009.

\bibitem{Lu97PGO}
F.~Lu and E.~Milios.
\newblock Globally consistent range scan alignment for environment mapping.
\newblock {\em Autonomous Robots}, pages 334--349, 1997.

\bibitem{RobotCarDatasetIJRR}
W.~Maddern, G.~Pascoe, C.~Linegar, and P.~Newman.
\newblock {1 Year, 1000km: The Oxford RobotCar Dataset}.
\newblock {\em The International Journal of Robotics Research (IJRR)},
  36(1):3--15, 2017.

\bibitem{Martinec07BA}
D.~Martinec and T.~Pajdla.
\newblock Robust rotation and translation estimation in multiview
  reconstruction.
\newblock In {\em Proceedings of IEEE Conference on Computer Vision and Pattern
  Recognition (CVPR)}, 2007.

\bibitem{Melekhov17Hourglass}
I.~Melekhov, J.~Ylioinas, J.~Kannala, and E.~Rahtu.
\newblock Image-based localization using hourglass networks.
\newblock {\em arXiv}, abs/1703,07971, 2017.

\bibitem{Montemerlo02}
M.~Montemerlo, S.~Thrun, D.~Koller, and B.~Wegbreit.
\newblock {FastSLAM}: A factored solution to the simultaneous localization and
  mapping problem.
\newblock In {\em Eighteenth National Conference on Artificial Intelligence},
  pages 593--598, 2002.

\bibitem{Mur-ArtalMT15}
R.~Mur{-}Artal, J.~M.~M. Montiel, and J.~D. Tard{\'{o}}s.
\newblock {ORB-SLAM:} a versatile and accurate monocular {SLAM} system.
\newblock {\em CoRR}, abs/1502.00956, 2015.

\bibitem{Newcombe11}
R.~A. Newcombe, S.~J. Lovegrove, and A.~J. Davison.
\newblock {DTAM}: Dense tracking and mapping in real-time.
\newblock In {\em Proceedings of IEEE International Conference on Computer
  Vision (ICCV)}, pages 2320--2327, 2011.

\bibitem{Nister04visualodometry}
D.~Nistér, O.~Naroditsky, and J.~Bergen.
\newblock Visual odometry.
\newblock In {\em Proceedings of IEEE Conference on Computer Vision and Pattern
  Recognition (CVPR)}, pages 652--659, 2004.

\bibitem{Moreno13SLAMp}
R.~Salas-Moreno, R.~Newcombe, H.~Strasdat, and P.~Kelly.
\newblock {SLAM++}: Simultaneous localization and mapping at the level of
  objects.
\newblock In {\em Proceedings of IEEE Conference on Computer Vision and Pattern
  Recognition (CVPR)}, 2013.

\bibitem{Sattler2011}
T.~Sattler, B.~Leibe, and L.~Kobbelt.
\newblock Fast image-based localization using direct 2d-3d matching.
\newblock In {\em Proceedings of IEEE International Conference on Computer
  Vision (ICCV)}, 2011.

\bibitem{Sattler2017}
T.~Sattler, B.~Leibe, and L.~Kobbelt.
\newblock Efficient and effective prioritized matching for large-scale
  image-based localization.
\newblock {\em IEEE Transactions on Pattern Analysis and Machine Intelligence
  (TPAMI)}, 39(9):1744--1756, 2017.

\bibitem{Shotton13Scene7}
J.~Shotton, B.~Glocker, C.~Zach, S.~Izadi, A.~Criminisi, and A.~Fitzgibbon.
\newblock Scene coordinate regression forests for camera relocalization in
  {RGBD} images.
\newblock In {\em Proceedings of IEEE Conference on Computer Vision and Pattern
  Recognition (CVPR)}, 2013.

\bibitem{Strasdat12}
H.~Strasdat, J.~M.~M. Montiel, and A.~J. Davison.
\newblock Visual {SLAM}: Why filter?
\newblock {\em Image Vision Computing}, 30(2):65--77, Feb. 2012.

\bibitem{Su15}
H.~Su, C.~Qi, Y.~Li, and L.~Guibas.
\newblock Render for {CNN}: Viewpoint estimation in images using {CNN}s trained
  with rendered {3D} model views.
\newblock In {\em Proceedings of IEEE International Conference on Computer
  Vision (ICCV)}, 2015.

\bibitem{Taguchi2013may}
Y.~Taguchi, Y.-D. Jian, S.~Ramalingam, and C.~Feng.
\newblock Point-plane {SLAM} for hand-held 3d sensors.
\newblock In {\em IEEE International Conference on Robotics and Automation
  (ICRA)}, pages 5182--5189, 2013.

\bibitem{Thrun02a}
S.~Thrun.
\newblock Robotic mapping: A survey.
\newblock In G.~Lakemeyer and B.~Nebel, editors, {\em Exploring Artificial
  Intelligence in the New Millenium}. Morgan Kaufmann, 2002.

\bibitem{Thrun05}
S.~Thrun and M.~Montemerlo.
\newblock The {GraphSLAM} algorithm with applications to large-scale mapping of
  urban structures.
\newblock {\em International Journal on Robotics Research}, 25(5/6):403--430,
  2005.

\bibitem{Triggs99}
B.~Triggs, P.~F. McLauchlan, R.~I. Hartley, and A.~W. Fitzgibbon.
\newblock Bundle adjustment - a modern synthesis.
\newblock In {\em Proceedings of the International Workshop on Vision
  Algorithms: Theory and Practice}, ICCV '99, pages 298--372, 2000.

\bibitem{Walch17LSTM}
F.~Walch, C.~Hazirbas, L.~Leal-Taixe, T.~Sattler, S.~Hilsenbeck, and
  D.~Cremers.
\newblock Image-based localization using {LSTM}s for structured feature
  correlation.
\newblock In {\em Proceedings of IEEE International Conference on Computer
  Vision (ICCV)}, 2017.

\bibitem{Zhou2015StructSLAM}
H.~Zhou, D.~Zou, L.~Pei, R.~Ying, P.~Liu, and W.~Yu.
\newblock {StructSLAM}: Visual {SLAM} with building structure lines.
\newblock {\em IEEE Transactions on Vehicular Technology}, 64(4):1364--1375,
  2015.

\end{thebibliography}
}

\end{document}